\documentclass{article}
\pdfoutput=1
\usepackage[dvipsnames, table]{xcolor}
\usepackage{color}
\usepackage{tikz,pgfplots,pgfplotstable}
\usepackage{float}
\usepackage{amsthm}
\usepackage{diagbox}
\usepackage{xspace}
\usepackage{multirow}
\usepackage{tabularx}
\usepackage{xr}
\usepackage{caption}
\usepackage{pifont}
\usepackage{amsmath}
\usepackage{hyperref}
\usepackage{wrapfig}

\newcommand{\hh}[1]{#1}
\newcommand{\E}[1]{\mathbb{E}_{P_t}\left[#1\right]}
\newcommand{\Var}[1]{\mathrm{Var}_{P_t}\left(#1\right)}

\newcommand{\method}[0]{PeTTA\xspace}
\newcommand{\setting}[0]{recurring\xspace}
\newcommand{\Setting}[0]{Recurring\xspace}

\newcommand{\cmark}{\ding{51}}%
\newcommand{\xmark}{\ding{55}}%
\definecolor{ClrHighlight}{RGB}{210, 255, 255}
\definecolor{green01270}{RGB}{80, 200, 120}
\definecolor{ForestGreen}{RGB}{34,139,34}
\definecolor{darkgray}{RGB}{113, 121, 126}

\newtheorem{theorem}{Theorem}
\newtheorem{lemma}{Lemma}
\newtheorem{corollary}{Corollary}
\newtheorem{assumption}{Assumption}
\newtheorem{definition}{Definition}

\newcommand{\changes}[1]{\textcolor{black}{#1}}
\newcommand{\justification}[1]{\textcolor{blue}{#1}}
\definecolor{cvprblue}{rgb}{0,0.20,0.74}
\hypersetup{
    colorlinks=true,
    linkcolor=cvprblue,
    filecolor=magenta,      
    urlcolor=cyan,
    }
    
\newcommand{\Sec}[1]{{Sec.}~#1}
\newcommand{\Tab}[1]{{Tab.}~#1}
\newcommand{\Fig}[1]{{Fig.}~#1}
\newcommand{\Appdx}[1]{{Appdx.}~#1}
\newcommand{\Eq}[1]{{Eq.}~#1}
\newcommand{\Alg}[1]{{Alg.}~#1}

\usepackage{algorithm} 
\usepackage{algorithmic}  
\usepackage[algo2e]{algorithm2e} 
\usetikzlibrary{fit,calc}
\newcommand*{\tikzmk}[1]{\tikz[remember picture,overlay,] \node (#1) {};\ignorespaces}
\newcommand{\boxit}[1]{\tikz[remember picture,overlay]{\node[yshift=3pt,fill=#1,opacity=.25,fit={(StartHere)($(EndHere)+(.95\linewidth,.8\baselineskip)$)}] {};}\ignorespaces}
\colorlet{codehighlightpink}{red!40}
\colorlet{codehighlightblue}{cyan!60}

\usepackage[nonatbib, final]{neurips_2024}
\usepackage[numbers]{natbib}

\usepackage[utf8]{inputenc} %
\usepackage[T1]{fontenc}    %
\usepackage{hyperref}       %
\usepackage{url}            %
\usepackage{booktabs}       %
\usepackage{amsfonts}       %
\usepackage{nicefrac}       %
\usepackage{microtype}      %
\usepackage{xcolor}         %
\usepackage{minitoc}        %
\usepackage{comment}
\title{Persistent Test-time Adaptation in  \\ \Setting Testing Scenarios}

\author{%
  Trung-Hieu Hoang$^{1}$ \qquad Duc Minh Vo$^{2}$ \qquad Minh N. Do$^{1,3}$\\
  $^{1}$Department of Electrical and Computer Engineering, University of Illinois at Urbana-Champaign\\
  $^{2}$The University of Tokyo\\
  $^{3}$VinUni-Illinois Smart Health Center, VinUniversity \\
  \texttt{\{hthieu, minhdo\}@illinois.edu} \qquad \texttt{vmduc@nlab.ci.i.u-tokyo.ac.jp}\\
}

\begin{document}
\maketitle

\doparttoc %
\faketableofcontents %

\begin{abstract}
Current test-time adaptation (TTA) approaches aim to adapt \changes{a machine learning model} to environments that change continuously.
Yet, it is unclear whether TTA methods can maintain their adaptability over prolonged periods.
To answer this question, we introduce a diagnostic setting - \textbf{\setting TTA} where environments not only change but also recur over time, creating an extensive data stream.
This setting allows us to examine the error accumulation of TTA models, in the most basic scenario, when they are regularly exposed to previous testing environments.
Furthermore, we simulate a TTA process on a simple yet representative \textbf{$\epsilon$-perturbed Gaussian Mixture Model Classifier}, deriving theoretical insights into the dataset- and algorithm-dependent factors contributing to gradual performance degradation.
Our investigation leads us to propose \textbf{persistent TTA (\method)}, which senses when the model is diverging towards collapse and adjusts the adaptation strategy, striking a balance between the dual objectives of adaptation and model collapse prevention.
The supreme stability of \method over existing approaches, in the face of lifelong TTA scenarios, has been demonstrated over comprehensive experiments on various benchmarks. Our project page is available at \url{https://hthieu166.github.io/petta}.

\end{abstract}

\vspace*{-1.5\baselineskip}
\section{Introduction}

Machine learning (ML) models have demonstrated significant achievements in various areas~\cite{he2015delving, mildenhall2020nerf, alec2021_clip, isensee_nnu-net_2021}. Still, they are inherently susceptible to distribution-shift~\cite{10.5555/1462129, pmlr-v37-ganin15, pmlr-v97-recht19a, 9710159, blaas2023considerations} (also known as the divergence between the training and testing environments), leading to a significant degradation in model performance.
The ability to deviate from the conventional testing setting appears as a crucial aspect in boosting ML models' adaptability when confronted with a new testing environment that has been investigated~\cite{Li_2018_CVPR, pmlr-v119-sun20b, Ganin2017_domain}.
Among \changes{common} domain generalization methods~\cite{jindong2021_dg_survey, yusuke2021_t3a, ahuja2021invariance}, \textit{test-time adaptation (TTA)} takes the most challenging yet rewarding path that
\changes{leverages} unlabeled data available at test time for self-supervised adaptation prior to the final inference~\cite{wang2021tent, nguyen2023tipi, chen2022contrastive, niu2023towards, Wang_2022_CVPR}.

\newsavebox{\figb}
\sbox{\figb}{%
    \input{figures/intro/error_rotta_ours}
 }
 
\begin{figure}[ht!]
    \pgfplotsset{every x tick label/.append style={font=\tiny, yshift=0.5ex}}
    \pgfplotsset{every y tick label/.append style={font=\tiny, xshift=0.5ex}}
    \hspace{-10pt}
    \begin{tikzpicture}
    \draw (0.6, 0) node[inner sep=0, anchor=south west] {\includegraphics[width=5.7cm]{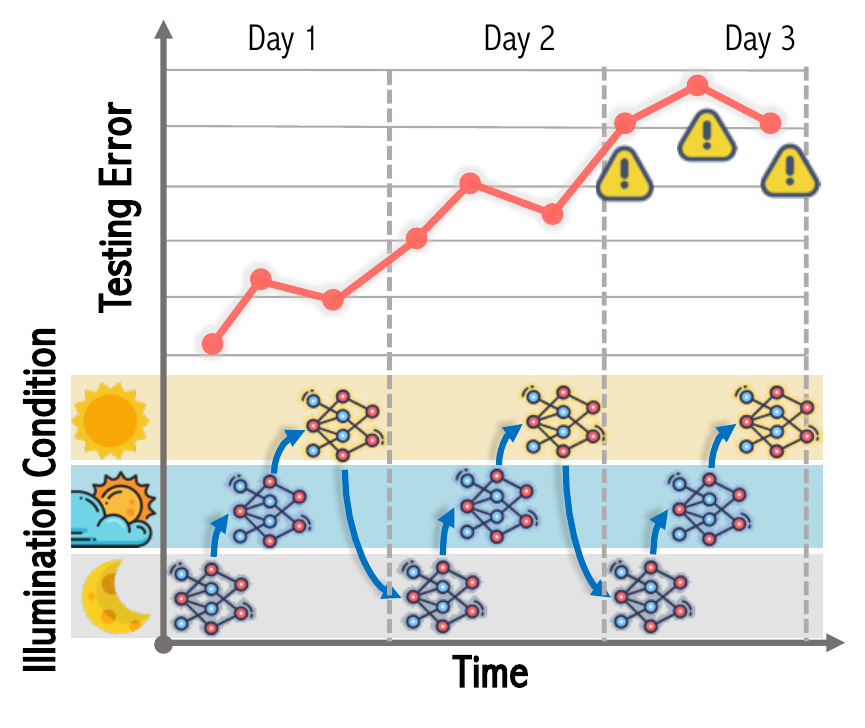}};
    \draw (6.0, 0) node[inner sep=0, anchor=south west] {\usebox \figb};
    \end{tikzpicture}
    \caption{
     \textit{\Setting Test-time Adaption (TTA)}.
    (left) Testing environments may change recurringly and preserving adaptability when visiting \textit{the same} testing condition is not guaranteed. 
    (right) The testing error of RoTTA~\cite{yuan2023robust} progressively raises (performance degradation) and exceeds the error of the source model (no TTA) while our \method demonstrates its stability when adapting to the test set of CIFAR-10-C~\cite{hendrycks2019robustness} 20 times. 
    The \textbf{bold} lines denote the running mean and the \textcolor{darkgray}{shaded} lines in the background represent the testing error on each domain (excluding the source model, for clarity).
    }
    \label{fig:fig1_introduction}
     \vspace*{-1.1\baselineskip}
\end{figure}

Early TTA studies have concentrated on a simply ideal adaptation scenario where the test samples come from a fixed single domain~\cite{wang2021tent, nguyen2023tipi, niu2023towards}.
As a result, such an assumption is far from the ever-changing and complex testing environments. %
To confront continually changing environments~\cite{Wang_2022_CVPR, döbler2023robust}, Yuan \textit{et al.}~\cite{yuan2023robust} proposed a \textit{practical TTA} scenario where distribution changing and correlative sampling occur~\cite{gong2022note} simultaneously.
Though practical TTA is more realistic than \changes{what the previous assumptions have made}, it still assumes that any environment only appears once in the data stream, \changes{a condition which does not hold true}.
\changes{Taking a surveillance camera as an example,} it might accommodate varying lighting conditions recurringly day after day (Fig.~\ref{fig:fig1_introduction}-left).
\changes{Based on this reality,} we hypothesize that the recurring of those conditions may reveal the error accumulation phenomenon in TTA, resulting in performance degradation over a long period.
To verify our hypothesis, we simulate a \setting testing environment and observe the increasing error rate by recurringly adapting to the test set of CIFAR-10-C~\cite{hendrycks2019robustness} multiple times.
We showcase the testing error of RoTTA~\cite{yuan2023robust} after 20 cycles of adaptation in Fig.~\ref{fig:fig1_introduction}-right.
As expected, RoTTA can successfully adapt and deliver encouraging outcomes within the first few passes. 
However, this advantage \changes{is short-lived} as our study \changes{uncovers a significant issue}:\textit{ TTA approaches in this setting may experience severe and persistent degradation in performance}.
Consequently, the testing error of RoTTA gradually escalates over time and quickly surpasses the model without adaptation. 
This \changes{result} confirms the risk of TTA deployment in our illustrative scenario, as an algorithm might work well in the first place and gradually degenerate.
Therefore, ensuring sustainable quality is crucial for real-world applications, \changes{especially given the \setting nature of testing environments}.

This study examines whether the adaptability of a TTA algorithm persists over an extended testing stream. Specifically, in the most basic scenario, where the model returns to a previously encountered testing environment after undergoing various adjustments.
We thus propose a more general testing scenario than the practical TTA~\cite{yuan2023robust}, namely \textit{\setting TTA}, where the environments not only change gradually but also recur in a correlated manner over time.
We first \changes{analyze} a simulation \changes{using} the \textit{$\epsilon-$perturbed Gaussian Mixture Model Classifier ($\epsilon-$GMMC)} on a synthesized dataset and derive \changes{a} theoretical analysis to confirm \changes{our findings, offering insights to tackle} similar \changes{issues} in deep neural networks. The analysis provides hints for reasoning the success of many recent robust continual TTA approaches~\cite{yuan2023robust, döbler2023robust, Wang_2022_CVPR, gong2022note} and
leading us to propose a simple yet effective baseline to avoid performance degradation, namely \textit{Persistent TTA (\method)}. \method continuously monitors the chance of collapsing and adjusts the adaptation strategy on the fly, striking a balance between the two objectives: \textit{adaptation} and \textit{\changes{collapse prevention}}. Our contributions can be summarized as follows:

\vspace*{-0.5\baselineskip}
\begin{itemize}
\item First, this work \textit{proposes a testing scenario - \setting TTA}, a simple yet sufficient setup for diagnosing the overlooked \textit{gradual performance degradation} phenomenon of TTA. 
\item Second, we formally\textit{ define the phenomenon of TTA collapsing and undertake a theoretical analysis} on an $\epsilon$-GMMC, shedding light on dataset-dependent and algorithm-dependent factors that contribute to the error accumulation during TTA processes.
\item Third, we \textit{introduce persistent TTA (\method)} - a simple yet effective adaptation scheme that surpasses all baseline models and demonstrates a persisting performance. %
\end{itemize}
For \changes{more} context on related work, readers are directed to visit our discussions in \Appdx{\ref{sec:related_work}}.

\vspace*{-0.5\baselineskip}
\vspace*{-0.4\baselineskip}
\section{Background}
\label{sec:background}
\vspace*{-0.4\baselineskip}
\noindent \textbf{Test-time Adaptation (TTA). } A TTA algorithm operates on an ML classifier $f_t: \mathcal{X} \rightarrow \mathcal{Y}$ with parameter $\theta_t \in \Theta$ (parameter space) gradually changing over time ($t \in \mathcal{T} $) that maps an input image $\boldsymbol{x} \in \mathcal{X}$ to a category (label) $y \in \mathcal{Y}$. 
Let the capital letters $(X_t, Y_t) \in \mathcal{X} \times \mathcal{Y}$ denote a pair of \textit{random variables} with the joint distribution $P_t(\boldsymbol{x},y) \in \mathcal{P}_d, t \in \mathcal{T}$. Here, $\mathcal{P}_d$ belongs to collection of $D$ sets of testing scenarios (domains) $\{\mathcal{P}_d\}_{d=1}^D$.  The covariate shift~\cite{10.5555/1462129} is assumed: $P_t(\boldsymbol{x})$ and $P_{t'}(\boldsymbol{x})$ could be different but $P_t(y|\boldsymbol{x}) = P_{t'}(y|\boldsymbol{x})$ holds $\forall t \neq t'$.
At $t=0$, $\theta_0$ is initialized by a supervised model trained on $P_0 \in \mathcal{P}_0$ (source dataset). The model then explores an online stream of testing data. For each $t>0$, 
it receives  $X_t$ (typically in form of  a batch of $N_t$ testing samples) for adapting itself $f_{t-1} \rightarrow f_t$ before making the final prediction $f_t\left(X_t\right)$.

\noindent \textbf{TTA with Mean Teacher Update. } To achieve a stable optimization process, the main (\textit{teacher}) model $f_t$ are updated indirectly through a \textit{student} model with parameters $\theta_t'$~\cite{wang2021tent, yuan2023robust, döbler2023robust,gong2022note, antti2017_mean_teachers}.  At first, the teacher model in the previous step introduces a \textit{pseudo label}~\cite{hyun2013_pseudo} $\hat Y_t$ for each $X_t$: 
\begin{align}
    \label{eq:pseudo_label}
    \hat Y_t = f_{t-1}(X_t). %
\end{align}
With a classification loss $\mathcal{L}_{\mathrm{CLS}}$ (e.g., cross-entropy~\cite{NIPS2004_96f2b50b}), and a model parameters regularizer $\mathcal{R}$, the student model is first updated with a generic optimization operator \texttt{Optim}, followed by an exponential moving average (EMA) update of the teacher model parameter $\theta_{t-1}$:
\begin{align}
    \theta_t'&= \underset{\theta' \in \Theta}{\text{\texttt{Optim} }}  \mathbb{E}_{P_t}\left[\mathcal{L}_{\mathrm{CLS}} \left(\hat Y_t, X_t; \theta'\right) \right] + \lambda \mathcal{R}(\theta'), \label{eq:general_opti_step}\\ \ %
    \theta_t &= (1-\alpha) \theta_{t-1} + \alpha \theta'_t ,\label{eq:general_teacher_update}
\end{align}
with $\alpha \in (0,1)$ - the update rate of EMA, \changes{and} $\lambda \in \mathbb{R}^+$ - the weighting coefficient of the regularization term, are the two hyper-parameters. 

\noindent \textbf{Practical TTA. } In practical TTA~\cite{yuan2023robust}, two characteristics of the aforementioned distribution of data stream are noticeable. Firstly, $P_t$'s can be partitioned by $t_d$'s in which $\left\{P_t\right\}_{t=t_{d-1}}^{t_d} \subset \mathcal{P}_d$. Here, each partition of consecutive steps follows the same underlying distribution which will \textit{change continually through $D$ domains}~\cite{Wang_2022_CVPR} ($\mathcal{P}_1 \rightarrow \mathcal{P}_2 \cdots \rightarrow \mathcal{P}_D$). Secondly, the category distribution in each testing batch is \textit{temporally correlated}~\cite{gong2022note}. This means within a batch, a small subset of categories is dominant over others, making the marginal distribution $P_t(y) = 0, \forall y \not \in \mathcal{Y}_t \subset \mathcal{Y}$ even though the category distribution over all batches are balanced. Optimizing under this low intra-batch diversity ($|\mathcal{Y}_t| \ll |\mathcal{Y}|$) situation can slowly degenerate the model \cite{malik2022_paramter-free}.

\vspace*{-0.6\baselineskip}
\section{\Setting TTA and Theoretical Analysis}
\vspace*{-0.3\baselineskip}
This section conducts a theoretical analysis on a concrete failure case of a simple TTA model. The results presented at the end of \Sec{\ref{ssec:egmmc_analysis}} will elucidate the factors contributing to the collapse (\Sec{\ref{ssec:model_collapse}}), explaining existing good practices (\Sec{\ref{ssec:existing_solutions}}) and give insights into potential solutions (\Sec{\ref{ssec:pers_tta}}).
\vspace*{-0.4\baselineskip}
\subsection{\Setting TTA and Model Collapse}
\label{ssec:model_collapse}
\vspace*{-0.3\baselineskip}
\noindent \textbf{\Setting TTA.} To study the gradual performance degradation (or model collapse), we propose a \textit{new testing scenario based on practical TTA}~\cite{yuan2023robust}. Conducting a single pass through $D$ distributions, as done in earlier studies \cite{yuan2023robust, Wang_2022_CVPR}, may not effectively identify the degradation.
\hh{To promote consistency}, our \setting TTA performs\textit{ revisiting the previous distributions $K$ times} to compare the incremental error versus the previous visits. For example, a sequence with $K=2$ could be $\mathcal{P}_1 \rightarrow  \mathcal{P}_2 \rightarrow \cdots  \rightarrow \mathcal{P}_D \rightarrow \mathcal{P}_1 \rightarrow  \mathcal{P}_2 \rightarrow \cdots \rightarrow \mathcal{P}_D$. \Appdx{\ref{appdx:repeating_tta}} extends our justifications on constructing \setting TTA.

\begin{definition}[\textbf{Model Collapse}]
\label{def:model_collapse}
A model is said to be \textit{collapsed} from step $\tau \in \mathcal{T}, \tau < \infty$ if there exists a non-empty subset of categories $\Tilde{\mathcal{Y}} \subset \mathcal{Y}$ such that $\Pr\{Y_t \in \Tilde{\mathcal{Y}}\} > 0$ but the marginal $\Pr\{\hat Y_t \in \mathcal{\Tilde{Y}}\}$ converges to zero in probability: 
\begin{align*}
    \underset{t \to \tau}{\lim} \Pr\{\hat Y_t \in \Tilde{\mathcal{Y}}\} = 0. %
\end{align*}%
\end{definition}

Here, upon collapsing, a model tends to \textit{ignore} almost categories in $\Tilde{\mathcal{Y}}$. As it is irrecoverable once collapsed, the only remedy would be resetting all parameters back to $\theta_0$.

\vspace*{-0.5\baselineskip}
\subsection{Simulation of Failure and Theoretical Analysis}
\label{ssec:egmmc_analysis}
Collapsing behavior varies across datasets and the adaptation processes. Formally studying this phenomenon on a particular real dataset and a TTA algorithm is challenging. Therefore, we propose a theoretical analysis on $\epsilon$-perturbed binary Gaussian Mixture Model Classifier ($\epsilon$-GMMC) that shares the typical characteristics \textit{by construction} and demonstrates the \textit{same collapsing pattern} in action (Sec.~\ref{sec:eps_gmmc_result}) as observed on real continual TTA processes (Sec.~\ref{ssec:result_real_data}).

\noindent \textbf{Simulated Testing Stream. } Observing a testing stream with $(X_t,Y_t) \in \mathcal{X} \times \mathcal{Y} = \mathbb{R} \times \{0,1\}$ and the underlying joint distribution $P_t(x,y) = p_{y,t} \cdot \mathcal{N}(x; \mu_y, \sigma_y^2)$. The main task is predicting $X_t$ was sampled from cluster 0 or 1 (negative or positive). Conveniently, 
let $p_{y,t} \overset{\Delta}{=} P_t(y) = \Pr(Y_t = y)$ and $\hat p_{y,t} \overset{\Delta}{=} \Pr(\hat Y_t = y)$ be the marginal distribution of the true label $Y_t$ and pseudo label $\hat{Y}_t$.

\noindent \textbf{GMMC and TTA. } GMMC first implies an \textit{equal prior} distribution by construction which is desirable for the actual TTA algorithms (e.g.,  category-balanced sampling strategies in~\cite{yuan2023robust, gong2022note}). Thus, it simplifies $f_t$ into a maximum likelihood estimation $f_t(x) = \text{argmax}_{y \in \mathcal{Y}} \Pr(x|y; \theta_t)$ with $\Pr(x|y; \theta_t) = \mathcal{N}(x; \hat\mu_{y,t}, \hat\sigma^2_{y,t})$. The goal is estimating a set of parameters $\theta_t = \{\hat\mu_{y,t}, \hat \sigma^2_{y,t}\}_{y \in \mathcal{Y}}$.
A perfect classifier $\theta_{0} = \{\mu_{y}, \sigma^2_{y}\}_{y \in \mathcal{Y}}$ is initialized at $t=0$. For the consecutive steps, the simplicity of GMMC allows solving the \texttt{Optim} (for finding $\theta'_t$, Eq.~\ref{eq:general_opti_step}) perfectly by computing the empirical mean and variance of new samples, approximating $\mathbb{E}_{P_t}$. 
The mean teacher update (Eq.~\ref{eq:general_teacher_update}) for GMMC is: 
{
\begin{align}
    \label{eq:update_gmmc}
    \hat \mu_{y,t} = \begin{cases} 
    (1-\alpha) \hat \mu_{y,t-1} +  \alpha \mathbb{E}_{P_t}\left[X_t|\hat Y_t\right] & \text{if } \hat{Y}_t = y \\
    \hat \mu_{y,t-1}  & \text{otherwise}
    \end{cases}.
\end{align}

}
\vspace{-0.4\baselineskip}
The update of $\hat \sigma^2_{y,t}$ is similar. $\hat{Y}_t = f_{t-1}(X_t)$ can be interpreted as a \textit{pseudo label} (Eq.~\ref{eq:pseudo_label}). 

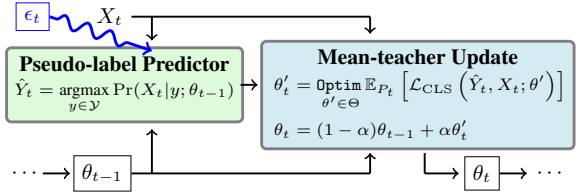
\begin{figure}
  \begin{minipage}[c]{0.55\textwidth}
    \resizebox{\textwidth}{!}{
    \begin{tikzpicture}[
    rec2/.style={rectangle, draw, minimum width=30, minimum height=25, very thick, rounded corners=2, draw=gray,fill=lightgreen},
    rec3/.style={rectangle, draw, minimum width=30, minimum height=25, very thick, rounded corners=2, draw=gray,fill=lightblue},
    decoration = {snake,   %
                    pre length=3pt,post length=7pt,%
                    },
    ]
    \definecolor{lightgreen}{RGB}{221,250,218}
    \definecolor{lightblue}{RGB}{212, 235, 242}
    \tikzstyle{every node}=[font=\small];
    \node[rec2, align=center] (a) at (-0.4,0,0) {\textbf{Pseudo-label Predictor} \\\scriptsize{$ \hat Y_t=\underset{y \in \mathcal{Y}}{\text{argmax}} \Pr(X_t|y; \theta_{t-1})$}};
    \node at (-0.3,1.0,0) [left] {\small{$X_t$}};
    \draw[thick,->] (1.3,0,0) -- (1.55,0,0) node[left] {};
    \node[rec3, align=center] (a) at (3.9,-0.15) {\textbf{Mean-teacher Update} \\
    \scriptsize{
    $\begin{aligned}
        \theta_t'&= \underset{\theta' \in \Theta}{\text{\texttt{Optim} }}  \mathbb{E}_{P_t}\left[\mathcal{L}_{\mathrm{CLS}}\left(\hat Y_t, X_t; \theta'\right)\right] \\
        \theta_t &= (1-\alpha) \theta_{t-1} + \alpha \theta_t'
    \end{aligned}$}
    };

    \node at (-1.5,1.0,0) [left, blue, draw=blue] {\small{$\epsilon_t$}};
    \draw[very thick, ->, draw=blue, decorate] (-1.5,1.0,0) -- (0.0, 0.5,0) node[left] {};
    
    \draw[thick,->] (0,1.0,0) -- (0,0.65,0) node[left] {};
    \draw[thick,->] (3.2,1.0,0) -- (3.2,0.75,0) node[left] {};
    \draw[thick,->] (0,-1.3,0) -- (0,-0.65,0) node[left] {};
    \draw[thick,->] (3.2,-1.3,0) -- (3.2,-1.0,0) node[left] {};
    \draw[thick,-] (-0.3,-1.3,0) -- (3.2,-1.3,0) node[left] {};
    \draw[thick,-] (-0.3,1.0,0) -- (3.2,1.0,0) node[left] {};
    \draw[thick,->] (-1.6,-1.3,0) -- (-1.2,-1.3,0) node[left] {};
    
    \node at (-1.5,-1.3,0) [left] {\small{$\cdots$}};
    \node at (-0.3,-1.3,0) [left, draw=black] {\small{$\theta_{t-1}$}};
    \draw[thick,-] (4.0,-1.3,0) -- (4.0,-1.0,0) node[left] {};
    
    \draw[thick,->] (4.0,-1.3,0) -- (4.5,-1.3,0) node[left] {};
    \node at (5.1,-1.3,0) [left, draw=black] {\small{$\theta_t$}};
    \draw[thick,->] (5.1,-1.3,0) -- (5.5,-1.3,0) node[left] {};
    \node at (6.2,-1.3,0) [left] {\small{$\cdots$}};

    \end{tikzpicture}}
  \end{minipage}\hfill
  \begin{minipage}[c]{0.43\textwidth}
    
    \caption{
    $\epsilon$-perturbed binary Gaussian Mixture Model Classifier, imitating a continual TTA algorithm for theoretical analysis. Two main components include a pseudo-label  predictor (Eq.~\ref{eq:pseudo_label}), and a mean teacher update (Eqs.~\ref{eq:general_opti_step}, \ref{eq:general_teacher_update}). The \textit{predictor is perturbed} for retaining a false negative rate of $\epsilon_t$ to simulate an undesirable TTA testing stream.}
    \label{fig:diagram_egmmc}
  \end{minipage}
  \vspace*{-1.0\baselineskip}
\end{figure}

\noindent \textbf{$\epsilon$-GMMC. } Severe distribution shifts or low intra-batch category diversity of \setting TTA/practical TTA \textit{both result in an increase in the error rate of the predictor}. 
Instead of directly modeling the dynamic changes of $p_{y,t}$ (which can be complicated depending on the dataset), we study an $\epsilon-$pertubed GMMC ($\epsilon-$GMMC), where $p_{y,t}$ is \textit{assumed to be static} (defined below) and the pseudo-label predictor of this model is \textit{perturbed} to \textit{simulate undesirable effects of the testing stream} on the predictor.
Two kinds of errors appear in a binary classifier~\cite{banerjee_hypothesis_2009}. Let 
\begin{align}
\label{eq:fnr}
\epsilon_t = \Pr\{Y_t = 1| \hat Y_t = 0 \}  
\end{align}
be the false negative rate (FNR) of the model at step $t$. Without loss of generality, we study the \textit{increasing type II collapse of $\epsilon$-GMMC}. By intentionally flipping the true positive pseudo labels in simulation, an FNR of $\epsilon_t$ is maintained (\Fig{\ref{fig:diagram_egmmc}}).

\begin{assumption}[\textbf{Static Data Stream}]
\label{as:static_stream}
The marginal distribution of the true label follows the same Bernoulli distribution $\mathrm{Ber}(p_0)$:
$p_{0,t} = p_0$, ($p_{1,t} = p_1 = 1-p_0), \forall t \in \mathcal{T}$.
\end{assumption}

\begin{lemma}[\textbf{Increasing FNR}] Under Assumption~\ref{as:static_stream}, a binary $\epsilon$-GMMC would collapsed (Def.~\ref{def:model_collapse}) with $\underset{t \to \tau}{\lim} \hat p_{1,t} = 0$ (or  $\underset{t \to \tau}{\lim} \hat p_{0,t} = 1$, equivalently) if and only if $\underset{t \to \tau}{\lim} \epsilon_{t} = p_1$.
\label{lmm:increasing_fnr}
\end{lemma}%

Lemma~\ref{lmm:increasing_fnr} states the negative correlation between $\hat p_{1,t}$ and $\epsilon_t$. Unsurprisingly, towards the collapsing point where all predictions are zeros, the FNR also increases at every step and eventually reaches the highest possible FNR of $p_1$. 

\begin{lemma}[\textbf{$\epsilon$-GMMC After Collapsing}]
\label{lmm:collapsed}
For a binary $\epsilon$-GMMC model, with Assumption~\ref{as:static_stream}, if $\underset{t \to \tau}{\lim} \hat p_{1,t} = 0$ (collapsing), the cluster 0 in  GMMC converges in distribution to a single-cluster GMMC with parameters:
\begin{align*}%
     \mathcal{N}(\hat \mu_{0,t}, \hat \sigma_{0,t}^2) \overset{d.}{\rightarrow}  \mathcal{N}&(p_0 \mu_0 + p_1 \mu_1, p_0\sigma_{0}^2+p_1\sigma_{1}^2+p_0p_1(\mu_0 - \mu_1)^2).
\end{align*}%
\end{lemma}

Lemma~\ref{lmm:collapsed} states the resulting $\epsilon-$GMMC after collapsing. Cluster 0 now \textit{covers the whole data distribution} (and assigning label 0 for all samples).  Furthermore, \textit{collapsing happens when $\hat \mu_{0,t}$ moves toward $\mu_1$}. We next investigate the factors and conditions for this undesirable convergence. 

\begin{theorem}[\textbf{Convergence of $\epsilon-$GMMC}]
\label{thm:cvg}
For a binary $\epsilon$-GMMC model, with Assumption~\ref{as:static_stream}, let the distance from $\hat \mu_{0,t}$ toward $\mu_1$ is $d_{t}^{0\rightarrow1} = \left|\E{\hat \mu_{0,t}} - \mu_1\right|$, then:
\[ d_{t}^{0\rightarrow1} - d_{t-1}^{0\rightarrow1} \leq \alpha \cdot p_0 \cdot \left( |\mu_0-\mu_1| - \frac{ d_{t-1}^{0\rightarrow1}}{1-\epsilon_t} \right).
\]%
\end{theorem}%

From Thm.~\ref{thm:cvg}, we observe that the distance $d_t^{0\rightarrow1}$'s converges (also indicating the convergence to the distribution in Lemma~\ref{lmm:collapsed}) if $d_t^{0\rightarrow1} < d_{t-1}^{0\rightarrow1}$. The model collapse happens when this condition holds for a sufficiently long period.  

\begin{corollary}[\textbf{A Condition for $\epsilon-$GMMC Collapse}]
\label{corollary:condition}
With fixed $p_0$, $\alpha, \mu_0, \mu_1$, $\epsilon-$GMMC is collapsed if there exists a sequence of $\{\epsilon_t\}_{\tau - \Delta_\tau}^{\tau}$ ($\tau \ge \Delta_\tau > 0$) such that: 
\begin{align*}
    p_1 \geq \epsilon_t > 1 - \frac{d_{t-1}^{0\rightarrow1}}{\left|\mu_0 - \mu_1\right|}, \quad t \in [\tau - \Delta_\tau, \tau].
\end{align*}%
\end{corollary}

Corollary~\ref{corollary:condition} introduces a condition $\epsilon$-GMMC collapse. Here, $\epsilon_t$'s are non-decreasing, $\underset{t \to \tau}{\lim} \epsilon_{t} = p_1$. 

\noindent \textbf{Remarks. } Thm.~\ref{thm:cvg} concludes two sets of factors contributing to collapse:
(i)~\textit{data-dependent factors}: the prior data distribution ($p_0$), the nature difference between two categories ($|\mu_0 - \mu_1|$); and 
(ii)~\textit{algorithm-dependent factors}: the update rate ($\alpha$), the FNR at each step ($\epsilon_t$). 
$\epsilon$-GMMC analysis sheds light on explaining model collapse on real datasets (Sec.~\ref{ssec:result_real_data}), reasons the existing approaches (Sec.~\ref{ssec:existing_solutions}) and motivates the development of our baseline (Sec.~\ref{ssec:pers_tta}).

\vspace*{-0.5\baselineskip}
\subsection{Connection to Existing Solutions}
\vspace*{-0.4\baselineskip}
\label{ssec:existing_solutions}

Prior TTA algorithms have already incorporated implicit mechanisms to mitigate model collapse. 
The theoretical results in the previous section explain the rationale behind these effective strategies.

\noindent \textbf{Regularization Term for $\theta_t$.} Knowing that $f_0$ is always well-behaved, an attempt is restricting the divergence of $\theta_t$ from $\theta_0$, e.g. using $\mathcal{R}(\theta_t) \overset{\Delta}{=} \|\theta_0 - \theta_t \|_2^2$ regularization~\cite{niu2022efficient}. The key idea is introducing a penalty term to avoid an extreme divergence as happening in Thm.~\ref{thm:cvg}.

\noindent \textbf{Memory Bank for Harmonizing $P_t(x)$.} Upon receiving $X_t$, samples in this batch are selectively updated to a memory bank $\mathcal{M}$ (which already contains a subset of some instances of $X_{t'}, t'<t$ in the previous steps). By keeping a balanced number of samples from each category, distribution $P_t^{\mathcal{M}}(y)$ of samples in $\mathcal{M}$ is expected to have less zero entries than $P_t(y)$, making the optimization step over $P_t^{\mathcal{M}}$ more desirable. From Thm.~\ref{thm:cvg},  $\mathcal{M}$ moderates the extreme value of the category distribution ($p_0$ term) which typically appears on batches with low intra-batch category diversity.

\vspace*{-0.6\baselineskip}
\section{Persistent Test-time Adaptation (PeTTA)}
\vspace*{-0.5\baselineskip}
\label{ssec:pers_tta}
Now we introduce our \textit{ Persistent TTA (\method)} approach. Further inspecting Thm.~\ref{thm:cvg}, while $\epsilon_t$ (Eq.~\ref{eq:fnr}) is not computable without knowing the true labels, the measure of divergence from the initial distribution (analogously to $d_{t-1}^{0\rightarrow1}$ term) can provide hints to fine-tune the adaptation process. 

\noindent \textbf{Key Idea. } A proper adjustment toward the TTA algorithm can \textit{break the chain of monotonically increasing} $\epsilon_t$'s in Corollary~\ref{corollary:condition} to prevent the model collapse. In the mean teacher update, the larger value of $\lambda$ (Eq.~\ref{eq:general_opti_step}) prioritizes the task of preventing collapse on one hand but also limits its adaptability to the new testing environment. Meanwhile, $\alpha$ (Eq.~\ref{eq:general_teacher_update}) controls the weight on preserving versus changing the model from the previous step. Drawing inspiration from the exploration-exploitation tradeoff~\cite{Rhee2018, Katehakis1987TheMB} encountered in reinforcement learning~\cite{richard2018_reinforcement}, we introduce a mechanism for \textit{adjusting $\lambda$ and $\alpha$ on the fly, balancing between the two primary objectives: adaptation and preventing model collapse}. Our strategy is prioritizing collapse prevention (increasing $\lambda$) and preserving the model from previous steps (decreasing $\alpha$) when there is a significant deviation from $\theta_0$.

In~\cite{niu2022efficient, yuan2023robust, Wang_2022_CVPR}, $\lambda$ and $\alpha$ were fixed through hyper-parameter tuning. This is suboptimal due to varying TTA environments and the lack of validation set~\cite{zhao2023on}. Furthermore, Thm.~\ref{thm:cvg} suggests the convergence rate quickly escalates when $\epsilon_t$ increases, making constant $\lambda, \alpha$ insufficient to prevent collapse.    

\noindent \textbf{Sensing the Divergence of $\theta_t$}.
We first equip \method with a mechanism for \textit{measuring its divergence from} $\theta_0$. 
Since $f_t(\boldsymbol{x}) = \text{argmax }_{y \in \mathcal{Y}} \Pr(y|\boldsymbol{x}; \theta_t)$,  %
we can decompose  $\Pr(y|\boldsymbol{x}; \theta_t) = [h \left( \phi_{\theta_t} (\boldsymbol{x}) \right)]_y$,  %
with $\phi_{\theta_t}(\cdot)$ is a $\theta_t$-parameterized deep feature extractor followed by a \textit{fixed} classification head (a linear and softmax layer) $h(\cdot)$. The operator $[\cdot]_y$ extracts the $y$\textsuperscript{th} component of a vector. 

Since $h (\cdot)$ remains unchanged, instead of comparing the divergence in the parameter space ($\Theta$) or between the output probability $\Pr(y|\boldsymbol{x}; \theta_t)$ and $\Pr(y|\boldsymbol{x}; \theta_0)$, we suggest an \textit{inspection over the feature embedding space} that preserves a \textit{maximum amount of information} in our case (data processing inequality~\cite{cover2006_elements}). 
Inspired by~\cite{li2017revisiting} and under Gaussian assumption, the \hh{Mahalanobis distance} of the first moment of the feature embedding vectors is compared. Let $\boldsymbol{z} = \phi_{\theta_t}(\boldsymbol{x})$, we keep track of a collection of the running mean of feature vector $\boldsymbol{z}$: $\{\boldsymbol{\hat{\mu}}_t^y\}_{y\in \mathcal{Y}}$ in which $\boldsymbol{\hat{\mu}}_t^y$ is EMA updated with vector $\boldsymbol{z}$ if $f_t(\boldsymbol{x}) = y$.
The divergence of $\theta_t$ at step $t$, evaluated on class $y$ is defined as:
\begin{align}
    \gamma_{t}^y = 1 - \exp\left(-(\boldsymbol{\hat{\mu}}_t^y - \boldsymbol{\mu}_0^y)^T \left(\boldsymbol{\Sigma}^y_0\right)^{-1} (\boldsymbol{\hat{\mu}}_t^y - \boldsymbol{\mu}_0^y) \right),
    \label{eq:divg_sensing}
\end{align}
where $\boldsymbol{\mu}_0^y$ and $\boldsymbol{\Sigma}_0^y$ are the pre-computed empirical mean and covariant matrix of feature vectors in the source dataset ($P_0$). 
The covariant matrix here is diagonal for simplicity. 
In practice, without directly accessing the training set, we assume a small set of unlabeled samples can be drawn from the source distribution for empirically computing these values (visit \Appdx{\ref{appdx:mean_cov_source_dataset}} for further details).

Here, we implicitly expect the independence of each entry in $\boldsymbol{z}$ and TTA approaches \textit{learn to align feature vectors of new domains back to the source domain} ($P_0$). Therefore, the accumulated statistics of these feature vectors at each step should be concentrated near the vectors of the initial model. The value of $\gamma_{t}^y \in [0,1]$ is close to 0 when $\theta_t = \theta_0$ and increases exponentially as $\boldsymbol{\hat{\mu}}_t^y$ diverging from $\boldsymbol{\mu}_0^y$.

\noindent \textbf{Adaptive Regularization and Model Update. }
With $\alpha_0$, $\lambda_0$ are initial values, utilizing $\gamma_{t}^y$ derived in \Eq{\ref{eq:divg_sensing}}, a pair of $(\lambda_t$, $\alpha_t$) is \textit{adaptively} chosen at each step:
\begin{align}
    \bar \gamma_t &= \frac{1}{|\hat{\mathcal{Y}}_t|} \sum_{y \in \hat{\mathcal{Y}}_t} \gamma_t^{y},  \quad  \hat{\mathcal{Y}}_t = \left\{\hat Y_t^{(i)} | i=1, \cdots, N_t\right\}; \nonumber \\
    \lambda_t &= \bar \gamma_t \cdot \lambda_0, \qquad \alpha_t = (1-\bar \gamma_t) \cdot \alpha_0, \label{eq:adaptive_lambda_alpha}
\end{align}
$\hat{\mathcal{Y}}_t$ is a set of unique pseudo labels in a testing batch ($\hat Y_t^{(i)}$ is the $i$\textsuperscript{th} realization of $\hat Y_t$).

\noindent \textbf{Anchor Loss. }  Penalizing the divergence with regular vector norms in high-dimensional space ($\Theta$) is insufficient (curse of dimensionality~\cite{Bellman:1957, tanin2019_high_dim_regularization}), especially with a large model and limited samples. \textit{Anchor loss} $\mathcal{L}_{\mathrm{AL}}$ can nail down the similarity between $f_t$ and $f_0$ in the probability space~\cite{li2016_learning_wo_forgetting, döbler2023robust}:
\begin{align}
    \label{eq:anchor_loss}
    \mathcal{L}_{\mathrm{AL}}(X_t; \theta) = -\sum_{y\in \mathcal{Y}}  \Pr(y|X_t; \theta_0) \log  \Pr(y|X_t; \theta),
    \vspace{-0.1\baselineskip}%
\end{align}%
which is equivalent to minimizing the KL divergence $D_{KL}\left(\Pr(y|X_t; \theta_0)\| \Pr(y|X_t; \theta)\right)$.

\noindent \textbf{Persistent TTA.} Having all the ingredients, we design our approach, \method, following the convention setup of the mean teacher update, with the category-balanced memory bank and the robust batch normalization layer from~\cite{yuan2023robust}. \Appdx{\ref{appdx:petta_pseudo_code}} introduces the \textit{pseudo code} of \method. For $\mathcal{L}_{\mathrm{CLS}}$, either the self-training scheme~\cite{döbler2023robust} or the regular cross-entropy~\cite{NIPS2004_96f2b50b} is adopted. With $\mathcal{R}(\theta)$, cosine similarity or L2 distance are both valid metrics for measuring the distance between $\theta$ and $\theta_0$ in the parameter space. Fisher regularizer coefficient~\cite{niu2022efficient, kirkpatrick2017_fisher} can also be used, optionally.
To sum up, the teacher model update of \method is an \textit{elaborated version} of EMA with $\lambda_t, \alpha_t$ (Eq.~\ref{eq:adaptive_lambda_alpha}) and $\mathcal{L}_{\mathrm{AL}}$ (Eq.~\ref{eq:anchor_loss}):%
\begin{align*}
    \nonumber
    \theta_t'&= \underset{\theta' \in \Theta}{\text{\texttt{Optim} }}  \mathbb{E}_{P_t}\left[\mathcal{L}_{\mathrm{CLS}} \left(\hat{Y}_t, X_t; \theta'\right) + \mathcal{L}_{\mathrm{AL}} \left(X_t; \theta'\right) \right] + \lambda_t \mathcal{R}(\theta'), \\
    \theta_t &= (1-\alpha_t) \theta_{t-1}  + \alpha_t \theta'_t.
\end{align*}

\section{Experimental Results}
\newsavebox{\figegmmca}
\sbox{\figegmmca}{%
    \begin{tikzpicture}

\definecolor{darkgray176}{RGB}{176,176,176}
\definecolor{lightgray204}{RGB}{204,204,204}
\tikzstyle{every node}=[font=\footnotesize]
\begin{axis}[
height=4.2cm, width = 4.6cm,
tick align=inside,
tick pos=left,
x grid style={darkgray176},
xmin=-0.5, xmax=30,
xtick style={color=black},
y grid style={darkgray176},
ymajorgrids,
ymin=0, ymax=1,
xmin=-0.5, xmax=29.5,
xtick={0,5,11,17,23,29},
xticklabels={0,120,240,360,480, 600},
ytick style={color=black},
xlabel={Testing Step ($t$)},
every axis x label/.style={
    at={(axis description cs:0.5,-0.25)},
    align = center},
every axis y label/.style={
    rotate=90,
    at={(axis description cs:-3,0.5)},
    align = center},
outer sep=0pt,
]
\draw[draw=none,fill=green01270] (axis cs:-0.4,0) rectangle (axis cs:0.4,0.509);
\draw[draw=none,fill=red] (axis cs:-0.4,0.509) rectangle (axis cs:0.4,1);
\draw[draw=none,fill=green01270] (axis cs:0.6,0) rectangle (axis cs:1.4,0.494);
\draw[draw=none,fill=red] (axis cs:0.6,0.494) rectangle (axis cs:1.4,1);
\draw[draw=none,fill=green01270] (axis cs:1.6,0) rectangle (axis cs:2.4,0.4635);
\draw[draw=none,fill=red] (axis cs:1.6,0.4635) rectangle (axis cs:2.4,1);
\draw[draw=none,fill=green01270] (axis cs:2.6,0) rectangle (axis cs:3.4,0.474);
\draw[draw=none,fill=red] (axis cs:2.6,0.474) rectangle (axis cs:3.4,1);
\draw[draw=none,fill=green01270] (axis cs:3.6,0) rectangle (axis cs:4.4,0.458);
\draw[draw=none,fill=red] (axis cs:3.6,0.458) rectangle (axis cs:4.4,1);
\draw[draw=none,fill=green01270] (axis cs:4.6,0) rectangle (axis cs:5.4,0.484);
\draw[draw=none,fill=red] (axis cs:4.6,0.484) rectangle (axis cs:5.4,1);
\draw[draw=none,fill=green01270] (axis cs:5.6,0) rectangle (axis cs:6.4,0.476);
\draw[draw=none,fill=red] (axis cs:5.6,0.476) rectangle (axis cs:6.4,1);
\draw[draw=none,fill=green01270] (axis cs:6.6,0) rectangle (axis cs:7.4,0.476);
\draw[draw=none,fill=red] (axis cs:6.6,0.476) rectangle (axis cs:7.4,1);
\draw[draw=none,fill=green01270] (axis cs:7.6,0) rectangle (axis cs:8.4,0.4675);
\draw[draw=none,fill=red] (axis cs:7.6,0.4675) rectangle (axis cs:8.4,1);
\draw[draw=none,fill=green01270] (axis cs:8.6,0) rectangle (axis cs:9.4,0.474);
\draw[draw=none,fill=red] (axis cs:8.6,0.474) rectangle (axis cs:9.4,1);
\draw[draw=none,fill=green01270] (axis cs:9.6,0) rectangle (axis cs:10.4,0.4835);
\draw[draw=none,fill=red] (axis cs:9.6,0.4835) rectangle (axis cs:10.4,1);
\draw[draw=none,fill=green01270] (axis cs:10.6,0) rectangle (axis cs:11.4,0.4635);
\draw[draw=none,fill=red] (axis cs:10.6,0.4635) rectangle (axis cs:11.4,1);
\draw[draw=none,fill=green01270] (axis cs:11.6,0) rectangle (axis cs:12.4,0.466);
\draw[draw=none,fill=red] (axis cs:11.6,0.466) rectangle (axis cs:12.4,1);
\draw[draw=none,fill=green01270] (axis cs:12.6,0) rectangle (axis cs:13.4,0.4505);
\draw[draw=none,fill=red] (axis cs:12.6,0.4505) rectangle (axis cs:13.4,1);
\draw[draw=none,fill=green01270] (axis cs:13.6,0) rectangle (axis cs:14.4,0.424);
\draw[draw=none,fill=red] (axis cs:13.6,0.424) rectangle (axis cs:14.4,1);
\draw[draw=none,fill=green01270] (axis cs:14.6,0) rectangle (axis cs:15.4,0.466);
\draw[draw=none,fill=red] (axis cs:14.6,0.466) rectangle (axis cs:15.4,1);
\draw[draw=none,fill=green01270] (axis cs:15.6,0) rectangle (axis cs:16.4,0.4715);
\draw[draw=none,fill=red] (axis cs:15.6,0.4715) rectangle (axis cs:16.4,1);
\draw[draw=none,fill=green01270] (axis cs:16.6,0) rectangle (axis cs:17.4,0.4725);
\draw[draw=none,fill=red] (axis cs:16.6,0.4725) rectangle (axis cs:17.4,1);
\draw[draw=none,fill=green01270] (axis cs:17.6,0) rectangle (axis cs:18.4,0.4565);
\draw[draw=none,fill=red] (axis cs:17.6,0.4565) rectangle (axis cs:18.4,1);
\draw[draw=none,fill=green01270] (axis cs:18.6,0) rectangle (axis cs:19.4,0.439);
\draw[draw=none,fill=red] (axis cs:18.6,0.439) rectangle (axis cs:19.4,1);
\draw[draw=none,fill=green01270] (axis cs:19.6,0) rectangle (axis cs:20.4,0.4435);
\draw[draw=none,fill=red] (axis cs:19.6,0.4435) rectangle (axis cs:20.4,1);
\draw[draw=none,fill=green01270] (axis cs:20.6,0) rectangle (axis cs:21.4,0.454);
\draw[draw=none,fill=red] (axis cs:20.6,0.454) rectangle (axis cs:21.4,1);
\draw[draw=none,fill=green01270] (axis cs:21.6,0) rectangle (axis cs:22.4,0.474);
\draw[draw=none,fill=red] (axis cs:21.6,0.474) rectangle (axis cs:22.4,1);
\draw[draw=none,fill=green01270] (axis cs:22.6,0) rectangle (axis cs:23.4,0.489);
\draw[draw=none,fill=red] (axis cs:22.6,0.489) rectangle (axis cs:23.4,1);
\draw[draw=none,fill=green01270] (axis cs:23.6,0) rectangle (axis cs:24.4,0.4735);
\draw[draw=none,fill=red] (axis cs:23.6,0.4735) rectangle (axis cs:24.4,1);
\draw[draw=none,fill=green01270] (axis cs:24.6,0) rectangle (axis cs:25.4,0.484);
\draw[draw=none,fill=red] (axis cs:24.6,0.484) rectangle (axis cs:25.4,1);
\draw[draw=none,fill=green01270] (axis cs:25.6,0) rectangle (axis cs:26.4,0.5025);
\draw[draw=none,fill=red] (axis cs:25.6,0.5025) rectangle (axis cs:26.4,1);
\draw[draw=none,fill=green01270] (axis cs:26.6,0) rectangle (axis cs:27.4,0.5);
\draw[draw=none,fill=red] (axis cs:26.6,0.5) rectangle (axis cs:27.4,1);
\draw[draw=none,fill=green01270] (axis cs:27.6,0) rectangle (axis cs:28.4,0.4995);
\draw[draw=none,fill=red] (axis cs:27.6,0.4995) rectangle (axis cs:28.4,1);
\draw[draw=none,fill=green01270] (axis cs:28.6,0) rectangle (axis cs:29.4,0.5015);
\draw[draw=none,fill=red] (axis cs:28.6,0.5015) rectangle (axis cs:29.4,1);
\end{axis}

\end{tikzpicture}
 }
 
 \newsavebox{\figegmmcb}
 \sbox{\figegmmcb}{%
    \begin{tikzpicture}

\definecolor{darkgray176}{RGB}{176,176,176}
\definecolor{lightgray204}{RGB}{204,204,204}
\tikzstyle{every node}=[font=\footnotesize]
\begin{axis}[
height=4.2cm, width = 4.6cm,
tick align=inside,
tick pos=left,
x grid style={darkgray176},
xtick style={color=black},
xmin=-0.5, xmax=29.5,
xtick={0,5,11,17,23,29},
xticklabels={0,120,240,360,480, 600},
y grid style={darkgray176},
ymajorgrids,
ymin=0, ymax=1,
ytick style={color=black},
xlabel={Testing Step ($t$)},
every axis x label/.style={
    at={(axis description cs:0.5,-0.25)},
    align = center},
every axis y label/.style={
    rotate=90,
    at={(axis description cs:-0.18,0.5)},
    align = center},
outer sep=0pt,
]
\draw[draw=none,fill=green01270] (axis cs:-0.4,0) rectangle (axis cs:0.4,0.509);
\draw[draw=none,fill=red] (axis cs:-0.4,0.509) rectangle (axis cs:0.4,1);
\draw[draw=none,fill=green01270] (axis cs:0.6,0) rectangle (axis cs:1.4,0.5405);
\draw[draw=none,fill=red] (axis cs:0.6,0.5405) rectangle (axis cs:1.4,1);
\draw[draw=none,fill=green01270] (axis cs:1.6,0) rectangle (axis cs:2.4,0.5915);
\draw[draw=none,fill=red] (axis cs:1.6,0.5915) rectangle (axis cs:2.4,1);
\draw[draw=none,fill=green01270] (axis cs:2.6,0) rectangle (axis cs:3.4,0.6265);
\draw[draw=none,fill=red] (axis cs:2.6,0.6265) rectangle (axis cs:3.4,1);
\draw[draw=none,fill=green01270] (axis cs:3.6,0) rectangle (axis cs:4.4,0.672);
\draw[draw=none,fill=red] (axis cs:3.6,0.672) rectangle (axis cs:4.4,1);
\draw[draw=none,fill=green01270] (axis cs:4.6,0) rectangle (axis cs:5.4,0.769);
\draw[draw=none,fill=red] (axis cs:4.6,0.769) rectangle (axis cs:5.4,1);
\draw[draw=none,fill=green01270] (axis cs:5.6,0) rectangle (axis cs:6.4,0.851);
\draw[draw=none,fill=red] (axis cs:5.6,0.851) rectangle (axis cs:6.4,1);
\draw[draw=none,fill=green01270] (axis cs:6.6,0) rectangle (axis cs:7.4,0.8675);
\draw[draw=none,fill=red] (axis cs:6.6,0.8675) rectangle (axis cs:7.4,1);
\draw[draw=none,fill=green01270] (axis cs:7.6,0) rectangle (axis cs:8.4,0.9045);
\draw[draw=none,fill=red] (axis cs:7.6,0.9045) rectangle (axis cs:8.4,1);
\draw[draw=none,fill=green01270] (axis cs:8.6,0) rectangle (axis cs:9.4,0.929);
\draw[draw=none,fill=red] (axis cs:8.6,0.929) rectangle (axis cs:9.4,1);
\draw[draw=none,fill=green01270] (axis cs:9.6,0) rectangle (axis cs:10.4,0.934);
\draw[draw=none,fill=red] (axis cs:9.6,0.934) rectangle (axis cs:10.4,1);
\draw[draw=none,fill=green01270] (axis cs:10.6,0) rectangle (axis cs:11.4,0.941);
\draw[draw=none,fill=red] (axis cs:10.6,0.941) rectangle (axis cs:11.4,1);
\draw[draw=none,fill=green01270] (axis cs:11.6,0) rectangle (axis cs:12.4,0.949);
\draw[draw=none,fill=red] (axis cs:11.6,0.949) rectangle (axis cs:12.4,1);
\draw[draw=none,fill=green01270] (axis cs:12.6,0) rectangle (axis cs:13.4,0.9505);
\draw[draw=none,fill=red] (axis cs:12.6,0.9505) rectangle (axis cs:13.4,1);
\draw[draw=none,fill=green01270] (axis cs:13.6,0) rectangle (axis cs:14.4,0.9605);
\draw[draw=none,fill=red] (axis cs:13.6,0.9605) rectangle (axis cs:14.4,1);
\draw[draw=none,fill=green01270] (axis cs:14.6,0) rectangle (axis cs:15.4,0.962);
\draw[draw=none,fill=red] (axis cs:14.6,0.962) rectangle (axis cs:15.4,1);
\draw[draw=none,fill=green01270] (axis cs:15.6,0) rectangle (axis cs:16.4,0.966);
\draw[draw=none,fill=red] (axis cs:15.6,0.966) rectangle (axis cs:16.4,1);
\draw[draw=none,fill=green01270] (axis cs:16.6,0) rectangle (axis cs:17.4,0.9685);
\draw[draw=none,fill=red] (axis cs:16.6,0.9685) rectangle (axis cs:17.4,1);
\draw[draw=none,fill=green01270] (axis cs:17.6,0) rectangle (axis cs:18.4,0.9685);
\draw[draw=none,fill=red] (axis cs:17.6,0.9685) rectangle (axis cs:18.4,1);
\draw[draw=none,fill=green01270] (axis cs:18.6,0) rectangle (axis cs:19.4,0.9685);
\draw[draw=none,fill=red] (axis cs:18.6,0.9685) rectangle (axis cs:19.4,1);
\draw[draw=none,fill=green01270] (axis cs:19.6,0) rectangle (axis cs:20.4,0.9685);
\draw[draw=none,fill=red] (axis cs:19.6,0.9685) rectangle (axis cs:20.4,1);
\draw[draw=none,fill=green01270] (axis cs:20.6,0) rectangle (axis cs:21.4,0.9705);
\draw[draw=none,fill=red] (axis cs:20.6,0.9705) rectangle (axis cs:21.4,1);
\draw[draw=none,fill=green01270] (axis cs:21.6,0) rectangle (axis cs:22.4,0.974);
\draw[draw=none,fill=red] (axis cs:21.6,0.974) rectangle (axis cs:22.4,1);
\draw[draw=none,fill=green01270] (axis cs:22.6,0) rectangle (axis cs:23.4,0.9745);
\draw[draw=none,fill=red] (axis cs:22.6,0.9745) rectangle (axis cs:23.4,1);
\draw[draw=none,fill=green01270] (axis cs:23.6,0) rectangle (axis cs:24.4,0.9745);
\draw[draw=none,fill=red] (axis cs:23.6,0.9745) rectangle (axis cs:24.4,1);
\draw[draw=none,fill=green01270] (axis cs:24.6,0) rectangle (axis cs:25.4,0.9795);
\draw[draw=none,fill=red] (axis cs:24.6,0.9795) rectangle (axis cs:25.4,1);
\draw[draw=none,fill=green01270] (axis cs:25.6,0) rectangle (axis cs:26.4,0.9805);
\draw[draw=none,fill=red] (axis cs:25.6,0.9805) rectangle (axis cs:26.4,1);
\draw[draw=none,fill=green01270] (axis cs:26.6,0) rectangle (axis cs:27.4,0.9805);
\draw[draw=none,fill=red] (axis cs:26.6,0.9805) rectangle (axis cs:27.4,1);
\draw[draw=none,fill=green01270] (axis cs:27.6,0) rectangle (axis cs:28.4,0.983);
\draw[draw=none,fill=red] (axis cs:27.6,0.983) rectangle (axis cs:28.4,1);
\draw[draw=none,fill=green01270] (axis cs:28.6,0) rectangle (axis cs:29.4,0.9825);
\draw[draw=none,fill=red] (axis cs:28.6,0.9825) rectangle (axis cs:29.4,1);
\end{axis}

\end{tikzpicture}
 }

\newsavebox{\cvgdista}
\sbox{\cvgdista}{%
    \begin{tikzpicture}

\definecolor{darkgray176}{RGB}{176,176,176}
\definecolor{lightgray204}{RGB}{204,204,204}
 \tikzstyle{every node}=[font=\footnotesize]
\begin{axis}[
height=3.7 cm, width = 4.0cm,
tick align=inside,
tick pos=left,
x grid style={darkgray176},
xlabel={$x$},
xmajorgrids,
xmin=-4.1, xmax=4.1,
xtick style={color=black},
y grid style={darkgray176},
yticklabels={,,},
ymajorgrids,
ymin=-0.01, ymax=1,
ytick style={color=black},
every axis y label/.style={
    rotate=90,
    at={(axis description cs:-0.25,0.5)},
    align = center},
every axis x label/.style={
    at={(axis description cs:0.5,-0.22)},
    align = center},
]
\addplot [very thick, green01270]
table {%
-5 8.15633436295066e-07
-4.77551020408163 2.59883160701895e-06
-4.55102040816327 7.84900457690352e-06
-4.3265306122449 2.24700659965526e-05
-4.10204081632653 6.09743861728494e-05
-3.87755102040816 0.000156835310950634
-3.6530612244898 0.000382378605476483
-3.42857142857143 0.000883683292550308
-3.20408163265306 0.00193576675246321
-2.97959183673469 0.00401941391915387
-2.75510204081633 0.00791089756768381
-2.53061224489796 0.0147584963969908
-2.30612244897959 0.0260982741787718
-2.08163265306122 0.0437456413894904
-1.85714285714286 0.0695042099156696
-1.63265306122449 0.104674466643797
-1.40816326530612 0.149425158715459
-1.18367346938776 0.20219016108941
-0.959183673469388 0.259328133409745
-0.73469387755102 0.315277206083725
-0.510204081632653 0.363319607013306
-0.285714285714286 0.396860999303723
-0.0612244897959187 0.410904910022184
0.163265306122449 0.403271530369905
0.387755102040816 0.375151873134421
0.612244897959184 0.330803425796379
0.836734693877551 0.276494311702548
1.06122448979592 0.219056289434707
1.28571428571429 0.164504800634671
1.51020408163265 0.117099421297099
1.73469387755102 0.0790103971400655
1.95918367346939 0.050532067087173
2.18367346938776 0.0306339646368153
2.40816326530612 0.0176032425349584
2.63265306122449 0.00958816418119304
2.85714285714286 0.00495030063426061
3.08163265306122 0.0024225960828854
3.30612244897959 0.0011237863949466
3.53061224489796 0.000494128442185127
3.75510204081633 0.000205944054579125
3.97959183673469 8.13601963787785e-05
4.20408163265306 3.04668841313527e-05
4.42857142857143 1.08142751936968e-05
4.6530612244898 3.63848104201061e-06
4.87755102040816 1.16036905703626e-06
5.10204081632653 3.50772457395346e-07
5.3265306122449 1.00509741202654e-07
5.55102040816327 2.72988371360499e-08
5.77551020408163 7.02802768632397e-09
6 1.71504730415169e-09
};
\addplot [very thick, red]
table {%
-5 9.09020063059665e-12
-4.77551020408163 4.27121744839922e-11
-4.55102040816327 1.908174003721e-10
-4.3265306122449 8.10536071873506e-10
-4.10204081632653 3.27351849788691e-09
-3.87755102040816 1.25702909801386e-08
-3.6530612244898 4.5894848662881e-08
-3.42857142857143 1.59320122936981e-07
-3.20408163265306 5.2585426801997e-07
-2.97959183673469 1.65024434435306e-06
-2.75510204081633 4.92401290500621e-06
-2.53061224489796 1.39694141002515e-05
-2.30612244897959 3.76812481656073e-05
-2.08163265306122 9.66407848246271e-05
-1.85714285714286 0.000235658818483301
-1.63265306122449 0.000546380288635655
-1.40816326530612 0.00120446568097066
-1.18367346938776 0.00252453802514588
-0.959183673469388 0.00503103666669728
-0.73469387755102 0.00953281419189513
-0.510204081632653 0.0171740551514851
-0.285714285714286 0.0294179671002189
-0.0612244897959187 0.0479115854286877
0.163265306122449 0.0741919021991198
0.387755102040816 0.109234682191159
0.612244897959184 0.152915909617934
0.836734693877551 0.203532081928443
1.06122448979592 0.257573486825526
1.28571428571429 0.309925648661676
1.51020408163265 0.354569977739159
1.73469387755102 0.385686538370638
1.95918367346939 0.398891774934659
2.18367346938776 0.392250730730017
2.40816326530612 0.366741891747304
2.63265306122449 0.326020838434264
2.85714285714286 0.275561334085298
3.08163265306122 0.221451831739017
3.30612244897959 0.169210907528505
3.53061224489796 0.122932159881883
3.75510204081633 0.0849162386800441
3.97959183673469 0.0557704393246208
4.20408163265306 0.0348261496361587
4.42857142857143 0.0206773527792518
4.6530612244898 0.0116727314128747
4.87755102040816 0.00626524599309867
5.10204081632653 0.00319736224034227
5.3265306122449 0.00155143518200395
5.55102040816327 0.000715753547605386
5.77551020408163 0.000313965163089874
6 0.000130944570350831
};
\addplot [very thick, green01270, dashed]
table {%
-5 5.5327148846111e-13
-4.77551020408163 6.60132548181025e-12
-4.55102040816327 7.01289328037777e-11
-4.3265306122449 6.6334054203146e-10
-4.10204081632653 5.58661933720613e-09
-3.87755102040816 4.18923619245184e-08
-3.6530612244898 2.7970088545788e-07
-3.42857142857143 1.66274657995718e-06
-3.20408163265306 8.80098970320818e-06
-2.97959183673469 4.14772725282993e-05
-2.75510204081633 0.000174045217983662
-2.53061224489796 0.000650260281814441
-2.30612244897959 0.00216314478195185
-2.08163265306122 0.00640703617833715
-1.85714285714286 0.0168967038419461
-1.63265306122449 0.0396752831991141
-1.40816326530612 0.0829490289822978
-1.18367346938776 0.154410135892568
-0.959183673469388 0.255925482996946
-0.73469387755102 0.377680433288023
-0.510204081632653 0.496259310052655
-0.285714285714286 0.580585355109475
-0.0612244897959187 0.604779015837016
0.163265306122449 0.56091955227078
0.387755102040816 0.463209727568354
0.612244897959184 0.340586940467778
0.836734693877551 0.222972621071211
1.06122448979592 0.129971554880291
1.28571428571429 0.0674556478548901
1.51020408163265 0.0311717761256198
1.73469387755102 0.0128256093263552
1.95918367346939 0.00469859154822405
2.18367346938776 0.00153260630187301
2.40816326530612 0.000445109382076139
2.63265306122449 0.000115100210446147
2.85714285714286 2.65007835261521e-05
3.08163265306122 5.43268518438213e-06
3.30612244897959 9.91616149941216e-07
3.53061224489796 1.61155774310427e-07
3.75510204081633 2.33196151422415e-08
3.97959183673469 3.00448551073718e-09
4.20408163265306 3.44660946916966e-10
4.42857142857143 3.52036153804114e-11
4.6530612244898 3.20151606687853e-12
4.87755102040816 2.59237205718625e-13
5.10204081632653 1.869012496755e-14
5.3265306122449 1.19977637654714e-15
5.55102040816327 6.85743398478874e-17
5.77551020408163 3.4897650755134e-18
6 1.58126251703247e-19
};
\addplot [very thick, red, dashed]
table {%
-5 1.30466783095434e-25
-4.77551020408163 4.25738517089374e-24
-4.55102040816327 1.24405726632234e-22
-4.3265306122449 3.25530916921187e-21
-4.10204081632653 7.62778688915025e-20
-3.87755102040816 1.60051334814965e-18
-3.6530612244898 3.00728432230015e-17
-3.42857142857143 5.05992547169239e-16
-3.20408163265306 7.62374199160208e-15
-2.97959183673469 1.02860042723133e-13
-2.75510204081633 1.24273820745475e-12
-2.53061224489796 1.34451930170671e-11
-2.30612244897959 1.30259339579813e-10
-2.08163265306122 1.13006947164391e-09
-1.85714285714286 8.77921841468324e-09
-1.63265306122449 6.10746525474884e-08
-1.40816326530612 3.80470155828943e-07
-1.18367346938776 2.12243611910342e-06
-0.959183673469388 1.06023734980876e-05
-0.73469387755102 4.74270389869378e-05
-0.510204081632653 0.000189978022669943
-0.285714285714286 0.000681451819174608
-0.0612244897959187 0.00218887714947111
0.163265306122449 0.00629596144644235
0.387755102040816 0.0162165017221752
0.612244897959184 0.0374030248043779
0.836734693877551 0.0772521754510826
1.06122448979592 0.142879246503335
1.28571428571429 0.236636658684074
1.51020408163265 0.350953305980835
1.73469387755102 0.466091401776202
1.95918367346939 0.554303071424084
2.18367346938776 0.590306980537449
2.40816326530612 0.562941146022307
2.63265306122449 0.480731418980677
2.85714285714286 0.36761774049535
3.08163265306122 0.251735726849816
3.30612244897959 0.154364592408525
3.53061224489796 0.0847627357577554
3.75510204081633 0.0416789414444585
3.97959183673469 0.0183519761182137
4.20408163265306 0.00723608118345398
4.42857142857143 0.00255492682486428
4.6530612244898 0.000807807612883965
4.87755102040816 0.000228713530282849
5.10204081632653 5.79869429299279e-05
5.3265306122449 1.31650650524656e-05
5.55102040816327 2.67651847590903e-06
5.77551020408163 4.87272522506477e-07
6 7.94379519162247e-08
};
\end{axis}

\end{tikzpicture}
 }
 
 \newsavebox{\cvgdistb}
 \sbox{\cvgdistb}{%
    \begin{tikzpicture}

\definecolor{darkgray176}{RGB}{176,176,176}
\definecolor{lightgray204}{RGB}{204,204,204}
\tikzstyle{every node}=[font=\footnotesize]
\begin{axis}[
legend cell align={left},
legend style={
  fill=white,
  fill opacity=0,
  text opacity=1,
  at={(0.0,-0.70)},
  anchor=south west,
  draw=none,
  column sep = 2,
  legend columns=2,
  font=\footnotesize
},
height=3.7cm, width = 4.0cm,
tick align=inside,
tick pos=left,
x grid style={darkgray176},
xlabel={$ x $},
xmajorgrids,
xmin=-4.1, xmax=4.1,
xtick style={color=black},
y grid style={darkgray176},
ylabel={Probability density},
ymajorgrids,
ymin=-0.01, ymax=1,
ytick style={color=black},
every axis y label/.style={
    rotate=90,
    at={(axis description cs:-0.3,0.5)},
    align = center},
every axis x label/.style={
    at={(axis description cs:0.5,-0.22)},
    align = center},
]
\addplot [very thick, green01270]
table {%
-5 8.15633436295066e-07
-4.77551020408163 2.59883160701895e-06
-4.55102040816327 7.84900457690352e-06
-4.3265306122449 2.24700659965526e-05
-4.10204081632653 6.09743861728494e-05
-3.87755102040816 0.000156835310950634
-3.6530612244898 0.000382378605476483
-3.42857142857143 0.000883683292550308
-3.20408163265306 0.00193576675246321
-2.97959183673469 0.00401941391915387
-2.75510204081633 0.00791089756768381
-2.53061224489796 0.0147584963969908
-2.30612244897959 0.0260982741787718
-2.08163265306122 0.0437456413894904
-1.85714285714286 0.0695042099156696
-1.63265306122449 0.104674466643797
-1.40816326530612 0.149425158715459
-1.18367346938776 0.20219016108941
-0.959183673469388 0.259328133409745
-0.73469387755102 0.315277206083725
-0.510204081632653 0.363319607013306
-0.285714285714286 0.396860999303723
-0.0612244897959187 0.410904910022184
0.163265306122449 0.403271530369905
0.387755102040816 0.375151873134421
0.612244897959184 0.330803425796379
0.836734693877551 0.276494311702548
1.06122448979592 0.219056289434707
1.28571428571429 0.164504800634671
1.51020408163265 0.117099421297099
1.73469387755102 0.0790103971400655
1.95918367346939 0.050532067087173
2.18367346938776 0.0306339646368153
2.40816326530612 0.0176032425349584
2.63265306122449 0.00958816418119304
2.85714285714286 0.00495030063426061
3.08163265306122 0.0024225960828854
3.30612244897959 0.0011237863949466
3.53061224489796 0.000494128442185127
3.75510204081633 0.000205944054579125
3.97959183673469 8.13601963787785e-05
4.20408163265306 3.04668841313527e-05
4.42857142857143 1.08142751936968e-05
4.6530612244898 3.63848104201061e-06
4.87755102040816 1.16036905703626e-06
5.10204081632653 3.50772457395346e-07
5.3265306122449 1.00509741202654e-07
5.55102040816327 2.72988371360499e-08
5.77551020408163 7.02802768632397e-09
6 1.71504730415169e-09
};
\addlegendentry{$\mathcal{N}(\mu_0, \sigma_0)$}
\addplot [very thick, red]
table {%
-5 9.09020063059665e-12
-4.77551020408163 4.27121744839922e-11
-4.55102040816327 1.908174003721e-10
-4.3265306122449 8.10536071873506e-10
-4.10204081632653 3.27351849788691e-09
-3.87755102040816 1.25702909801386e-08
-3.6530612244898 4.5894848662881e-08
-3.42857142857143 1.59320122936981e-07
-3.20408163265306 5.2585426801997e-07
-2.97959183673469 1.65024434435306e-06
-2.75510204081633 4.92401290500621e-06
-2.53061224489796 1.39694141002515e-05
-2.30612244897959 3.76812481656073e-05
-2.08163265306122 9.66407848246271e-05
-1.85714285714286 0.000235658818483301
-1.63265306122449 0.000546380288635655
-1.40816326530612 0.00120446568097066
-1.18367346938776 0.00252453802514588
-0.959183673469388 0.00503103666669728
-0.73469387755102 0.00953281419189513
-0.510204081632653 0.0171740551514851
-0.285714285714286 0.0294179671002189
-0.0612244897959187 0.0479115854286877
0.163265306122449 0.0741919021991198
0.387755102040816 0.109234682191159
0.612244897959184 0.152915909617934
0.836734693877551 0.203532081928443
1.06122448979592 0.257573486825526
1.28571428571429 0.309925648661676
1.51020408163265 0.354569977739159
1.73469387755102 0.385686538370638
1.95918367346939 0.398891774934659
2.18367346938776 0.392250730730017
2.40816326530612 0.366741891747304
2.63265306122449 0.326020838434264
2.85714285714286 0.275561334085298
3.08163265306122 0.221451831739017
3.30612244897959 0.169210907528505
3.53061224489796 0.122932159881883
3.75510204081633 0.0849162386800441
3.97959183673469 0.0557704393246208
4.20408163265306 0.0348261496361587
4.42857142857143 0.0206773527792518
4.6530612244898 0.0116727314128747
4.87755102040816 0.00626524599309867
5.10204081632653 0.00319736224034227
5.3265306122449 0.00155143518200395
5.55102040816327 0.000715753547605386
5.77551020408163 0.000313965163089874
6 0.000130944570350831
};
\addlegendentry{$\mathcal{N}(\mu_1, \sigma_1)$}
\addplot [very thick, green01270, dashed]
table {%
-5 5.56035921031721e-06
-4.77551020408163 1.26616499659821e-05
-4.55102040816327 2.79207154754873e-05
-4.3265306122449 5.96226902738403e-05
-4.10204081632653 0.000123294975556063
-3.87755102040816 0.000246903914177656
-3.6530612244898 0.000478805748987857
-3.42857142857143 0.000899165268139344
-3.20408163265306 0.00163519095206425
-2.97959183673469 0.00287969314815056
-2.75510204081633 0.00491103168482386
-2.53061224489796 0.00811050776770215
-2.30612244897959 0.0129709605491148
-2.08163265306122 0.0200883847475393
-1.85714285714286 0.0301277488539034
-1.63265306122449 0.0437559515198429
-1.40816326530612 0.0615398407361249
-1.18367346938776 0.0838155094044663
-0.959183673469388 0.110545532278701
-0.73469387755102 0.141190926103276
-0.510204081632653 0.174630928846576
-0.285714285714286 0.209162741982648
-0.0612244897959187 0.242603060011211
0.163265306122449 0.27249403996532
0.387755102040816 0.296392033570453
0.612244897959184 0.312194200165602
0.836734693877551 0.318443156541352
1.06122448979592 0.314548627935268
1.28571428571429 0.300879400775545
1.51020408163265 0.2787057323377
1.73469387755102 0.250004676445725
1.95918367346939 0.217169658784262
2.18367346938776 0.182683343260687
2.40816326530612 0.148815284291465
2.63265306122449 0.117393732508741
2.85714285714286 0.0896790645269099
3.08163265306122 0.0663416109202129
3.30612244897959 0.0475258334732769
3.53061224489796 0.0329702512238332
3.75510204081633 0.0221494810893544
3.97959183673469 0.0144096569395668
4.20408163265306 0.00907804877941022
4.42857142857143 0.00553834706984509
4.6530612244898 0.00327202497171307
4.87755102040816 0.00187198272357879
5.10204081632653 0.001037136332624
5.3265306122449 0.000556440389890644
5.55102040816327 0.000289101422981312
5.77551020408163 0.000145455650792985
6 7.08695624861426e-05
};
\addlegendentry{$\mathcal{N}(\hat\mu_{0}, \hat\sigma_{0})$}
\addplot [very thick, red, dashed]
table {%
-5 0
-4.77551020408163 0
-4.55102040816327 0
-4.3265306122449 0
-4.10204081632653 0
-3.87755102040816 0
-3.6530612244898 0
-3.42857142857143 0
-3.20408163265306 0
-2.97959183673469 0
-2.75510204081633 0
-2.53061224489796 0
-2.30612244897959 0
-2.08163265306122 0
-1.85714285714286 0
-1.63265306122449 0
-1.40816326530612 0
-1.18367346938776 0
-0.959183673469388 0
-0.73469387755102 0
-0.510204081632653 0
-0.285714285714286 0
-0.0612244897959187 0
0.163265306122449 0
0.387755102040816 0
0.612244897959184 0
0.836734693877551 0
1.06122448979592 0
1.28571428571429 0
1.51020408163265 0
1.73469387755102 7.41243574677536e-284
1.95918367346939 2.45853258397811e-122
2.18367346938776 1.38742777929253e-27
2.40816326530612 13.3218393894224
2.63265306122449 2.176393356402e-38
2.85714285714286 6.049646247589e-144
3.08163265306122 2.86115905597521e-316
3.30612244897959 0
3.53061224489796 0
3.75510204081633 0
3.97959183673469 0
4.20408163265306 0
4.42857142857143 0
4.6530612244898 0
4.87755102040816 0
5.10204081632653 0
5.3265306122449 0
5.55102040816327 0
5.77551020408163 0
6 0
};
\addlegendentry{$\mathcal{N}(\hat\mu_{1}, \hat\sigma_{1})$}
\end{axis}

\end{tikzpicture}
 }

\newsavebox{\cvgdist}
\sbox{\cvgdist}{%
    \input{figures/egmmc/dist_to_m1}
 }
 
 \newsavebox{\cvgeps}
 \sbox{\cvgeps}{%
    \input{figures/egmmc/eps_t}
 }

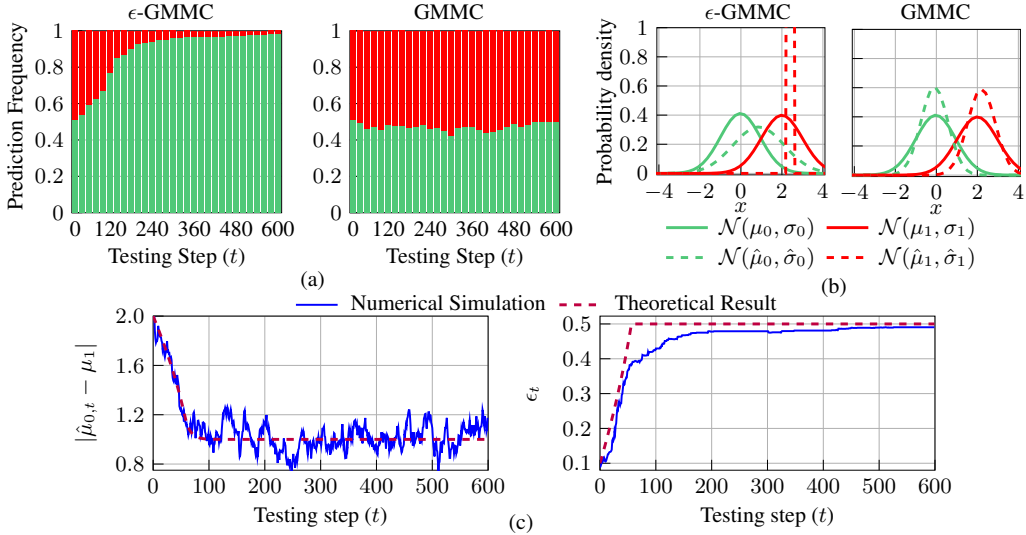
\begin{figure*}[t!]
    \pgfplotsset{every x tick label/.append style={font=\footnotesize, yshift=0.5ex}}
    \pgfplotsset{every y tick label/.append style={font=\footnotesize, xshift=0.5ex}}
    \centering    
    \resizebox{\textwidth}{!}{%
        \begin{tikzpicture}
        \draw (4.0, 0.5) node[inner sep=0] {\usebox \figegmmca};
        \draw (0.0, 0.5) node[inner sep=0] {\usebox \figegmmcb};
        
        \draw (9.5, -0.6) node[inner sep=0, anchor=south west] {\usebox \cvgdista};
        \draw (6.0, -1.4) node[inner sep=0, anchor=south west] {\usebox \cvgdistb};

        \draw (-1.5,-5.1) node[inner sep=0, anchor=south west] {\usebox \cvgdist};
        \draw (5, -5.1) node[inner sep=0, anchor=south west] {\usebox \cvgeps};
        \node at (-1.7, 1.7, 2) [below, rotate=90] {\footnotesize{Prediction Frequency}};
        \node at (4.0,   2.2 ) [above] {\footnotesize{GMMC}};
        \node at (0.0, 2.2) [above] {\footnotesize{$\epsilon$-GMMC}};

        \node at (8.3, 2.2) [above] {\footnotesize{$\epsilon$-GMMC}};
        \node at (11.0, 2.2) [above] {\footnotesize{GMMC}};
        
        \node at (2, -1.7) [above] {\small{ (a) }};
        \node at (9.5, -1.8) [above] {\small{ (b)}};
        \node at (5, -5.2) [above] {\small{ (c)}};
        \end{tikzpicture}
    }
    \vspace*{-1.0\baselineskip}
    \caption{Simulation result on $\epsilon$-perturbed Gaussian Mixture Model Classifier ($\epsilon$-GMMC) and GMMC (perturbed-free). 
    (a) Histogram of model predictions through time. A similar prediction frequency pattern is observed on CIFAR-10-C (Fig.~\ref{fig:cifar10-c-result}a-left). 
    (b) The probability density function of the two clusters after convergence versus the true data distribution. 
    The initial two clusters of $\epsilon$-GMMC collapsed into a single cluster with parameters stated in Lemma~\ref{lmm:collapsed}. In the perturbed-free, GMMC converges to the true data distribution. 
    (c) Distance toward $\mu_1$ ($\left|\E{\hat \mu_{0,t}} - \mu_1 \right|$) and false-negative rate ($\epsilon_t$) in simulation coincides with the result in Thm.~\ref{thm:cvg} (with $\epsilon_t$ following Corollary~\ref{corollary:condition}).
    \vspace*{-1.2\baselineskip}
    }
    \label{fig:egmmc_results}
\end{figure*}

\vspace*{-0.5\baselineskip}
\subsection{\texorpdfstring{$\epsilon-$}GMMC Simulation Result}
\vspace*{-0.5\baselineskip}
\label{sec:eps_gmmc_result}

\noindent \textbf{Simulation Setup. } A total of 6000 samples from two Gaussian distributions: $\mathcal{N}(\mu_0=0, \sigma_0^2 =1)$ and $\mathcal{N}(\mu_1=2, \sigma_1^2=1)$ with $p_0=p_1=\frac{1}{2}$ are synthesized and gradually released in a batch of $B=10$ samples. For evaluation, an independent set of 2000 samples following the same distribution is used for computing the prediction frequency, and the false negative rate (FNR). $\epsilon-$GMMC update follows Eq.~\ref{eq:update_gmmc} with $\alpha = 5e^{-2}$. To simulate model collapse, the predictor is intercepted and 10\% of the true-postive pseudo labels at each testing step are randomly flipped (Corollary~\ref{corollary:condition}).

\noindent \textbf{Simulation Result. } In action, both the likelihood of predicting class 0 (\Fig{\ref{fig:egmmc_results}}a-left) and the $\epsilon_t$ (\Eq{\ref{eq:fnr}}) (\Fig{\ref{fig:egmmc_results}}c-right, solid line) gradually increases over time as expected (Lemma~\ref{lmm:increasing_fnr}). After collapsing, $\epsilon$-GMMC merges the two initial clusters, resulting in a single one (\Fig{\ref{fig:egmmc_results}}b-left) with parameters that match Lemma~\ref{lmm:collapsed}. The distance from $\hat \mu_{0,t}$ (initialized at $\mu_0$) towards $\mu_1$ converges (\Fig{\ref{fig:egmmc_results}}c-left, solid line), coincided with the analysis in Thm.~\ref{thm:cvg} when $\epsilon_t$ is chosen following Corollary~\ref{corollary:condition} (\Fig{\ref{fig:egmmc_results}}c, dashed line).
GMMC (perturbed-free) stably produces accurate predictions (\Fig{\ref{fig:egmmc_results}}a-right) and approximates the true data distribution (\Fig{\ref{fig:egmmc_results}}b-right). The simulation empirically validates our analysis (Sec.~\ref{ssec:egmmc_analysis}), confirming the vulnerability of TTA models when the pseudo labels are inaccurately estimated.

\vspace*{-0.5\baselineskip}
\subsection{Setup - Benchmark Datasets}

\label{ssec:exp_setup}
\vspace*{-0.5\baselineskip}
\noindent \textbf{Datasets.}
We benchmark the performance on \textit{four} TTA classification tasks. Specifically, CIFAR10 $\rightarrow$ CIFAR10-C, CIFAR100 $\rightarrow$ CIFAR100-C, and ImageNet $\rightarrow$ ImageNet-C~\cite{hendrycks2019robustness} are three corrupted images classification tasks (corruption level 5, the most severe). Additionally, we incorporate DomainNet~\cite{peng2019moment} with 126 categories from four domains for the task \textit{real} $\rightarrow$ \textit{clipart, painting,} \textit{sketch}. %

\noindent \textbf{Compared Methods.} Besides \method,
the following algorithms are investigated: CoTTA~\cite{Wang_2022_CVPR}, EATA~\cite{niu2022efficient}, RMT \cite{döbler2023robust}, MECTA~\cite{hong2023mecta}, RoTTA~\cite{yuan2023robust}, ROID~\cite{marsden2024universal} and TRIBE~\cite{Su_Xu_Jia_2024}.
Noteworthy, only RoTTA is specifically designed for the practical TTA setting while others fit the continual TTA setting in general.
A parameter-free approach: LAME~\cite{malik2022_paramter-free} and a reset-based approach (i.e., reverting the model to the source model after adapting to every $1,000$ images): RDumb~\cite{press2023rdumb} are also included.

\noindent \textbf{\Setting TTA.} Following the practical TTA setup, multiple testing scenarios from each testing set will gradually change from one to another while the Dirichlet distribution ($\mathrm{Dir}(0.1)$ for CIFAR10-C, DomainNet, and ImageNet-C, and $\mathrm{Dir}(0.01)$ for CIFAR100-C) generates category temporally correlated batches of data. For all experiments, we set the number of revisits $K=20$ (times)  as this number is sufficient to fully observe the gradual degradation on existing TTA baselines.

\noindent \textbf{Implementation Details.} We use \texttt{PyTorch}~\cite{adam2019_pytorch} for implementation. \texttt{RobustBench}~\cite{croce2021robustbench} and \texttt{torchvision}~\cite{torchvision2016} provide pre-trained source models. Hyper-parameter choices are kept as close as possible to the original selections of authors. \changes{Visit \Sec{\ref{sec:experimental_details}} for more implementation details.}
Unless otherwise noted, for all \method experiments, the EMA update rate for robust batch normalization~\cite{yuan2023robust} and feature embedding statistics is set to $5e^{-2}$; $\alpha_0 = 1e^{-3}$ and cosine similarity regularizer is used. On CIFAR10/100-C  and ImageNet-C we use the self-training loss in~\cite{döbler2023robust} for $\mathcal{L}_{\mathrm{CLS}}$ and $\lambda_0=10$ while the regular cross-entropy loss~\cite{pmlr-v37-ganin15} and $\lambda_0=1$ (severe domain shift requires prioritizing adaptability) are applied in DomainNet experiments. 
\changes{In \Appdx{\ref{sec:petta_hyper_params}}, we provide a sensitivity analysis on the choice of hyper-parameter $\lambda_0$ in \method.}

\vspace*{-0.3\baselineskip}
\subsection{Result - Benchmark Datasets}
\vspace*{-0.2\baselineskip}
\label{ssec:result_real_data}

\noindent \textbf{\Setting TTA Performance. } \Fig{\ref{fig:fig1_introduction}}-right presents the testing error on CIFAR-10-C in \setting TTA setting. RoTTA~\cite{yuan2023robust} exhibits promising performance in the first several visits but soon raises and eventually exceeds the source model (no TTA). 
The classification error of compared methods on CIFAR-10$\rightarrow$CIFAR-10-C, and ImageNet $\rightarrow$ ImageNet-C~\cite{hendrycks2019robustness} tasks are shown in \Tab{\ref{tab:cifar-10-performance}}, and \Tab{\ref{tab:imagenet-c-performance}}. \Appdx{~\ref{appdx:additional_main_results}} provides the results on the other two datasets.
The observed performance degradation of CoTTA~\cite{Wang_2022_CVPR}, \changes{EATA~\cite{niu2022efficient}}, RoTTA~\cite{yuan2023robust}, \changes{and TRIBE~\cite{Su_Xu_Jia_2024}} \textit{confirms the risk of error accumulation} for an extensive period. While RMT~\cite{döbler2023robust}, MECTA~\cite{hong2023mecta}, \changes{and ROID~\cite{marsden2024universal}}  remain stable, they failed to adapt to the temporally correlated test stream at the beginning, with a higher error rate than the source model.
LAME~\cite{malik2022_paramter-free} (parameter-free TTA) and RDumb~\cite{press2023rdumb} (reset-based TTA) do not suffer from collapsing. However, their performance is lagging behind, and knowledge accumulation is limited in these approaches that could potentially favor a higher performance as achieved by \method.
\changes{Furthermore, LAME~\cite{malik2022_paramter-free} is highly constrained by the source model, and selecting a precise reset frequency in RDumb~\cite{press2023rdumb} is challenging in practice (see \Appdx{\ref{appdx:model_reset}} for a further discussion).}

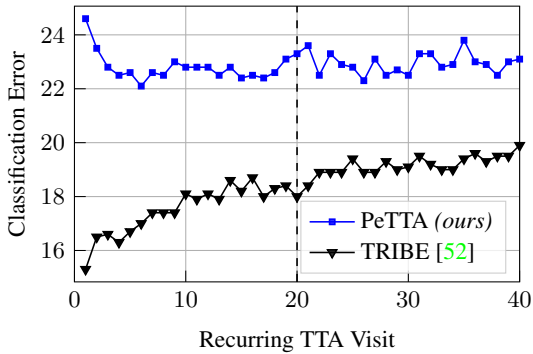
\begin{wrapfigure}[17]{L}{0.50\linewidth}  %
\vspace{-1.0\baselineskip}
\begin{minipage}{\linewidth} 
    \begin{tikzpicture}

\definecolor{darkgray176}{RGB}{176,176,176}
\definecolor{lightgray204}{RGB}{204,204,204}
\tikzstyle{every node}=[font=\footnotesize]
\begin{axis}[
legend cell align={left},
legend style={
  fill opacity=0.8,
  draw opacity=1,
  text opacity=1,
  at={(0.97,0.03)},
  anchor=south east,
  draw=lightgray204
},
tick align=outside,
tick pos=left,
x grid style={darkgray176},
xlabel={Recurring TTA Visit},
xmajorgrids,
height=5.25cm, width = 7.5 cm,
xmin=0, xmax=40.0,
xtick style={color=black},
y grid style={darkgray176},
ylabel={Classification Error},
ymajorgrids,
ymin=14.835, ymax=25.065,
ytick style={color=black},
xtick align=inside,
ytick align=inside,
every axis y label/.style={
    rotate=90,
    at={(axis description cs:-0.13,0.5)},
    align = center},
]
\addplot [semithick, blue, mark=square*, mark size=1, mark options={solid}]
table {%
1 24.6
2 23.5
3 22.8
4 22.5
5 22.6
6 22.1
7 22.6
8 22.5
9 23
10 22.8
11 22.8
12 22.8
13 22.5
14 22.8
15 22.4
16 22.5
17 22.4
18 22.6
19 23.1
20 23.3
21 23.6
22 22.5
23 23.3
24 22.9
25 22.8
26 22.3
27 23.1
28 22.5
29 22.7
30 22.5
31 23.3
32 23.3
33 22.8
34 22.9
35 23.8
36 23
37 22.9
38 22.5
39 23
40 23.1
};
\addlegendentry{\method \textit{(ours)}}
\addplot [semithick, black, mark=triangle*, mark size=2, mark options={solid,rotate=180}]
table {%
1 15.3
2 16.5
3 16.6
4 16.3
5 16.7
6 17
7 17.4
8 17.4
9 17.4
10 18.1
11 17.9
12 18.1
13 17.9
14 18.6
15 18.2
16 18.7
17 18
18 18.3
19 18.4
20 18
21 18.4
22 18.9
23 18.9
24 18.9
25 19.4
26 18.9
27 18.9
28 19.3
29 19
30 19.1
31 19.5
32 19.2
33 19
34 19
35 19.4
36 19.6
37 19.3
38 19.5
39 19.5
40 19.9
};
\addlegendentry{TRIBE~\cite{Su_Xu_Jia_2024}}
\addplot [semithick, black, dashed, forget plot]
table {%
20 14.835
20 25.065
};
\end{axis}

\end{tikzpicture}
    \vspace{-1.0\baselineskip}
    \caption{Classification error of TRIBE~\cite{Su_Xu_Jia_2024} and \method \textit{(ours)} of the task CIFAR-10$\rightarrow$CIFAR10-C task in \textit{\setting} TTA with 40 visits.}
    \label{fig:tribe_petta}
\end{minipage} %
\end{wrapfigure} %

In average, \method \textit{outperforms \changes{almost every} baseline approaches} 
and \textit{persists across 20 visits} over the three datasets. 
\changes{The only exception is at the case of TRIBE~\cite{Su_Xu_Jia_2024} on CIFAR-10-C. While this state-of-the-art model provides stronger adaptability, outweighing the \method, and baseline RoTTA~\cite{yuan2023robust} in several recurrences, the risk of the model collapsing still presents in TRIBE~\cite{Su_Xu_Jia_2024}.
This can be clearly observed when we increase the observation period to 40 recurring visits in \Fig{\ref{fig:tribe_petta}}}.  
As the degree of freedom for adaptation in \method is more constrained, it takes a bit longer for adaptation but remains stable afterward. \Fig{\ref{fig:cifar10-c-result}}b-bottom exhibits the confusion matrix at the last visit with satisfactory accuracy. \changes{The same results are also observed when shuffling the order of domain shifts within each recurrence (\Appdx{\ref{appdx:repeating_random_orders}}), or extending the number of recurrences to 40 visits (\Appdx{\ref{sec:results_recurring40}}).}

\begin{table*}[t!]
    \caption{Average classification error of the task CIFAR-10 $\rightarrow$ CIFAR-10-C in \textit{\setting TTA}. The lowest error is in \textbf{bold},$^{(*)}$average value across 5 runs (different random seeds) is reported for \method.}
    \label{tab:cifar-10-performance}
    \resizebox{\textwidth}{!}{
    \begin{tabular}{r|cccccccccccccccccccc|c}
\toprule
& \multicolumn{20}{l}{ \textit{\Setting TTA visit} $\xrightarrow{\hspace*{5cm}}$ } \\
\textbf{Method} &              1 &              2 &              3 &              4 &              5 &              6 &              7 &              8 &              9 &             10 &             11 &             12 &             13 &             14 &             15 &             16 &             17 &             18 &             19 &             20 &   \textbf{Avg} \\
\midrule
Source                                      & \multicolumn{20}{c|}{43.5}  & 43.5 \\
\midrule
LAME~\cite{malik2022_paramter-free}  & \multicolumn{20}{c|}{31.1} & 31.1 \\
\midrule
CoTTA~\cite{Wang_2022_CVPR} &           82.2 &           85.6 &           87.2 &           87.8 &           88.2 &           88.5 &           88.7 &           88.7 &           88.9 &           88.9 &           88.9 &           89.2 &           89.2 &           89.2 &           89.1 &           89.2 &           89.2 &           89.1 &           89.3 &           89.3 &           88.3 \\
EATA~\cite{niu2022efficient}  &  81.6 &  87.0 &  88.7 &  88.7 &  88.9 &  88.7 &  88.6 &  89.0 &  89.3 &  89.6 &  89.5 &  89.6 &  89.7 &  89.7 &  89.3 &  89.6 &  89.6 &  89.8 &  89.9 &  89.4 &          88.8 \\
RMT~\cite{döbler2023robust} &           77.5 &           76.9 &           76.5 &           75.8 &           75.5 &           75.5 &           75.4 &           75.4 &           75.5 &           75.3 &           75.5 &           75.6 &           75.5 &           75.5 &           75.7 &           75.6 &           75.7 &           75.6 &           75.7 &           75.8 &           75.8 \\
MECTA~\cite{hong2023mecta}  &           72.2 &           82.0 &           85.2 &           86.3 &           87.0 &           87.3 &           87.3 &           87.5 &           88.1 &           88.8 &           88.9 &           88.9 &           88.6 &           89.1 &           88.7 &           88.8 &           88.5 &           88.6 &           88.3 &           88.8 &           86.9 \\
RoTTA~\cite{yuan2023robust} &           24.6 &           25.5 &           29.6 &           33.6 &           38.2 &           42.8 &           46.2 &           50.6 &           52.2 &           54.1 &           56.5 &           57.5 &           59.4 &           60.2 &           61.7 &           63.0 &           64.8 &           66.1 &           68.2 &           70.3 &           51.3 \\
RDumb~\cite{press2023rdumb} &           31.1 &           32.1 &           32.3 &           31.6 &           31.9 &           31.8 &           31.8 &           31.9 &           31.9 &           32.1 &           31.7 &           32.0 &           32.5 &           32.0 &           31.9 &           31.6 &           31.9 &           31.4 &           32.3 &           32.4 &           31.9 \\
ROID~\cite{marsden2024universal}  &  72.7 &  72.6 &  73.1 &  72.4 &  72.7 &  72.8 &  72.7 &  72.7 &  72.9 &  72.8 &  72.9 &  72.9 &  72.8 &  72.5 &  73.0 &  72.8 &  72.5 &  72.5 &  72.7 &  72.7 &          72.7 \\
TRIBE~\cite{Su_Xu_Jia_2024} &  \textbf{15.3} &  \textbf{16.6} &  \textbf{16.6} &  \textbf{16.3} &  \textbf{16.7} &  \textbf{17.0} &  \textbf{17.3} &  \textbf{17.4} &  \textbf{17.4} &  \textbf{18.0} &  \textbf{17.9} &  \textbf{18.0} &  \textbf{17.9} &  \textbf{18.6} &  \textbf{18.2} &  \textbf{18.8} &  \textbf{18.0} &  \textbf{18.2} &  \textbf{18.4} &  \textbf{18.0} &          \textbf{17.5} \\
\hline
\rowcolor{ClrHighlight}
\method \textit{(ours)}$^{(*)}$     & 24.3 &  23.0 & 22.6 &  22.4 &  22.4 &  22.5 &  22.3 &  22.5 &  22.8 &  22.8 &  22.6 & 22.7 & 22.7 &  22.9 &  22.6 & 22.7 &  22.6 &  22.8 &  22.9 &  23.0 &  22.8 \\

\bottomrule
\end{tabular}

   }
   \vspace*{-0.5\baselineskip}
\end{table*}

\begin{table*}[t!]
    \caption{Average classification error of the task ImageNet $\rightarrow$ ImageNet-C in \textit{\setting TTA} scenario.}
    \label{tab:imagenet-c-performance}
    \resizebox{\textwidth}{!}{
    \begin{tabular}{r|cccccccccccccccccccc|c}
\toprule
& \multicolumn{20}{l}{ \textit{\Setting TTA visit} $\xrightarrow{\hspace*{5cm}}$ } \\
\textbf{Method} &              1 &              2 &              3 &              4 &              5 &              6 &              7 &              8 &              9 &             10 &             11 &             12 &             13 &             14 &             15 &             16 &             17 &             18 &             19 &             20 &   \textbf{Avg} \\
\midrule
Source                                      & \multicolumn{20}{c|}{82.0}  & 82.0 \\
\midrule
LAME~\cite{malik2022_paramter-free}   & \multicolumn{20}{c|}{80.9} & 80.9 \\
\midrule
CoTTA~\cite{Wang_2022_CVPR} &           98.6 &           99.1 &           99.4 &           99.4 &           99.5 &           99.5 &           99.5 &           99.5 &           99.6 &           99.7 &           99.6 &           99.6 &           99.6 &           99.6 &           99.6 &           99.6 &           99.6 &           99.6 &           99.7 &           99.7 &           99.5 \\
EATA~\cite{niu2022efficient}  &  60.4 &  59.3 &  65.4 &  72.6 &  79.1 &  84.2 &  88.7 &  92.7 &  95.2 &  96.9 &  97.7 &  98.1 &  98.4 &  98.6 &  98.7 &  98.8 &  98.8 &  98.9 &  98.9 &  99.0 &          89.0 \\
RMT~\cite{döbler2023robust} &           72.3 &           71.0 &           69.9 &           69.1 &           68.8 &           68.5 &           68.4 &           68.3 &           70.0 &           70.2 &           70.1 &           70.2 &           72.8 &           76.8 &           75.6 &           75.1 &           75.1 &           75.2 &           74.8 &           74.7 &           71.8 \\
MECTA~\cite{hong2023mecta}  &           77.2 &           82.8 &           86.1 &           87.9 &           88.9 &           89.4 &           89.8 &           89.9 &           90.0 &           90.4 &           90.6 &           90.7 &           90.7 &           90.8 &           90.8 &           90.9 &           90.8 &           90.8 &           90.7 &           90.8 &           89.0 \\
RoTTA~\cite{yuan2023robust} &           68.3 &           62.1 &           61.8 &           64.5 &           68.4 &           75.4 &           82.7 &           95.1 &           95.8 &           96.6 &           97.1 &           97.9 &           98.3 &           98.7 &           99.0 &           99.1 &           99.3 &           99.4 &           99.5 &           99.6 &           87.9 \\
RDumb~\cite{press2023rdumb} &           72.2 &           73.0 &           73.2 &           72.8 &           72.2 &           72.8 &           73.3 &           72.7 &           71.9 &           73.0 &           73.2 &           73.1 &           72.0 &           72.7 &           73.3 &           73.1 &           72.1 &           72.6 &           73.3 &           73.1 &           72.8 \\
ROID~\cite{marsden2024universal}  &  62.7 &  62.3 &  62.3 &  62.3 &  62.5 &  62.3 &  62.4 &  62.4 &  62.3 &  62.6 &  62.5 &  62.3 &  62.5 &  62.4 &  62.5 &  62.4 &  62.4 &  62.5 &  62.4 &  62.5 &          62.4 \\
TRIBE~\cite{Su_Xu_Jia_2024} & \textbf{63.6} &  64.0 &  64.9 &  67.8 &  69.6 &  71.7 &  73.5 &  75.5 &  77.4 &  79.8 &  85.0 &  96.5 &  99.4 &  99.8 &  99.9 &  99.8 &  99.8 &  99.9 &  99.9 &  99.9 &          84.4 \\
\hline
\rowcolor{ClrHighlight}
\method \textit{(ours)}$^{(*)}$ &  65.3 &  \textbf{61.7} &  \textbf{59.8} &  \textbf{59.1} &  \textbf{59.4} &  \textbf{59.6} &  \textbf{59.8} &  \textbf{59.3} &  \textbf{59.4} &  \textbf{60.0} &  \textbf{60.3} &  \textbf{61.0} &  \textbf{60.7} &  \textbf{60.4} &  \textbf{60.6} &  \textbf{60.7} &  \textbf{60.8} &  \textbf{60.7} &  \textbf{60.4} &  \textbf{60.2} &  \textbf{60.5} \\
\bottomrule
\end{tabular}

   }
    \vspace*{-0.5\baselineskip}
\end{table*}

\begin{table*}[t!]
    \caption{Average classification error on CCC~\cite{press2023rdumb} setting. Each column presents the average error within an adaptation interval (e.g., the second column provides the average error between the $6701$ and $13400$ adaptation steps). Each adaptation step here is performed on a mini-batch of $64$ images.}
    \label{tab:ccc_performance}
    \resizebox{\textwidth}{!}{
    \begin{tabular}{r|cccccccccccc|c}
    \toprule
    & \multicolumn{11}{l}{ \textit{CCC~\cite{press2023rdumb} Adaptation Step} $\xrightarrow{\hspace*{5cm}}$ } \\
    \textbf{Method} 
    &           6700 &          13400 &          20100 &          26800 &          33500 &          40200 &          46900 &          53600 &          60200 &          66800 &          73400 &          80000 &   \textbf{Avg} \\

    \midrule
    Source                      &           0.83 &           0.83 &           0.83 &           0.83 &           0.83 &           0.84 &           0.84 &           0.83 &           0.84 &           0.83 &           0.83 &           0.83 &           0.83 \\ \midrule
RoTTA~\cite{yuan2023robust} &            0.70 &           0.85 &           0.92 &           0.96 &           0.98 &            1.00 &            1.00 &            1.00 &            1.00 &            1.00 &            1.00 &            1.00 &           0.95 \\
RDumb~\cite{press2023rdumb} &           0.78 &           0.74 &           0.75 &           0.77 &           0.75 &           0.72 &           0.75 &           0.77 &           0.75 &           0.74 &           0.75 &           0.75 &           0.75 \\

    \midrule
    \rowcolor{ClrHighlight}
    
    \method \textit{(ours)}     &  \textbf{0.67} &  \textbf{0.63} &  \textbf{0.62} &  \textbf{0.65} &  \textbf{0.65} &  \textbf{0.64} &  \textbf{0.64} &  \textbf{0.68} &  \textbf{0.63} &  \textbf{0.63} &  \textbf{0.65} &  \textbf{0.65} &  \textbf{0.64} \\

    \bottomrule
    \end{tabular}
   }
\end{table*}

\changes{
\noindent \textbf{Continuously Changing Corruption (CCC)~\cite{press2023rdumb} Performance.} Under CCC~\cite{press2023rdumb}, \Tab{\ref{tab:ccc_performance}} reveals the supreme performance of \method over RoTTA~\cite{yuan2023robust} and RDumb~\cite{press2023rdumb}. Here, we report the average classification error between two consecutive adaptation step intervals. An adaptation step in this table corresponds to a mini-batch of data with 64 images. The model is adapted to $80,000$ steps in total with more than $5.1$M images, significantly longer than 20 \setting TTA visits. Undoubtedly, \method still achieves good performance where the corruptions are algorithmically generated, non-cyclic with two or more corruption types can happen simultaneously. This experiment also empirically justifies the construction of our \setting TTA as a diagnostic tool (\Appdx{\ref{appdx:repeating_tta_diagnostic}}) where similar observations are concluded on the two settings. Obviously, our \setting TTA is notably simpler than CCC~\cite{press2023rdumb}.
}

\newsavebox{\fighistrotta}
\sbox{\fighistrotta}{%
    \input{figures/rotta_hist}
 }
 
 \newsavebox{\fighistours}
 \sbox{\fighistours}{%
    \input{figures/petta_hist}
 }
    
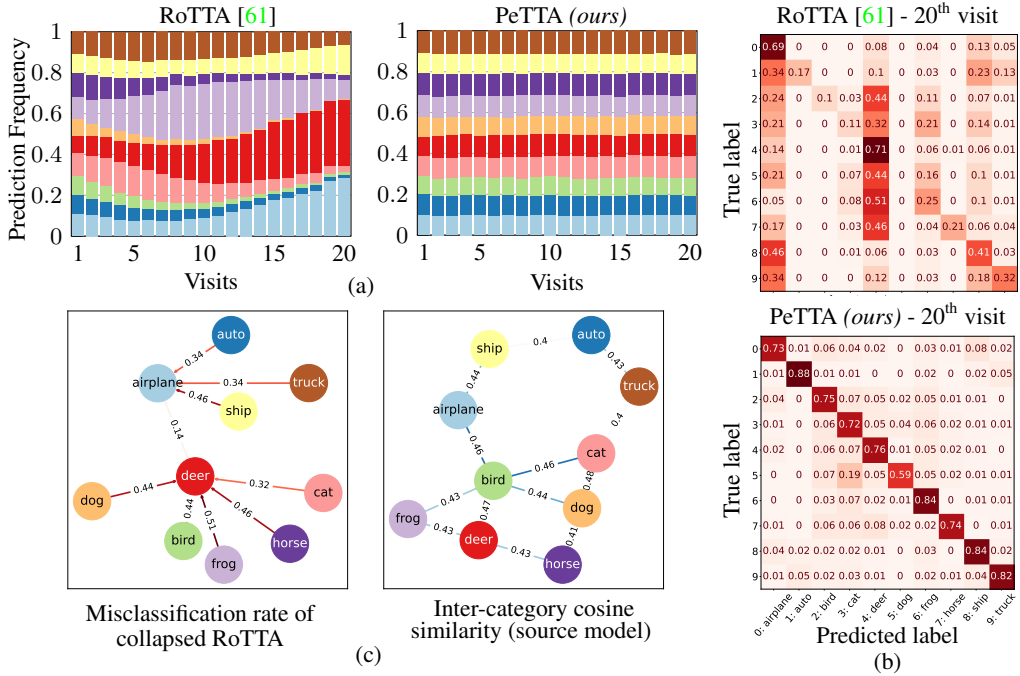
\begin{figure*}[t!]
    \pgfplotsset{every x tick label/.append style={font=\footnotesize, yshift=0.5ex}}
    \pgfplotsset{every y tick label/.append style={font=\footnotesize, xshift=0.5ex}}
    \centering    
        \begin{tikzpicture}

        \draw (3.2, -1.7) node[inner sep=0] {\includegraphics[width=3.8cm]{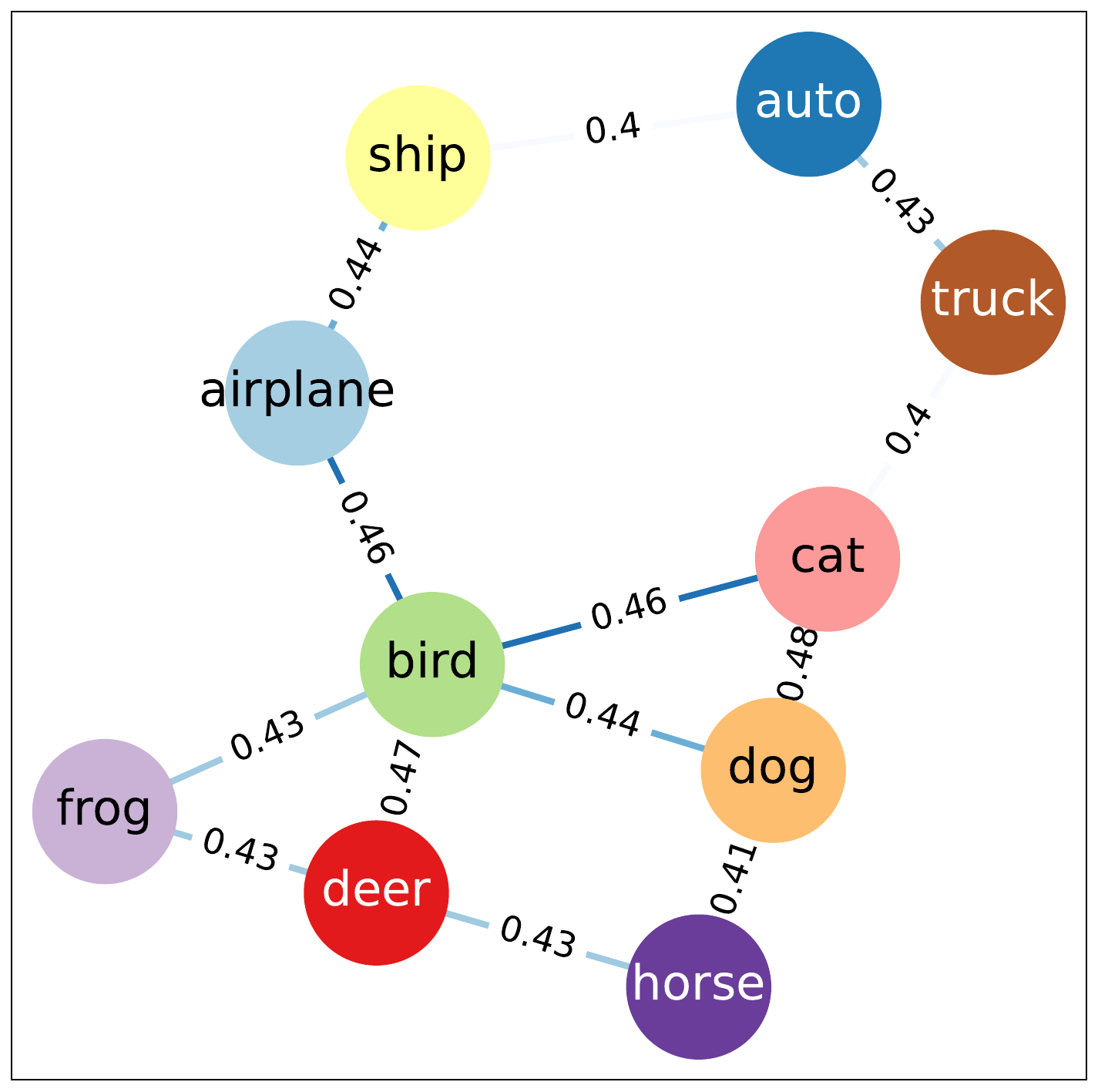}};
        \draw (-1, -1.7) node[inner sep=0] {\includegraphics[width=3.8cm]{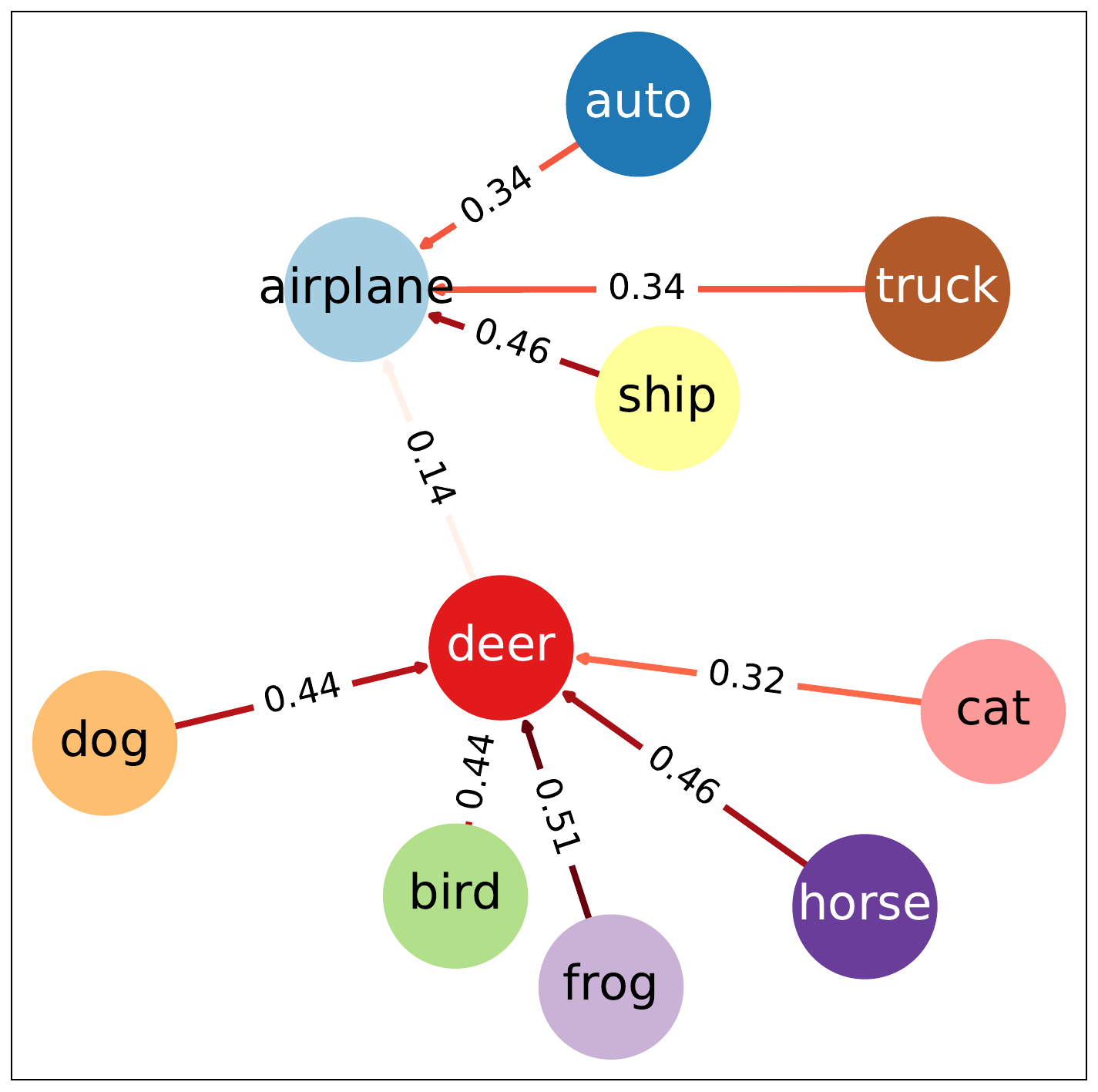}};

        \node at (3.3, -4.1) [above] {\footnotesize{ Inter-category cosine}};
        \node at (3.3, -4.4) [above, align=center] {\footnotesize{similarity (source model)}};
        \node at (-1.1, -4.1) [above, align=center] {\footnotesize{Misclassification  rate of}};
        \node at (-1.1, -4.5) [above, align=center] { \footnotesize{collapsed RoTTA}};

        \draw (8, 2.1) node[inner sep=0] {\includegraphics[width=3.6cm]{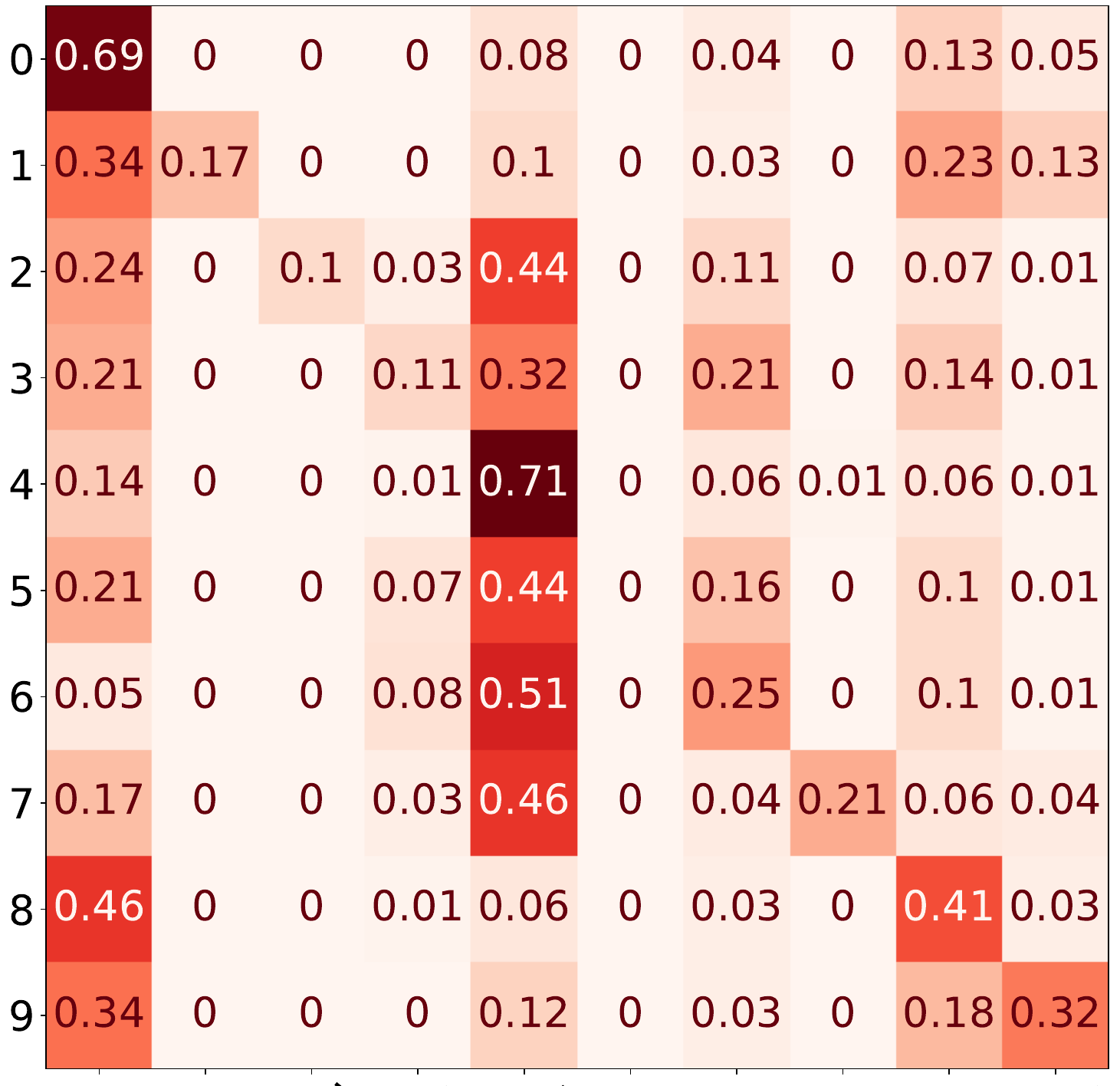}};
        \draw (8, -2.2) node[inner sep=0] {\includegraphics[width=3.6cm]{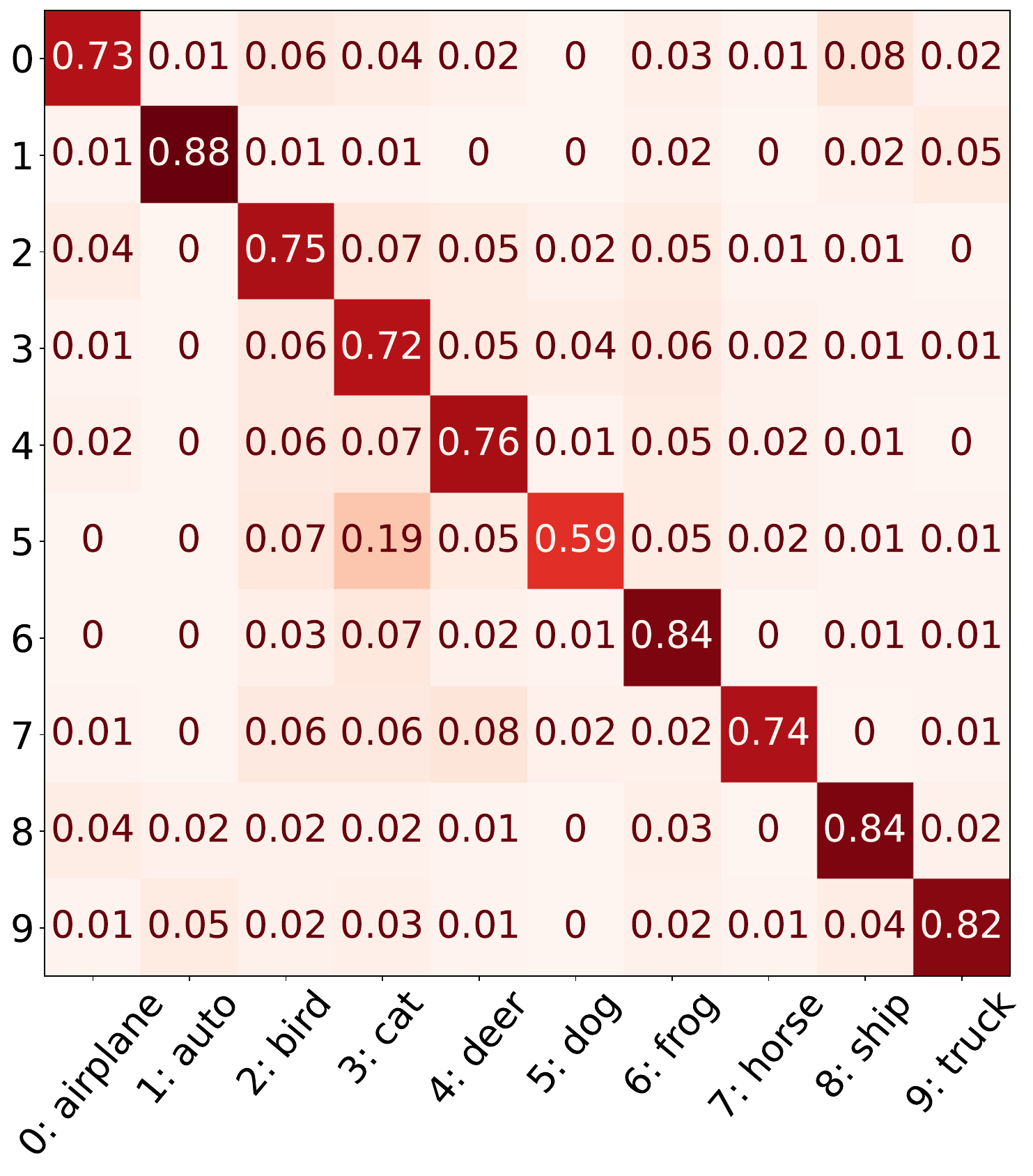}};
        
        \draw (-1.1, 2.2) node[inner sep=0] {\usebox \fighistrotta};
        \draw (3.5, 2.2) node[inner sep=0] {\usebox \fighistours};
        
        \node at (-0.9, 3.8) [above] {\footnotesize{ RoTTA \cite{yuan2023robust} }};
        \node at (3.7, 3.8) [above] {\footnotesize{\method \textit{(ours)}}};

        \node at (8.0, -0.2) [above] {\footnotesize{ \method \textit{(ours)} - 20\textsuperscript{th} visit}};
        \node at (8.0, 3.8) [above] {\footnotesize{ RoTTA \cite{yuan2023robust} - 20\textsuperscript{th} visit}};
        \node at (8.0, -4.4) [above] {\footnotesize{ Predicted label}};
        
        \node at (1.0, 0.2) [above] {\small{ (a) }};
        \node at (8.0, -4.8) [above] {\small{ (b)}};
        \node at (1.1, -4.7) [above] {\small{ (c)}};
         
        \node at (6.5, 2.8, 2) [below, rotate=90] {\footnotesize{True label}};
        \node at (6.5, -1.2, 2) [below, rotate=90] {\footnotesize{True label}};

        \node at (-3, 3.2, 2) [below, rotate=90] {\footnotesize{Prediction Frequency}};
        \end{tikzpicture}
    \caption{\Setting TTA (20 visits) on CIFAR-10$\rightarrow$CIFAR10-C task. (a) Histogram of model predictions (10 labels are color-coded). \method achieves a persisting performance while RoTTA~\cite{yuan2023robust} degrades. (b) Confusion matrix at the last visit, RoTTA classifies all samples into a few categories (e.g., \textit{0: airplane}, \textit{4: deer}). (c) Force-directed graphs showing (left) the most prone to misclassification pairs (arrows indicating the portion and pointing from the true to the misclassified category); (right) similar categories tend to be easily collapsed. Edges denote the average cosine similarity of feature vectors (source model), only the highest similar pairs are shown. Best viewed in color.}
    \label{fig:cifar10-c-result}
    \vspace*{-1.5\baselineskip}
\end{figure*}

\noindent \textbf{Collapsing Pattern. } The rise in classification error (\Fig{\ref{fig:fig1_introduction}}-right) can be reasoned by the prediction frequency of RoTTA~\cite{yuan2023robust} in an \setting TTA setting (\Fig{\ref{fig:cifar10-c-result}}a-left). Similar to $\epsilon$-GMMC, the likelihood of receiving predictions on certain categories gradually increases and dominates the others. Further inspecting the confusion matrix of a collapsed model (\Fig{\ref{fig:cifar10-c-result}}b-top) reveals two major groups of categories are formed and a single category within each group represents all members, thereby becoming dominant. To see this, \Fig{\ref{fig:cifar10-c-result}}c-left simplifies the confusion matrix by only visualizing the top prone-to-misclassified pair of categories. Here, label \textit{deer} is used for almost every living animal while \textit{airplane} represents transport vehicles. 
The similarity between categories in the feature space of the source model (\Fig{\ref{fig:cifar10-c-result}}c-right) is correlated with the likelihood of being merged upon collapsing.
As distance in feature space is analogous to $|\mu_0 - \mu_1|$ (Thm.~\ref{thm:cvg}), closer clusters are at a higher risk of collapsing. 
This explains and showcases that the collapsing behavior is predictable up to some extent.

\vspace*{-0.4\baselineskip}
\subsection{Ablation Study}
\vspace*{-0.4\baselineskip}

\begin{table}[t!]
    \parbox{.53\linewidth}{
        
    \centering
    \vspace*{-0.5\baselineskip}
    \caption{Average (across 20 visits) error of multiple variations of \method: without (w/o) $\mathcal{R}(\theta), \mathcal{L}_{\mathrm{AL}}$;
    $\mathcal{L}_{\mathrm{AL}}$ only;
    fixed regularization coefficient $\lambda$; adaptive coefficient $\lambda_t$, update rate $\alpha_t$; using anchor loss $\mathcal{L}_{\mathrm{AL}}$. 
    }
    \label{tab:ablation_study}
    \resizebox{.52\textwidth}{!}{
    \begin{tabular}{l|c|c|c|c}
         \toprule
         \textbf{Method}                                         & \textbf{CF-10-C} & \textbf{CF-100-C} & \textbf{DN}  & \textbf{IN-C} \\
         \midrule
         Baseline w/o $\mathcal{R}(\theta), \mathcal{L}_{\mathrm{AL}}$                      & 42.6             & 63.0              & 77.9  & 93.4  \\
         \midrule
         $\mathcal{R}(\theta)$ fixed $\lambda = 0.1\lambda_0$    & 43.3             & 65.0              & 80.0  & 92.5   \\
         $\mathcal{R}(\theta)$ fixed $\lambda = \lambda_0$       & 42.0             & 64.6              & 66.6  & 92.9  \\
         \midrule

        $\mathcal{L}_{\mathrm{AL}}$ only & 25.4             & 56.5              & 47.5  & 68.1 \\
         \hline \midrule
         \method~- $\lambda_t$                                   & 27.1             & 55.0              & 59.7  & 92.7   \\
         \method~- $\lambda_t$ + $\alpha_t$                      & 23.9             & 41.4              & 44.5  & 75.7   \\
         \method~- $\lambda_t$ + $\mathcal{L}_{\mathrm{AL}}$     & 26.2             & 36.3              & 43.2  & 62.0   \\
         
         \midrule
         \method~- $\lambda_t$  + $\alpha_t$  + $\mathcal{L}_{\mathrm{AL}}$  & \textbf{22.8}  & \textbf{35.1} & \textbf{42.9}  & \textbf{60.5}  \\
         \bottomrule
    \end{tabular}
    }
    \vspace*{-0.25\baselineskip}

    }
    \hfill
    \parbox{.45\linewidth}{
        \centering
\vspace*{-0.5\baselineskip}
\caption{Average (across 20 visits) error of \method. \method favors various choices of regularizers $\mathcal{R}(\theta)$: L2 and cosine similarity in conjunction with Fisher~\cite{kirkpatrick2017_fisher, niu2022efficient} coefficient.}
\label{tab:regularizers}
\resizebox{.45\textwidth}{!}{
\begin{tabular}{c|c|c|c|c|c}
    \toprule
      \multicolumn{2}{c|}{\textbf{Method}}     &   \multirow{2}{*}{\textbf{CF-10-C}}  & \multirow{2}{*}{\textbf{CF-100-C}} & \multirow{2}{*}{\textbf{DN}} & \multirow{2}{*}{\textbf{IN-C}}\\
      \cline{1-2}
        $\mathcal{R}(\theta)$ & Fisher& & &\\
    \midrule
      \multirow{2}{*}{L2}       &\xmark&  23.0              & 35.6           & 43.1             & 70.8  \\
                                &\cmark&  22.7              & 36.0           & 43.9             & 70.0  \\
    \midrule
      \multirow{2}{*}{Cosine}   &\xmark&  22.8              & {\textbf{35.1}}  & {\textbf{42.9}}    & \textbf{60.5} \\
                                &\cmark&  \textbf{22.6}     & 35.9           &  43.3            & 63.8  \\
     \bottomrule
\end{tabular}}
\vspace*{0.1\baselineskip}

    \\
    \footnotesize{CF: CIFAR, DN: DomainNet, IN: ImageNet}
    }
    \vspace*{-0.6\baselineskip}
\end{table}

\noindent \textbf{Effect of Each Component. } \Tab{~\ref{tab:ablation_study}} gives an ablation study on \method, highlighting the use of a regularization term ($\mathcal{R}(\theta)$) with a fixed choice of $\lambda, \alpha$ not only fails to mitigate model collapse but may also introduce a negative effect (rows 2-3). 
\changes{Trivially applying the anchor loss ($\mathcal{L}_{\mathrm{AL}}$) alone is also incapable of eliminating the lifelong performance degradation in continual TTA (row 4)}. Within \method, adopting the adaptive $\lambda_t$ scheme alone (row 5) or in conjunction with either $\alpha_t$ or anchor loss $\mathcal{L}_{\mathrm{AL}}$ (rows 6-7) partially stabilizes the performance. Under the drastic domain shifts with a larger size of categories or model parameters (e.g., on CIFAR-100-C, DomainNet, ImageNet-C), restricting $\alpha_t$ adjustment limits the ability of \method to stop undesirable updates while a common regularization term without $\mathcal{L}_{\mathrm{AL}}$ is insufficient to guide the adaptation. Thus, leveraging all elements secures the persistence of \method (row 8).

\noindent \textbf{Various Choices of Regularizers. } The design of \method is not coupled with any specific regularization term. Demonstrated in Tab.~\ref{tab:regularizers}, \method works well for the two common choices: L2 and cosine similarity. The conjunction use of Fisher coefficent~\cite{kirkpatrick2017_fisher, niu2022efficient} for weighting the model parameter importance is also studied.
While the benefit (in terms of improving accuracy) varies across datasets, \method accommodates all choices, as the model collapse is not observed in any of the options.

\section{Discussions and Conclusion}
\label{sec:discussions}
\noindent \textbf{On a Potential Risk of TTA in Practice. } We provide empirical and theoretical evidence on the risk of deploying continual TTA algorithms. Existing studies fail to detect this issue with \textit{a single pass per test set}.  
The \setting TTA could be conveniently adopted as a \textit{straightforward evaluation}, where its challenging test stream magnifies the error accumulation that a model might encounter in practice. 

\noindent \textbf{Limitations. } \method takes one step toward mitigating the gradual performance degradation of TTA. Nevertheless, a complete elimination of error accumulation cannot be guaranteed rigorously through regularization. Future research could delve deeper into expanding our efforts to develop an algorithm that achieves error accumulation-free by construction.
Furthermore, as tackling the challenge of the temporally correlated testing stream is not the focus of \method, using a small memory bank as in~\cite{yuan2023robust, gong2022note} is necessary. It also assumes the features statistics from the source distribution are available (\Appdx{\ref{appdx:using_mem_bank}, \ref{appdx:mean_cov_source_dataset}}). These constraints potentially limit its scalability in real-world scenarios.

\noindent \textbf{Conclusion}. 
Towards trustworthy and reliable TTA applications, we rigorously study the \textit{performance degradation problem of TTA}. 
The proposed \textit{\setting TTA} setting highlights the limitations of modern TTA methods, which struggle to prevent the error accumulation when continuously adapting to demanding test streams. 
Theoretically inspecting a failure case of \textit{$\epsilon-$GMMC} paves the road for designing \method - a simple yet efficient solution that continuously assesses the model divergence for harmonizing the TTA process, balancing adaptation, and collapse prevention.%

\small
\section*{Acknowledgements}
This work was supported by the Jump ARCHES Endowment through the Health Care Engineering Systems Center, JSPS/MEXT KAKENHI JP24K20830, ROIS NII Open Collaborative Research 2024-24S1201, in part by the National Institute of Health (NIH) under Grant R01 AI139401, and in part by the Vingroup Innovation Foundation under Grant VINIF.2021.DA00128.
\bibliographystyle{plainnat}
\bibliography{egbib}

\normalsize
\newpage
\appendix
\begin{center}
\large{\textbf{Persistent Test-time Adaptation in \Setting Testing Scenarios \\ Technical Appendices}}    
\end{center}
\addcontentsline{toc}{section}{} %

\part{} %
\begin{table}[h!]
    \parttoc %
\end{table}

\newpage
\setcounter{theorem}{0}
\setcounter{lemma}{0}
\setcounter{corollary}{0}
\setcounter{assumption}{0}
\setcounter{definition}{0}
\vspace*{-0.3\baselineskip}
\section{Related Work}
\vspace*{-0.3\baselineskip}
\label{sec:related_work}

\noindent \textbf{Towards Robust and Practical TTA. } While forming the basis, early single-target TTA approaches~\cite{pmlr-v119-sun20b, wang2021tent, nguyen2023tipi, niu2023towards, liang2020we} is far from practice. 
Observing the dynamic of many testing environments, a continual TTA setting is proposed where an ML model continuously adapts to a sequence of multiple shifts~\cite{marsden2022gradual, Wang_2022_CVPR}. Meanwhile, recent studies~\cite{gong2022note, malik2022_paramter-free} point out that the category distribution realistic streams is highly temporally correlated. Towards real-world TTA setting, Yuan \textit{et al.}~\cite{yuan2023robust} launch the \textit{practical TTA} which considers the simultaneous occurrence of the two aforementioned challenges. 

For a robust and gradual adaptation, an update via the mean teacher~\cite{antti2017_mean_teachers} mechanism is exploited in many continual TTA algorithms~\cite{Wang_2022_CVPR, yuan2023robust, döbler2023robust, hong2023mecta}. To moderate the temporally correlated test stream, common approaches utilize a small memory bank for saving a category-balanced subset of testing samples~\cite{gong2022note, yuan2023robust}, inspired by the replay methods~\cite{riemer2018learning, rahaf2019_online_cl} to avoid forgetting in the task of continual learning~\cite{sen2023_theory_cl, NEURIPS2019_e562cd9c, 9349197}. Our study emphasizes another perspective: beyond a supreme performance, a desirable TTA should also \textit{sustain it for an extended duration}. 

\noindent \textbf{Temporal Performance Degradation.} By studying the quality of various ML models across multiple industry applications~\cite{vela_temporal_2022, YOUNG2022100070}  the issue of AI ``aging" with the temporal model degradation progress, even with data coming from a stable process has been confirmed. In TTA, the continuous changes of model parameters through gradient descent aggravate the situation, \hh{as also recently noticed in~\cite{press2023rdumb}}. \hh{Apart from observation, we attempt to investigate and provide \textit{theoretical} insights towards the mechanism of this phenomenon.}

\noindent \textbf{Accumulated Errors in TTA. }
In TTA, the issue of accumulated error has been \hh{briefly acknowledged}. 
Previous works strive to avoid drastic changes to model parameters as a good practice. Up to some degree, it helps to avoid performance degradation. Nevertheless, it is still \textit{ unclear whether their effectiveness truly eliminates the risk}.
To preserve in-distribution performance, regularization~\cite{kirkpatrick2017_fisher, niu2022efficient} or replaying of training samples at test-time~\cite{döbler2023robust} have been used. 
\hh{Other studies explore reset (recovering the initial model parameters) strategies~\cite{Wang_2022_CVPR, press2023rdumb}, periodically or upon the running entropy loss approaches a threshold~\cite{niu2023towards}. Unfortunately, knowledge accumulated in the preceding steps will vanish, and a bad heuristic choice of threshold or period leads to highly frequent model resets. Noteworthy, tuning those hyper-parameters is exceedingly difficult due to the unavailability of the validation set~\cite{zhao2023on}. LAME~\cite{malik2022_paramter-free} suggests a post-processing step for adaptation (without updating the parameters). This approach, however, still limits the knowledge accumulation. 
Our \method is \textit{reset-free} by achieving an adaptable continual test-time training.}

\section{Proof of Lemmas and Theorems}
\label{sec:convergence_proof}
In this section, we prove the theoretical results regarding the $\epsilon-$perturbed Gaussian Mixture Model Classifier ($\epsilon-$GMMC) introduced in \Sec{\ref{ssec:egmmc_analysis}}. We first briefly summarize the definition of model collapse and the static data stream assumption:

\begin{definition}[\textbf{Model Collapse}]
\label{def:model_collapse}
A model is said to be \textit{collapsed} from step $\tau \in \mathcal{T}, \tau < \infty$ if there exists a non-empty subset of categories $\Tilde{\mathcal{Y}} \subset \mathcal{Y}$ such that $\Pr\{Y_t \in \Tilde{\mathcal{Y}}\} > 0$ but the marginal $\Pr\{\hat Y_t \in \mathcal{\Tilde{Y}}\}$ converges to zero in probability: 
\begin{align*}
    \underset{t \to \tau}{\lim} \Pr\{\hat Y_t \in \Tilde{\mathcal{Y}}\} = 0. %
\end{align*}%
\end{definition}

\begin{assumption}[\textbf{Static Data Stream}]
\label{as:static_stream}
The marginal distribution of the true label follows the same Bernoulli distribution $\mathrm{Ber}(p_0)$:
$p_{0,t} = p_0$, ($p_{1,t} = p_1 = 1-p_0), \forall t \in \mathcal{T}$.
\end{assumption}

\noindent \textbf{Preliminary. } Following the same set of notations introduced in the main text, recall that we denoted $p_{y,t} \overset{\Delta}{=} \Pr\{Y_t = y\}$, $\hat p_{y,t} \overset{\Delta}{=} \Pr \{\hat Y_t = y\}$ (marginal distribution of the true label $Y_t$ and pseudo label $\hat Y_t$ receiving label $y$, respectively) and $\epsilon_t$ = $\Pr\{Y_t = 1| \hat Y_t = 0 \}$ (the false negative rate (FNR) of $\epsilon-$GMMC). At testing step $t$, we obtain the following relations:
\begin{align}
    \E{X_t | \hat Y_t = 0}   &= (1-\epsilon_t) \mu_0 + \epsilon_t \mu_1 \label{eq:EXt_yt=0}, \\
    \E{X_t | \hat Y_t = 1}   &= \mu_1 \label{eq:EXt_yt=1}, \\
    \Var{X_t | \hat Y_t = 0} &= (1 - \epsilon_t) \sigma_0^2+ \epsilon_t \sigma_1^2 + \epsilon_t(1-\epsilon_t)(\mu_0 - \mu_1)^2 \label{eq:VarXt_yt=0}, \ \\
    \Var{X_t | \hat Y_t = 1} &= \sigma_1^2 \label{eq:VarXt_yt=1}.
\end{align}

In addition, under Assumption~\ref{as:static_stream}, the marginal distribution $P_t(x)$ (also referred as \textit{data distribution} in our setup) is:
\begin{align}
    \label{eq:dist_ptx}
    P_t(x) = \mathcal{N}(x; p_0 \mu_0 + p_1 \mu_1, p_0\sigma_{0}^2+p_1\sigma_{1}^2+p_0p_1(\mu_0 - \mu_1)^2) \qquad \forall t \in \mathcal{T}.
\end{align}

\subsection{Proof of Lemma 1}
\begin{lemma}[\textbf{Increasing FNR}] Under Assumption~\ref{as:static_stream}, a binary $\epsilon$-GMMC would collapsed (Def.~\ref{def:model_collapse}) with $\underset{t \to \tau}{\lim} \hat p_{1,t} = 0$ (or  $\underset{t \to \tau}{\lim} \hat p_{0,t} = 1$, equivalently) if and only if $\underset{t \to \tau}{\lim} \epsilon_{t} = p_1$.
\label{lmm:increasing_fnr}
\end{lemma}%

\begin{proof} Under Assumption~\ref{as:static_stream}, we have $\E{X_t} = p_0 \mu_0 + (1-p_0) \mu_1$. Also note that:
\begin{align}
    \E{X_t} &= \E{\E{X_t|\hat Y_t}} \nonumber \\
        &= \E{X_t|\hat Y_t = 0} \hat p_{0,t} + \E{X_t|\hat Y_t = 1} \hat p_{1,t}  \label{eq:cond_E}  \\
        &= \left[(1-\epsilon_t)\mu_0 + \epsilon_t \mu_1\right] \hat p_{0,t} + \mu_1 (1 - \hat p_{0,t}) \nonumber \\
        &= \left[(1-\epsilon_t)  \hat p_{0,t} \right] \mu_0  +  \left[1 - \hat p_{0,t} (1 - \epsilon_t) \right]  \mu_1 \nonumber \\
        &= p_0 \mu_0 + (1-p_0) \mu_1 \nonumber,
\end{align}
where the second equality follows Eqs.~\ref{eq:EXt_yt=0}-\ref{eq:EXt_yt=1}. Therefore:
\begin{align}
    \label{eq:pred_fnr_correlation}
    \hat p_{0,t} = \frac{p_0}{1-\epsilon_t}.
\end{align}
Eq.~\ref{eq:pred_fnr_correlation} shows positive correlation between $\hat p_{0,t}$ and $\epsilon_t$. Given $\underset{t \to \tau}{\lim} \epsilon_{t} = p_1$, taking the limit introduces:
\begin{align*}
     \underset{t \to \tau}{\lim}\hat p_{0,t} =  \underset{t \to \tau}{\lim}  \frac{p_0}{1-\epsilon_t} = \frac{p_0}{1 - p_1} = 1.
\end{align*}
Similarly, having $\underset{t \to \tau}{\lim}\hat p_{0,t}= 1$, the false negative rate $\epsilon_t$ when $t \rightarrow \tau$ is:
\begin{align*}
     \underset{t \to \tau}{\lim} \epsilon_{t} = 1 - p_0 = p_1.
\end{align*}
Since $\hat p_{0,t} + \hat p_{1,t} =1$, $\underset{t \to \tau}{\lim}\hat p_{1,t}= 0$, equivalently. Towards the collapsing point, the model tends to predict a single label (class 0 in the current setup). In addition, the FNR of the model $\epsilon_t$ also raises correspondingly. 
\end{proof}

\subsection{Proof of Lemma 2. }
\begin{lemma}[\textbf{$\epsilon$-GMMC After Collapsing}]
\label{lmm:collapsed}
For a binary $\epsilon$-GMMC model, with Assumption~\ref{as:static_stream}, if $\underset{t \to \tau}{\lim} \hat p_{1,t} = 0$ (collapsing), the cluster 0 in  GMMC converges in distribution to a single-cluster GMMC with parameters:
\begin{align*}%
     \mathcal{N}(\hat \mu_{0,t}, \hat \sigma_{0,t}^2) \overset{d.}{\rightarrow}  \mathcal{N}&(p_0 \mu_0 + p_1 \mu_1, p_0\sigma_{0}^2+p_1\sigma_{1}^2+p_0p_1(\mu_0 - \mu_1)^2).
\end{align*}%
\end{lemma}

\begin{proof} 
From Eqs.~\ref{eq:EXt_yt=0}-\ref{eq:EXt_yt=1}, under the increasing type II collapse of $\epsilon-$GMMC setting, the perturbation does not affect the approximation of $\mu_1$. Meanwhile, when $\epsilon_t$ increases, one can expect that $\hat \mu_{0,t}$ moves further away from $\mu_0$ toward $\mu_1$. Frist, the mean teacher model of GMMC (\Eq{\ref{eq:update_gmmc}}, main text) gives:
\begin{align*}
   \E{\hat \mu_{0,t} | \hat Y_t = 1} &= \mathbb{E}_{P_{t-1}}\left[{\hat \mu_{0,t-1}}\right], \\
   \E{\hat \mu_{0,t} | \hat Y_t = 0} &= (1-\alpha) \mathbb{E}_{P_{t-1}}\left[\hat\mu_{0,t-1}|\hat Y_t=0\right]  + \alpha \E{X_t | \hat Y_t= 0} \\
   &= (1-\alpha) \mathbb{E}_{P_{t-1}}\left[\hat \mu_{0,t-1}\right] + \alpha \left(\E{X_i | \hat Y_t = 0}\right), \\
   \E{\hat \mu_{1,t} | \hat Y_t = 1} &=  (1-\alpha) \mathbb{E}_{P_{t-1}}\left[\hat \mu_{1, t-1}| \hat Y_t = 1\right] + \alpha \E{X_t | \hat Y_t =1} \\ 
    &= (1-\alpha) \mathbb{E}_{P_{t-1}}\left[\hat \mu_{1,t-1}\right] + \alpha \left(\E{X_i | \hat Y_t = 1}\right), \\
   \E{\hat \mu_{1,t} | \hat Y_t = 0} &= \mathbb{E}_{P_{t-1}}\left[\hat \mu_{1,t-1}\right].
\end{align*}
By defining $u_{y,t} = \E{\hat \mu_{y,t}}$, we obtain the following recurrence relation between $u_{0,t}$ and $u_{0,t-1}$:
\begin{align}
    u_{0,t} &= \E{\hat \mu_{0,t} | \hat Y_t = 0} \hat p_{0,t} + \E{\hat \mu_{0,t} | \hat Y_t = 1} \hat p_{1,t} \nonumber \\
            &= \left((1-\alpha) u_{0,t-1}  + \alpha \E{X_t | \hat Y_t= 0} \right)\hat p_{0,t} + u_{0,t-1} \hat p_{1,t} \nonumber \\
            &= \left[(1-\alpha) \hat p_{0,t} + \hat p_{1,t}\right] u_{0,t-1}   + \alpha \hat p_{0,t} \E{X_t | \hat Y_t= 0} \nonumber \\
            &= (1 -\alpha \hat p_{0,t} ) u_{0,t-1} + \alpha \hat p_{0,t} \E{X_t | \hat Y_t= 0} \nonumber \\
            &= (1 -\alpha \hat p_{0,t} ) u_{0,t-1} + \alpha \hat p_{0,t} \left[ (1-\epsilon_t) \mu_0 + \epsilon_t \mu_1 \right]. \label{eq:u0t}
\end{align}
Given $\underset{t \to \tau}{\lim} \hat p_{0,t} = 1$, it follows that $\underset{t \to \tau}{\lim} \epsilon_{0,t} = p_1$ by Lemma~\ref{lmm:increasing_fnr}. From this point:
\begin{align*}
    u_{0,t} =  (1-\alpha) u_{0,t-1} + \alpha \left(p_0 \mu_0 + p_1 \mu_1 \right) \qquad \forall t>\tau.
\end{align*}
Taking the limit $t \rightarrow \infty$:
\begin{align*}
    \underset{t \to \infty}{\lim} u_{0,t} 
    &= \underset{t \to \infty}{\lim} (1-\alpha) u_{0,t-1} + \alpha \left(p_0 \mu_0 + p_1 \mu_1 \right) \\
    &= \underset{t \to \infty}{\lim} (1-\alpha)^t \hat \mu_{0,0} + \alpha \sum_{i=1}^{t} (1-\alpha)^{i-1}  \left(p_0 \mu_0 + p_1 \mu_1 \right)\\
    &= \underset{t \to \infty}{\lim} (1-\alpha)^{t} \hat \mu_{0,0} + (1 - (1-\alpha)^t) (p_0 \mu_0 + p_1 \mu_1) \\
    &= p_0 \mu_0 + p_1 \mu_1.
\end{align*}
The second equation is obtained by solving the recurrence relation. When $\underset{t \to \tau}{\lim} \hat p_{0,t} = 1$, $\{\hat \mu_{y,t}\}_{y \in \{0,1\}}$ becomes a deterministic values.
Hence, giving $u_{y,t} = \E{\hat \mu_{y,t}} = \hat \mu_{0,t} (\forall t > \tau)$ and
\begin{align}
    \label{eq:lim_mu0t}
    \underset{t \to \infty}{\lim} \hat \mu_{0,t} = \underset{t \to \infty}{\lim} u_{0,t} = p_0 \mu_0 + p_1 \mu_1.
\end{align} 

\noindent Repeating the steps above with Eqs.~\ref{eq:VarXt_yt=0}-\ref{eq:VarXt_yt=1} in place of Eqs.~\ref{eq:EXt_yt=0}-\ref{eq:EXt_yt=1}, we obtain a similar result for $\sigma_{0,t}^2$:
\begin{align}
    \label{eq:lim_sigma0t}
    \underset{t \to \infty}{\lim} \hat \sigma_{0,t}^2 =  p_0\sigma_{0}^2+p_1\sigma_{1}^2+p_0p_1(\mu_0 - \mu_1)^2.
\end{align}
By \textit{Lévy's continuity theorem} (p. 302, ~\cite{parthasarathy_introduction_2005}), from Eqs.~\ref{eq:lim_mu0t}-\ref{eq:lim_sigma0t}, when $t \rightarrow \infty$, the estimated distribution of the first cluster $\mathcal{N}(x; \hat \mu_{0,t} \hat \sigma_{0,t}^2)$ converges to the whole data distribution $P_t (x)$ (Eq.~\ref{eq:dist_ptx}) when collapsing. 

\end{proof}

\subsection{Proof of Theorem 1 and Corollary 1.}
\begin{theorem}[\textbf{Convergence of $\epsilon-$GMMC}]
\label{thm:cvg}
For a binary $\epsilon$-GMMC model, with Assumption~\ref{as:static_stream}, let the distance from $\hat \mu_{0,t}$ toward $\mu_1$ is $d_{t}^{0\rightarrow1} = \left|\E{\hat \mu_{0,t}} - \mu_1\right|$, then:
\[ d_{t}^{0\rightarrow1} - d_{t-1}^{0\rightarrow1} \leq \alpha \cdot p_0 \cdot \left( |\mu_0-\mu_1| - \frac{ d_{t-1}^{0\rightarrow1}}{1-\epsilon_t} \right).
\]%
\end{theorem}%

\begin{proof}
Substituting  Eq.~\ref{eq:pred_fnr_correlation} into $\hat p_{0,t}$ of Eq.~\ref{eq:u0t} gives:
\begin{align*}
    u_{0,t} 
    &= \left(1 - \frac{\alpha p_0}{1-\epsilon_t}\right) u_{0,t-1} +  \frac{\alpha p_0}{1-\epsilon_t} \left[ (1-\epsilon_t) \mu_0 + \epsilon_t \mu_1 \right].
\end{align*}
Hence, we have the distance from $u_{0,t}$ toward $\mu_1$:
\begin{align*}
    |u_{0,t} - \mu_{1}| 
    &= \left| \left(1 - \frac{\alpha p_0}{1-\epsilon_t}\right) u_{0,t-1} + \alpha p_0 \mu_0 + \frac{\alpha p_0 \epsilon_t \mu_1}{1 - \epsilon_t} - \mu_1 \right| \\
    &= \left| \left(1-\frac{\alpha p_0}{1-\epsilon_t} \right) (u_{0,t-1}-\mu_1) + \alpha p_0 \mu_0 + \frac{\alpha p_0 \epsilon_t \mu_1}{1 - \epsilon_t} - \frac{\alpha p_0 \mu_1}{1-\epsilon_t} \right| \\
    &= \left| \left(1-\frac{\alpha p_0}{1-\epsilon_t} \right) (u_{0,t-1}-\mu_1) + \alpha p_0 \mu_0 - \frac{\alpha p_0 \mu_1 (1 - \epsilon_t)}{1 - \epsilon_t} \right| \\
    &= \left| \left(1-\frac{\alpha p_0}{1-\epsilon_t} \right) (u_{0,t-1}-\mu_1) + \alpha p_0 (\mu_0 - \mu_1) \right| \\
    &\leq \left(1-\frac{\alpha p_0}{1-\epsilon_t} \right) |u_{0,t-1}- \mu_1| + \alpha p_0 |\mu_0-\mu_1|.\\
\end{align*}
The last inequality holds due to the triangle inequality. Equivalently, 
\begin{align*}
    |u_{0,t} - \mu_{1}| - |u_{0,{t-1}} - \mu_{1}|                   &\leq \alpha \cdot p_0 \cdot \left( |\mu_0-\mu_1| - \frac{ |u_{0,t-1}- \mu_1|}{1-\epsilon_t} \right).
\end{align*}
Let  $d_{t}^{0\rightarrow1} = \left|\E{\hat \mu_{0,t}} - \mu_1\right|$, we conclude that:
\begin{align*}
    d_{t}^{0\rightarrow1} - d_{t-1}^{0\rightarrow1} &\leq  \alpha \cdot p_0 \cdot \left( |\mu_0-\mu_1| - \frac{  d_{t-1}^{0\rightarrow1}}{1-\epsilon_t} \right) .
\end{align*}

\end{proof}

\begin{corollary}[\textbf{A Condition for $\epsilon-$GMMC Collapse}]
\label{corollary:condition}
With fixed $p_0$, $\alpha, \mu_0, \mu_1$, $\epsilon-$GMMC is collapsed if there exists a sequence of $\{\epsilon_t\}_{\tau - \Delta_\tau}^{\tau}$ ($\tau \ge \Delta_\tau > 0$) such that: 
\begin{align*}
    p_1 \geq \epsilon_t > 1 - \frac{d_{t-1}^{0\rightarrow1}}{\left|\mu_0 - \mu_1\right|}, \quad t \in [\tau - \Delta_\tau, \tau].
\end{align*}%
\end{corollary}

\begin{proof}
    Initialized at $\mu_0$, $\epsilon$-GMMC is collapsing when $\hat \mu_{0,t}$ converges to the mid-point $p_0 \mu_0 + p_1 \mu_1$ (Lemma~\ref{lmm:collapsed}), i.e., moving closer to $\mu_1$. From Thm.~\ref{thm:cvg}, the distance towards $\mu_1$  
     $d_{t}^{0\rightarrow1} < d_{t-1}^{0\rightarrow1}$ if
    \begin{align*}
    |\mu_0-\mu_1| -  \frac{ |u_{0,t-1}- \mu_1|}{1-\epsilon_t} < 0
    \Leftrightarrow |\mu_0-\mu_1|  <   \frac{ |u_{0,t-1}- \mu_1|}{1-\epsilon_t} 
    \Leftrightarrow  \epsilon_t > 1 - \frac{ |u_{0,t-1}- \mu_1|}{|\mu_0-\mu_1|}.
    \end{align*}
    When there exists this sequence $\{\epsilon_t\}_{\tau - \Delta_\tau}^{\tau}$ $(\tau \ge \Delta_\tau > 0)$ it follows that $d_{t}^{0\rightarrow1} < d_{t-1}^{0\rightarrow1}$ and $\epsilon_t > \epsilon_{t-1}$ is guaranteed $\forall t \in [\tau - \Delta_\tau, \tau]$. Hence, $\underset{t \to \tau}{\lim} \epsilon_{t} = p_1$ (model collapsed, by Lemma~\ref{lmm:increasing_fnr}). 
\end{proof}

\section{Further Justifications on Gaussian Mixture Model Classifier}

One may notice that in $\epsilon$-GMMC (Sec. 4.2), the classifier is defined $f_t(x) = \text{argmax}_{y \in \mathcal{Y}} \Pr(x|y; \theta_t)$ (maximum likelihood estimation) while in general, $f_t(x) = \text{argmax}_{y \in \mathcal{Y}} \Pr(y|x; \theta_t)$ (maximum a posterior estimation), parameterized by a neural network. In this case, since the \textit{equal prior} (i.e., $\Pr(y; \theta_t) = \Pr(y'; \theta_t), \forall y,y'\in \mathcal{C}$) is enforced in $\epsilon$-GMMC, the two definitions are \textit{equivalent}.
\begin{proof}
    Having:
    \begin{align*}
        \text{argmax}_{y \in \mathcal{Y}} \Pr(y|x; \theta_t) 
        &= \text{argmax}_{y \in \mathcal{Y}} \frac{\Pr(x|y; \theta_t) \Pr(y; \theta_t)}{\sum_{y' \in \mathcal{Y}}\Pr(x|y'; \theta_t) \Pr(y'; \theta_t)} \\
        &= \text{argmax}_{y \in \mathcal{Y}} \Pr(x|y; \theta_t).
    \end{align*}
    We conclude that the two definitions are equivalent. In fact, it is well-known that maximum likelihood estimation is a special case of maximum a posterior estimation when the prior is uniform.
\end{proof}

\section{Further Justifications on the \Setting Testing Scenario}
\label{appdx:repeating_tta}

\subsection{\Setting TTA Follows the Design of a Practical TTA Stream}
Note that in \setting TTA, besides the recurrence of environments (or corruptions) as in~\cite{Wang_2022_CVPR, niu2022efficient}, the distribution of class labels is also temporally correlated (non-i.i.d.) as suggested by~\cite{gong2022note, yuan2023robust} to reflect the practical testing stream better. In short, \setting TTA is formed by recurring the environments of \textit{practical TTA} scenario introduced in~\cite{yuan2023robust} multiple times (readers are encouraged to visit the original paper for additional motivations on this scenario).

\subsection{\Setting TTA as a Diagnostic Tool}
\label{appdx:repeating_tta_diagnostic}
Noticeably, CoTTA~\cite{Wang_2022_CVPR} also performed 10-round repetition across multiple domain shifts to simulate a lifelong TTA testing stream just like our \setting TTA.
However, the key difference is CoTTA assumes the distribution of class labels is {i.i.d.}, which does not hold in many real-life testing scenarios as argued in~\cite{gong2022note, yuan2023robust}. Our \setting TTA lifts this assumption and allows temporally correlated ({non-i.i.d.}) label distribution (more challenging, more practical).
This extension allows \textit{\setting TTA} to spot the risk of model collapse on CoTTA~\cite{Wang_2022_CVPR} and other methods. The \textit{over-simplicity} of the repeating scheme in CoTTA for spotting performance degradation is also suggested in~\cite{press2023rdumb}. Clearly, it seems not to be a problem at first glance in \Tab{5} of~\cite{Wang_2022_CVPR} (CoTTA’s 10-round repetition), but in fact, the risk in CoTTA remains, as explored in our scenario and also on CCC~\cite{press2023rdumb}.

The construction of our \setting TTA is notably simple - a technical effort to extend the testing stream. However, this simplicity is on purpose, \textit{serving as a diagnostic tool for lifelong continual TTA}. Counterintuitively, our experiments on four different tasks with the latest methods verify that even if the model is exposed to the same environment \textit{(the most basic case)}, their adaptability and performance are still consistently reduced (demonstrated visually in \Fig{\ref{fig:fig1_introduction}}, quantitatively in \Sec{\ref{ssec:result_real_data}}). 

We believe that the extensive testing stream by recurrence in our setup is a \textit{simple yet sufficient scenario} to demonstrate the vulnerability of existing continual TTA methods when facing the issue of model collapse (compared to CCC~\cite{press2023rdumb}, a notably \textit{more complicated scenario} than our \setting TTA). Indeed, \setting shifts are sufficient to show this failure mode and any lifelong TTA method should necessarily be able to handle \setting conditions.

\subsection{\Setting TTA with Random Orders}
\label{appdx:repeating_random_orders}
\newrobustcmd*{\mycircle}[1]{\tikz{\filldraw[draw=#1,fill=#1] (0,0) circle [radius=0.075cm];}}
\newrobustcmd*{\mysquare}[1]{\tikz{\filldraw[draw=#1,fill=#1] (0,0) --
(0.2cm,0) -- (0.2cm,0.2cm) -- (0cm,0.2cm);}}

\begin{figure}[t]
    \pgfplotsset{every x tick label/.append style={font=\tiny, yshift=0.5ex}}
    \pgfplotsset{every y tick label/.append style={font=\tiny, xshift=0.5ex}}
    \centering
        \begin{tikzpicture}
        
        \def \h{4.9}
        \def \u{0.6}
        \def \w{3.4}
        \def \ox{0.2} 
        \def \imgwidth{3.35}
        \def \r{0} 
        
        \draw (\ox + \w * 0 - 0.1, -\h * \r) node[inner sep=0] {\includegraphics[width=\imgwidth cm]{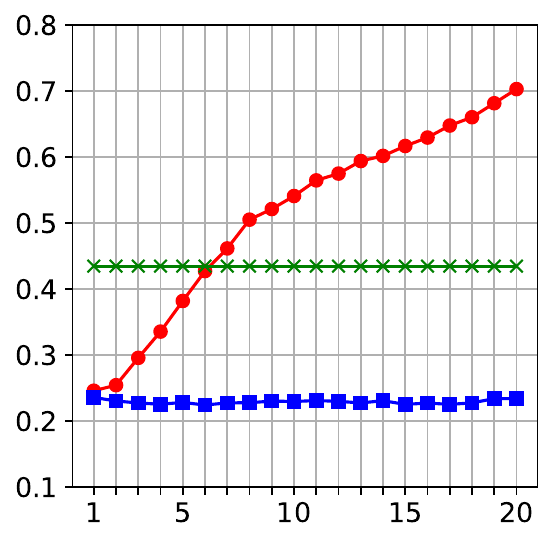}};
        \draw (\ox + \w * 1, -\h * \r) node[inner sep=0] {\includegraphics[width=\imgwidth cm]{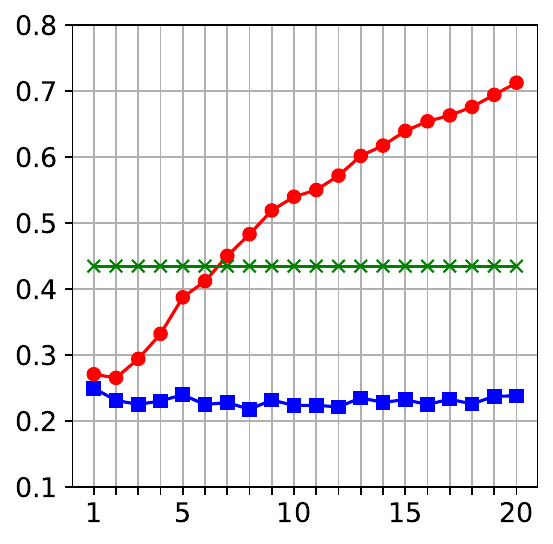}};
        \draw (\ox + \w * 2, -\h * \r) node[inner sep=0] {\includegraphics[width=\imgwidth cm]{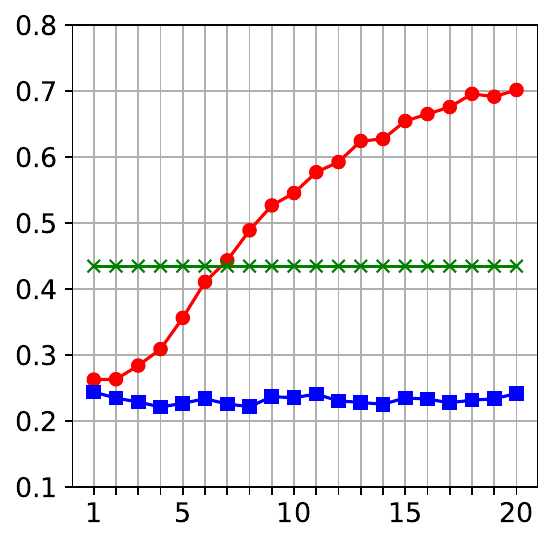}};
        \draw (\ox + \w * 3 ,-\h * \r) node[inner sep=0] {\includegraphics[width=\imgwidth cm]{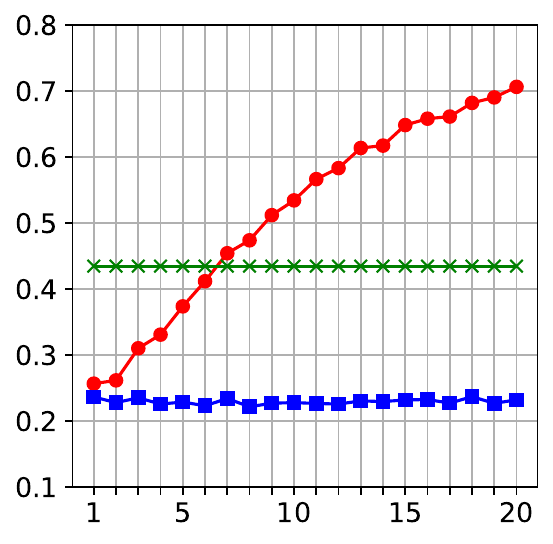}};

        \node at (\ox+ 0.2 + \w * 0, -\h * \r + \imgwidth * 0.45) [above] {\scriptsize{Same-order}};
        \node at (\ox+ 0.2 + \w * 1, -\h * \r + \imgwidth * 0.45) [above] {\scriptsize{Random-order (\textit{seed=0})}};
        \node at (\ox+ 0.2 + \w * 2, -\h * \r + \imgwidth * 0.45) [above] {\scriptsize{Random-order (\textit{seed=1})}};
        \node at (\ox+ 0.2 + \w * 3, -\h * \r + \imgwidth * 0.45) [above] {\scriptsize{Random-order (\textit{seed=2})}};
        
        \node at (-\imgwidth * 0.26, -\h * \r + 0.8, 2) [rotate=90] {\scriptsize{Testing  Error}};

        \node at (\ox + 0.2 + \w * 0, -\h * \r - \imgwidth * 0.45) [below] {\scriptsize{\Setting TTA visit}};
        \node at (\ox + 0.2 + \w * 1, -\h * \r - \imgwidth * 0.45) [below] {\scriptsize{\Setting TTA visit}};
        \node at (\ox + 0.2 + \w * 2, -\h * \r - \imgwidth * 0.45) [below] {\scriptsize{\Setting TTA visit}};
        \node at (\ox + 0.2 + \w * 3, -\h * \r - \imgwidth * 0.45) [below] {\scriptsize{\Setting TTA visit}};
        
        \node at (0.38 * \linewidth, -\h * \r - \imgwidth * 0.6) [below] {(a) CIFAR-10 $\rightarrow$ CIFAR-10-C task.};
        \draw (\ox + 0.1 + \w * 0.7, \h * \r -0.3 * \imgwidth, 2) -- (\ox + 0.1 + \w * 0.7, \h * \r + 0.8 * \imgwidth, 2);
        
        \def \r{1} 
        \draw (\ox + \w * 0, -\h * \r) node[inner sep=0] {\includegraphics[width=\imgwidth cm]{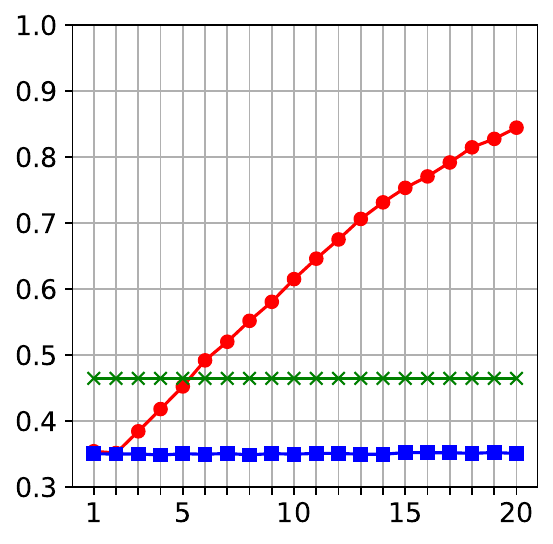}};
        \draw (\ox + \w * 1, -\h * \r) node[inner sep=0] {\includegraphics[width=\imgwidth cm]{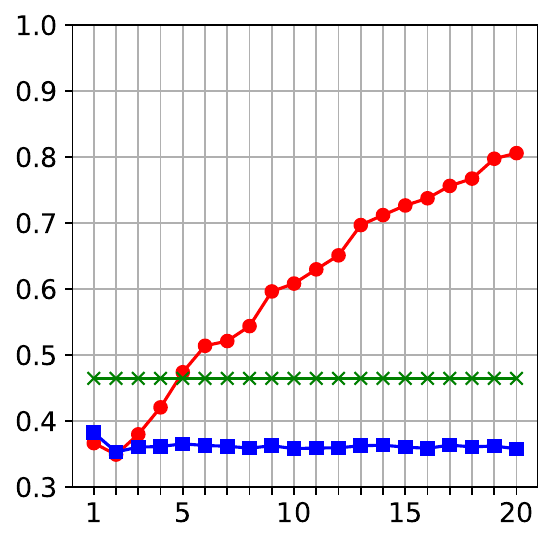}};
        \draw (\ox + \w * 2, -\h * \r) node[inner sep=0] {\includegraphics[width=\imgwidth cm]{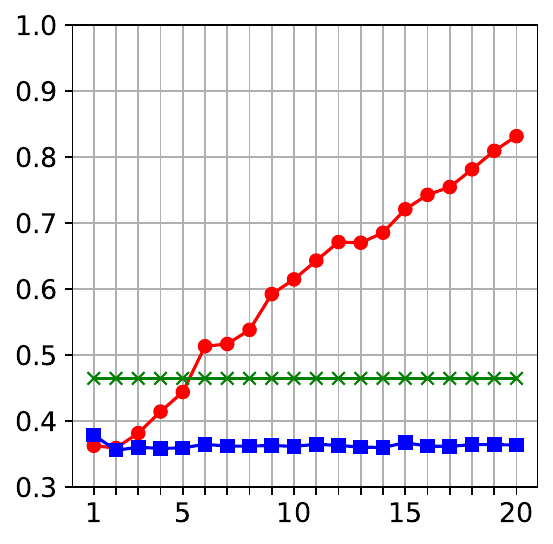}};
        \draw (\ox + \w * 3 ,-\h * \r) node[inner sep=0] {\includegraphics[width=\imgwidth cm]{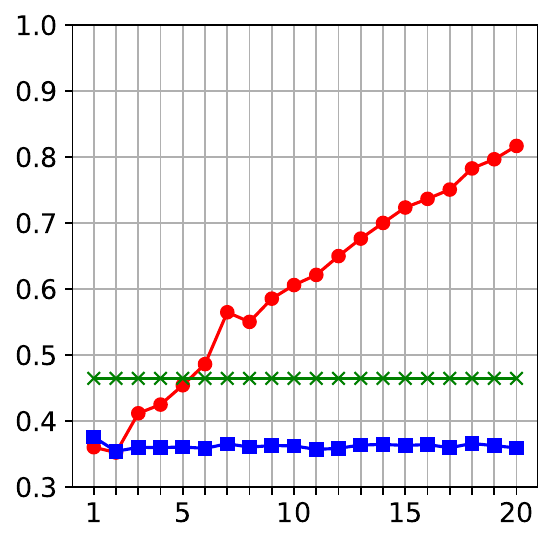}};

        \node at (\ox+ 0.2 + \w * 0, -\h * \r + \imgwidth * 0.45) [above] {\scriptsize{Same-order}};
        \node at (\ox+ 0.2 + \w * 1, -\h * \r + \imgwidth * 0.45) [above] {\scriptsize{Random-order (\textit{seed=0})}};
        \node at (\ox+ 0.2 + \w * 2, -\h * \r + \imgwidth * 0.45) [above] {\scriptsize{Random-order (\textit{seed=1})}};
        \node at (\ox+ 0.2 + \w * 3, -\h * \r + \imgwidth * 0.45) [above] {\scriptsize{Random-order (\textit{seed=2})}};
        
        \node at (-\imgwidth * 0.25, -\h * \r + 0.8, 2) [rotate=90] {\scriptsize{Testing  Error}};

        \node at (\ox + 0.2 + \w * 0, -\h * \r - \imgwidth * 0.45) [below] {\scriptsize{\Setting TTA visit}};
        \node at (\ox + 0.2 + \w * 1, -\h * \r - \imgwidth * 0.45) [below] {\scriptsize{\Setting TTA visit}};
        \node at (\ox + 0.2 + \w * 2, -\h * \r - \imgwidth * 0.45) [below] {\scriptsize{\Setting TTA visit}};
        \node at (\ox + 0.2 + \w * 3, -\h * \r - \imgwidth * 0.45) [below] {\scriptsize{\Setting TTA visit}};

        \node at (0.38 * \linewidth, -\h * \r - \imgwidth * 0.6) [below] {(b) CIFAR-100 $\rightarrow$ CIFAR-100-C task.};
        \draw (\ox + 0.1 + \w * 0.7, -\h * \r -0.3 * \imgwidth, 2) -- (\ox + 0.1 + \w * 0.7, -\h * \r + 0.8 * \imgwidth, 2);
        
        \def \r{2}
        \draw (\ox + \w * 0, -\h * \r) node[inner sep=0] {\includegraphics[width=\imgwidth cm]{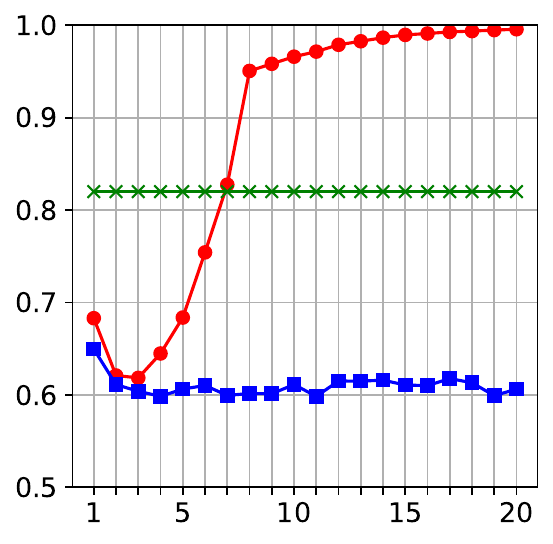}};
        \draw (\ox + \w * 1, -\h * \r) node[inner sep=0] {\includegraphics[width=\imgwidth cm]{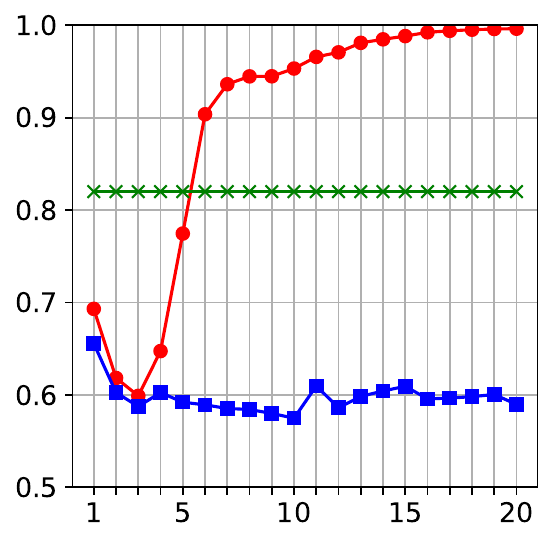}};
        \draw (\ox + \w * 2, -\h * \r) node[inner sep=0] {\includegraphics[width=\imgwidth cm]{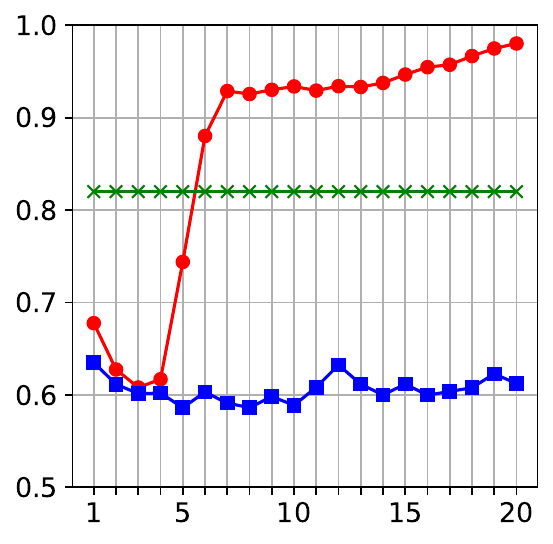}};
        \draw (\ox + \w * 3 ,-\h * \r) node[inner sep=0] {\includegraphics[width=\imgwidth cm]{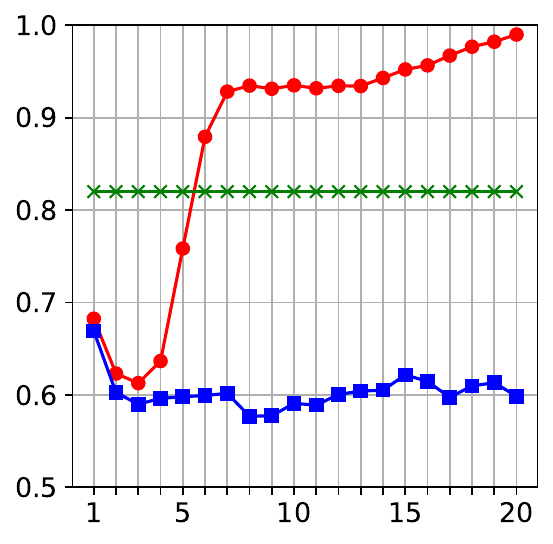}};

        \node at (\ox+ 0.2 + \w * 0, -\h * \r + \imgwidth * 0.45) [above] {\scriptsize{Same-order}};
        \node at (\ox+ 0.2 + \w * 1, -\h * \r + \imgwidth * 0.45) [above] {\scriptsize{Random-order (\textit{seed=0})}};
        \node at (\ox+ 0.2 + \w * 2, -\h * \r + \imgwidth * 0.45) [above] {\scriptsize{Random-order (\textit{seed=1})}};
        \node at (\ox+ 0.2 + \w * 3, -\h * \r + \imgwidth * 0.45) [above] {\scriptsize{Random-order (\textit{seed=2})}};
        
        \node at (-\imgwidth * 0.25, -\h * \r + 0.8, 2) [rotate=90] {\scriptsize{Testing  Error}};

        \node at (\ox + 0.2 + \w * 0, -\h * \r - \imgwidth * 0.45) [below] {\scriptsize{\Setting TTA visit}};
        \node at (\ox + 0.2 + \w * 1, -\h * \r - \imgwidth * 0.45) [below] {\scriptsize{\Setting TTA visit}};
        \node at (\ox + 0.2 + \w * 2, -\h * \r - \imgwidth * 0.45) [below] {\scriptsize{\Setting TTA visit}};
        \node at (\ox + 0.2 + \w * 3, -\h * \r - \imgwidth * 0.45) [below] {\scriptsize{\Setting TTA visit}};

        \node at (0.38 * \linewidth, -\h * \r - \imgwidth * 0.6) [below] {(c) ImageNet $\rightarrow$ ImageNet-C task.};
        \draw (\ox + 0.1 + \w * 0.7, -\h * \r -0.3 * \imgwidth, 2) -- (\ox + 0.1 + \w * 0.7, -\h * \r + 0.8 * \imgwidth, 2);
        \end{tikzpicture}
    \vspace*{-0.5\baselineskip}
    \caption{\Setting TTA with different order of corruptions. This figure plots the testing error of two TTA approaches: \textcolor{red}{RoTTA -\mycircle{red}- } ~\cite{yuan2023robust}, and, \textcolor{blue}{\method -\mysquare{blue}-} \textit{(ours)}, and \textcolor{ForestGreen}{source model-$\times$-} as a reference performance under our \setting TTA (with 20 visits) across three TTA tasks. On the \textit{same-order} experiments (column 1), the same order of image corruptions is applied for all 20 visits. Meanwhile, in \textit{random-order}, this order is reshuffled at the beginning of each visit (columns 2-4). Random-order experiments are redone three times with different random seeds. Here, we empirically validate that using the same order of domain shifts (image corruptions) in our \setting TTA is sufficient to showcase the model collapse and evaluate the persistence of our \method. Best viewed in color.}
    \label{fig:random_order_comparision}
\end{figure}

Recall that in \Sec{\ref{ssec:model_collapse}}, \textit{\setting TTA} is constructed by repeating \textit{the same} sequence of $D$ distributions $K$ times.  For example, a sequence with $K=2$ could be $\mathcal{P}_1 \rightarrow  \mathcal{P}_2 \rightarrow \cdots  \rightarrow \mathcal{P}_D \rightarrow \mathcal{P}_1 \rightarrow  \mathcal{P}_2 \rightarrow \cdots \rightarrow \mathcal{P}_D$.
For simplicity and consistency that promote reproducibility, the \textit{same order of image corruptions} (following~\cite{yuan2023robust}) is used for all recurrences. This section presents supplementary experimental findings indicating that \textit{the order of image corruptions} within each recurrence, indeed, \textit{does not affect} the demonstration of TTA model collapse and the performance of our \method. 

\noindent \textbf{Experiment Setup.} We refer to the setting \textit{same-order} as using one order of image corruptions in~\cite{yuan2023robust} for all recurrences (specifically, on CIFAR-10/100-C and ImageNet-C: \textit{motion $\rightarrow$ snow $\rightarrow$ fog $\rightarrow$ shot $\rightarrow$ defocus $\rightarrow$ contrast $\rightarrow$ zoom $\rightarrow$ brightness $\rightarrow$ frost $\rightarrow$ elastic $\rightarrow$ glass $\rightarrow$ gaussian $\rightarrow$ pixelated $\rightarrow$ jpeg $\rightarrow$ impulse}). 
Conversely, in \textit{random-order}, the order of image corruptions is randomly shuffled at the beginning of each recurrence. Hence, the corruption orders across $K$ recurrences are now entirely different. We redo the experiment of the second setting three times (with different random seeds = $0, 1, 2$). 
Nevertheless, different TTA methods are ensured to be evaluated on the same testing stream, since it is fixed after generation. Without updating its parameters, the performance of the \textit{source model} is trivially independent of the order of corruptions. 

\noindent \textbf{Experimental Result.} The experimental results are visualized in \Fig{\ref{fig:random_order_comparision}}. The first column plots the experiments under the \textit{same-order}, while the remaining three columns plot the experiments in the \textit{random-order} setting, with varying random seeds. Note that the message conveyed by each sub-figure entirely matches that of \Fig{\ref{fig:fig1_introduction}-right}.

\noindent \textbf{Discussions.} Clearly, a similar collapsing pattern is observed in all three TTA tasks, with three combinations of 20 image corruption orders. This pattern also matches the easiest setting using the \textit{same order} of image corruptions we promoted in \textit{\setting TTA}.

\section{Further Justifications on Persistent TTA (\method)}
\subsection{Pseudo Code}
\label{appdx:petta_pseudo_code}
We summarize the key steps of our proposed \method in \Alg{\ref{alg:petta}}, with the key part (lines 4-13) \textcolor{blue}{highlighted in blue}. Our approach fits well in the general workflow of a TTA algorithm, \textit{enhancing the regular mean-teacher update step}. \Appdx{\ref{appdx:novelty_petta}} elaborates more on our contributions in \method, distinguishing them from other components proposed in previous work. 
The notations and definitions of all components follow the main text (described in detail in \Sec{\ref{ssec:pers_tta}}). On line 8 of Alg.~\ref{alg:petta}, as a shorthand notation, $\phi_{\theta_{t-1}}(X_t |\hat Y_t = y)$ denotes the empirical mean of all feature vectors of $X_t^{(i)}$ (extracted by $\phi_{\theta_{t-1}} \left(X_t^{(i)}\right)$) if $\hat Y_t^{(i)} = y, i = 1, \cdots, N_t$ in the current testing batch.  
\begin{algorithm}[t]
    \LinesNumbered
    \KwIn{Classification model $f_t$ and its deep feature extractor $\phi_{\theta_t}$, both parameterized by $\theta_t \in \Theta$. Testing stream $\left\{X_t\right\}_{t=0}^T$, initial model parameter $(\theta_0)$, initial update rate $(\alpha_0)$, regularization term coefficient $(\lambda_0)$, empirical mean $\left(\{\boldsymbol{{\mu}}_0^y\}_{y \in \mathcal{Y}}\right)$ and covariant matrix $\left( \{\boldsymbol{\Sigma}_0^y \}_{y \in \mathcal{Y}}\right)$ of feature vectors in the training set, $\boldsymbol{\hat{\mu}}_t^y$ EMA update rate $(\nu)$.}

    $\boldsymbol{\hat{\mu}}_0^y \leftarrow \boldsymbol{{\mu}}_0^y, \forall y \in \mathcal{Y}$ \tcp*[r]{Initialization}
    
    \For{$t \in [1, \cdots, T]$}{ 

        $\hat Y_t \leftarrow f_{t-1}(X_t)$  \tcp*[r]{Obtaining pseudo-labels for all samples in $X_t$}

        \tikzmk{StartHere}
        \textcolor{blue}{\tcp{Persistent TTA (\method)}}
        $\hat{\mathcal{Y}}_t \leftarrow \left\{\hat Y_t^{(i)} | i=1, \cdots, N_t\right\}$ \tcp*[r]{Set of (unique) pseudo-labels in $X_t$}

        $\bar \gamma_t \leftarrow 0$ ;
        
        \For{$y \in \hat{\mathcal{Y}_t}$}{
            $\gamma_{t}^y \leftarrow 1 - \exp\left(-(\boldsymbol{\hat{\mu}}_t^y - \boldsymbol{\mu}_0^y)^T \left(\boldsymbol{\Sigma}^y_0\right)^{-1} (\boldsymbol{\hat{\mu}}_t^y - \boldsymbol{\mu}_0^y) \right)$
            \tcp*[r]{Divergence sensing term on category $y$}
            
            $\bar \gamma_t \leftarrow \bar \gamma_t + \frac{\gamma_t^{y}}{|\hat{\mathcal{Y}}_t|}$
            \tcp*[r]{Average divergence sensing term for step $t$}

            $\boldsymbol{\hat{\mu}}_t^y \leftarrow (1-\nu) \boldsymbol{\hat{\mu}}_{t-1}^y + \nu \phi_{\theta_{t-1}}(X_t |\hat Y_t = y)$
            \tcp*[r]{EMA update of $\boldsymbol{\hat{\mu}}_t^y$ for samples with $\hat Y_t = y$ }
            
        }

        $\lambda_t \leftarrow \bar \gamma_t \cdot \lambda_0$ \tcp*[r]{Computing adaptive regularization term coefficient}

        $\alpha_t \leftarrow (1 - \bar \gamma_t) \cdot \alpha_0$ \tcp*[r]{Computing adaptive update rate}
        \tikzmk{EndHere}
        \boxit{codehighlightblue}
        \textcolor{blue}{\tcp{Regular Mean-teacher Update}}

        $\theta_t' \leftarrow \underset{\theta' \in \Theta}{\text{\texttt{Optim} }}  \mathbb{E}_{P_t}\left[\mathcal{L}_{\mathrm{CLS}} \left(\hat{Y}_t, X_t; \theta'\right) + \mathcal{L}_{\mathrm{AL}} \left(X_t; \theta'\right) \right] + \lambda_t \mathcal{R}(\theta')$ \tcp*[r]{Student model update}

        $\theta_t \leftarrow (1-\alpha_t) \theta_{t-1}  + \alpha_t \theta'_t.$ \tcp*[r]{Teacher model update}

        \textcolor{blue}{\tcp{Final prediction}}
        
            \textbf{$\texttt{yeild } f_t(X_t)$} \tcp*[r]{Returning the final inference with updated model $f_t$}
    }
    \caption{Persistent TTA (\method)}
    \label{alg:petta}
\end{algorithm}

\subsection{Anchor Loss}
\textbf{KL Divergence Minimization-based Interpretation of Anchor Loss}. In \Sec{\ref{ssec:pers_tta}}, we claimed that minimizing the anchor loss $\mathcal{L}_{\mathrm{AL}}$ is equivalent to minimizing the relative entropy (or KL divergence) between the output probability of two models parameterized by $\theta_0$ and $\theta$.
\begin{proof} Having:
\begin{align*}
    D_{KL}\left(\Pr (y | X_t; \theta_0) || \Pr (y | X_t; \theta)\right)
    &=  \sum_{y \in \mathcal{Y}} \Pr (y | X_t; \theta_0) \log \frac{\Pr (y | X_t; \theta_0)}{\Pr (y | X_t; \theta)} \\ 
    &= \underbrace{-\sum_{y \in \mathcal{Y}}  \Pr (y | X_t; \theta_0)  \log \Pr (y | X_t; \theta)}_{\mathcal{L}_{\mathrm{AL}}(X_t; \theta)} - \underbrace{H(\Pr (y | X_t; \theta_0))}_{\text{constant}}.
\end{align*}    
Hence,
\begin{align*}
    \underset{\theta \in \Theta}{\text{argmin }} \mathcal{L}_{\mathrm{AL}}(X_t; \theta) =  \underset{\theta \in \Theta}{\text{argmin }}  D_{KL}\left(\Pr (y | X_t; \theta_0) || \Pr (y | X_t; \theta)\right).
\end{align*}%
\end{proof}
Intuitively, a desirable TTA solution should be able to adapt to novel testing distributions on the one hand, but it should \textit{not} significantly diverge from the initial model. $\mathcal{L}_{\mathrm{AL}}$ fits this purpose, constraining the KL divergence between two models at each step.

\noindent \textbf{Connections between Anchor Loss and Regularizer Term. } While supporting the same objective (collapse prevention by avoiding the model significantly diverging from the source model), the major difference between Anchor loss ($\mathcal{L}_{\mathrm{AL}}$) and the Regularizer term ($\mathcal{R}(\theta)$) is that the anchor loss operates on the probability space of model prediction while the regularizer term works on the model parameter spaces. \Tab{\ref{tab:ablation_study}} (lines 1 and 5) summarizes the ablation study when each of them is eliminated. We see the role of the regularization term is crucial for avoiding model collapse, while the anchor loss guides the adaptation under the drastic domain shift.
Nevertheless, fully utilizing all components is suggested for maintaining TTA persistence.

\subsection{The Use of the Memory Bank}
\label{appdx:using_mem_bank}
\noindent \textbf{The size of Memory Bank.} The size of the memory bank in \method is \textit{relatively small, equal to the size of one mini-batch for update} (64 images, specifically).

\noindent \textbf{The Use of the Memory Bank in \method is Fair with Respect To the Compared Methods.}
Our directly comparable method - RoTTA~\cite{yuan2023robust} also takes this advantage (referred to as category-balanced sampling, Sec. 3.2 of~\cite{yuan2023robust}). Hence, the comparison between \method and RoTTA \textit{is fair} in terms of additional memory usage.
Noteworthy, the use of a memory bank is a \textit{common practice} in TTA literature (e.g.,~\cite{gong2022note, chen2022contrastive, yuan2023robust}), especially in situations where the class labels are temporally correlated or non-i.i.d. distributed (as we briefly summarized in \Appdx{\ref{sec:related_work}} - Related Work section).
CoTTA~\cite{Wang_2022_CVPR}, EATA~\cite{niu2022efficient} and MECTA~\cite{hong2023mecta} (compared method) assume labels are i.i.d. distributed. Hence, a memory bank is unnecessary, but their performance under temporally correlated label distribution has dropped significantly as a trade-off.
The RMT~\cite{döbler2023robust} (compared method) does not require a memory bank but it needs to cache a portion of the source training set for replaying (Sec. 3.3 in~\cite{döbler2023robust}) which even requires \textit{more} resources than the memory bank.

\noindent \textbf{Eliminating the Need for a Memory Bank.} As addressing the challenge of temporally correlated label distribution on the testing stream is not the focus of \method, we have conveniently adopted the use of the memory bank proposed in~\cite{yuan2023robust}. Since this small additional memory requirement is not universally applied in every real-world scenario, we believe that this is a reasonable assumption, and commonly adopted in TTA practices.
Nevertheless, exploring alternative ways for reducing the memory size (e.g., storing the embedded features instead of the original image) would be an interesting future direction.

\subsection{Empirical Mean and Covariant Matrix of Feature Vectors on the Source Dataset}
\label{appdx:mean_cov_source_dataset}

\noindent \textbf{Two Ways of Computing $\mu_0^y$ and $\Sigma_0^y$ in Practice.}
One may notice that in \method, computing $\gamma_t^y$ requires the \textit{pre-computed empirical mean ($\mu_0^y$) and covariance ($\Sigma_0^y$) of the source dataset}. This requirement may not be met in real-world situations where the source data is unavailable. In practice, the empirical mean and covariance matrix computed on the source distribution can be provided in the following two ways:
\begin{enumerate}
    \item Most ideally, these values are computed directly by inference on the entire training set once the model is fully trained. They will be provided alongside the source-distribution pre-trained model as a pair for running TTA.
    \item With only the source pre-trained model available, assume we can sample a set of unlabeled data from the source distribution. The (pseudo) labels for them are obtained by inferring from the source model. Since the source model is well-performed in this case, using pseudo is approximately as good as the true label. 
\end{enumerate}

\noindent \textbf{Accessing the Source Distribution Assumption in TTA. }
In fact, the second way is typically assumed to be possible in previous TTA methods such as EATA~\cite{niu2022efficient}, and MECTA~\cite{hong2023mecta} (a compared method) to estimate a Fisher matrix (for anti-forgetting regularization purposes).
Our work - \method \textit{follows the same second setup} as the previous approaches mentioned above.
A variation of RMT~\cite{döbler2023robust} (a compared method) approach even requires having the fully labeled source data available at test-time for source replaying (Sec. 3.3 of~\cite{döbler2023robust}). This variation is used for comparison in our experiments.

We believe that having the empirical mean and covariant matrix pre-computed on a portion of the source distribution in \method \textit{is a reasonable assumption}. Even in the ideal way, revealing the statistics might not severely violate the risk of data privacy leakage or require notable additional computing resources.

\noindent \textbf{Number of Samples Needed for Computation.}
To elaborate more on the feasibility of setting (2) mentioned above, we perform a small additional experiment on the performance of \method while varying the number of samples used for computing the empirical mean and covariant matrix on the source distribution.
In this setting, we use the test set of CIFAR-10, CIFAR-100, DomainNet validation set of ImageNet (original images, without corruption, or the \textit{real} domain test set of DomainNet),  representing samples from the source distribution. The total number of images is $10,000$ in CIFAR-10/A00, $50,000$ in ImageNet, and $69,622$ in DomainNet. We randomly sample $25\%, 50\%, 75\%$, and $100\%$ of the images in this set to run \method for 20 rounds of \setting. The result is provided in \Tab{\ref{tab:source_data_portion}} below. 

\begin{table}[h]
    \centering
    \caption{ Average classification error of PeTTA (across 20 visits) with varying sizes of source samples used for computing feature empirical mean ($\mu_0^y$) and covariant matrix ($\Sigma_0^y$).}
    \label{tab:source_data_portion}
    \resizebox{0.7\textwidth}{!}{
    \begin{tabular}{c|ccc|c}
        \toprule
        \textbf{TTA Task} & \textbf{25\%} & \textbf{50\%}	&  \textbf{75\%}	& \textbf{100\%} \\
        \midrule
        CIFAR-10 $\rightarrow$ CIFAR-10-C   & 22.96 & 22.99 & 23.03 & 22.75 \\
        CIFAR-100 $\rightarrow$ CIFAR-100-C & 35.01	& 35.11	& 35.09	& 35.15 \\
        DomainNet: \textit{real} $\rightarrow$ \textit{clip} $\rightarrow$ \textit{paint} $\rightarrow$ \textit{sketch} & 43.18 & 43.12 & 43.15 & 42.89 \\
        ImageNet $\rightarrow$ ImageNet-C   & 61.37 & 59.68 & 61.05 & 60.46 \\
        
        \bottomrule
    \end{tabular}}
\end{table}
The default choice of \method is using $100\%$ samples of the validation set of the source dataset. However, we showcase that it is possible to reduce the number of unlabeled samples from the source distribution to compute the empirical mean and covariant matrix for \method, without significantly impacting its performance.

\subsection{Novelty of \method}
\label{appdx:novelty_petta}
\method is composed of multiple components. Among them, the anchor loss is an existing idea (examples of previous work utilizing this idea are~\cite{li2016_learning_wo_forgetting, döbler2023robust}). Similarly, the mean-teacher update; and regularization are well-established techniques and very useful for the continual or gradual TTA scenario. Hence, we do not aim to improve or alternate these components.

Nevertheless, the novelty of our contribution is the \textit{sensing of the divergence and adaptive model update}, in which the importance of minimizing the loss (adaptation) and regularization (collapse prevention) is changed adaptively. In short, we propose a harmonic way of combining those elements adaptively to achieve a persistent TTA process. 

The design of \method draws inspiration from a theoretical analysis (\Sec{\ref{ssec:egmmc_analysis}}), empirically surpassing both the conventional reset-based approach~\cite{press2023rdumb} (\Appdx{\ref{appdx:model_reset}}) and other continual TTA approaches~\cite{yuan2023robust, döbler2023robust, Wang_2022_CVPR, hong2023mecta, malik2022_paramter-free} on our proposed \setting TTA (\Sec{\ref{ssec:model_collapse}}, \Appdx{\ref{appdx:additional_main_results}}), as well as the previously established CCC~\cite{press2023rdumb} benchmark. 

\section{Additional Experimental Results of PeTTA}

\subsection{Performance of \method Versus Compared Methods}
\label{appdx:additional_main_results}

\noindent \textbf{Performance on CIFAR-100-C and Domainnet Datasets.} Due to the length constraint, the classification errors on the tasks CIFAR-100$\rightarrow$CIFAR-100-C, and \textit{real} $\rightarrow$ \textit{clipart, painting, sketch} of DomainNet are provided in \Tab{\ref{tab:cifar-100-performance}} and  \Tab{\ref{tab:domainnet-performance}}. To prevent model collapse, the adaptability of \method is more constrained. As a result, it requires more time for adaptation initially (e.g., in the first visit) but remains stable thereafter.  Generally, consistent trends and observations are identified across all four TTA tasks. 
\begin{table*}[ht!]
    \caption{Average classification error of the task CIFAR-100 $\rightarrow$ CIFAR-100-C in \textit{\setting TTA} scenario. The lowest error is highlighted in \textbf{bold}, $^{(*)}$average value across 5 runs (different random seeds) is reported for \method.}
    \label{tab:cifar-100-performance}
    \resizebox{\textwidth}{!}{
    \begin{tabular}{r|cccccccccccccccccccc|c}
\toprule
& \multicolumn{20}{l}{  \textit{\Setting TTA visit} $\xrightarrow{\hspace*{5cm}}$ } \\
\textbf{Method} &              1 &              2 &              3 &              4 &              5 &              6 &              7 &              8 &              9 &             10 &             11 &             12 &             13 &             14 &             15 &             16 &             17 &             18 &             19 &             20 &   \textbf{Avg} \\
\midrule
Source                                      & \multicolumn{20}{c|}{46.5} & 46.5 \\
\midrule
LAME~\cite{malik2022_paramter-free}   & \multicolumn{20}{c|}{40.5} & 40.5 \\
\midrule
CoTTA~\cite{Wang_2022_CVPR} &           53.4 &           58.4 &           63.4 &           67.6 &           71.4 &           74.9 &           78.2 &           81.1 &           84.0 &           86.7 &           88.8 &           90.7 &           92.3 &           93.5 &           94.7 &           95.6 &           96.3 &           97.0 &           97.3 &           97.6 &           83.1 \\
EATA~\cite{niu2022efficient}  &  88.5 &  95.0 &  96.8 &  97.3 &  97.4 &  97.2 &  97.2 &  97.3 &  97.4 &  97.5 &  97.5 &  97.5 &  97.6 &  97.7 &  97.7 &  97.7 &  97.8 &  97.8 &  97.7 &  97.7 &          96.9 \\
RMT~\cite{döbler2023robust} &           50.5 &           48.6 &           47.9 &           47.4 &           47.3 &           47.1 &           46.9 &           46.9 &           46.6 &           46.8 &           46.7 &           46.5 &           46.5 &           46.6 &           46.5 &           46.5 &           46.5 &           46.5 &           46.5 &           46.5 &           47.1 \\
MECTA~\cite{hong2023mecta}  &           44.8 &           44.3 &           44.6 &           43.1 &           44.8 &           44.2 &           44.4 &           43.8 &           43.8 &           43.9 &           44.6 &           43.8 &           44.4 &           44.6 &           43.9 &           44.2 &           43.8 &           44.4 &           44.9 &           44.2 &           44.2 \\
RoTTA~\cite{yuan2023robust} &  35.5 &           35.2 &           38.5 &           41.9 &           45.3 &           49.2 &           52.0 &           55.2 &           58.1 &           61.5 &           64.6 &           67.5 &           70.7 &           73.2 &           75.4 &           77.1 &           79.2 &           81.5 &           82.8 &           84.5 &           61.4 \\
RDumb~\cite{press2023rdumb} &           36.7 &           36.7 &           36.6 &           36.6 &           36.7 &           36.8 &           36.7 &           36.5 &           36.6 &           36.5 &           36.7 &           36.6 &           36.5 &           36.7 &           36.5 &           36.6 &           36.6 &           36.7 &           36.6 &           36.5 &           36.6 \\
ROID~\cite{marsden2024universal}  &  76.4 &  76.4 &  76.2 &  76.2 &  76.3 &  76.1 &  75.9 &  76.1 &  76.3 &  76.3 &  76.6 &  76.3 &  76.8 &  76.7 &  76.6 &  76.3 &  76.2 &  76.0 &  75.9 &  76.0 &          76.3 \\
TRIBE~\cite{Su_Xu_Jia_2024} &  \textbf{33.8} &  \textbf{33.3} &  35.3 &  \textbf{34.9} &  35.3 &  \textbf{35.1} &  37.1 &  37.2 &  37.2 &  39.1 &  39.2 &  41.1 &  41.0 &  43.1 &  45.1 &  45.1 &  45.0 &  44.9 &  44.9 &  44.9 &          39.6 \\
\hline
\rowcolor{ClrHighlight}
\method \textit{(ours)}$^{(*)}$     & 35.8 &  34.4 &  \textbf{34.7} &  35.0 &  \textbf{35.1} &  \textbf{35.1} &  \textbf{35.2} &  \textbf{35.3} &  \textbf{35.3} &  \textbf{35.3} &  \textbf{35.2} &  \textbf{35.3} &  \textbf{35.2} &  \textbf{35.2} &  \textbf{35.1} &  \textbf{35.2} &  \textbf{35.2} &  \textbf{35.2} &  \textbf{35.2} &  \textbf{35.2} &  \textbf{35.1} \\
\bottomrule
\end{tabular}

   }
\end{table*}
\begin{table*}[ht!]
    \caption{Average classification error of the task \textit{real} $\rightarrow$ \textit{clipart} $\rightarrow$ \textit{painting} $\rightarrow$ \textit{sketch} on DomainNet dataset in \textit{\setting TTA} scenario.}
    \label{tab:domainnet-performance}
    \resizebox{\textwidth}{!}{
    \begin{tabular}{r|cccccccccccccccccccc|c}
\toprule
& \multicolumn{20}{l}{  \textit{Episodic TTA visit} $\xrightarrow{\hspace*{5cm}}$ } \\
\textbf{Method} &              1 &              2 &              3 &              4 &              5 &              6 &              7 &              8 &              9 &             10 &             11 &             12 &             13 &             14 &             15 &             16 &             17 &             18 &             19 &             20 &   \textbf{Avg} \\

\midrule
Source                                      & \multicolumn{20}{c|}{45.3} & 45.3 \\
\midrule
LAME~\cite{malik2022_paramter-free}   & \multicolumn{20}{c|}{45.6} & 45.6 \\
\midrule
CoTTA~\cite{Wang_2022_CVPR} &           96.2 &           97.1 &           97.4 &           97.8 &           98.1 &           98.2 &           98.4 &           98.4 &           98.4 &           98.5 &           98.6 &           98.6 &           98.6 &           98.6 &           98.6 &           98.7 &           98.7 &           98.7 &           98.7 &           98.7 &           98.3 \\
RMT~\cite{döbler2023robust} &           76.2 &           77.1 &           77.3 &           77.3 &           77.2 &           77.1 &           76.8 &           76.9 &           76.5 &           76.4 &           76.4 &           76.3 &           76.4 &           76.2 &           76.2 &           76.1 &           76.4 &           76.1 &           76.0 &           75.8 &           76.5 \\
MECTA~\cite{hong2023mecta}  &           94.6 &           98.4 &           98.6 &           98.8 &           99.1 &           99.0 &           99.0 &           99.0 &           99.0 &           99.0 &           99.0 &           99.0 &           99.0 &           99.0 &           99.0 &           99.0 &           99.0 &           99.0 &           99.0 &           99.0 &           98.7 \\
RoTTA~\cite{yuan2023robust} &           44.3 &           43.8 &           44.7 &           46.7 &           48.7 &           50.8 &           52.7 &           55.0 &           57.1 &           59.7 &           62.7 &           65.1 &           68.0 &           70.3 &           72.7 &           75.2 &           77.2 &           79.6 &           82.6 &           85.3 &           62.1 \\
RDumb~\cite{press2023rdumb} &           44.3 &           44.4 &           44.3 &           44.5 &           44.2 &           44.2 &           44.3 &           44.5 &           44.4 &           44.2 &           44.3 &           44.3 &           44.3 &           44.3 &           44.5 &           44.3 &           44.2 &           44.3 &           44.4 &           44.3 &           44.3 \\

\hline
\rowcolor{ClrHighlight}
\method \textit{(ours)}$^{(*)}$ &  \textbf{43.8} &  \textbf{42.6} &  \textbf{42.3} &  \textbf{42.3} &  \textbf{42.6} &  \textbf{42.8} &  \textbf{42.8} &  \textbf{43.0} &  \textbf{42.9} &  \textbf{42.9} &  \textbf{43.1} &  \textbf{43.0} &  \textbf{42.9} &  \textbf{43.0} &  \textbf{43.0} &  \textbf{43.1} &  \textbf{43.0} &  \textbf{42.8} &  \textbf{42.9} &  \textbf{42.9} &  \textbf{42.9} \\

\bottomrule
\end{tabular}

   }
   \vspace*{-0.5\baselineskip}
\end{table*}

\noindent \textbf{Standard Deviation of \method Performance Across Multiple Runs. } For \method experiments marked with (*) in \Tab{\ref{tab:cifar-10-performance}}, \Tab{\ref{tab:imagenet-c-performance}}, \Tab{\ref{tab:cifar-100-performance}}, and \Tab{\ref{tab:domainnet-performance}}, the average performance across five independent runs with different random seeds is reported. Due to the space constraint, the corresponding standard deviation values are now reported in \Tab{\ref{tab:std-petta-performance}}.
Generally, the average standard deviation across runs stays within $\pm 0.1 \%$ for small datasets (CIFAR-10-C, CIFAR-100-C) and $\pm 0.5\%$ for larger datasets (ImageNet-C, DomainNet).

\begin{table*}[ht!]
    \caption{Mean and standard deviation classification error of \method on the four datasets: CIFAR-10-C (CF-10-C), CIFAR-100-C (CF-100-C), DomainNet (DN), and ImageNet-C (IN-C) with \textit{\setting TTA} scenario. Each experiment is run 5 times with different random seeds. }
    \label{tab:std-petta-performance}
    \resizebox{\linewidth}{!}{
    \addtolength{\tabcolsep}{-0.4em}
    \begin{tabular}{r|cccccccccccccccccccc|c}
    \toprule
    & \multicolumn{20}{l}{ \textit{\Setting TTA visit} $\xrightarrow{\hspace*{5cm}}$ } \\
    \textbf{Dataset} &              1 &              2 &              3 &              4 &              5 &              6 &              7 &              8 &              9 &             10 &             11 &             12 &             13 &             14 &             15 &             16 &             17 &             18 &             19 &             20 &   \textbf{Avg} \\
    \midrule
    \multirow{2}{*}{CF-10-C} &       24.3 &       23.0 &       22.6 &       22.4 &       22.4 &       22.5 &       22.3 &       22.5 &       22.8 &       22.8 &       22.6 &       22.7 &       22.7 &       22.9 &       22.6 &       22.7 &       22.6 &       22.8 &       22.9 &       23.0 &         22.8 \\
    &  $\pm 0.4$ &  $\pm 0.3$ &  $\pm 0.4$ &  $\pm 0.3$ &  $\pm 0.3$ &  $\pm 0.3$ &  $\pm 0.4$ &  $\pm 0.2$ &  $\pm 0.3$ &  $\pm 0.4$ &  $\pm 0.4$ &  $\pm 0.2$ &  $\pm 0.1$ &  $\pm 0.3$ &  $\pm 0.5$ &  $\pm 0.2$ &  $\pm 0.2$ &  $\pm 0.3$ &  $\pm 0.4$ &  $\pm 0.5$ &    $\pm 0.1$ \\
    
    \midrule
    \multirow{2}{*}{CF-100-C}  &       35.8 &       34.4 &       34.7 &       35.0 &       35.1 &       35.1 &       35.2 &       35.3 &       35.3 &       35.3 &       35.2 &       35.3 &       35.2 &       35.2 &       35.1 &       35.2 &       35.2 &       35.2 &       35.2 &       35.2 &         35.1 \\
    &  $\pm 0.4$ &  $\pm 0.4$ &  $\pm 0.2$ &  $\pm 0.2$ &  $\pm 0.1$ &  $\pm 0.1$ &  $\pm 0.2$ &  $\pm 0.2$ &  $\pm 0.1$ &  $\pm 0.2$ &  $\pm 0.1$ &  $\pm 0.2$ &  $\pm 0.2$ &  $\pm 0.1$ &  $\pm 0.1$ &  $\pm 0.1$ &  $\pm 0.1$ &  $\pm 0.1$ &  $\pm 0.2$ &  $\pm 0.2$ &    $\pm 0.1$ \\
    
    \midrule
    \multirow{2}{*}{DN} &       43.8 &       42.6 &       42.3 &       42.3 &       42.6 &       42.8 &       42.8 &       43.0 &       42.9 &       42.9 &       43.1 &       43.0 &       42.9 &       43.0 &       43.0 &       43.1 &       43.0 &       42.8 &       42.9 &       42.9 &         42.9 \\
    &  $\pm 0.1$ &  $\pm 0.1$ &  $\pm 0.2$ &  $\pm 0.2$ &  $\pm 0.3$ &  $\pm 0.3$ &  $\pm 0.3$ &  $\pm 0.4$ &  $\pm 0.4$ &  $\pm 0.4$ &  $\pm 0.4$ &  $\pm 0.4$ &  $\pm 0.4$ &  $\pm 0.3$ &  $\pm 0.3$ &  $\pm 0.2$ &  $\pm 0.4$ &  $\pm 0.3$ &  $\pm 0.3$ &  $\pm 0.3$ &    $\pm 0.3$ \\

    \midrule
    
    \multirow{2}{*}{IN-C}&       65.3 &       61.7 &       59.8 &       59.1 &       59.4 &       59.6 &       59.8 &       59.3 &       59.4 &       60.0 &       60.3 &       61.0 &       60.7 &       60.4 &       60.6 &       60.7 &       60.8 &       60.7 &       60.4 &       60.2 &         60.5 \\
    &  $\pm 0.6$ &  $\pm 0.5$ &  $\pm 0.5$ &  $\pm 0.5$ &  $\pm 1.4$ &  $\pm 1.1$ &  $\pm 1.0$ &  $\pm 0.5$ &  $\pm 0.8$ &  $\pm 0.9$ &  $\pm 0.4$ &  $\pm 0.8$ &  $\pm 0.9$ &  $\pm 0.8$ &  $\pm 0.9$ &  $\pm 0.8$ &  $\pm 1.0$ &  $\pm 0.6$ &  $\pm 0.6$ &  $\pm 0.7$ &    $\pm 0.5$ \\

    \bottomrule
    \end{tabular}
    \addtolength{\tabcolsep}{0.4em}
   }
\end{table*}

\subsection{An Inspection of \method}
\begin{figure}[ht!]
    \pgfplotsset{every x tick label/.append style={font=\tiny, yshift=0.5ex}}
    \pgfplotsset{every y tick label/.append style={font=\tiny, xshift=0.5ex}}
    \centering    
        \begin{tikzpicture}

        \draw (0, 0) node[inner sep=0] {\includegraphics[width=\linewidth]{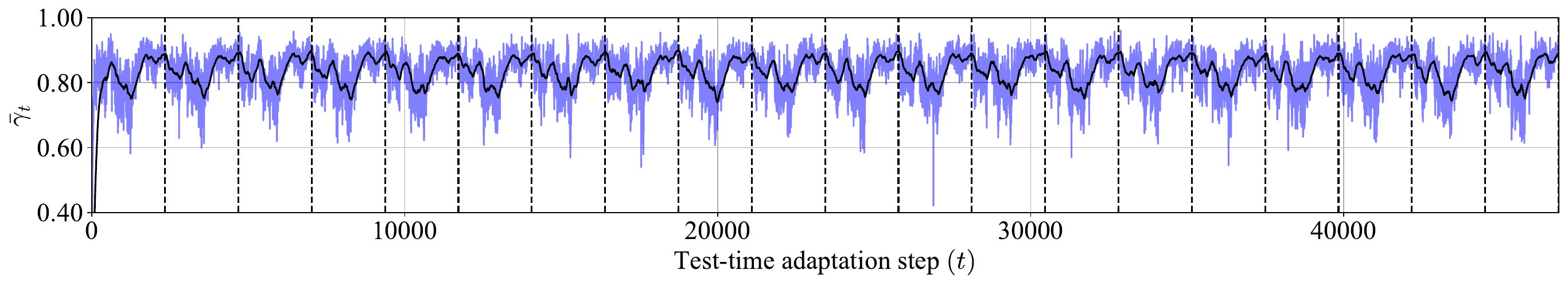}};
        \draw (0, -2.6) node[inner sep=0] {\includegraphics[width=\linewidth]{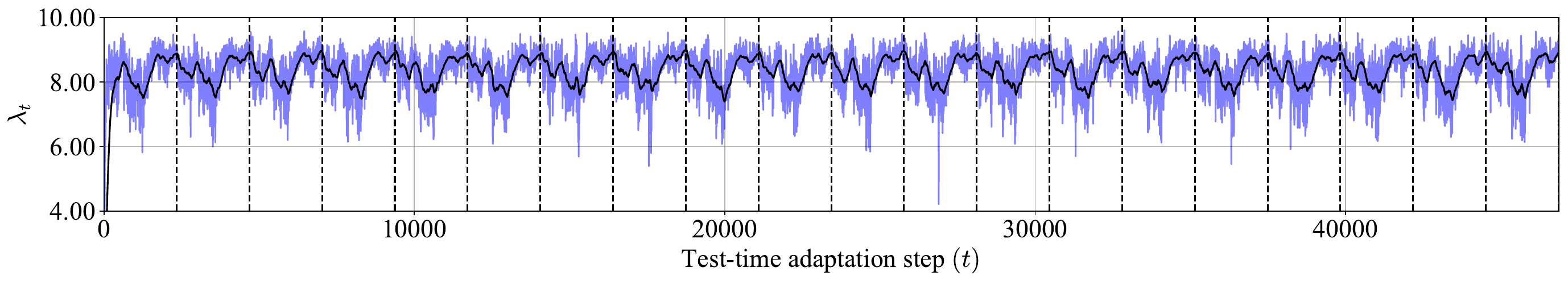}};
        \draw (0, -5.6) node[inner sep=0] {\includegraphics[width=\linewidth]{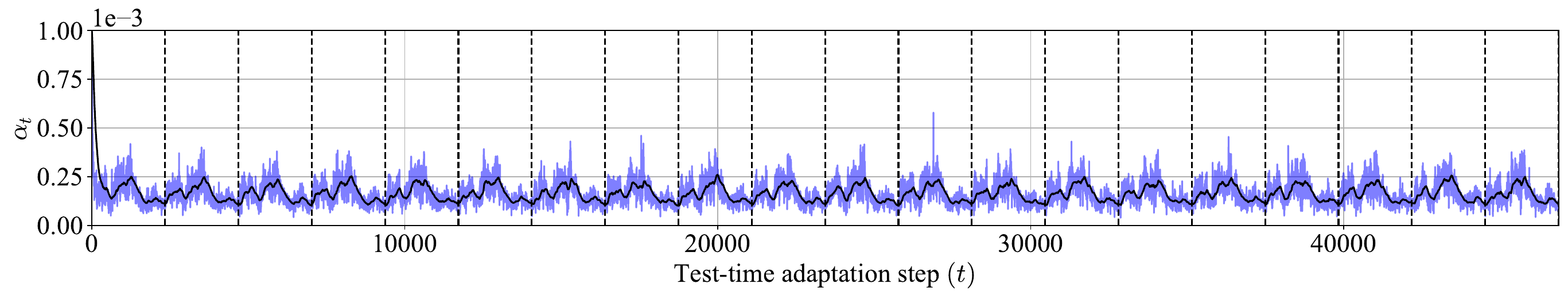}};

        \draw (0, -8.6) node[inner sep=0] {\includegraphics[width=\linewidth]{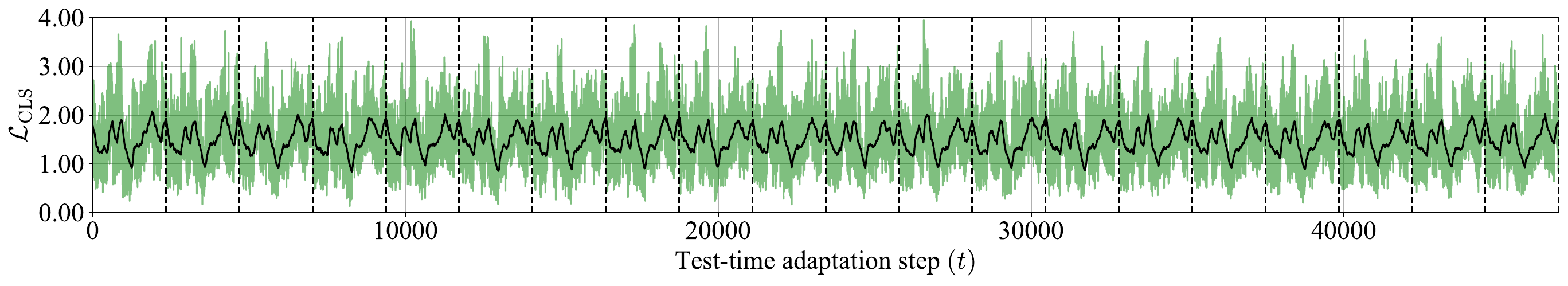}};
        \draw (0, -11.6) node[inner sep=0] {\includegraphics[width=\linewidth]{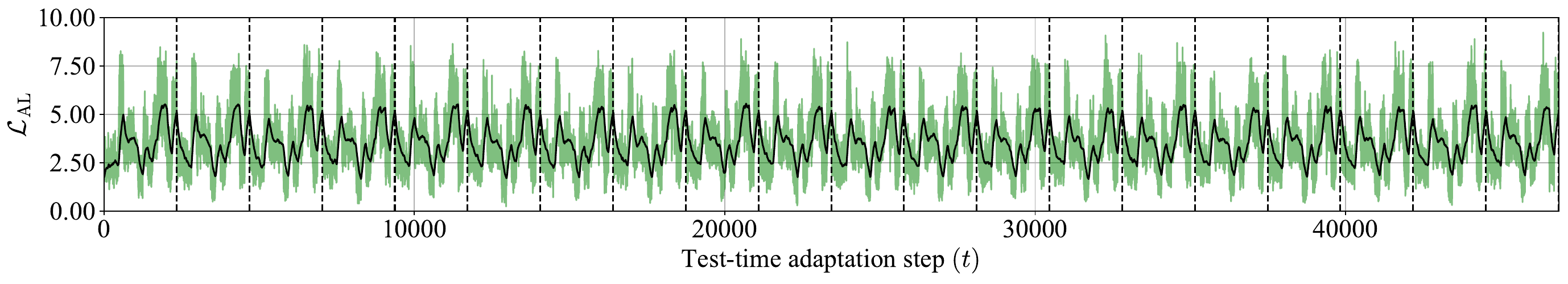}};
        \draw (0, -14.6) node[inner sep=0] {\includegraphics[width=\linewidth]{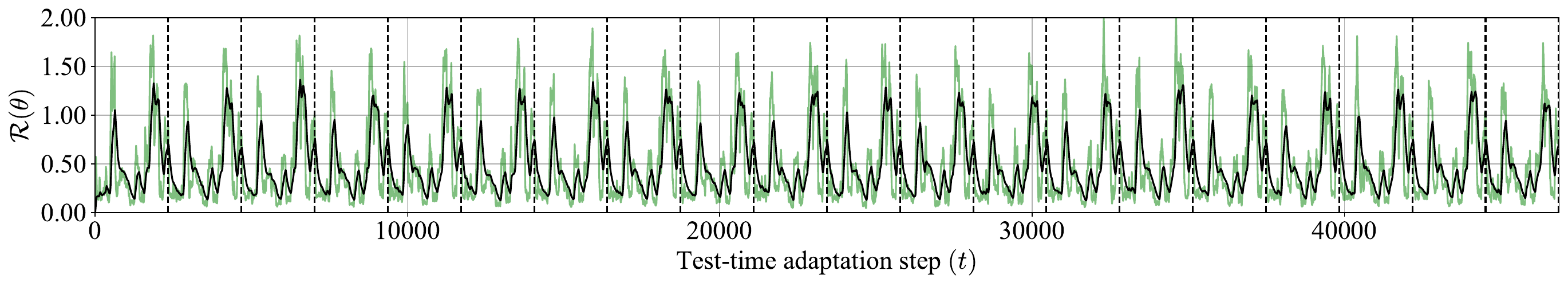}};

        \draw (0, -17.6) node[inner sep=0] {\includegraphics[width=\linewidth]{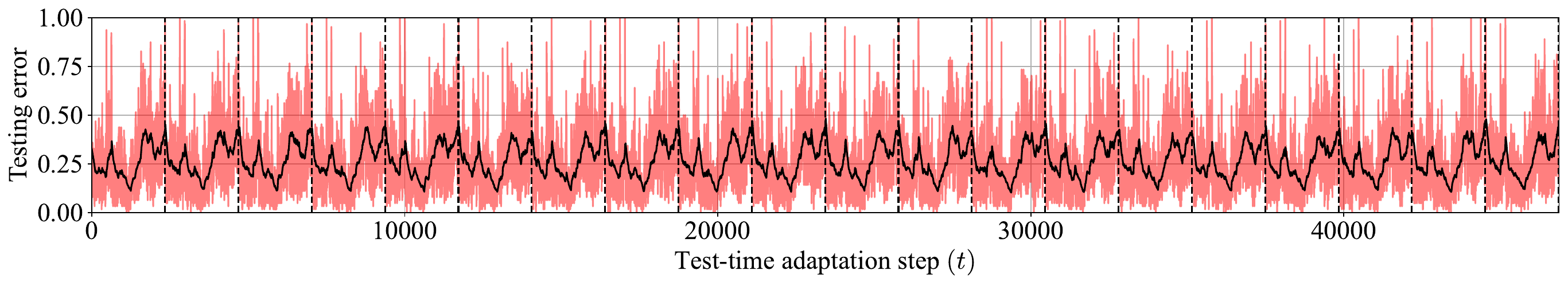}};
    
        \end{tikzpicture}
    \caption{An inspection of \method on the task CIFAR-10 $\rightarrow$ CIFAR-10-C~\cite{hendrycks2019robustness} in a \setting with 20 visits (visits are separated by the \textit{vertical dashed lines}). Here, we visualize (rows 1-3) the dynamic of \method \textcolor{blue}{ adaptive parameters ($\bar \gamma_t, \lambda_t, \alpha_t$)}, (rows 4-5) the value of the \textcolor{ForestGreen}{loss functions  ($\mathcal{L}_{\mathrm{CLS}}, \mathcal{L}_{\mathrm{AL}}$)} and (row 6) the value of the \textcolor{ForestGreen}{regularization term $\left(\mathcal{R}(\theta)\right)$} and (row 7) the classification \textcolor{red}{error rate} at each step. The \textbf{solid line} in the foreground of each plot denotes the running mean. The plots show an adaptive change of $\lambda_t, \alpha_t$ through time in \method, which stabilizes TTA performance, making \method achieve a persisting adaptation process in all observed values across 20 visits.} 
    \label{fig:petta_inspection}
    \vspace*{-1.0\baselineskip}
\end{figure}

In \Fig{\ref{fig:petta_inspection}}, we showcase an inspection of our \method on the task CIFAR-10 $\rightarrow$ CIFAR-10-C~\cite{hendrycks2019robustness} in a typical \setting TTA with 20 visits. 
Specifically, the visualizations of \method parameters ($\bar \gamma_t$, $\lambda_t$, and $\alpha_t$), adaptation losses ($\mathcal{L}_{\mathrm{CLS}}, \mathcal{L}_{\mathrm{AL}}$) and regularization term ($\mathcal{R}(\theta)$) are provided.
Here, we observe the values of adaptive parameters $\lambda_t$ and $\alpha_t$ continuously changing through time, as the testing scenarios evolve during \setting TTA. 
This proposed mechanism \textit{stabilizes} the value of the loss functions, and regularization term, balancing between the two primary objectives: adaptation and preventing model collapse. Thus, \textit{the error rate persists} as a result.
A similar pattern is observed on other datasets (CIFAR-100-C~\cite{hendrycks2019robustness} and DomainNet~\cite{peng2019moment}).

\subsection{Does Model Reset Help?}
\label{appdx:model_reset}
\noindent \textbf{Experiment Setup.} We use the term \textit{``model reset''} to represent the action of \textit{``reverting the current TTA model to the source model''}. This straightforward approach is named RDumb~\cite{press2023rdumb}. We thoroughly conducted experiments to compare the performance of RDumb with \method. The implementation of RDumb in this setting is as follows. We employ RoTTA~\cite{yuan2023robust} as the base test-time adaptor due to the characteristics of the practical TTA~\cite{yuan2023robust} stream. The model \textit{(including model parameters, the optimizer state, and the memory bank)} is reset after adapting itself to $T$ images.\footnote{A slight abuse of notation. $T$ here is the number of images between two consecutive resets, following the notation on \Sec{3} of~\cite{press2023rdumb}, \textit{not} the sample indices in our notations.} For each dataset, three values of this hyper-parameter $T$ are selected:
\begin{itemize}
    \item $T=1,000$: This is the value selected by the RDumb's authors~\cite{press2023rdumb}. Unless specifically stated, we use this value when reporting the performance of RDumb~\cite{press2023rdumb} in all other tables.
    \item $T=10,000$ (CIFAR-10/100-C), $T=5,000$ (ImageNet-C) and $T=24,237$ (DomainNet).\footnote{A subset of $5,000$ samples from ImageNet-C are selected following \texttt{RobustBench}~\cite{croce2021robustbench} for a consistent evaluation with other benchmarks.} This value is equal to the number of samples in the test set of a \textit{single corruption type}, i.e., the model is reset exactly after visiting each $\mathcal{P}_i$'s (see \Sec{\ref{ssec:model_collapse}} for notations). For DomainNet~\cite{peng2019moment}, since the number of images within each domain is unequal, the average number of images is used instead. 
    \item $T=150,000$ (CIFAR-10/100-C), $T=75,000$ (ImageNet-C) and $T=72,712$ (DomainNet). This number is equal to the number of samples \textit{in one recurrence} of our \setting TTA, i.e., the model is reset exactly after visiting $\mathcal{P}_1 \rightarrow \dots \rightarrow \mathcal{P}_D$. Here, $D=15$ - types of corruptions~\cite{hendrycks2019robustness} for CIFAR-10/100-C and ImageNet-C and $D=3$ for DomainNet (\textit{clipart, painting, sketch}). For example, the model is reset 20 times within a \textit{\setting TTA} setting with 20 recurrences under this choice of $T$. 
\end{itemize}

The second and the last reset scheme could be interpreted as assuming the model has access to \textit{an oracle model} with a capability of signaling the transitions between domains, or recurrences. Typically, this is \textit{an unrealistic capability in real-world scenarios}, and a desirable continual TTA algorithm should be able to operate independently without knowing when the domain shift happening.

\noindent \textbf{Experimental Results.} An empirical comparison between RDumb~\cite{press2023rdumb} and our \method are reported in \Tab{\ref{tab:cifar-10-model-reset}}, \Tab{\ref{tab:cifar-100-model-reset}}, \Tab{\ref{tab:domainnet-model-reset}} and \Tab{\ref{tab:imagenetc-model-reset}} for all four tasks. 

\begin{table*}[ht!]
    \caption{Average classification error comparison between RDumb~\cite{press2023rdumb} (a reset-based approach) with different reset frequencies and our \method on CIFAR-10 $\rightarrow$ CIFAR-10-C task.}
    \label{tab:cifar-10-model-reset}
    \resizebox{\textwidth}{!}{
    \begin{tabular}{r|cccccccccccccccccccc|c}
    \toprule
    & \multicolumn{20}{l}{ \textit{\Setting TTA visit} $\xrightarrow{\hspace*{5cm}}$ } \\
    \textbf{Reset Every} &              1 &              2 &              3 &              4 &              5 &              6 &              7 &              8 &              9 &             10 &             11 &             12 &             13 &             14 &             15 &             16 &             17 &             18 &             19 &             20 &   \textbf{Avg} \\
    \midrule
    
$T=1000$                &           31.1 &           32.1 &           32.3 &           31.6 &           31.9 &           31.8 &           31.8 &           31.9 &           31.9 &           32.1 &           31.7 &           32.0 &           32.5 &           32.0 &           31.9 &           31.6 &           31.9 &           31.4 &           32.3 &           32.4 &           31.9 \\
$T=10000$               &           25.8 &           25.9 &           26.5 &           26.1 &           26.4 &           25.4 &           25.8 &           25.8 &           26.1 &           26.2 &           26.1 &           26.1 &           26.1 &           26.1 &           26.1 &           25.9 &           25.5 &           25.5 &           25.7 &           26.2 &           26.0 \\
$T=150000$              &           24.8 &           25.3 &           24.3 &           24.1 &           25.3 &           25.4 &           25.4 &           24.5 &           25.0 &           24.9 &           25.0 &           24.8 &           25.0 &           24.5 &           24.9 &           24.1 &           24.0 &           24.7 &           24.9 &           24.4 &           24.8 \\

    \midrule
    \rowcolor{ClrHighlight}
    \method \textit{(ours)}$^{(*)}$     &  \textbf{24.3} &  \textbf{23.0} &  \textbf{22.6} &  \textbf{22.4} &  \textbf{22.4} &  \textbf{22.5} &  \textbf{22.3} &  \textbf{22.5} &  \textbf{22.8} &  \textbf{22.8} &  \textbf{22.6} &  \textbf{22.7} &  \textbf{22.7} &  \textbf{22.9} &  \textbf{22.6} &  \textbf{22.7} &  \textbf{22.6} &  \textbf{22.8} &  \textbf{22.9} &  \textbf{23.0} &  \textbf{22.8} \\
    
    \bottomrule
    \end{tabular}
   }
\end{table*}

\begin{table*}[ht!]
    \caption{Average classification error comparison between RDumb~\cite{press2023rdumb} (a reset-based approach) with different reset frequencies and our \method on CIFAR-100-C dataset.}
    \label{tab:cifar-100-model-reset}
    \resizebox{\textwidth}{!}{
    \begin{tabular}{r|cccccccccccccccccccc|c}
    \toprule
    & \multicolumn{20}{l}{ \textit{\Setting TTA visit} $\xrightarrow{\hspace*{5cm}}$ } \\
    \textbf{Reset Every} &              1 &              2 &              3 &              4 &              5 &              6 &              7 &              8 &              9 &             10 &             11 &             12 &             13 &             14 &             15 &             16 &             17 &             18 &             19 &             20 &   \textbf{Avg} \\
    \midrule
    
$T=1000$                &           36.7 &           36.7 &           36.6 &           36.6 &           36.7 &           36.8 &           36.7 &           36.5 &           36.6 &           36.5 &           36.7 &           36.6 &           36.5 &           36.7 &           36.5 &           36.6 &           36.6 &           36.7 &           36.6 &           36.5 &           36.6 \\
$T=10000$               &           43.5 &           43.6 &           43.7 &           43.7 &           43.4 &           43.5 &           43.6 &           43.4 &           43.5 &           43.6 &           43.8 &           43.5 &           43.5 &           43.6 &           43.4 &           43.6 &           43.5 &           43.8 &           43.7 &           43.6 &           43.6 \\
$T=150000$              &  \textbf{35.4} &           35.4 &           35.4 &           35.3 &           35.4 &           35.4 &           35.5 &           35.6 &           35.4 &           35.4 &           35.5 &  \textbf{35.3} &  \textbf{35.2} &           35.4 &  \textbf{35.1} &           35.8 &  \textbf{35.1} &           35.6 &           35.3 &           35.8 &           35.4 \\

    \midrule
    \rowcolor{ClrHighlight}
   \method \textit{(ours)}$^{(*)}$ &           35.8 &  \textbf{34.4} &  \textbf{34.7} &  \textbf{35.0} &  \textbf{35.1} &  \textbf{35.1} &  \textbf{35.2} &  \textbf{35.3} &  \textbf{35.3} &  \textbf{35.3} &  \textbf{35.2} &  \textbf{35.3} &  \textbf{35.2} &  \textbf{35.2} &  \textbf{35.1} &  \textbf{35.2} &           35.2 &  \textbf{35.2} &  \textbf{35.2} &  \textbf{35.2} &  \textbf{35.1} \\

    \bottomrule
    \end{tabular}
   }
\end{table*}

\begin{table*}[ht!]
    \caption{Average classification error comparison between RDumb~\cite{press2023rdumb} (a reset-based approach) with different reset frequencies and our \method on DomainNet dataset.}
    \label{tab:domainnet-model-reset}
    \resizebox{\textwidth}{!}{
    \begin{tabular}{r|cccccccccccccccccccc|c}
    \toprule
    & \multicolumn{20}{l}{ \textit{\Setting TTA visit} $\xrightarrow{\hspace*{5cm}}$ } \\
    \textbf{Reset Every} &              1 &              2 &              3 &              4 &              5 &              6 &              7 &              8 &              9 &             10 &             11 &             12 &             13 &             14 &             15 &             16 &             17 &             18 &             19 &             20 &   \textbf{Avg} \\
    \midrule
    
$T=1000$                &           44.3 &           44.4 &           44.3 &           44.5 &           44.2 &           44.2 &           44.3 &           44.5 &           44.4 &           44.2 &           44.3 &           44.3 &           44.3 &           44.3 &           44.5 &           44.3 &           44.2 &           44.3 &           44.4 &           44.3 &           44.3 \\
$T=24237$               &           44.1 &           44.3 &           43.9 &           44.2 &           44.1 &           44.3 &           44.2 &           44.4 &           44.1 &           44.1 &           44.0 &           44.3 &           44.1 &           44.0 &           44.0 &           44.2 &           44.1 &           44.1 &           44.1 &           44.4 &           44.1 \\
$T=72712$               &           44.3 &           44.3 &           44.0 &           44.3 &           44.1 &           44.3 &           44.2 &           44.4 &           44.2 &           44.1 &           44.0 &           44.1 &           44.2 &           44.1 &           44.1 &           44.1 &           44.1 &           44.0 &           44.0 &           44.3 &           44.2 \\

    \midrule
    \rowcolor{ClrHighlight}
    \method \textit{(ours)}$^{(*)}$ &  \textbf{43.8} &  \textbf{42.6} &  \textbf{42.3} &  \textbf{42.3} &  \textbf{42.6} &  \textbf{42.8} &  \textbf{42.8} &  \textbf{43.0} &  \textbf{42.9} &  \textbf{42.9} &  \textbf{43.1} &  \textbf{43.0} &  \textbf{42.9} &  \textbf{43.0} &  \textbf{43.0} &  \textbf{43.1} &  \textbf{43.0} &  \textbf{42.8} &  \textbf{42.9} &  \textbf{42.9} &  \textbf{42.9} \\

    \bottomrule
    \end{tabular}
   }
\end{table*}

\begin{table*}[ht!]
    \caption{Average classification error comparison between RDumb~\cite{press2023rdumb} (a reset-based approach) with different reset frequencies and our \method on ImageNet-C dataset.}
    \label{tab:imagenetc-model-reset}
    \resizebox{\textwidth}{!}{
    \begin{tabular}{r|cccccccccccccccccccc|c}
    \toprule
    & \multicolumn{20}{l}{ \textit{\Setting TTA visit} $\xrightarrow{\hspace*{5cm}}$ } \\
    \textbf{Reset Every} &              1 &              2 &              3 &              4 &              5 &              6 &              7 &              8 &              9 &             10 &             11 &             12 &             13 &             14 &             15 &             16 &             17 &             18 &             19 &             20 &   \textbf{Avg} \\
    \midrule
    
$T=1000$                &           72.2 &           73.0 &           73.2 &           72.8 &           72.2 &           72.8 &           73.3 &           72.7 &           71.9 &           73.0 &           73.2 &           73.1 &           72.0 &           72.7 &           73.3 &           73.1 &           72.1 &           72.6 &           73.3 &           73.1 &           72.8 \\
$T=5000$                &           70.2 &           70.8 &           71.6 &           72.1 &           72.4 &           72.6 &           72.9 &           73.1 &           73.2 &           73.6 &           73.7 &           73.9 &           74.0 &           74.0 &           74.3 &           74.1 &           74.1 &           73.8 &           73.5 &           71.9 &           73.0 \\
$T=75000$               &           67.0 &           67.1 &           67.2 &           67.5 &           67.5 &           67.6 &           67.8 &           67.6 &           67.6 &           67.6 &           67.5 &           67.7 &           67.6 &           67.9 &           68.1 &           67.9 &           67.4 &           67.5 &           67.7 &           67.5 &           67.6 \\

    \midrule
    \rowcolor{ClrHighlight}
    \method \textit{(ours)}$^{(*)}$ &  \textbf{65.3} &  \textbf{61.7} &  \textbf{59.8} &  \textbf{59.1} &  \textbf{59.4} &  \textbf{59.6} &  \textbf{59.8} &  \textbf{59.3} &  \textbf{59.4} &  \textbf{60.0} &  \textbf{60.3} &  \textbf{61.0} &  \textbf{60.7} &  \textbf{60.4} &  \textbf{60.6} &  \textbf{60.7} &  \textbf{60.8} &  \textbf{60.7} &  \textbf{60.4} &  \textbf{60.2} &  \textbf{60.5} \\
    \bottomrule
    \end{tabular}
   }
\end{table*}

\noindent \textbf{Discussions.} Across datasets and reset frequencies, our \method approach is always \textit{better} than RDumb~\cite{press2023rdumb}. The supreme performance holds even when RDumb has access to the oracle information that can reset the model exactly at the transition between each domain shift or recurrence. Importantly, this oracle information is typically unavailable in practice. 

Noteworthy, it is clear that the performance of RDumb varies when changing the choice of the reset frequency. For a given choice of $T$, the better performance on one dataset does not guarantee the same performance on other datasets. For example, $T=1,000$ - the best empirical value found by RDumb authors~\cite{press2023rdumb} on CCC, does not give the best performance on our \setting TTA scenario; the second choice of $T$ negatively impact the performance on many tasks; the third choice gives the best results, but knowing this exact recurrence frequency of the testing stream is unrealistic. The result highlights the challenge in practice when tuning this parameter (too slow/frequent), especially in the TTA setting where a validation set is unavailable. Our \method, in contrast, is reset-free.

\subsection{\method with 40 Recurring Visits}
\label{sec:results_recurring40}
\begin{figure}
  \begin{minipage}[b]{.5\linewidth}
    \centering
    \includegraphics[width=\linewidth]{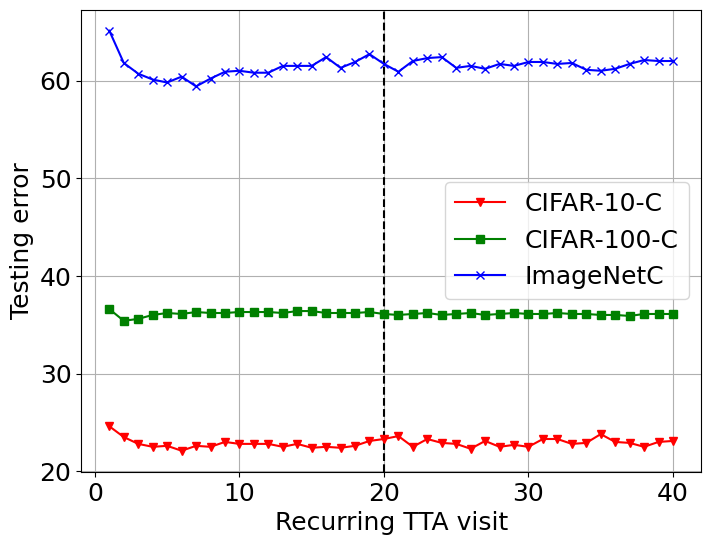}%
    \captionof{figure}{Testing error of PeTTA with 40 recurring TTA visits.}%
    \label{fig:recurring40}
  \end{minipage}\hfill
  \begin{minipage}[b]{.45\linewidth}
    \centering
    \resizebox{\textwidth}{!}{
    \begin{tabular}{c|c|c|c}
         \toprule
         \textbf{Total Visits}                                         & \textbf{CF-10-C} & \textbf{CF-100-C} & \textbf{IN-C} \\
         \midrule
         20 visits   & 22.8     & 35.1    & 60.5   \\
         40 visits   & 22.9     & 35.1    & 61.0   \\
         \bottomrule
        \end{tabular}}
    \captionof{table}{Average testing error of \method in recurring TTA with 20 and 40 visits. \method demonstrates its persistence over an extended testing time horizon beyond the 20$^{th}$ visit milestone (\Fig{\ref{fig:recurring40}}'s horizontal dashed line).}
    \label{tab:recurring40}
  \end{minipage}
\end{figure}
To demonstrate the persistence of \method over an even longer testing stream, in \Tab{\ref{tab:recurring40}} and \Fig{\ref{fig:recurring40}}, we provide the evaluation results of \method on \setting with 40 recurrences.

\subsection{The Sensitivity of Hyper-parameter Choices in \method}
\label{sec:petta_hyper_params}
\begin{table}[h!]
    \centering
    \caption{Sensitivity of \method with different choices of $\lambda_0$.}
    \label{tab:sensitivity_lambda0}
    \resizebox{.7\textwidth}{!}{
    \begin{tabular}{c|c|c|c|c|c}
         \toprule
         \textbf{Dataset}                                         & \textbf{$\lambda_0=1e^{0}$} & $\lambda_0=5e^{0}$ & $\lambda_0=1e^{1}$ &$\lambda_0=5e^{1}$ & $\lambda_0=1e^{2}$\\
         \midrule
         CIFAR-10-C   & 22.9     & 22.7    & 22.8  & 23.2 & 24.1 \\
         CIFAR-100-C   & 35.7     & 35.3    & 35.1  & 35.6 & 36.1 \\
         ImageNet-C   & 61.2     & 61.0    & 60.5  & 61.3 & 62.4 \\
         \bottomrule
        \end{tabular}}
\end{table}

There are two hyper-parameters in \method: $\alpha_0$ and $\lambda_0$.
The initial learning rate of $\alpha_0=1e^{-3}$ is used for all experiments. 
We do not tune this hyper-parameter, and the choice of $\alpha_0$ is universal across all datasets, following the previous works/compared methods (e.g., RoTTA~\cite{yuan2023robust}, CoTTA~\cite{Wang_2022_CVPR}).

Since $\lambda_0$ is more specific to \method, we included a sensitive analysis with different choices of $\lambda_0$ on \method, evaluated with images from CIFAR-10/100-C and ImageNet-C in \Tab{\ref{tab:sensitivity_lambda0}}. Overall, the choice of $\lambda_0$ 
 is not extremely sensitive, and while the best value is $1e^1$
 on most datasets, other choices such as $5e^0$ or $5e^1$ also produce roughly similar performance. Selecting $\lambda_0$
 is intuitive, the larger value of $\lambda_0$
 stronger prevents the model from collapsing but also limits its adaptability as a trade-off.

In action, $\lambda_0$ is an initial value and will be adaptively scaled with the sensing model divergence mechanism in \method, meaning it does not require careful tuning. More generally, this hyper-parameter can be tuned similarly to the hyper-parameters of other TTA approaches, via an additional validation set, or some accuracy prediction algorithm~\cite{lee2024_aetta} when labeled data is unavailable.

\subsection{More Details on the Ablation Study}

We provide the detailed classification error for each visit in the \setting TTA setting of each row entry in \Tab{\ref{tab:ablation_study}} (\method Ablation Study): \Tab{\ref{tab:cifar10_ablation_study}}, \Tab{\ref{tab:cifar100_ablation_study}}, \Tab{\ref{tab:domainnet_ablation_study}}, \Tab{\ref{tab:imagenetc_ablation_study}}; and \Tab{\ref{tab:regularizers}} (\method with various choices of regularizers): \Tab{\ref{tab:cifar10_regularizers}}, \Tab{\ref{tab:cifar100_regularizers}}, \Tab{\ref{tab:domainnet_regularizers}}, \Tab{\ref{tab:imagenetc_regularizers}}.   

    \begin{table*}[ht!]
    \caption{Average classification error of multiple variations of \method. Experiments on  CIFAR10 $\rightarrow$ CIFAR10-C~\cite{hendrycks2019robustness} task.}
    \label{tab:cifar10_ablation_study}
    \resizebox{\textwidth}{!}{
    \begin{tabular}{l|cccccccccccccccccccc|c}
\toprule
& \multicolumn{20}{l}{ \textit{Episodic TTA visit} $\xrightarrow{\hspace*{5cm}}$ } \\
\textbf{Method} &              1 &              2 &              3 &              4 &              5 &              6 &              7 &              8 &              9 &             10 &             11 &             12 &             13 &             14 &             15 &             16 &             17 &             18 &             19 &             20 &   \textbf{Avg} \\
\midrule

Baseline w/o $\mathcal{R}(\theta)$                 &           23.5 &           24.0 &           27.4 &           29.9 &           33.4 &           35.6 &           38.0 &           40.7 &           43.1 &           45.0 &           46.0 &           48.6 &           50.0 &           49.7 &           50.8 &           51.5 &           52.3 &           53.3 &           54.3 &           55.5 &           42.6 \\
\midrule
$\mathcal{R}(\theta)$ fixed $\lambda = 0.1\lambda_0$ &           23.5 &           24.0 &           27.2 &           29.8 &           33.4 &           35.3 &           37.9 &           40.5 &           43.3 &           45.3 &           46.8 &           49.3 &           50.9 &           51.0 &           52.1 &           53.2 &           54.0 &           54.8 &           56.0 &           57.6 &           43.3 \\
$\mathcal{R}(\theta)$ fixed $\lambda = \lambda_0$  &           23.5 &           23.6 &           26.2 &           28.4 &           31.6 &           33.5 &           36.4 &           38.7 &           41.1 &           43.1 &           44.8 &           47.6 &           49.3 &           49.5 &           50.9 &           52.1 &           53.1 &           54.2 &           55.6 &           57.0 &           42.0 \\
\midrule \midrule
\method - $\lambda_t$                              &           24.9 &           25.3 &           26.0 &           26.4 &           27.2 &           26.5 &           27.2 &           27.1 &           27.4 &           27.7 &           27.8 &           28.0 &           27.5 &           28.0 &           27.7 &           27.4 &           27.0 &           27.6 &           27.8 &           27.8 &           27.1 \\
\method - $\lambda_t + \alpha_t$                   &           25.5 &           24.5 &           23.7 &           23.1 &           23.2 &  \textbf{22.4} &           23.3 &           23.2 &           23.7 &           24.1 &           23.9 &           24.5 &           24.3 &           24.0 &           23.8 &           23.9 &           23.8 &           24.1 &           24.6 &           24.7 &           23.9 \\
\method - $\lambda_t + \mathcal{L}_{\mathrm{AL}}$           &  \textbf{23.3} &           23.9 &           24.6 &           25.3 &           26.2 &           25.9 &           26.4 &           26.6 &           26.9 &           26.6 &           26.7 &           26.7 &           26.7 &           26.8 &           26.8 &           27.2 &           26.9 &           26.9 &           26.8 &           27.0 &           26.2 \\
\midrule
\method $\alpha_t$  + $\mathcal{L}_{\mathrm{AL}}$  &           24.3 &  \textbf{23.0} &  \textbf{22.6} &  \textbf{22.4} &  \textbf{22.4} &           22.5 &  \textbf{22.3} &  \textbf{22.5} &  \textbf{22.8} &  \textbf{22.8} &  \textbf{22.6} &  \textbf{22.7} &  \textbf{22.7} &  \textbf{22.9} &  \textbf{22.6} &  \textbf{22.7} &  \textbf{22.6} &  \textbf{22.8} &  \textbf{22.9} &  \textbf{23.0} &  \textbf{22.8} \\

\bottomrule
\end{tabular}

   }
    \end{table*}

    \begin{table*}[ht!]
    \caption{Average classification error of multiple variations of \method. Experiments on CIFAR-100 $\rightarrow$ CIFAR100-C~\cite{hendrycks2019robustness} task.}
    \label{tab:cifar100_ablation_study}
    \resizebox{\textwidth}{!}{
    \begin{tabular}{l|cccccccccccccccccccc|c}
\toprule
& \multicolumn{20}{l}{ \textit{Episodic TTA visit} $\xrightarrow{\hspace*{5cm}}$ } \\
\textbf{Method} &              1 &              2 &              3 &              4 &              5 &              6 &              7 &              8 &              9 &             10 &             11 &             12 &             13 &             14 &             15 &             16 &             17 &             18 &             19 &             20 &   \textbf{Avg} \\
\midrule

Baseline w/o $\mathcal{R}(\theta)$                 &           40.2 &           46.3 &           51.2 &           54.4 &           57.3 &           59.4 &           61.3 &           62.6 &           63.9 &           65.1 &           66.3 &           67.1 &           68.1 &           68.9 &           69.6 &           70.3 &           71.1 &           71.6 &           72.4 &           72.9 &           63.0 \\
\midrule
$\mathcal{R}(\theta)$ fixed $\lambda = 0.1\lambda_0$ &           40.5 &           46.1 &           51.5 &           55.1 &           58.2 &           60.5 &           62.6 &           64.2 &           65.7 &           67.3 &           68.6 &           69.5 &           70.6 &           71.6 &           72.5 &           73.4 &           74.2 &           74.9 &           75.8 &           76.5 &           65.0 \\
$\mathcal{R}(\theta)$ fixed $\lambda = \lambda_0$  &           41.8 &           47.6 &           52.6 &           56.1 &           58.9 &           60.7 &           62.5 &           63.9 &           65.0 &           66.2 &           67.1 &           68.3 &           69.5 &           70.3 &           71.4 &           72.4 &           73.4 &           74.1 &           75.0 &           75.6 &           64.6 \\
\midrule \midrule
\method - $\lambda_t$                              &           39.4 &           43.4 &           46.6 &           49.1 &           51.0 &           52.6 &           53.8 &           54.7 &           55.7 &           56.5 &           57.1 &           57.7 &           58.3 &           58.8 &           59.3 &           59.9 &           60.6 &           61.0 &           61.6 &           62.1 &           55.0 \\
\method - $\lambda_t + \alpha_t$                   &           39.4 &           40.1 &           40.8 &           40.7 &           41.2 &           41.5 &           41.4 &           41.6 &           41.5 &           41.5 &           41.7 &           41.6 &           41.8 &           41.7 &           41.8 &           42.0 &           41.9 &           41.9 &           42.0 &           41.8 &           41.4 \\
\method - $\lambda_t + \mathcal{L}_{AL}$           &           36.2 &           35.6 &           35.7 &           36.1 &           36.2 &           36.4 &           36.4 &           36.5 &           36.2 &           36.2 &           36.6 &           36.5 &           36.5 &           36.6 &           36.5 &           36.6 &           36.5 &           36.5 &           36.3 &           36.5 &           36.3 \\
\midrule
\method $\lambda_t + \alpha_t$  + $\mathcal{L}_\mathrm{AL}$ &         \textbf{35.8} &  \textbf{34.4} &  \textbf{34.7} &  \textbf{35.0} &  \textbf{35.1} &  \textbf{35.1} &  \textbf{35.2} &  \textbf{35.3} &  \textbf{35.3} &  \textbf{35.3} &  \textbf{35.2} &  \textbf{35.3} &  \textbf{35.2} &  \textbf{35.2} &  \textbf{35.1} &  \textbf{35.2} &  \textbf{35.2} &  \textbf{35.2} &  \textbf{35.2} &  \textbf{35.2} &  \textbf{35.1} \\

\bottomrule
\end{tabular}

   }
    \end{table*}

    \begin{table*}[ht!]
    \caption{Average classification error of multiple variations of \method. Experiments on \textit{real} $\rightarrow$ \textit{clipart, painting, sketch} task from DomainNet~\cite{peng2019moment} task.}
    \label{tab:domainnet_ablation_study}
    \resizebox{\textwidth}{!}{
    \begin{tabular}{l|cccccccccccccccccccc|c}
\toprule
& \multicolumn{20}{l}{ \textit{\Setting TTA visit} $\xrightarrow{\hspace*{5cm}}$ } \\
\textbf{Method} &              1 &              2 &              3 &              4 &              5 &              6 &              7 &              8 &              9 &             10 &             11 &             12 &             13 &             14 &             15 &             16 &             17 &             18 &             19 &             20 &   \textbf{Avg} \\
\midrule

Baseline w/o $\mathcal{R}(\theta)$                 &           52.3 &           69.0 &           68.6 &           68.6 &           69.4 &           70.5 &           71.8 &           73.4 &           75.6 &           77.6 &           78.8 &           81.0 &           82.8 &           84.3 &           85.9 &           87.4 &           88.5 &           89.9 &           90.8 &           92.1 &           77.9 \\
\midrule
$\mathcal{R}(\theta)$ fixed $\lambda = 0.1\lambda_0$ &           52.5 &           70.0 &           69.8 &           70.0 &           71.1 &           72.5 &           74.6 &           76.1 &           77.8 &           80.4 &           81.9 &           83.5 &           85.2 &           87.2 &           89.1 &           90.2 &           91.5 &           93.2 &           94.1 &           94.9 &           80.0 \\
$\mathcal{R}(\theta)$ fixed $\lambda = \lambda_0$  &           54.6 &           69.8 &           63.7 &           56.0 &           61.7 &           76.4 &           70.4 &           62.5 &           58.2 &           76.0 &           73.6 &           66.8 &           58.6 &           62.3 &           80.8 &           75.5 &           67.0 &           59.9 &           59.3 &           78.3 &           66.6 \\

\midrule \midrule
\method - $\lambda_t$                              &           49.2 &           64.5 &           62.4 &           60.9 &           59.6 &           58.6 &           57.7 &           57.8 &           57.6 &           57.7 &           58.0 &           58.5 &           59.0 &           59.5 &           59.8 &           61.1 &           62.0 &           62.6 &           63.6 &           64.9 &           59.7 \\
\method - $\lambda_t + \alpha_t$                   &           43.9 &  \textbf{42.5} &  \textbf{42.3} &  \textbf{42.3} &  \textbf{42.6} &  \textbf{42.8} &           43.1 &           43.7 &           43.9 &           44.3 &           44.6 &           45.1 &           45.4 &           45.7 &           45.7 &           46.1 &           46.1 &           46.2 &           46.5 &           46.4 &           44.5 \\
\method - $\lambda_t + \mathcal{L}_{AL}$           &  \textbf{43.6} &  \textbf{42.5} &           42.6 &           42.6 &           42.9 &           43.0 &           43.3 &           43.4 &           43.1 &           43.2 &  \textbf{43.1} &           43.3 &           43.3 &           43.2 &           43.2 &           43.9 &           43.7 &           43.0 &           43.2 &           43.5 &           43.2 \\

\midrule
\method $\lambda_t + \alpha_t$  + $\mathcal{L}_\mathrm{AL}$ &            43.8 &           42.6 &  \textbf{42.3} &  \textbf{42.3} &  \textbf{42.6} &  \textbf{42.8} &  \textbf{42.8} &  \textbf{43.0} &  \textbf{42.9} &  \textbf{42.9} &  \textbf{43.1} &  \textbf{43.0} &  \textbf{42.9} &  \textbf{43.0} &  \textbf{43.0} &  \textbf{43.1} &  \textbf{43.0} &  \textbf{42.8} &  \textbf{42.9} &  \textbf{42.9} &  \textbf{42.9} \\

\bottomrule
\end{tabular}

   }
    \end{table*}

    \begin{table*}[ht!]
    \caption{Average classification error of multiple variations of \method. Experiments on ImageNet $\rightarrow$ ImageNet-C~\cite{hendrycks2019robustness} task.}
    \label{tab:imagenetc_ablation_study}
    \resizebox{\textwidth}{!}{
    \begin{tabular}{l|cccccccccccccccccccc|c}
\toprule
& \multicolumn{20}{l}{ \textit{\Setting TTA visit} $\xrightarrow{\hspace*{5cm}}$ } \\
\textbf{Method} &              1 &              2 &              3 &              4 &              5 &              6 &              7 &              8 &              9 &             10 &             11 &             12 &             13 &             14 &             15 &             16 &             17 &             18 &             19 &             20 &   \textbf{Avg} \\
\midrule

Baseline w/o $\mathcal{R}(\theta)$                 &           66.9 &           61.9 &           72.7 &           93.6 &           97.4 &           97.8 &           98.0 &           98.2 &           98.3 &           98.3 &           98.4 &           98.4 &           98.5 &           98.5 &           98.6 &           98.6 &           98.6 &           98.6 &           98.7 &           98.7 &           93.4 \\

\midrule
$\mathcal{R}(\theta)$ fixed $\lambda = 0.1\lambda_0$ &          65.5 &           70.9 &           79.1 &           85.2 &           90.3 &           92.6 &           95.8 &           95.8 &           95.4 &           97.3 &           96.9 &           97.7 &           97.9 &           98.2 &           98.0 &           98.7 &           98.6 &           98.4 &           98.4 &           98.7 &           92.5 \\

$\mathcal{R}(\theta)$ fixed $\lambda = \lambda_0$  &           66.5 &           62.1 &           73.0 &           93.5 &           97.0 &           97.2 &           97.5 &           97.5 &           97.6 &           97.5 &           97.7 &           97.7 &           97.7 &           97.8 &           97.9 &           97.9 &           98.0 &           98.0 &           98.0 &           97.9 &           92.9 \\

\midrule \midrule
\method - $\lambda_t$                              &           65.9 &           62.1 &           76.3 &           96.7 &           97.0 &           96.9 &           96.9 &           96.9 &           97.0 &           97.1 &           97.0 &           97.2 &           97.0 &           97.1 &           97.1 &           97.0 &           97.0 &           97.0 &           97.0 &           97.0 &           92.7 \\
\method - $\lambda_t + \alpha_t$                   &  \textbf{64.8} &           70.5 &           74.6 &           75.8 &           75.5 &           75.8 &           76.1 &           76.2 &           76.2 &           76.5 &           76.7 &           77.0 &           76.9 &           77.4 &           77.1 &           77.3 &           77.2 &           77.4 &           77.6 &           77.4 &           75.7 \\
\method - $\lambda_t + \mathcal{L}_{AL}$           &  \textbf{64.8} &  \textbf{61.1} &           60.0 &           59.8 &           60.4 &           60.4 &           61.2 &           61.2 &           61.8 &           61.9 &           62.1 &           62.2 &           62.1 &           62.9 &           62.1 &           62.8 &           62.7 &           62.1 &           62.8 &           66.6 &           62.0 \\
\midrule
\method \textit{(ours)}$^{(*)}$ &  \textbf{65.3} &  \textbf{61.7} &  \textbf{59.8} &  \textbf{59.1} &  \textbf{59.4} &  \textbf{59.6} &  \textbf{59.8} &  \textbf{59.3} &  \textbf{59.4} &  \textbf{60.0} &  \textbf{60.3} &  \textbf{61.0} &  \textbf{60.7} &  \textbf{60.4} &  \textbf{60.6} &  \textbf{60.7} &  \textbf{60.8} &  \textbf{60.7} &  \textbf{60.4} &  \textbf{60.2} &  \textbf{60.5} \\

\bottomrule
\end{tabular}

   }
    \end{table*}

\begin{table*}[ht!]
    \caption{Average classification error of \method with various choices of regularizers. Experiments on CIFAR-10 $\rightarrow$ CIFAR-10-C~\cite{hendrycks2019robustness} task.}
    \label{tab:cifar10_regularizers}
    \resizebox{\textwidth}{!}{
    \begin{tabular}{l|cccccccccccccccccccc|c}
\toprule
& \multicolumn{20}{l}{ \textit{Episodic TTA visit} $\xrightarrow{\hspace*{5cm}}$ } \\
\textbf{Method} &              1 &              2 &              3 &              4 &              5 &              6 &              7 &              8 &              9 &             10 &             11 &             12 &             13 &             14 &             15 &             16 &             17 &             18 &             19 &             20 &   \textbf{Avg} \\
\midrule
L2            &           25.6 &           24.8 &           23.8 &           23.1 &           23.2 &           22.7 &           23.0 &           22.7 &           22.7 &           22.7 &           22.8 &           22.7 &           22.8 &           22.7 &           22.5 &  \textbf{22.3} &  \textbf{22.2} &  \textbf{22.4} &  \textbf{22.7} &  \textbf{22.8} &           23.0 \\
L2+Fisher     &           25.2 &           23.7 &           22.5 &           21.8 &           22.3 &           21.5 &           22.3 &           22.1 &           22.5 &           22.8 &           22.6 &           22.6 &  \textbf{22.6} &           22.8 &           22.6 &           22.9 &           22.6 &           22.9 &           23.0 &           23.3 &           22.7 \\
\midrule
Cosine        &  \textbf{24.3} &  \textbf{23.0} &           22.6 &           22.4 &           22.4 &           22.5 &           22.3 &           22.5 &           22.8 &           22.8 &           22.6 &           22.7 &           22.7 &           22.9 &           22.6 &           22.7 &           22.6 &           22.8 &           22.9 &           23.0 &           22.8 \\
Cosine+Fisher &           25.1 &           23.8 &  \textbf{22.2} &  \textbf{21.6} &  \textbf{22.0} &  \textbf{21.4} &  \textbf{22.0} &  \textbf{21.8} &  \textbf{22.1} &  \textbf{22.3} &  \textbf{22.5} &  \textbf{22.4} &  \textbf{22.6} &  \textbf{22.6} &  \textbf{22.4} &           22.7 &           22.6 &           22.8 &           22.8 &           23.3 &  \textbf{22.6} \\

\bottomrule
\end{tabular}

   }
\end{table*}

\begin{table*}[ht!]
    \caption{Average classification error of \method with various choices of regularizers. Experiments on CIFAR-100 $\rightarrow$ CIFAR-100-C~\cite{hendrycks2019robustness} task.}
    \label{tab:cifar100_regularizers}
    \resizebox{\textwidth}{!}{
    \begin{tabular}{l|cccccccccccccccccccc|c}
\toprule
& \multicolumn{20}{l}{ \textit{\Setting TTA visit} $\xrightarrow{\hspace*{5cm}}$ } \\
\textbf{Method} &              1 &              2 &              3 &              4 &              5 &              6 &              7 &              8 &              9 &             10 &             11 &             12 &             13 &             14 &             15 &             16 &             17 &             18 &             19 &             20 &   \textbf{Avg} \\
\midrule

L2            &           36.9 &           35.5 &           35.5 &           35.5 &           35.7 &           35.6 &           35.6 &           35.5 &           35.5 &           35.4 &           35.6 &           35.5 &           35.7 &           35.7 &           35.7 &           35.7 &           35.8 &           35.5 &           35.4 &           35.5 &           35.6 \\
L2+Fisher     &           36.8 &           35.4 &           35.4 &           35.8 &           35.9 &           36.0 &           35.9 &           35.9 &           35.9 &           35.8 &           36.1 &           36.1 &           36.1 &           36.1 &           36.1 &           36.1 &           36.2 &           36.0 &           36.0 &           35.9 &           36.0 \\
\midrule
Cosine        &  \textbf{35.8} &  \textbf{34.4} &  \textbf{34.7} &  \textbf{35.0} &  \textbf{35.1} &  \textbf{35.1} &  \textbf{35.2} &  \textbf{35.3} &  \textbf{35.3} &  \textbf{35.3} &  \textbf{35.2} &  \textbf{35.3} &  \textbf{35.2} &  \textbf{35.2} &  \textbf{35.1} &  \textbf{35.2} &  \textbf{35.2} &  \textbf{35.2} &  \textbf{35.2} &  \textbf{35.2} &  \textbf{35.1} \\
Cosine+Fisher &           36.7 &           35.2 &           35.5 &           35.6 &           35.9 &           35.9 &           36.1 &           36.0 &           36.0 &           35.9 &           36.0 &           36.0 &           36.0 &           36.1 &           36.0 &           36.0 &           35.9 &           35.9 &           35.9 &           36.0 &           35.9 \\

\bottomrule
\end{tabular}

   }
\end{table*}

\begin{table*}[ht!]
    \caption{Average classification error of \method with various choices of regularizers. Experiments on \textit{real} $\rightarrow$ \textit{clipart, painting, sketch} task from DomainNet~\cite{peng2019moment} dataset.}
    \label{tab:domainnet_regularizers}
    \resizebox{\textwidth}{!}{
    \begin{tabular}{l|cccccccccccccccccccc|c}
\toprule
& \multicolumn{20}{l}{ \textit{\Setting TTA visit} $\xrightarrow{\hspace*{5cm}}$ } \\
\textbf{Method} &              1 &              2 &              3 &              4 &              5 &              6 &              7 &              8 &              9 &             10 &             11 &             12 &             13 &             14 &             15 &             16 &             17 &             18 &             19 &             20 &   \textbf{Avg} \\
\midrule
L2            &           43.8 &           42.7 &           42.5 &           42.4 &           42.8 &           42.9 &           43.0 &           43.1 &           43.1 &           43.2 &           43.4 &           43.3 &           43.2 &           43.3 &           43.2 &           43.2 &           43.4 &           43.0 &           43.1 &           43.1 &           43.1 \\
L2+Fisher     &           43.9 &           42.8 &           42.7 &           43.0 &           43.2 &           43.4 &           43.6 &           43.8 &           43.9 &           44.1 &           44.0 &           44.2 &           44.2 &           44.2 &           44.4 &           44.4 &           44.5 &           44.5 &           44.5 &           44.5 &           43.9 \\
\midrule
Cosine        &           43.8 &           42.6 &  \textbf{42.3} &  \textbf{42.3} &  \textbf{42.6} &  \textbf{42.8} &  \textbf{42.8} &  \textbf{43.0} &  \textbf{42.9} &  \textbf{42.9} &  \textbf{43.1} &  \textbf{43.0} &  \textbf{42.9} &  \textbf{43.0} &  \textbf{43.0} &  \textbf{43.1} &  \textbf{43.0} &  \textbf{42.8} &  \textbf{42.9} &  \textbf{42.9} &  \textbf{42.9} \\
Cosine+Fisher &  \textbf{43.7} &  \textbf{42.5} &           42.5 &           42.6 &           42.9 &           43.2 &           43.2 &           43.5 &           43.4 &           43.5 &           43.4 &           43.5 &           43.4 &           43.6 &           43.5 &           43.5 &           43.4 &           43.5 &           43.3 &           43.4 &           43.3 \\
\bottomrule
\end{tabular}

   }
\end{table*}

\begin{table*}[ht!]
    \caption{Average classification error of \method with various choices of regularizers. Experiments on ImageNet $\rightarrow$ ImageNet-C~\cite{hendrycks2019robustness} task.}
    \label{tab:imagenetc_regularizers}
    \resizebox{\textwidth}{!}{
    \begin{tabular}{l|cccccccccccccccccccc|c}
\toprule
& \multicolumn{20}{l}{ \textit{\Setting TTA visit} $\xrightarrow{\hspace*{5cm}}$ } \\
\textbf{Method} &              1 &              2 &              3 &              4 &              5 &              6 &              7 &              8 &              9 &             10 &             11 &             12 &             13 &             14 &             15 &             16 &             17 &             18 &             19 &             20 &   \textbf{Avg} \\
\midrule
L2            &           70.8 &           72.2 &           71.5 &           69.8 &           72.3 &           69.3 &           70.3 &           70.5 &           70.0 &           70.8 &           70.2 &           72.1 &           71.4 &           70.8 &           70.9 &           70.9 &           69.7 &           71.0 &           71.1 &           70.4 &           70.8 \\
L2+Fisher     &           70.5 &           70.0 &           69.5 &           69.4 &           69.6 &           69.9 &           69.2 &           69.3 &           72.2 &           70.4 &           71.0 &           70.5 &           71.7 &           71.5 &           71.3 &           68.4 &           68.6 &           68.8 &           68.7 &           68.7 &           70.0 \\
\midrule
Cosine        &           65.3 &  \textbf{61.7} &  \textbf{59.8} &  \textbf{59.1} &  \textbf{59.4} &  \textbf{59.6} &  \textbf{59.8} &  \textbf{59.3} &  \textbf{59.4} &  \textbf{60.0} &  \textbf{60.3} &  \textbf{61.0} &  \textbf{60.7} &  \textbf{60.4} &  \textbf{60.6} &  \textbf{60.7} &  \textbf{60.8} &  \textbf{60.7} &  \textbf{60.4} &  \textbf{60.2} &  \textbf{60.5} \\
Cosine+Fisher &  \textbf{65.1} &  \textbf{61.7} &           60.9 &           61.2 &           61.9 &           62.6 &           62.8 &           63.2 &           64.2 &           63.4 &           64.3 &           64.4 &           63.9 &           64.3 &           65.8 &           65.5 &           64.9 &           65.0 &           65.2 &           65.2 &           63.8 \\
\bottomrule
\end{tabular}

   }
\end{table*}

\begin{figure*}[ht!]
    \pgfplotsset{every x tick label/.append style={font=\tiny, yshift=0.5ex}}
    \pgfplotsset{every y tick label/.append style={font=\tiny, xshift=0.5ex}}
    \centering    
        \begin{tikzpicture}

        \draw (0, 0) node[inner sep=0] {\includegraphics[width=\linewidth]{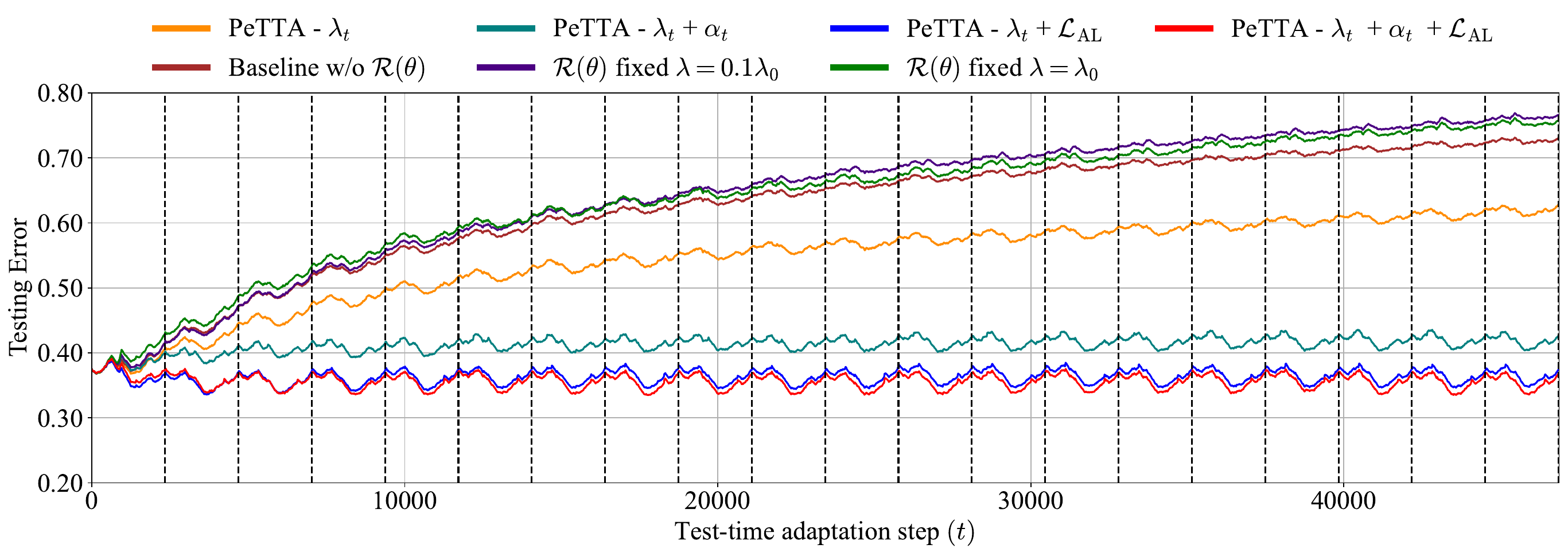}};
        \draw (0, -6) node[inner sep=0] {\includegraphics[width=\linewidth]{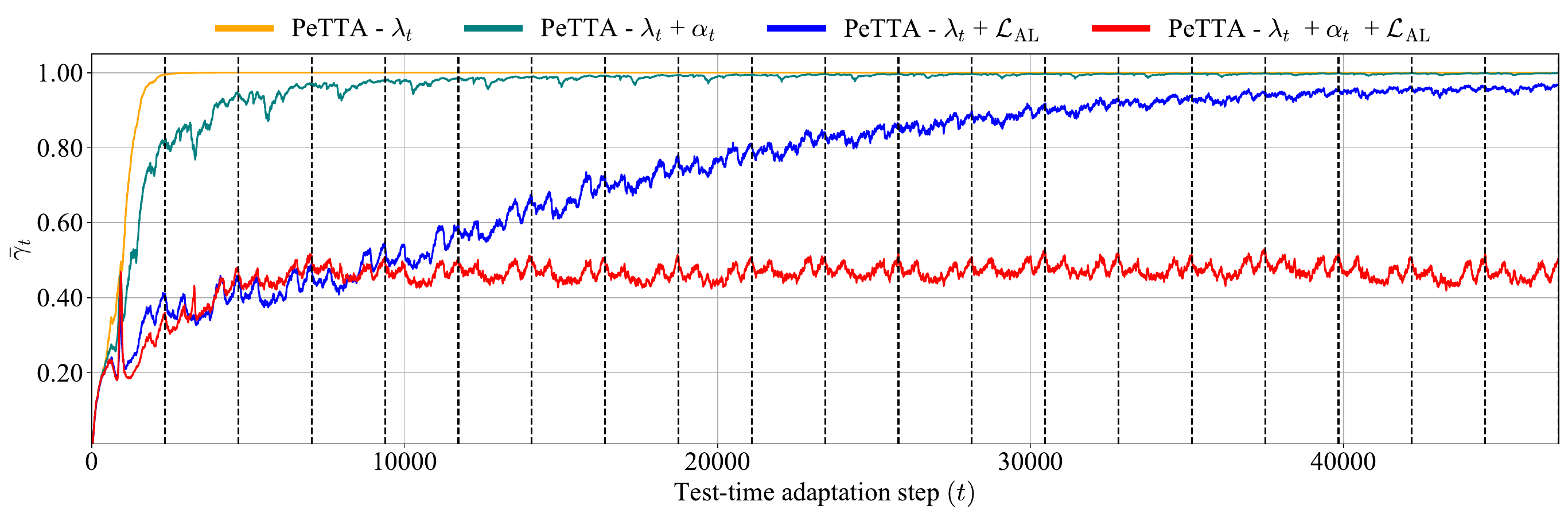}};
    
        \end{tikzpicture}
    \caption{An inspection on the ablation study of multiple variations of \method on the task CIFAR-100 $\rightarrow$ CIFAR-100-C~\cite{hendrycks2019robustness} in an episodic TTA with 20 visits (visits are separated by the vertical dashed lines). 
    \textbf{(top)}: testing error of multiple variations of \method. The performance of \method without \textcolor{Sepia}{(w/o) $\mathcal{R}(\theta)$}, or fixed regularization coefficient ($\textcolor{ForestGreen}{\lambda = \lambda_0}/\textcolor{Purple}{0.1\lambda_0}$) degrades through time (the top 3 lines). The degradation of \textcolor{orange}{\method~-$\lambda_t$}  is still happening but at a slower rate (justification below). 
    The performance of the other three variations persists through time with \textcolor{red}{\method~-$\lambda_t + \alpha_t + \mathcal{L}_{\mathrm{AL}}$} achieves the best performance. 
    \textbf{(bottom)}: changes of $\bar \gamma_t$ in multiple variations of \method. When limiting the degree of freedom in adjusting $\alpha_t$ or lacking of supervision from $\mathcal{L}_\mathrm{AL}$ (e.g., \textcolor{BlueGreen}{\method~-$\lambda_t$ + $\alpha_t$}, \textcolor{blue}{\method~-$\lambda_t$ + $\mathcal{L}_{\mathrm{AL}}$}, and especially \textcolor{orange}{\method~-$\lambda_t$}), the value of $\gamma_t$, unfortunately, escalates and eventually saturated. After this point, \method has the same effect as using a fixed regularization coefficient. Therefore, fully utilizing all components is necessary to preserve the persistence of \method. Best viewed in color.
    }
    \label{fig:petta_ablation_compare}
    \vspace*{-1.0\baselineskip}
\end{figure*}

\Fig{\ref{fig:petta_ablation_compare}} presents an additional examination of the ablation study conducted on the task CIFAR-100 $\rightarrow$ CIFAR-100-C~\cite{hendrycks2019robustness} for our \method approach. We plot the classification error (top) and the value of $\bar \gamma_t$ (bottom) for various \method variations. As the model diverges from the initial state, the value of $\bar \gamma_t$ increases. Unable to adjust $\alpha_t$ or constraint the probability space via $\mathcal{L}_{\mathrm{AL}}$ limits the ability of \method to prevent model collapse. In all variations with the model collapse in ablation studies, the rapid saturation of $\bar \gamma_t$ is all observed. Therefore, incorporating all components in \method is necessary.

\subsection{More Confusion Matrices in \Setting TTA Setting}
\begin{figure}[ht!]
    \pgfplotsset{every x tick label/.append style={font=\tiny, yshift=0.5ex}}
    \pgfplotsset{every y tick label/.append style={font=\tiny, xshift=0.5ex}}
    \centering
        \begin{tikzpicture}
        
        \def \h{4.4}
        \def \u{0.6}
        \def \w{3.4}
        \def \ox{0.3} 
        \def \imgwidth{3.35}
        \def \r{0} 
        \draw (\ox + \w * 0, -\h * \r) node[inner sep=0] {\includegraphics[width=\imgwidth cm]{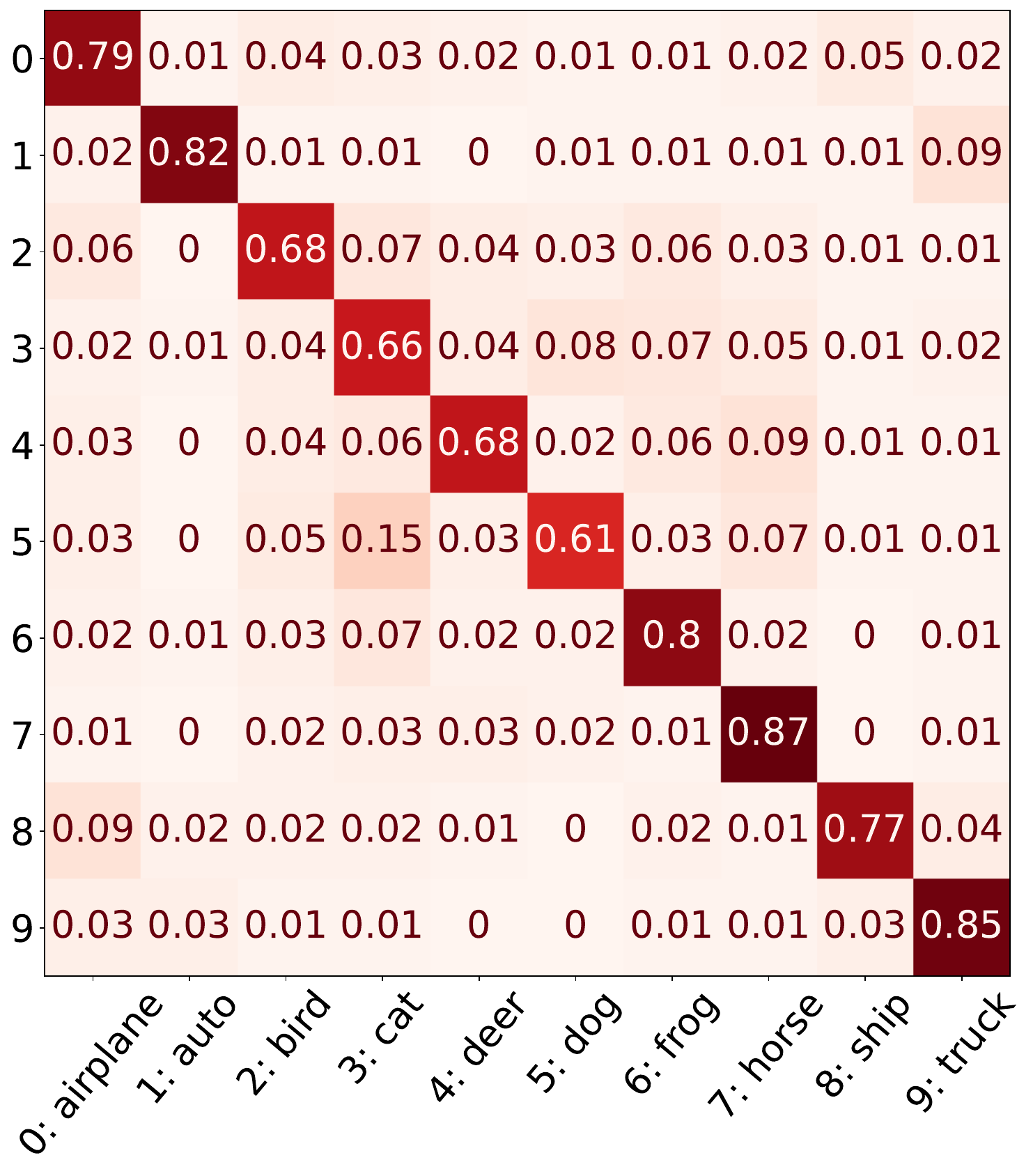}};
        \draw (\ox + \w * 1, -\h * \r) node[inner sep=0] {\includegraphics[width=\imgwidth cm]{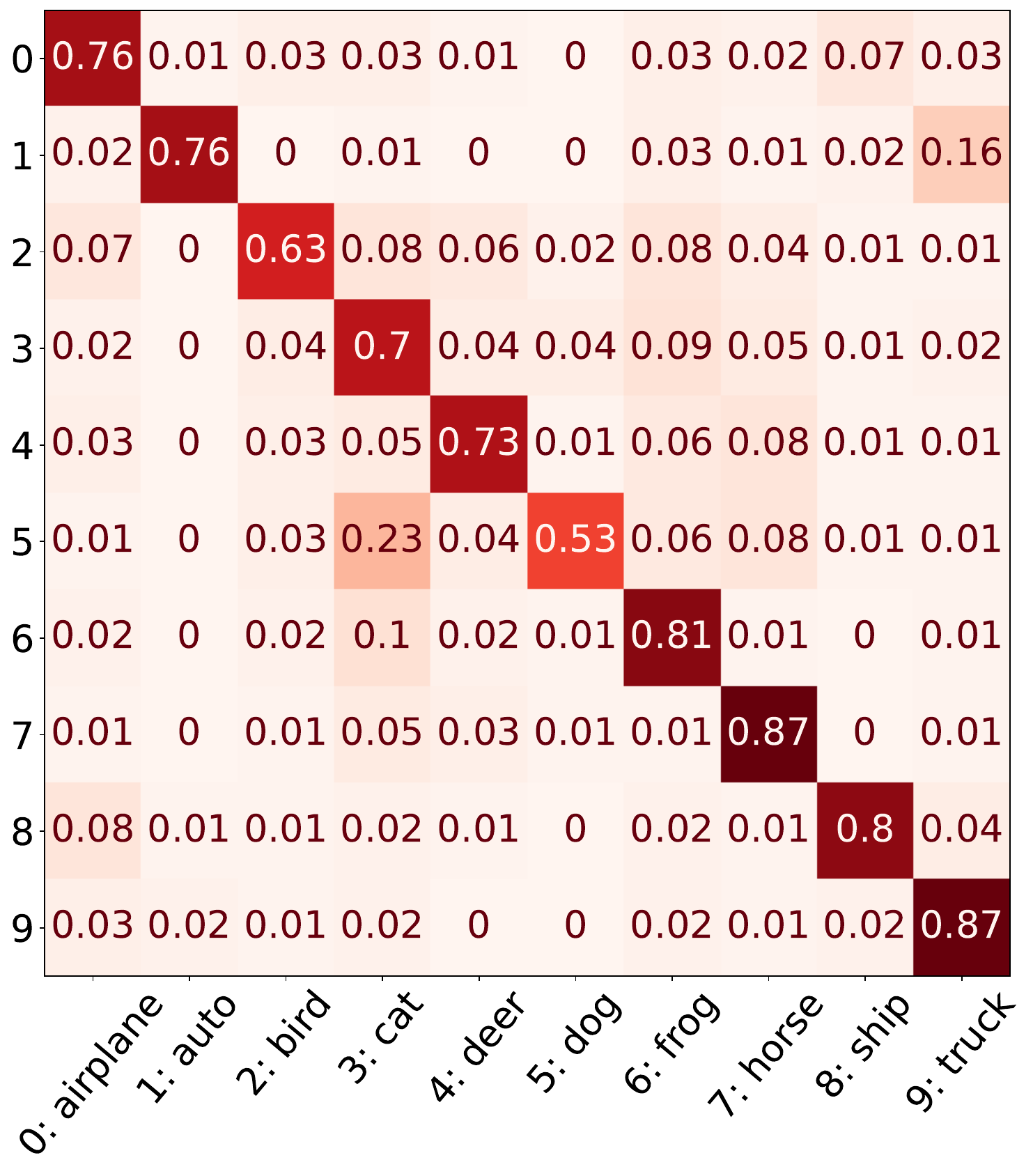}};
        \draw (\ox + \w * 2, -\h * \r) node[inner sep=0] {\includegraphics[width=\imgwidth cm]{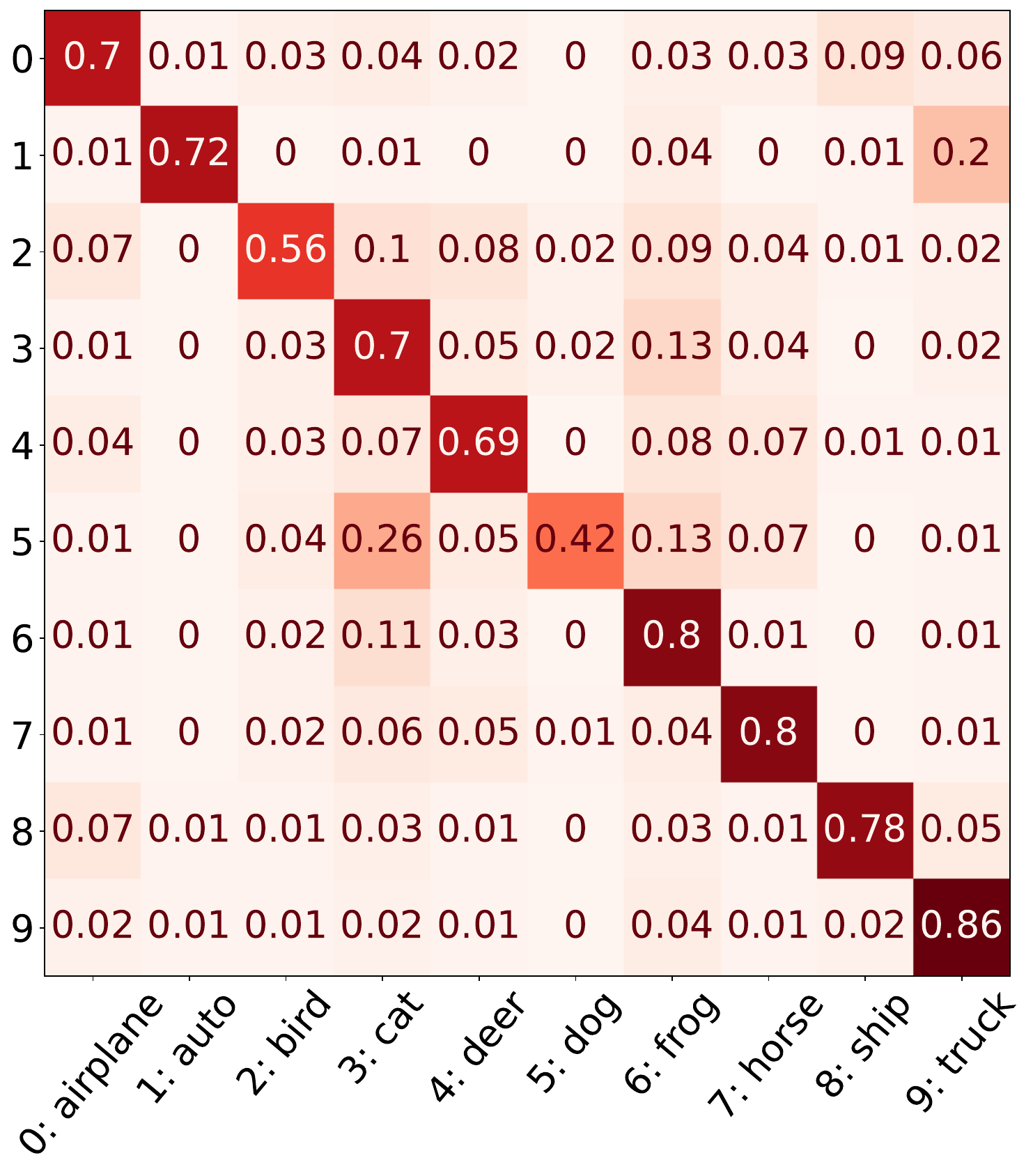}};
        \draw (\ox + \w * 3 ,-\h * \r) node[inner sep=0] {\includegraphics[width=\imgwidth cm]{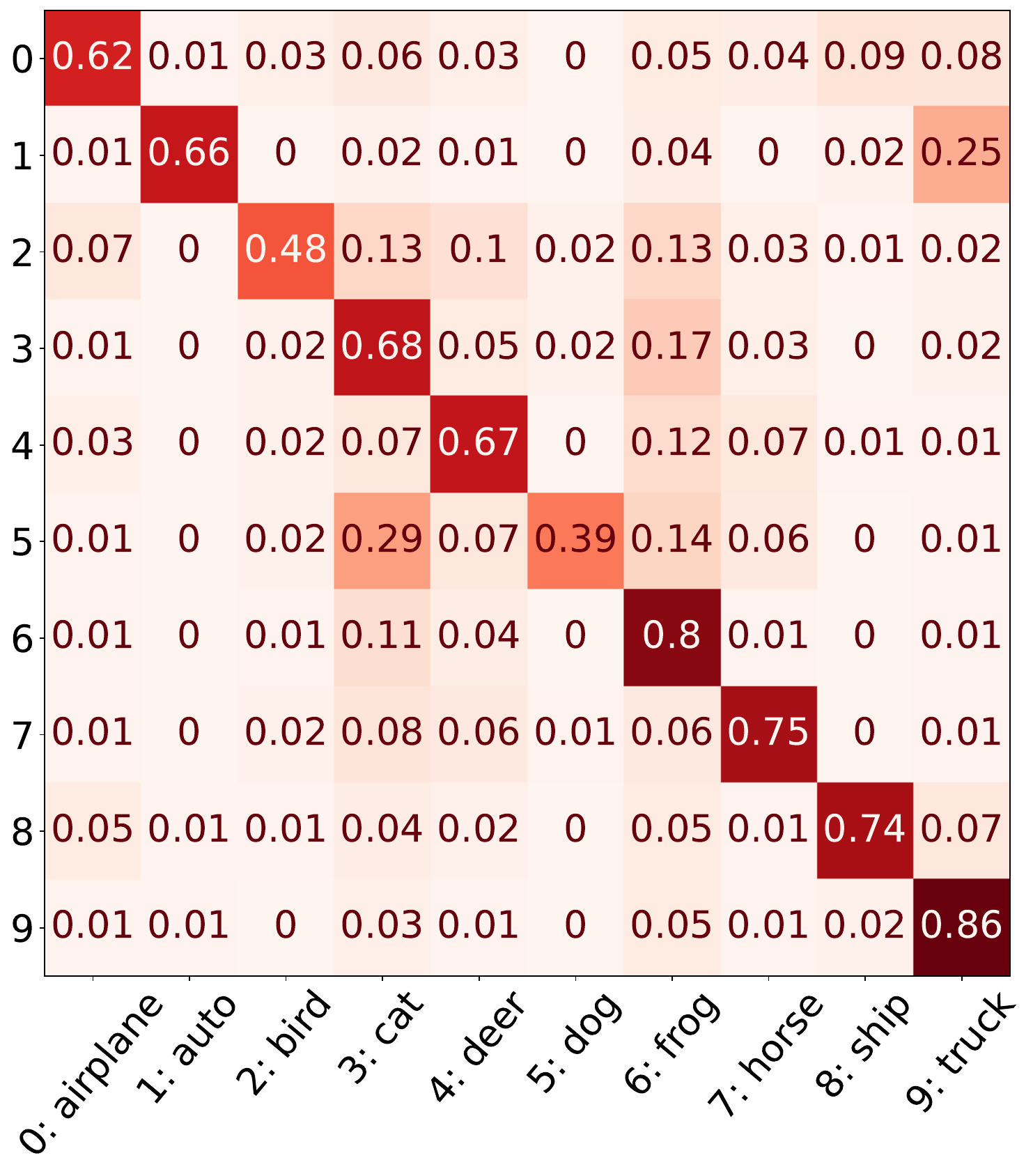}};

        \node at (-\imgwidth * 0.25, -\h * \r + 1, 2) [rotate=90] {\scriptsize{True label}};

        \node at (\ox + \w * 0, -\h * \r + \imgwidth * 0.55) [above] {\scriptsize{1\textsuperscript{st} visit}};
        \node at (\ox + \w * 1, -\h * \r + \imgwidth * 0.55) [above] {\scriptsize{2\textsuperscript{nd} visit}};
        \node at (\ox + \w * 2, -\h * \r + \imgwidth * 0.55) [above] {\scriptsize{3\textsuperscript{rd} visit}};
        \node at (\ox + \w * 3, -\h * \r + \imgwidth * 0.55) [above] {\scriptsize{4\textsuperscript{th} visit}};
        
        \def \r{1} 
        \draw (\ox + \w * 0, -\h * \r) node[inner sep=0] {\includegraphics[width=\imgwidth cm]{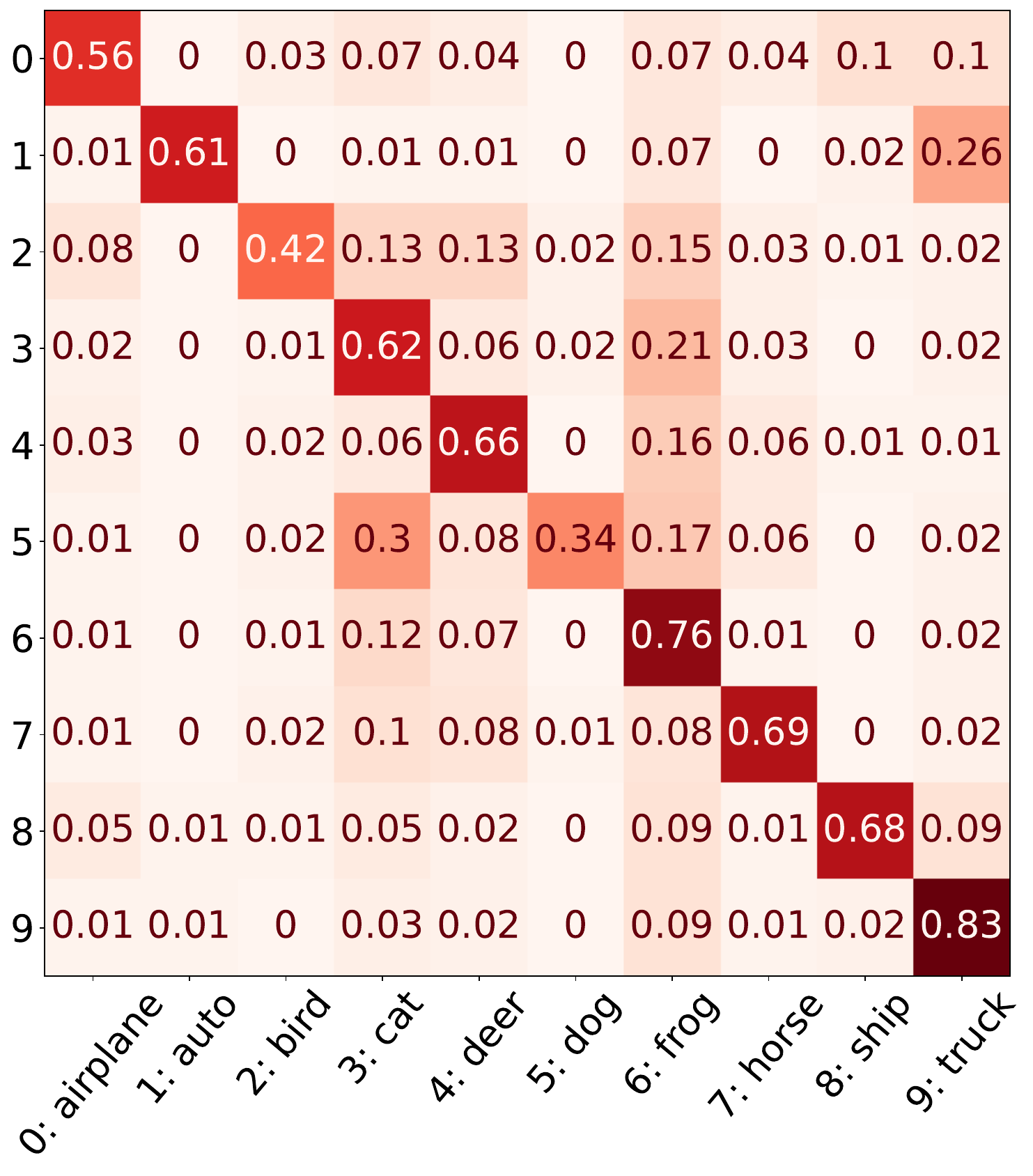}};
        \draw (\ox + \w * 1, -\h * \r) node[inner sep=0] {\includegraphics[width=\imgwidth cm]{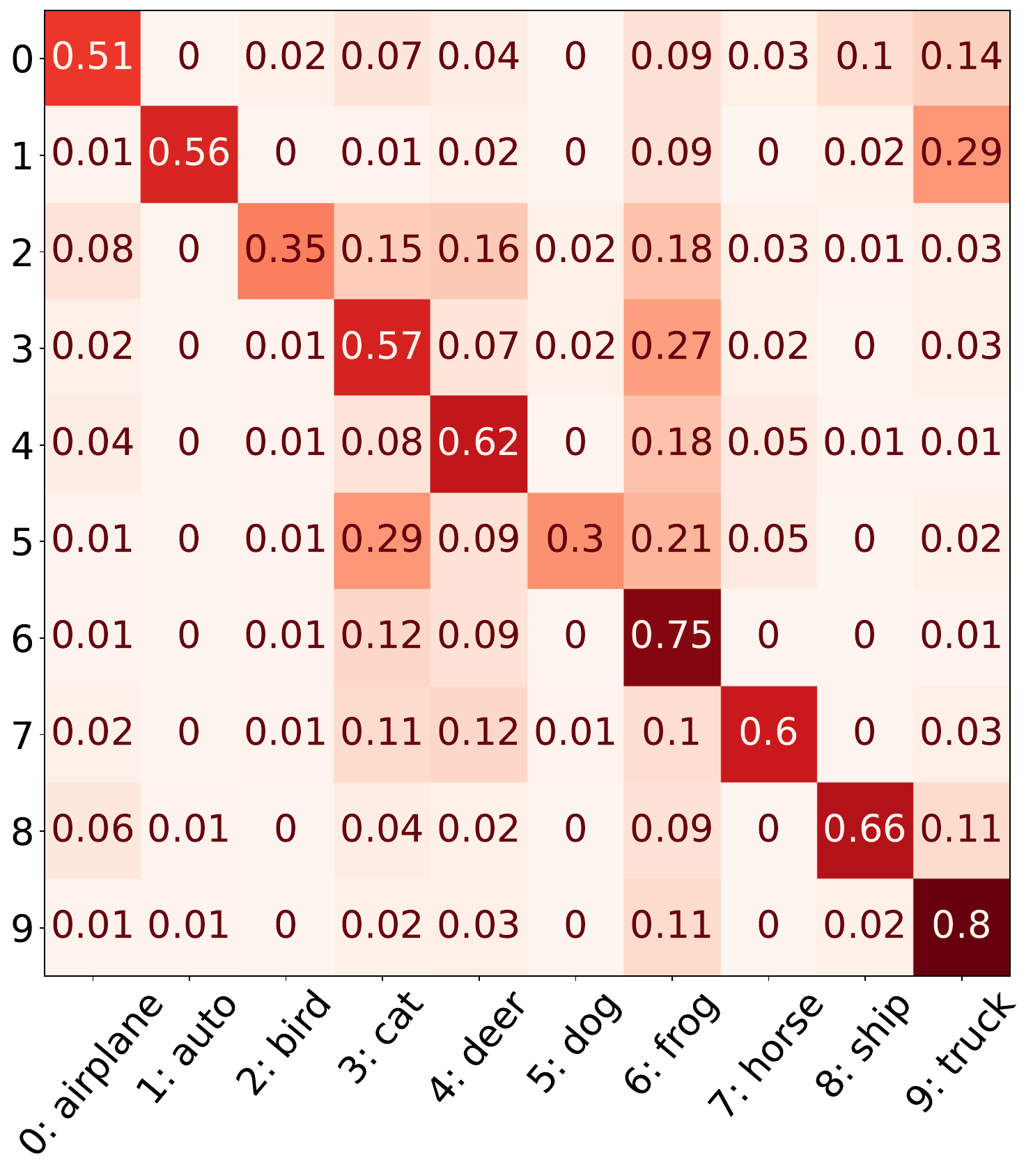}};
        \draw (\ox + \w * 2, -\h * \r) node[inner sep=0] {\includegraphics[width=\imgwidth cm]{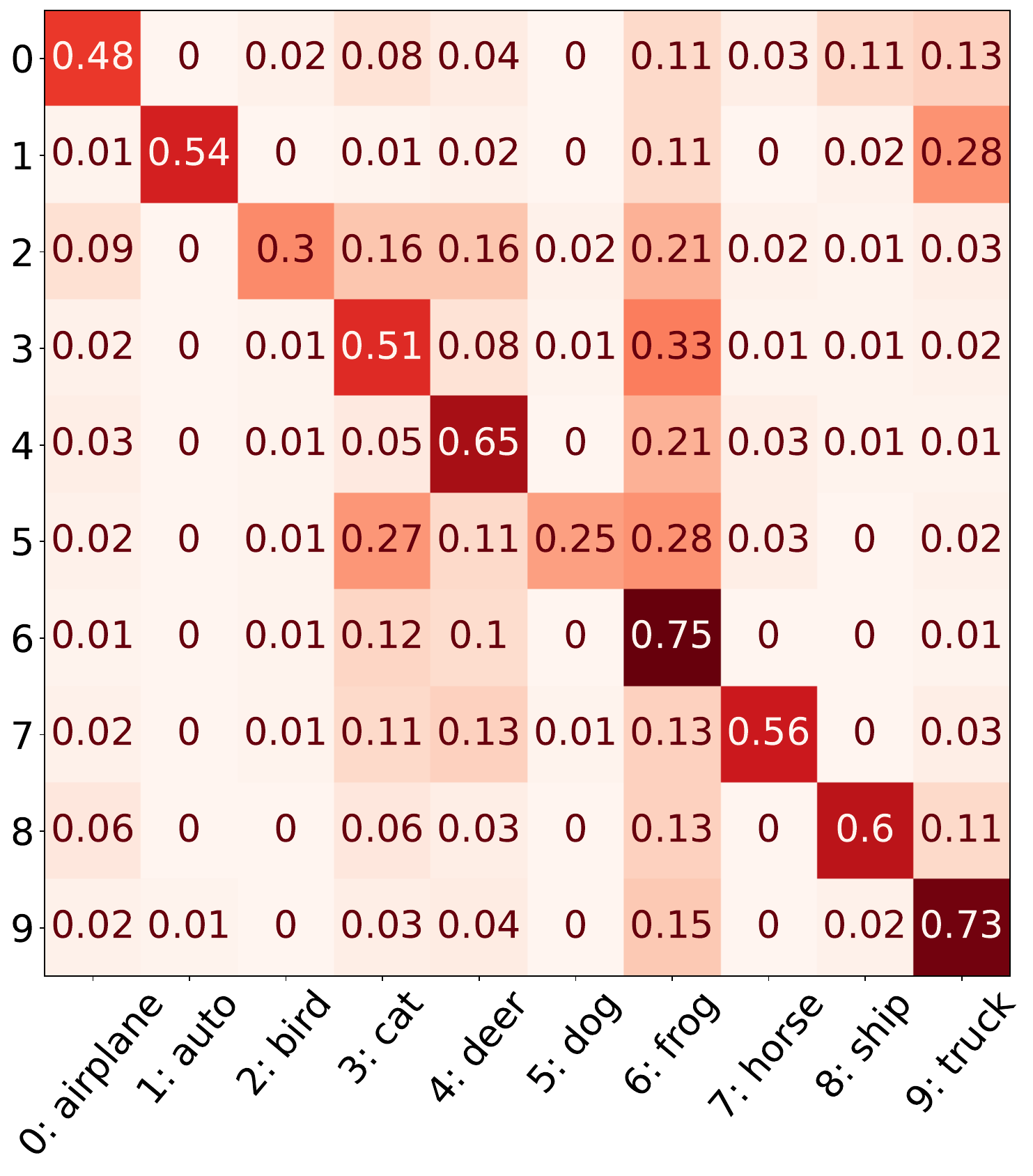}};
        \draw (\ox + \w * 3 ,-\h * \r) node[inner sep=0] {\includegraphics[width=\imgwidth cm]{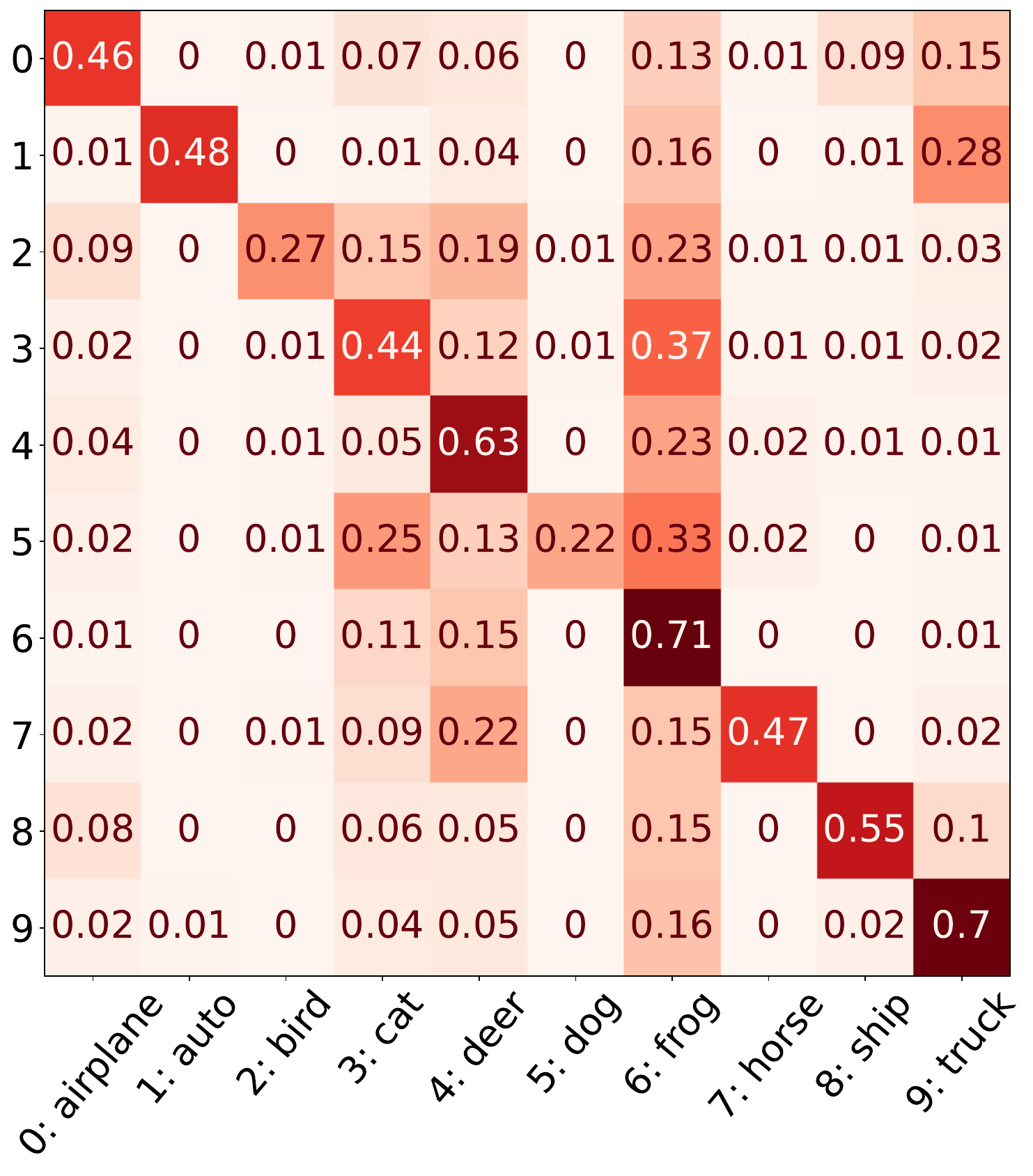}};

        \node at (-\imgwidth * 0.25, -\h * \r + 1, 2) [rotate=90] {\scriptsize{True label}};

        \node at (\ox + \w * 0, -\h * \r + \imgwidth * 0.55) [above] {\scriptsize{5\textsuperscript{th} visit}};
        \node at (\ox + \w * 1, -\h * \r + \imgwidth * 0.55) [above] {\scriptsize{6\textsuperscript{th} visit}};
        \node at (\ox + \w * 2, -\h * \r + \imgwidth * 0.55) [above] {\scriptsize{7\textsuperscript{th} visit}};
        \node at (\ox + \w * 3, -\h * \r + \imgwidth * 0.55) [above] {\scriptsize{8\textsuperscript{th} visit}};

        \def \r{2} 
        \draw (\ox + \w * 0, -\h * \r) node[inner sep=0] {\includegraphics[width=\imgwidth cm]{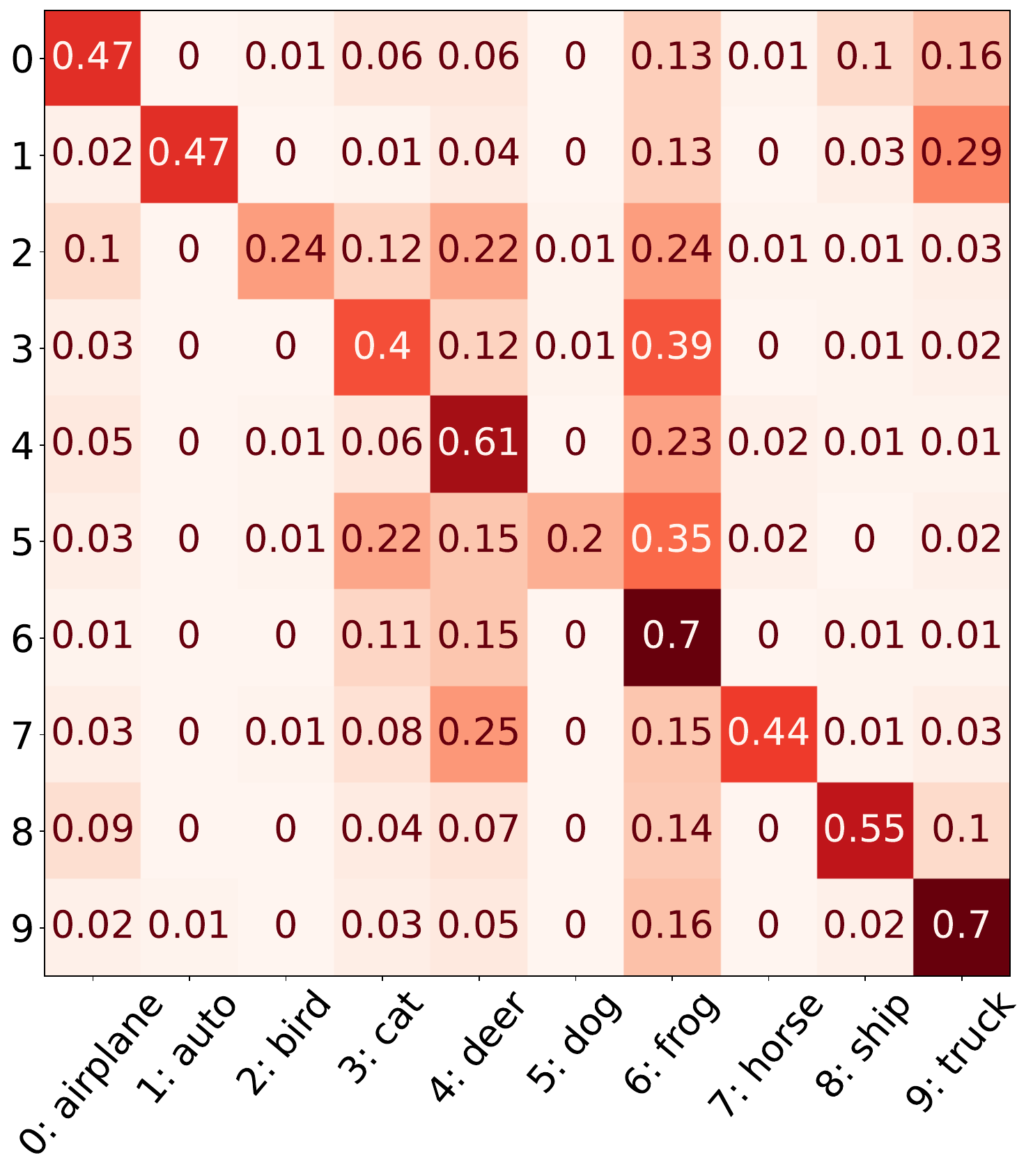}};
        \draw (\ox + \w * 1, -\h * \r) node[inner sep=0] {\includegraphics[width=\imgwidth cm]{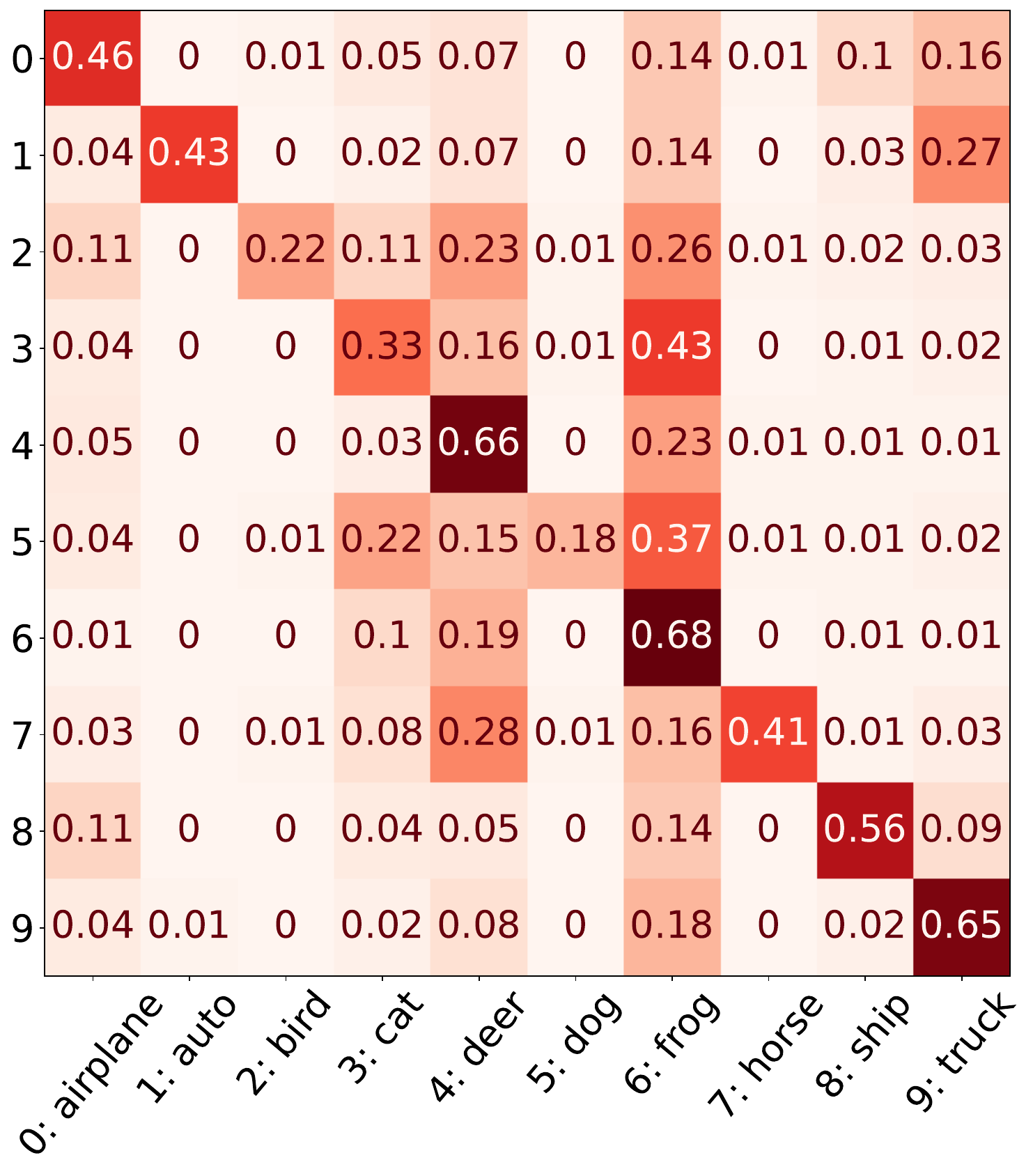}};
        \draw (\ox + \w * 2, -\h * \r) node[inner sep=0] {\includegraphics[width=\imgwidth cm]{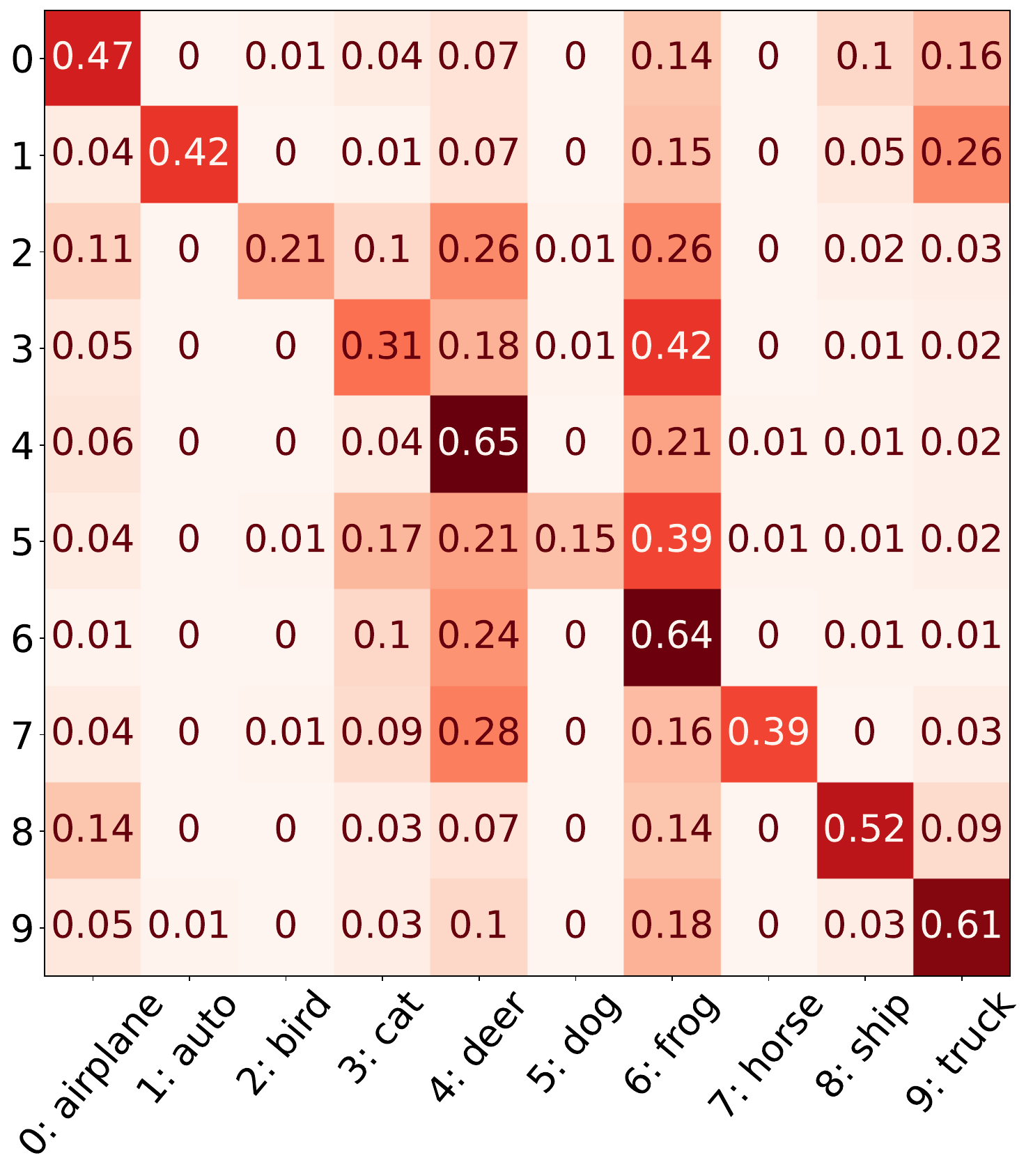}};
        \draw (\ox + \w * 3 ,-\h * \r) node[inner sep=0] {\includegraphics[width=\imgwidth cm]{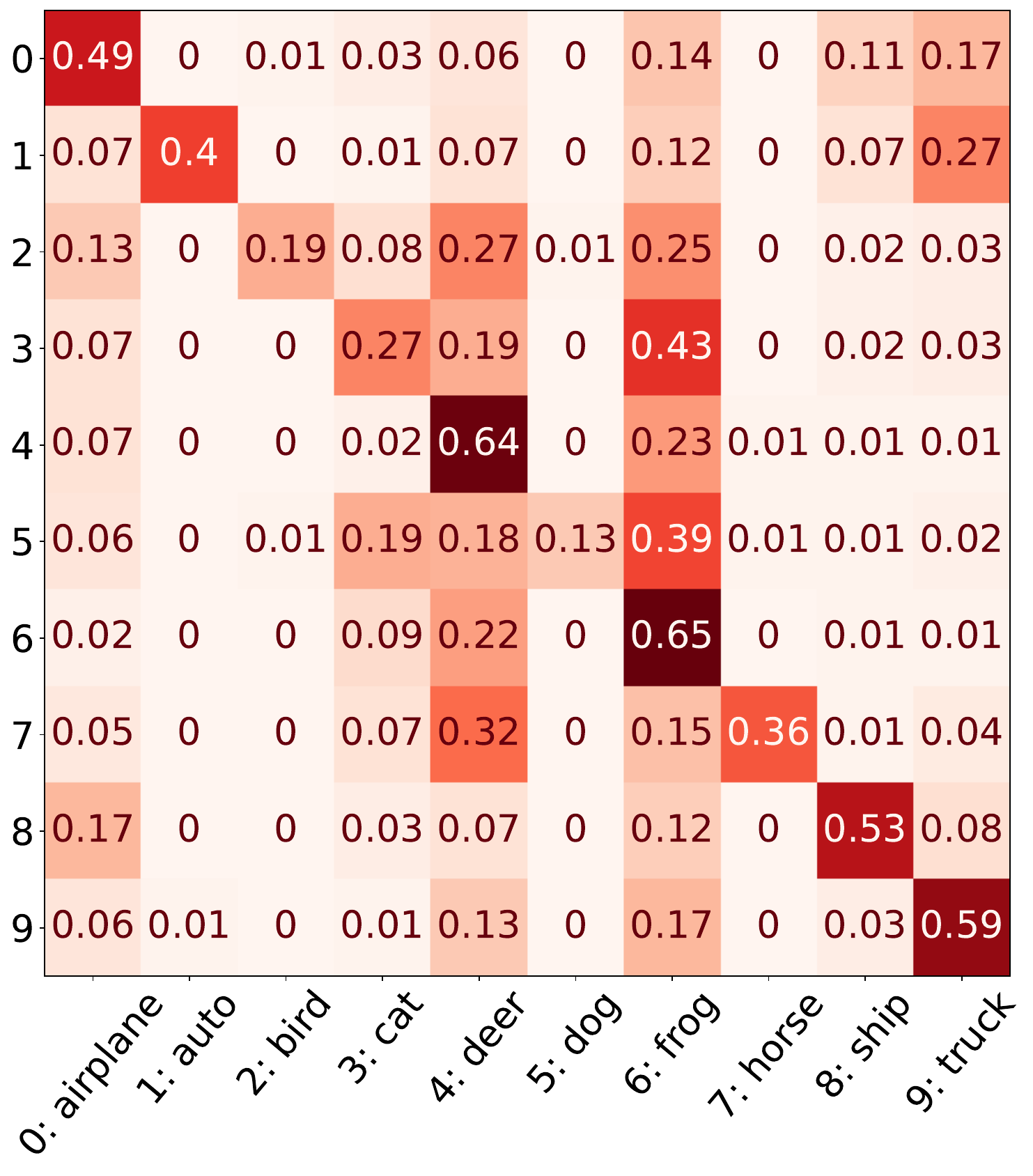}};

        \node at (-\imgwidth * 0.25, -\h * \r + 1, 2) [rotate=90] {\scriptsize{True label}};

        \node at (\ox + \w * 0, -\h * \r + \imgwidth * 0.55) [above] {\scriptsize{9\textsuperscript{th} visit}};
        \node at (\ox + \w * 1, -\h * \r + \imgwidth * 0.55) [above] {\scriptsize{10\textsuperscript{th} visit}};
        \node at (\ox + \w * 2, -\h * \r + \imgwidth * 0.55) [above] {\scriptsize{11\textsuperscript{th} visit}};
        \node at (\ox + \w * 3, -\h * \r + \imgwidth * 0.55) [above] {\scriptsize{12\textsuperscript{th} visit}};

        \def \r{3} 
        \draw (\ox + \w * 0, -\h * \r) node[inner sep=0] {\includegraphics[width=\imgwidth cm]{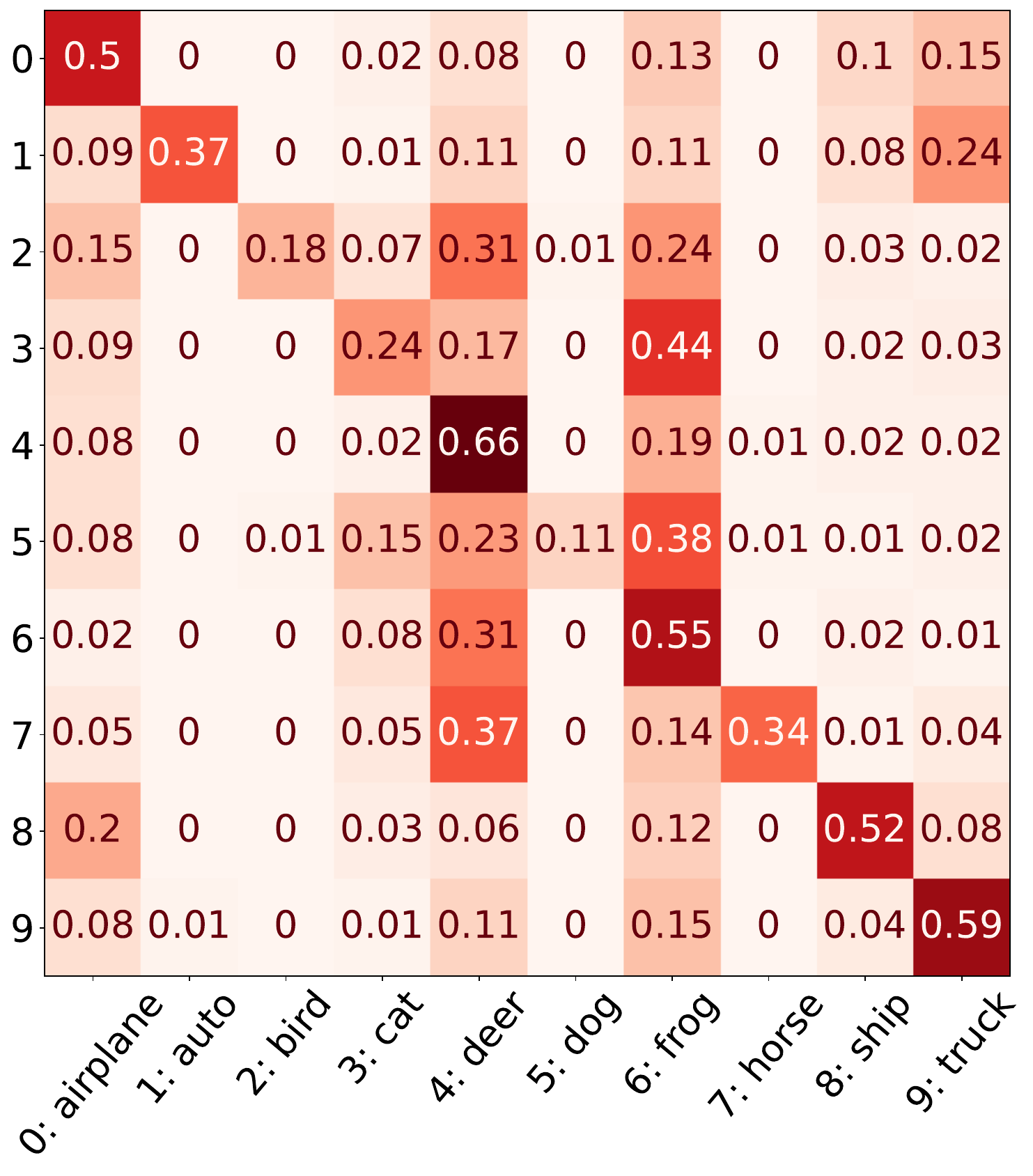}};
        \draw (\ox + \w * 1, -\h * \r) node[inner sep=0] {\includegraphics[width=\imgwidth cm]{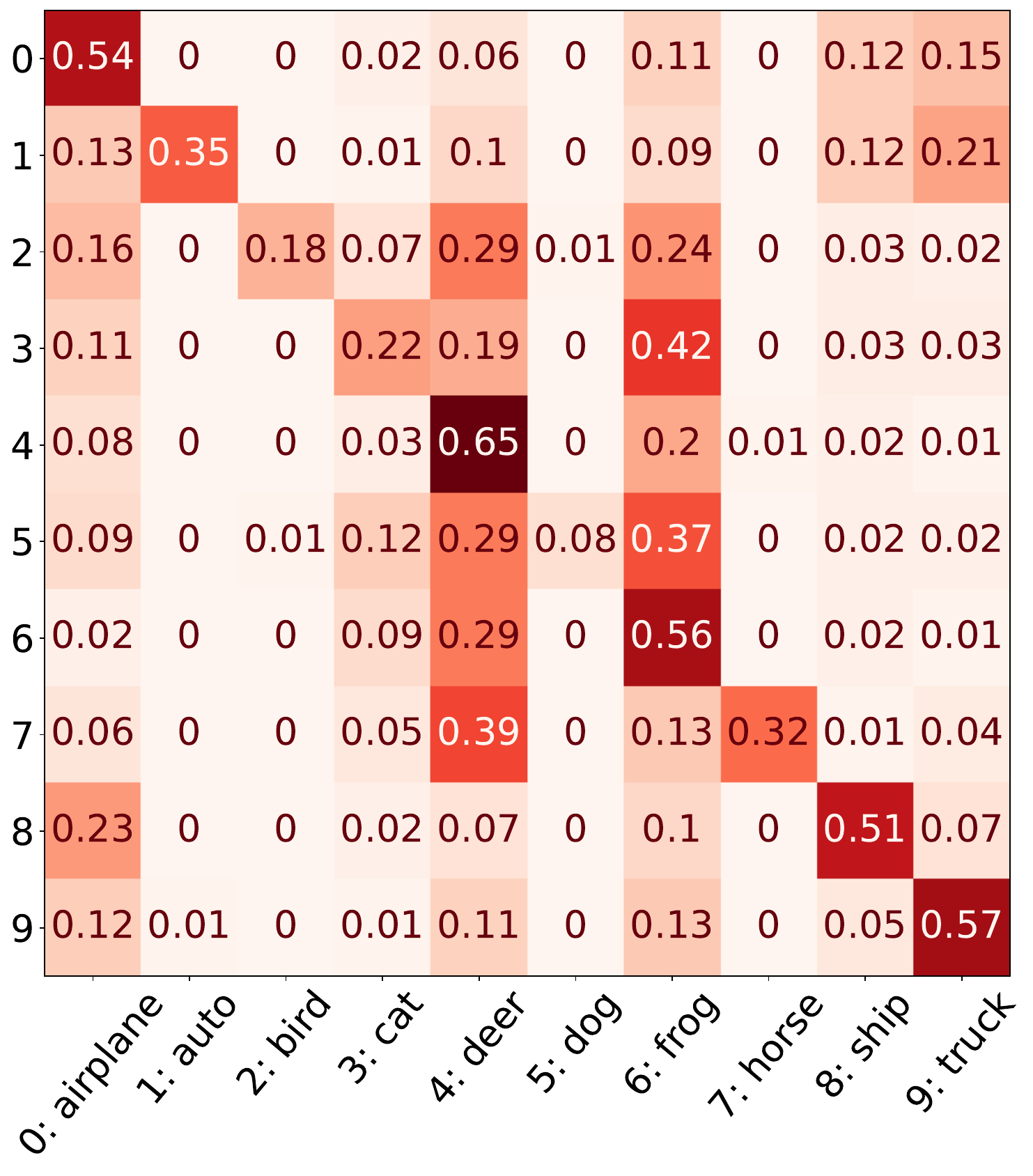}};
        \draw (\ox + \w * 2, -\h * \r) node[inner sep=0] {\includegraphics[width=\imgwidth cm]{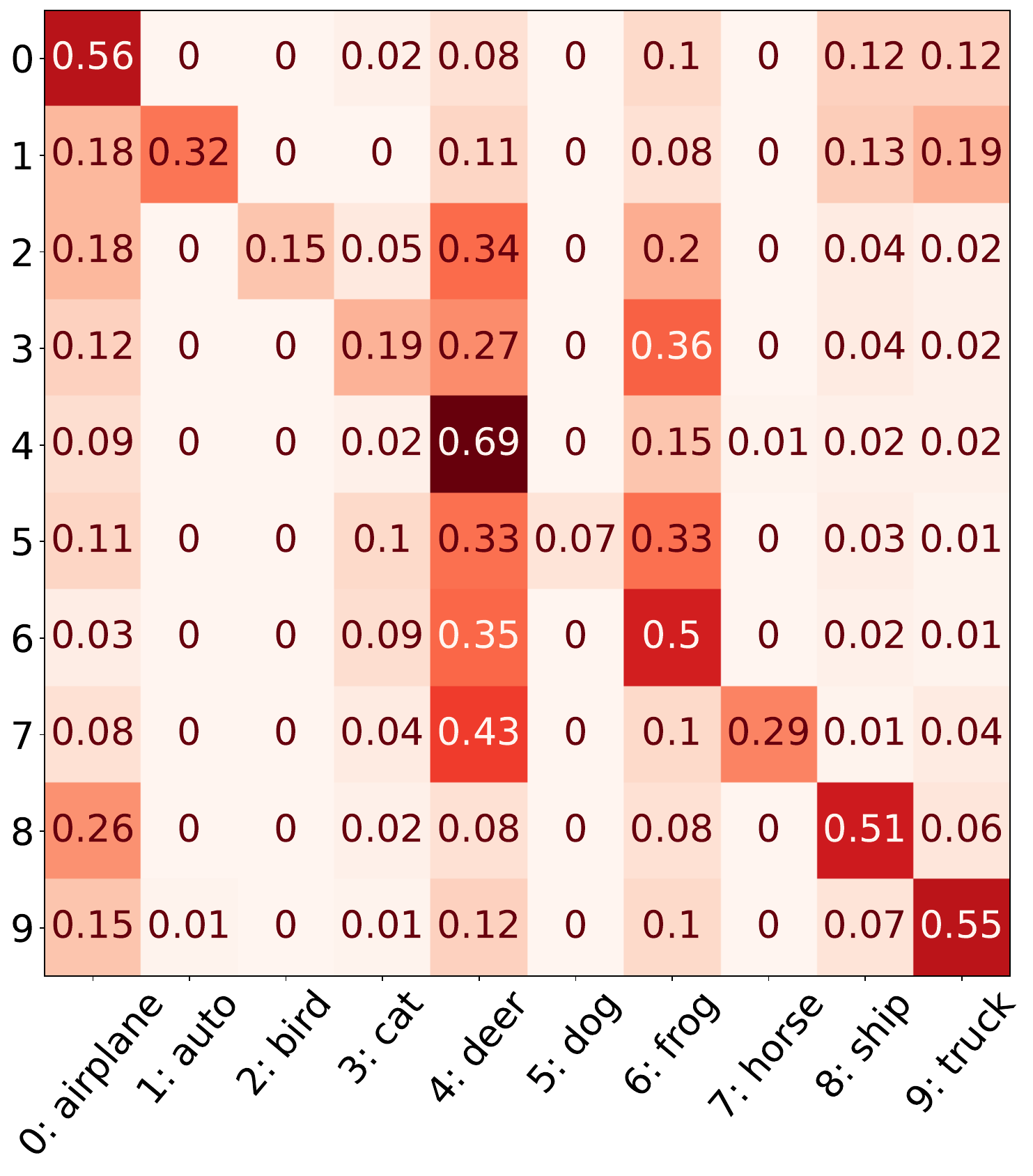}};
        \draw (\ox + \w * 3 ,-\h * \r) node[inner sep=0] {\includegraphics[width=\imgwidth cm]{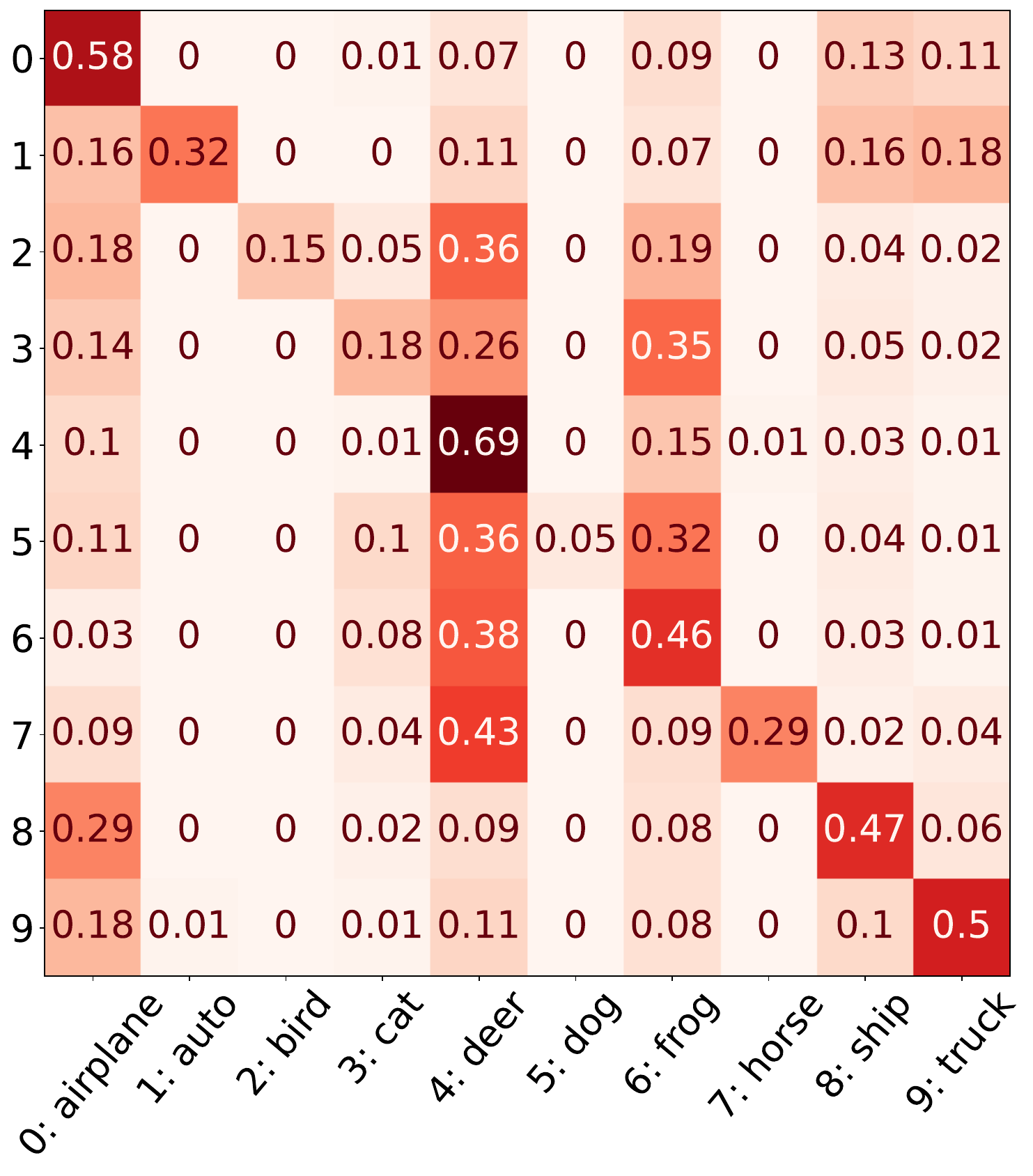}};

        \node at (-\imgwidth * 0.25, -\h * \r + 1, 2) [rotate=90] {\scriptsize{True label}};

        \node at (\ox + \w * 0, -\h * \r + \imgwidth * 0.55) [above] {\scriptsize{13\textsuperscript{th} visit}};
        \node at (\ox + \w * 1, -\h * \r + \imgwidth * 0.55) [above] {\scriptsize{14\textsuperscript{th} visit}};
        \node at (\ox + \w * 2, -\h * \r + \imgwidth * 0.55) [above] {\scriptsize{15\textsuperscript{th} visit}};
        \node at (\ox + \w * 3, -\h * \r + \imgwidth * 0.55) [above] {\scriptsize{16\textsuperscript{th} visit}};
        
        \def \r{4} 
        \draw (\ox + \w * 0, -\h * \r) node[inner sep=0] {\includegraphics[width=\imgwidth cm]{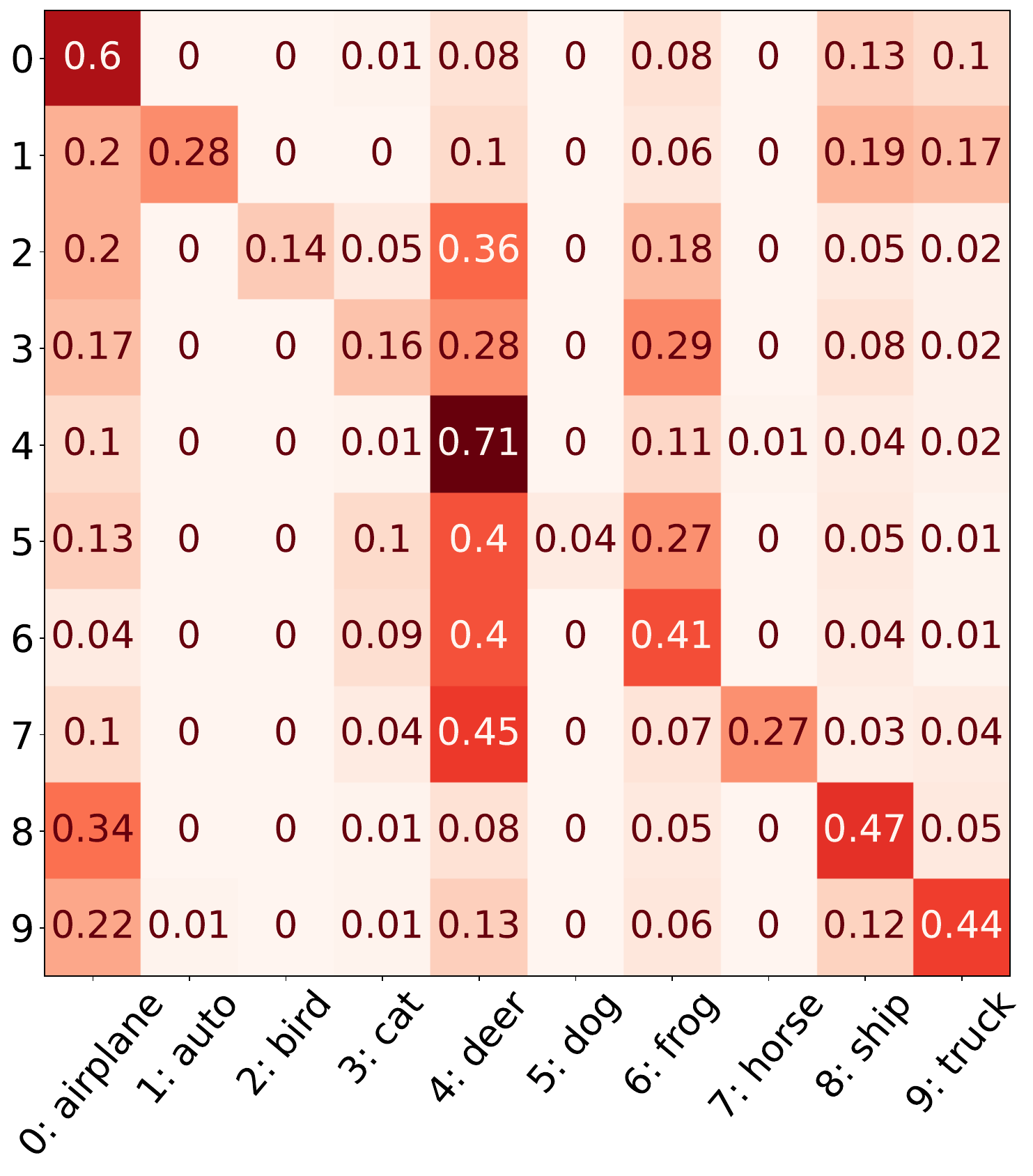}};
        \draw (\ox + \w * 1, -\h * \r) node[inner sep=0] {\includegraphics[width=\imgwidth cm]{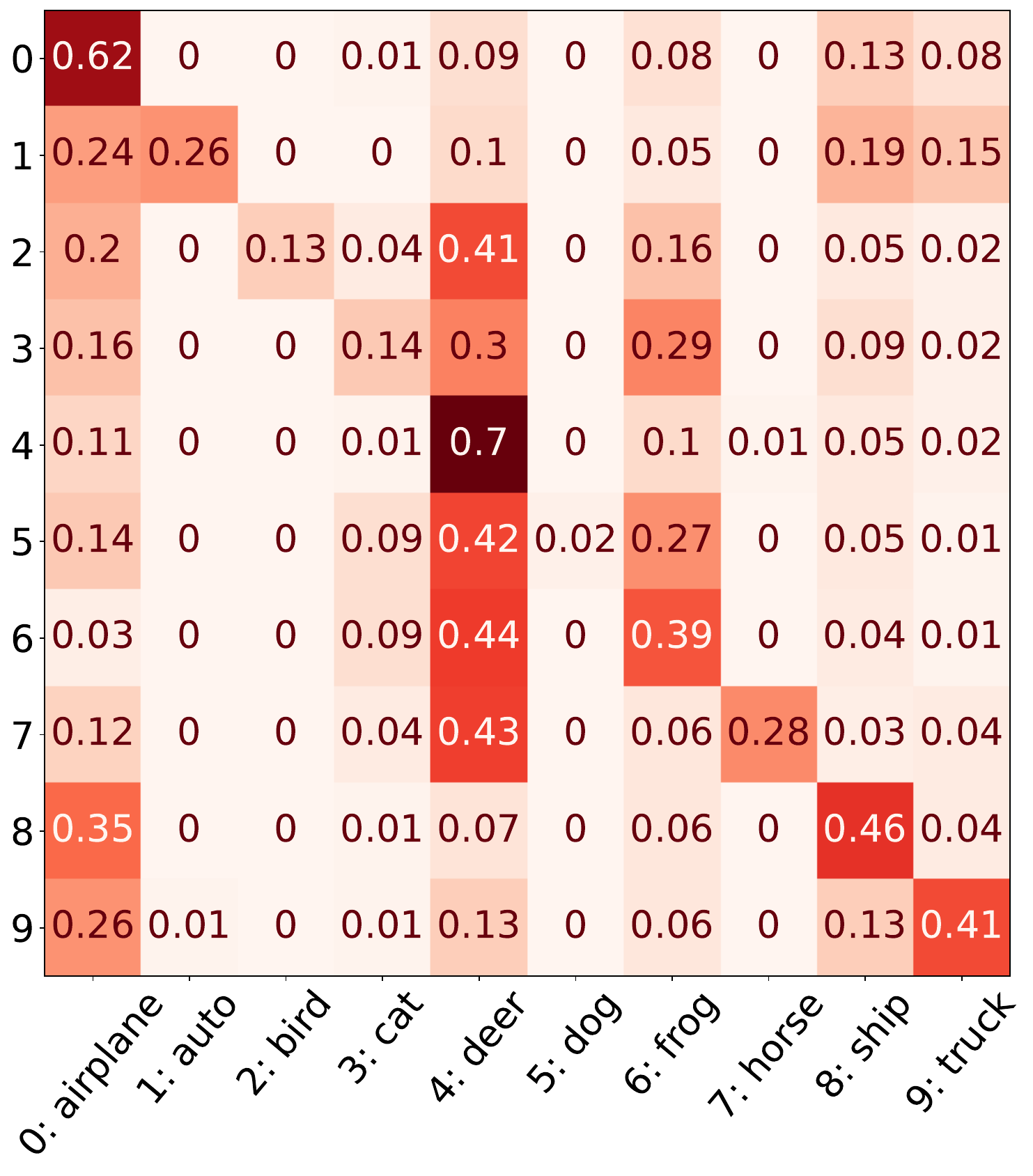}};
        \draw (\ox + \w * 2, -\h * \r) node[inner sep=0] {\includegraphics[width=\imgwidth cm]{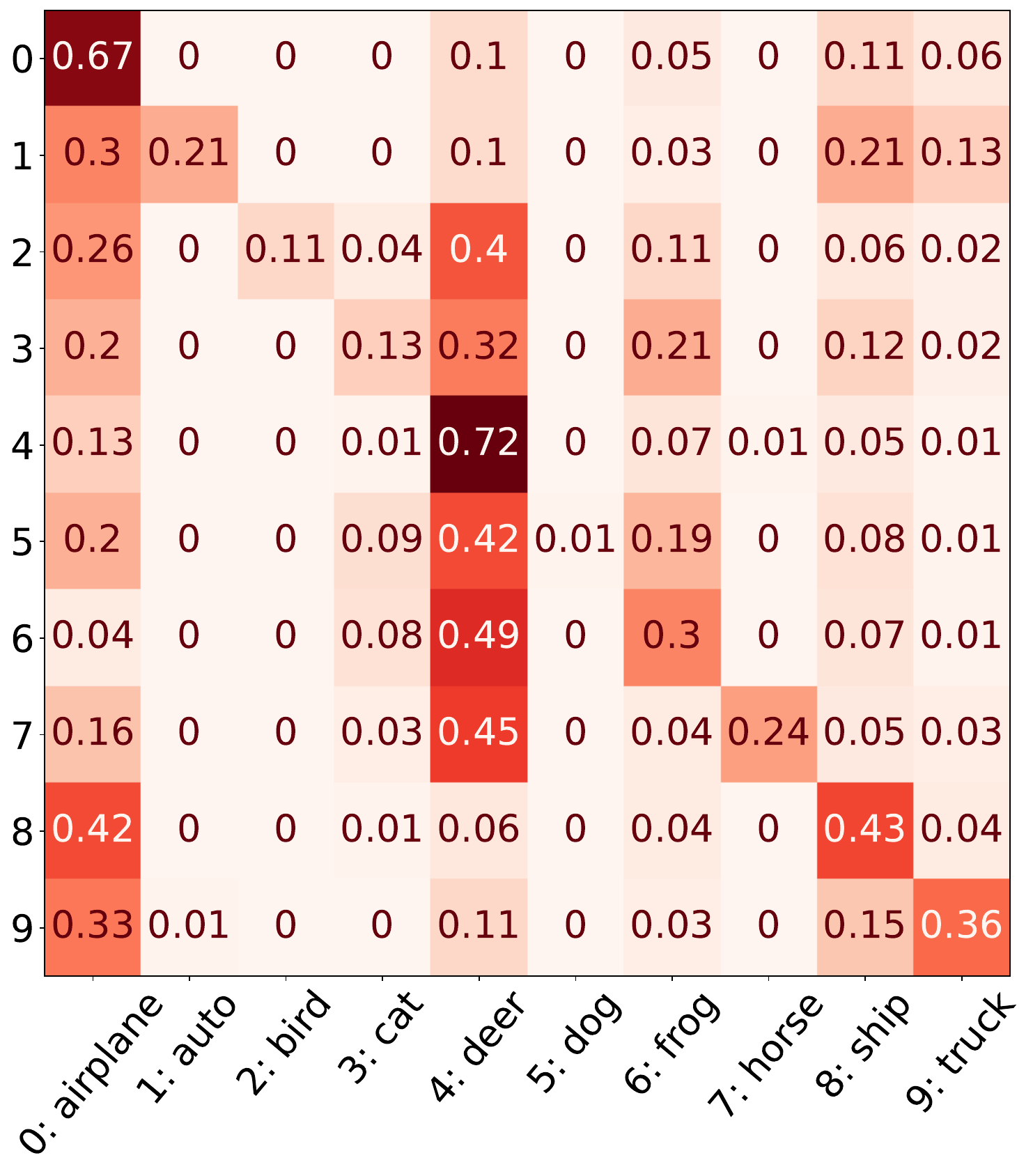}};
        \draw (\ox + \w * 3 ,-\h * \r) node[inner sep=0] {\includegraphics[width=\imgwidth cm]{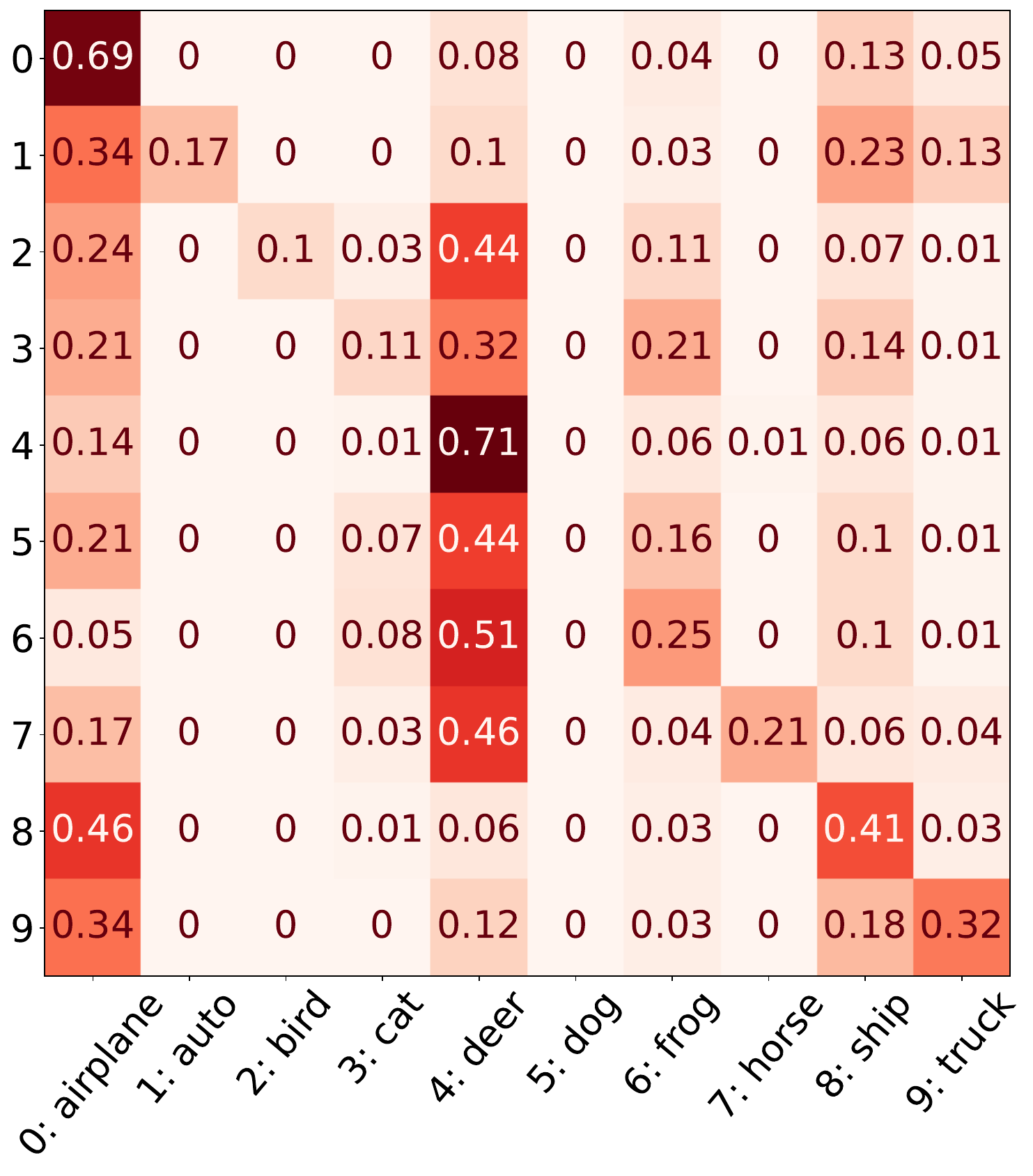}};

        \node at (\ox + \w * 0, -\h * \r - \imgwidth * 0.6) [above] {\scriptsize{Predicted label}};
        \node at (\ox + \w * 1, -\h * \r - \imgwidth * 0.6) [above] {\scriptsize{Predicted label}};
        \node at (\ox + \w * 2, -\h * \r - \imgwidth * 0.6) [above] {\scriptsize{Predicted label}};
        \node at (\ox + \w * 3, -\h * \r - \imgwidth * 0.6) [above] {\scriptsize{Predicted label}};
        
        \node at (-\imgwidth * 0.25, -\h * \r + 1, 2) [rotate=90] {\scriptsize{True label}};

        \node at (\ox + \w * 0, -\h * \r + \imgwidth * 0.55) [above] {\scriptsize{17\textsuperscript{th} visit}};
        \node at (\ox + \w * 1, -\h * \r + \imgwidth * 0.55) [above] {\scriptsize{18\textsuperscript{th} visit}};
        \node at (\ox + \w * 2, -\h * \r + \imgwidth * 0.55) [above] {\scriptsize{19\textsuperscript{th} visit}};
        \node at (\ox + \w * 3, -\h * \r + \imgwidth * 0.55) [above] {\scriptsize{20\textsuperscript{th} visit}};
        
        \end{tikzpicture}
    \vspace*{-0.5\baselineskip}
    \caption{The dynamic of the confusion matrix of RoTTA~\cite{yuan2023robust} in episodic TTA with 20 visits.}
    \vspace*{-\baselineskip}
    \label{fig:full_cfs_mat_rotta}
\end{figure}

\begin{figure*}[ht!]
    \pgfplotsset{every x tick label/.append style={font=\tiny, yshift=0.5ex}}
    \pgfplotsset{every y tick label/.append style={font=\tiny, xshift=0.5ex}}
    \centering
        \begin{tikzpicture}
        
        \def \h{4.4}
        \def \u{0.6}
        \def \w{3.4}
        \def \ox{0.3} 
        \def \imgwidth{3.35}
        \def \r{0} 
        \draw (\ox + \w * 0, -\h * \r) node[inner sep=0] {\includegraphics[width=\imgwidth cm]{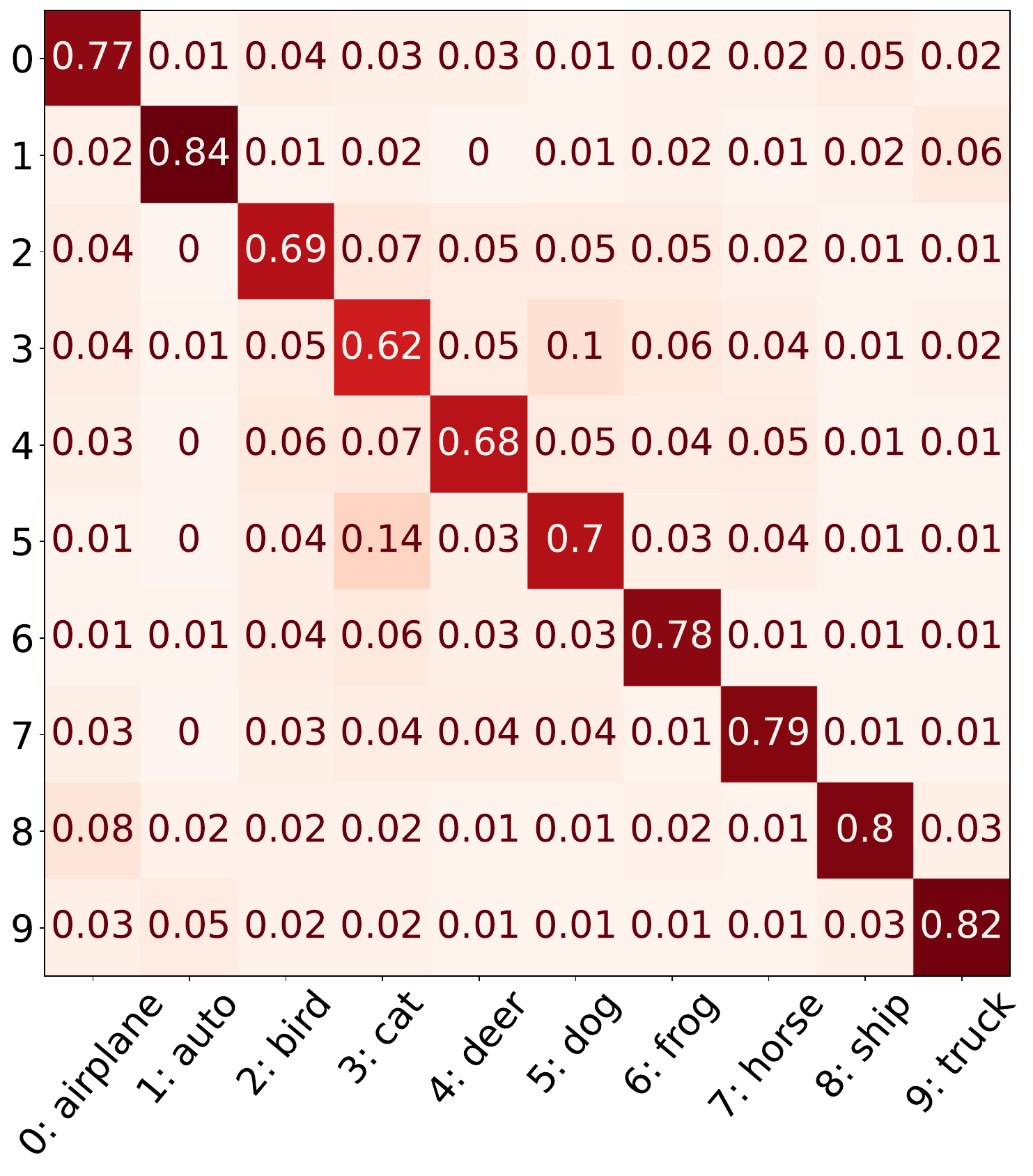}};
        \draw (\ox + \w * 1, -\h * \r) node[inner sep=0] {\includegraphics[width=\imgwidth cm]{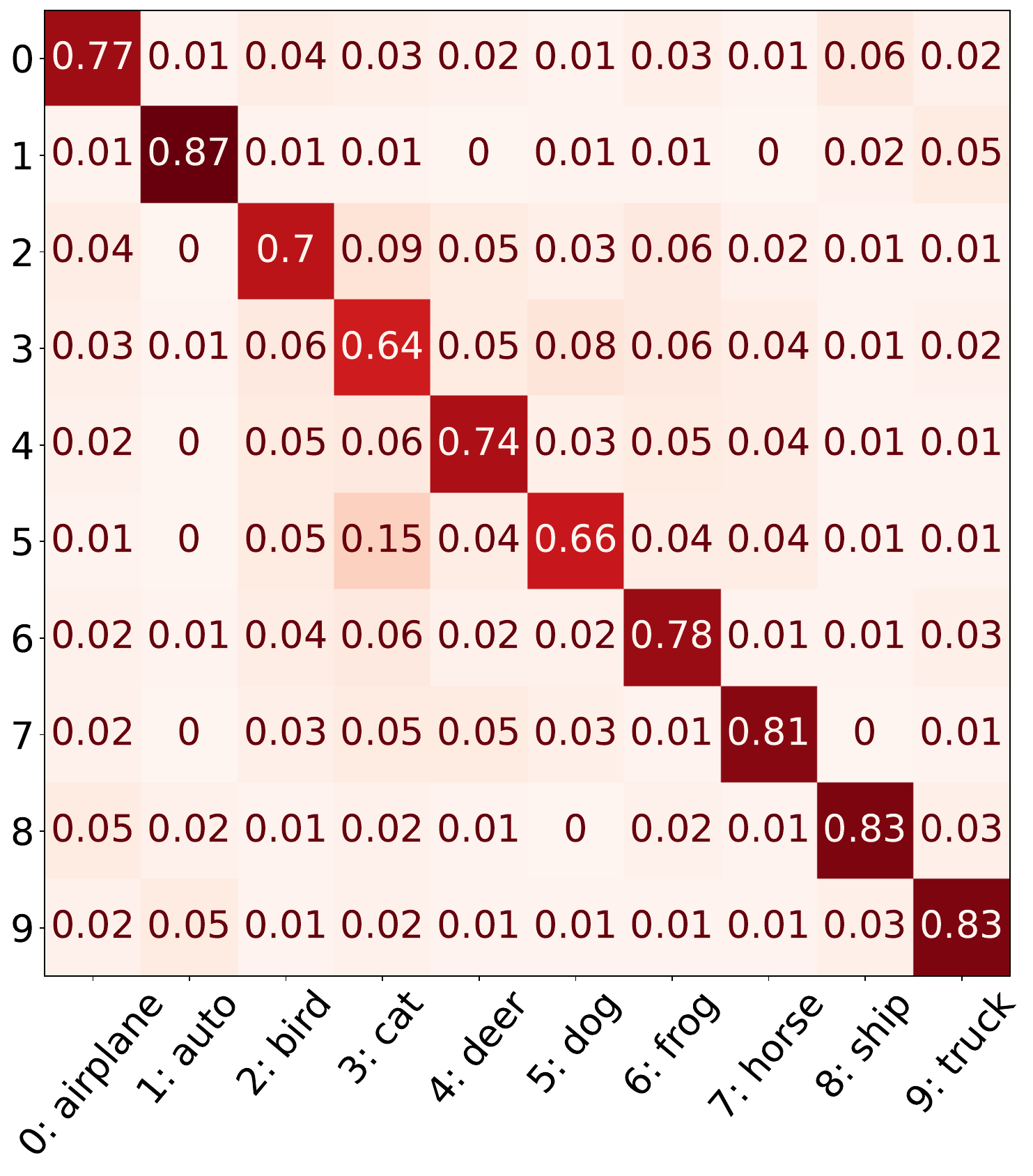}};
        \draw (\ox + \w * 2, -\h * \r) node[inner sep=0] {\includegraphics[width=\imgwidth cm]{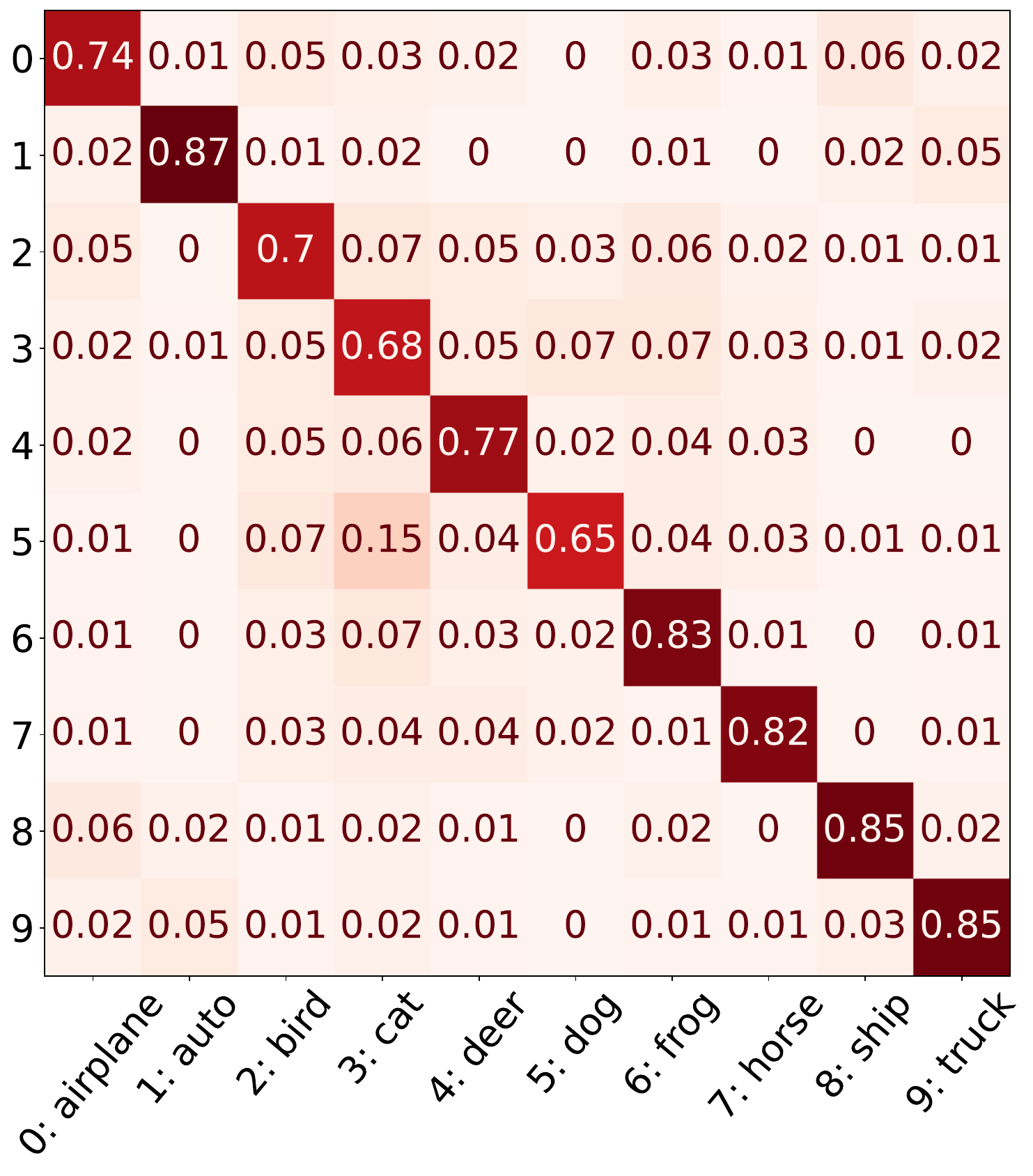}};
        \draw (\ox + \w * 3 ,-\h * \r) node[inner sep=0] {\includegraphics[width=\imgwidth cm]{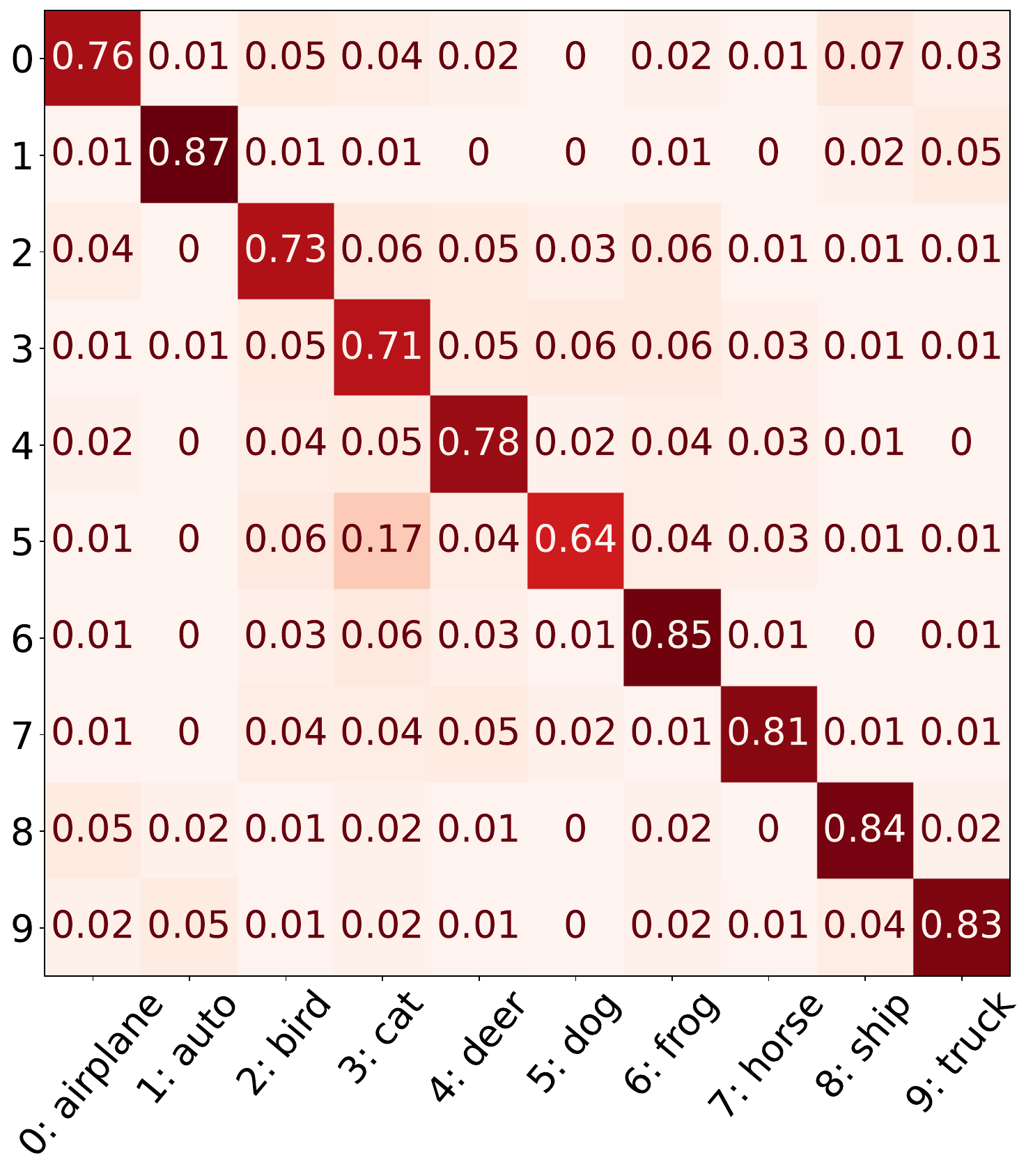}};

        \node at (-\imgwidth * 0.25, -\h * \r + 1, 2) [rotate=90] {\scriptsize{True label}};

        \node at (\ox + \w * 0, -\h * \r + \imgwidth * 0.55) [above] {\scriptsize{1\textsuperscript{st} visit}};
        \node at (\ox + \w * 1, -\h * \r + \imgwidth * 0.55) [above] {\scriptsize{2\textsuperscript{nd} visit}};
        \node at (\ox + \w * 2, -\h * \r + \imgwidth * 0.55) [above] {\scriptsize{3\textsuperscript{rd} visit}};
        \node at (\ox + \w * 3, -\h * \r + \imgwidth * 0.55) [above] {\scriptsize{4\textsuperscript{th} visit}};
        
        \def \r{1} 
        \draw (\ox + \w * 0, -\h * \r) node[inner sep=0] {\includegraphics[width=\imgwidth cm]{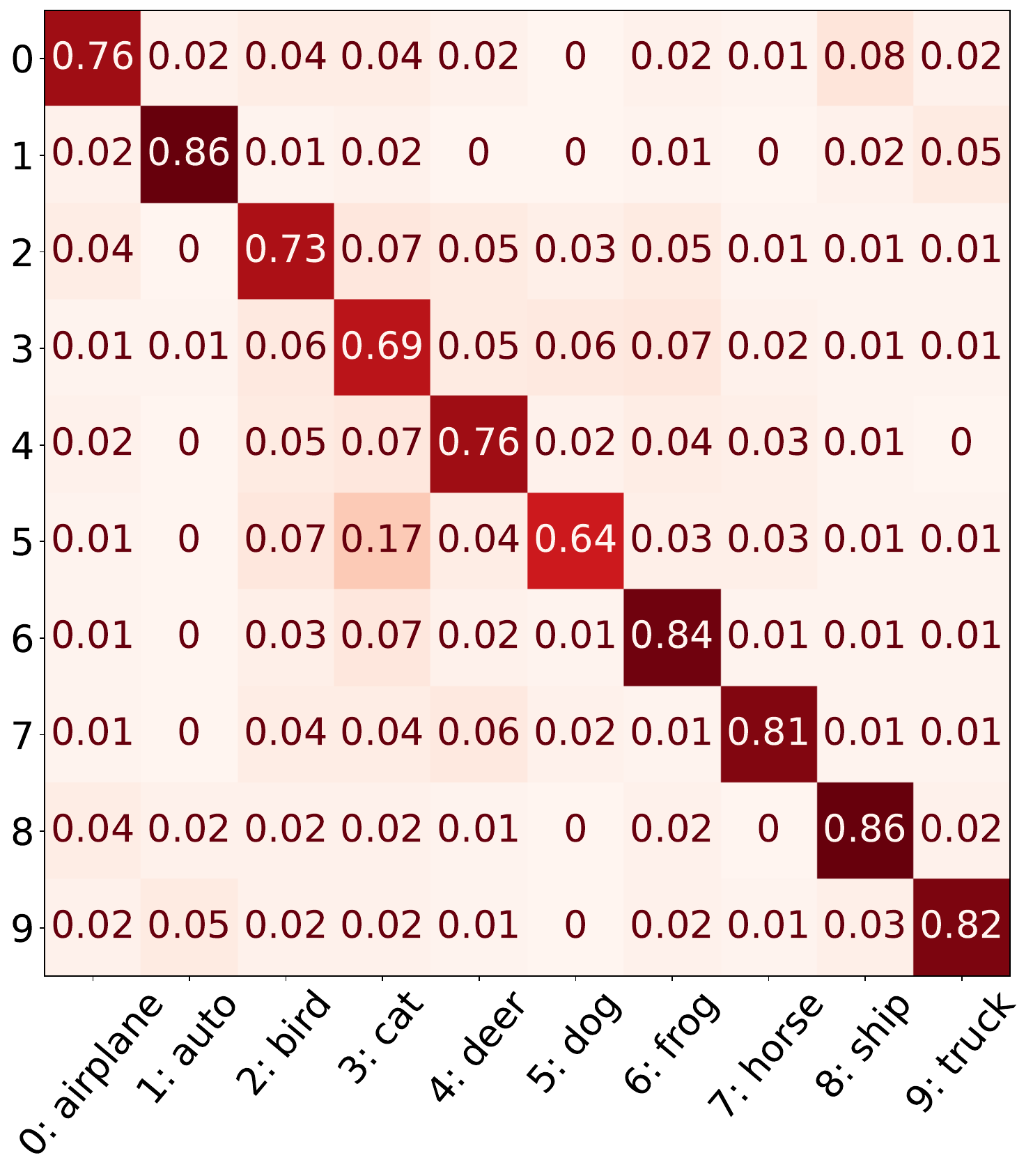}};
        \draw (\ox + \w * 1, -\h * \r) node[inner sep=0] {\includegraphics[width=\imgwidth cm]{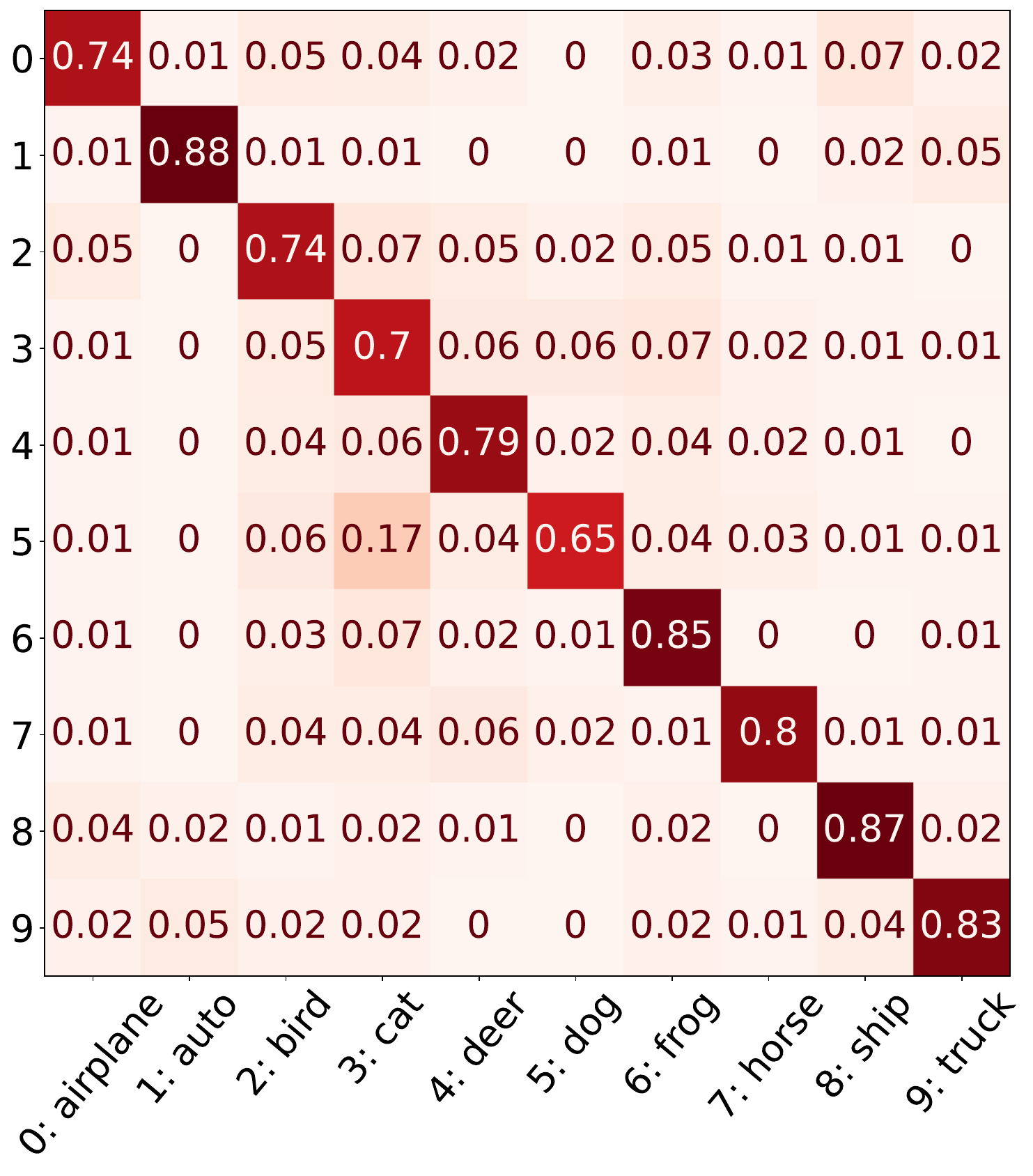}};
        \draw (\ox + \w * 2, -\h * \r) node[inner sep=0] {\includegraphics[width=\imgwidth cm]{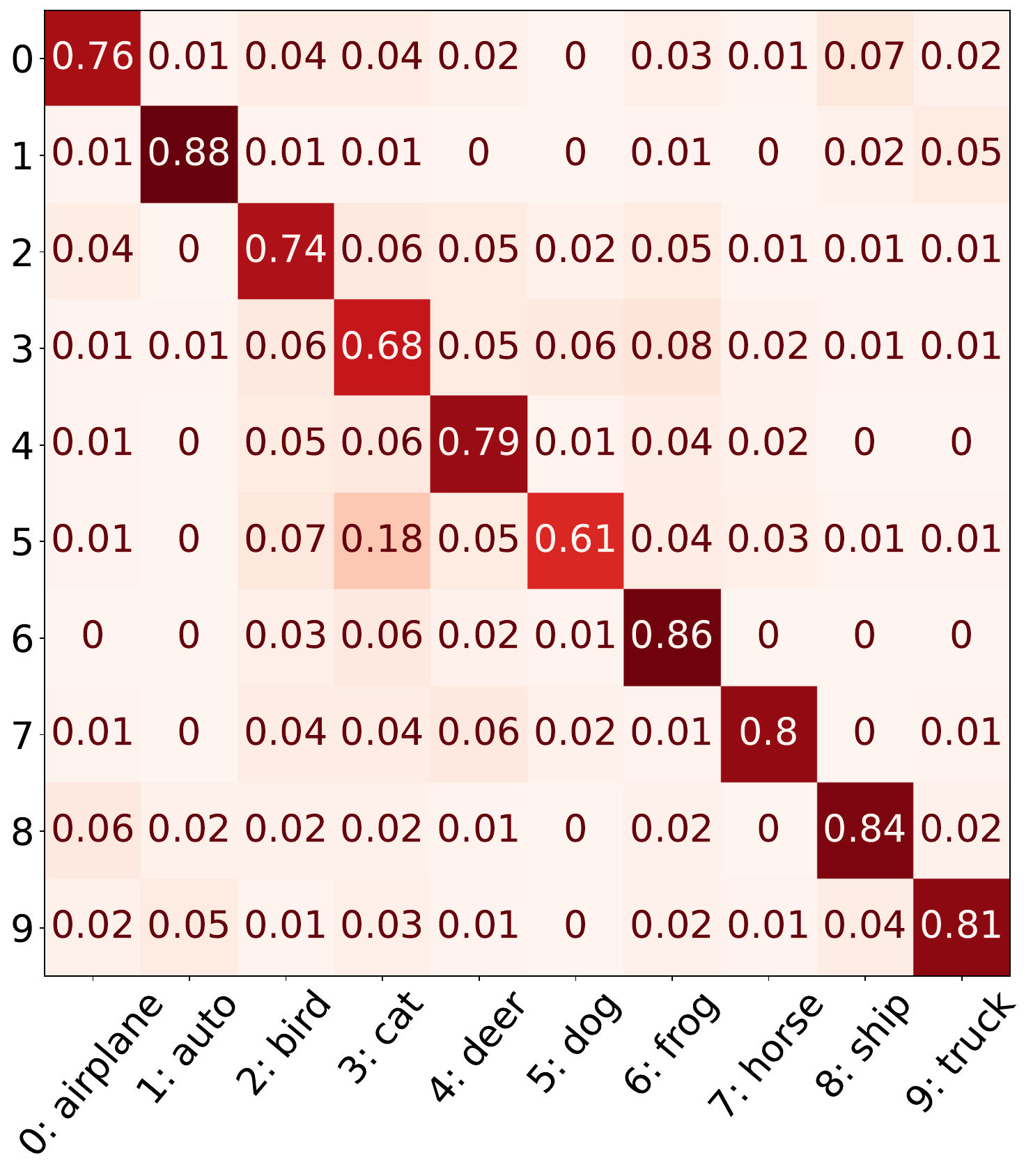}};
        \draw (\ox + \w * 3 ,-\h * \r) node[inner sep=0] {\includegraphics[width=\imgwidth cm]{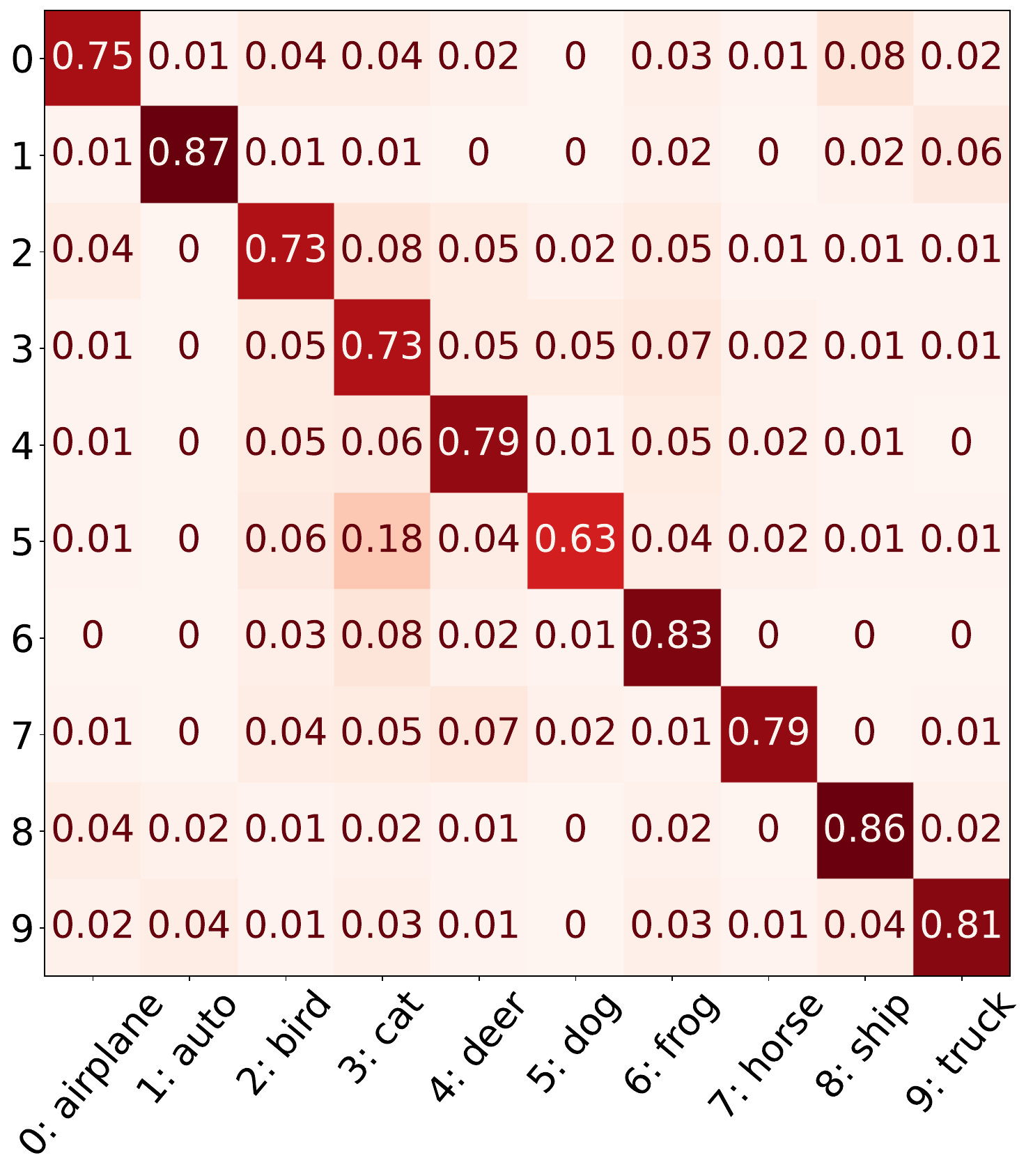}};

        \node at (-\imgwidth * 0.25, -\h * \r + 1, 2) [rotate=90] {\scriptsize{True label}};

        \node at (\ox + \w * 0, -\h * \r + \imgwidth * 0.55) [above] {\scriptsize{5\textsuperscript{th} visit}};
        \node at (\ox + \w * 1, -\h * \r + \imgwidth * 0.55) [above] {\scriptsize{6\textsuperscript{th} visit}};
        \node at (\ox + \w * 2, -\h * \r + \imgwidth * 0.55) [above] {\scriptsize{7\textsuperscript{th} visit}};
        \node at (\ox + \w * 3, -\h * \r + \imgwidth * 0.55) [above] {\scriptsize{8\textsuperscript{th} visit}};

        \def \r{2} 
        \draw (\ox + \w * 0, -\h * \r) node[inner sep=0] {\includegraphics[width=\imgwidth cm]{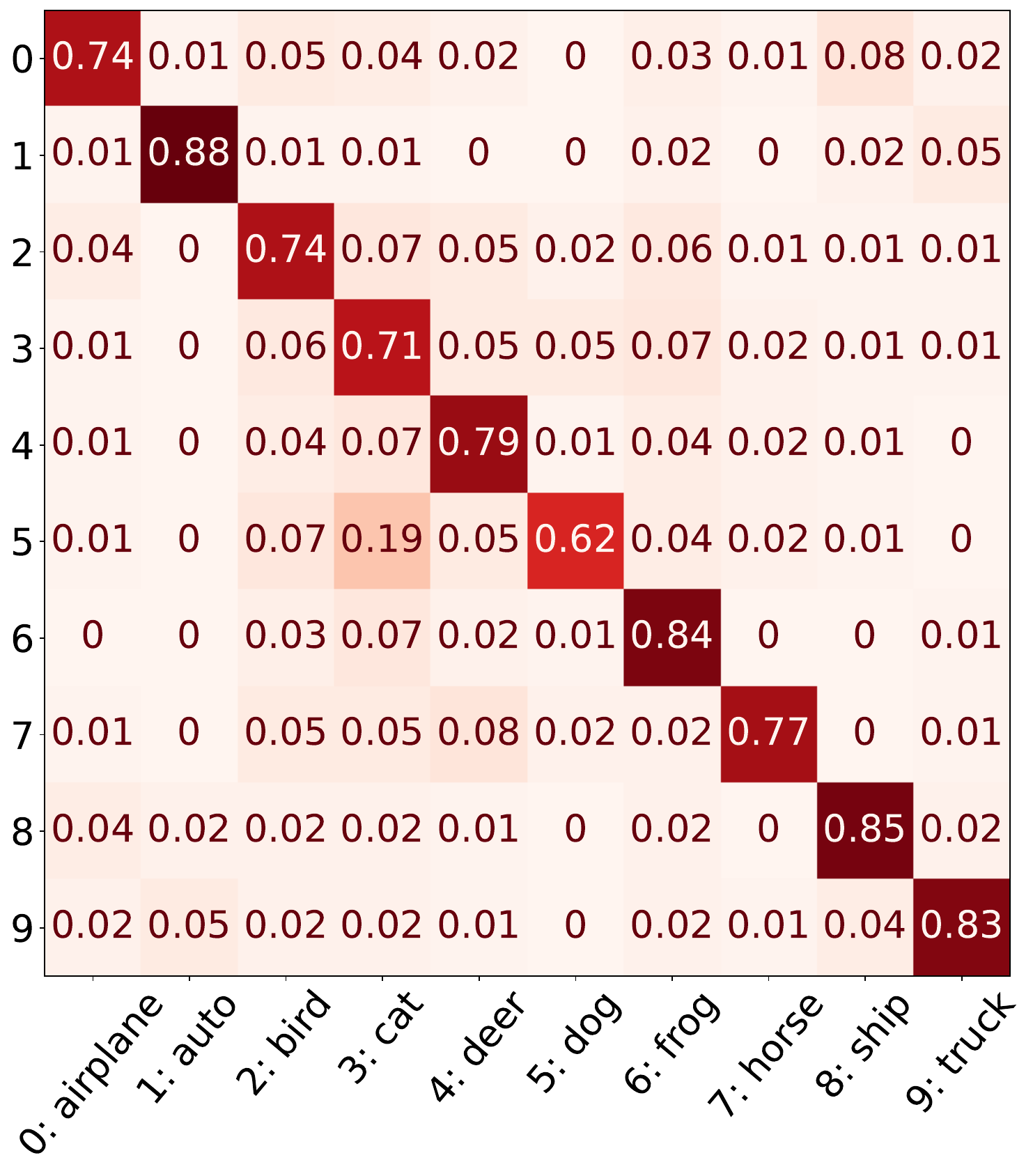}};
        \draw (\ox + \w * 1, -\h * \r) node[inner sep=0] {\includegraphics[width=\imgwidth cm]{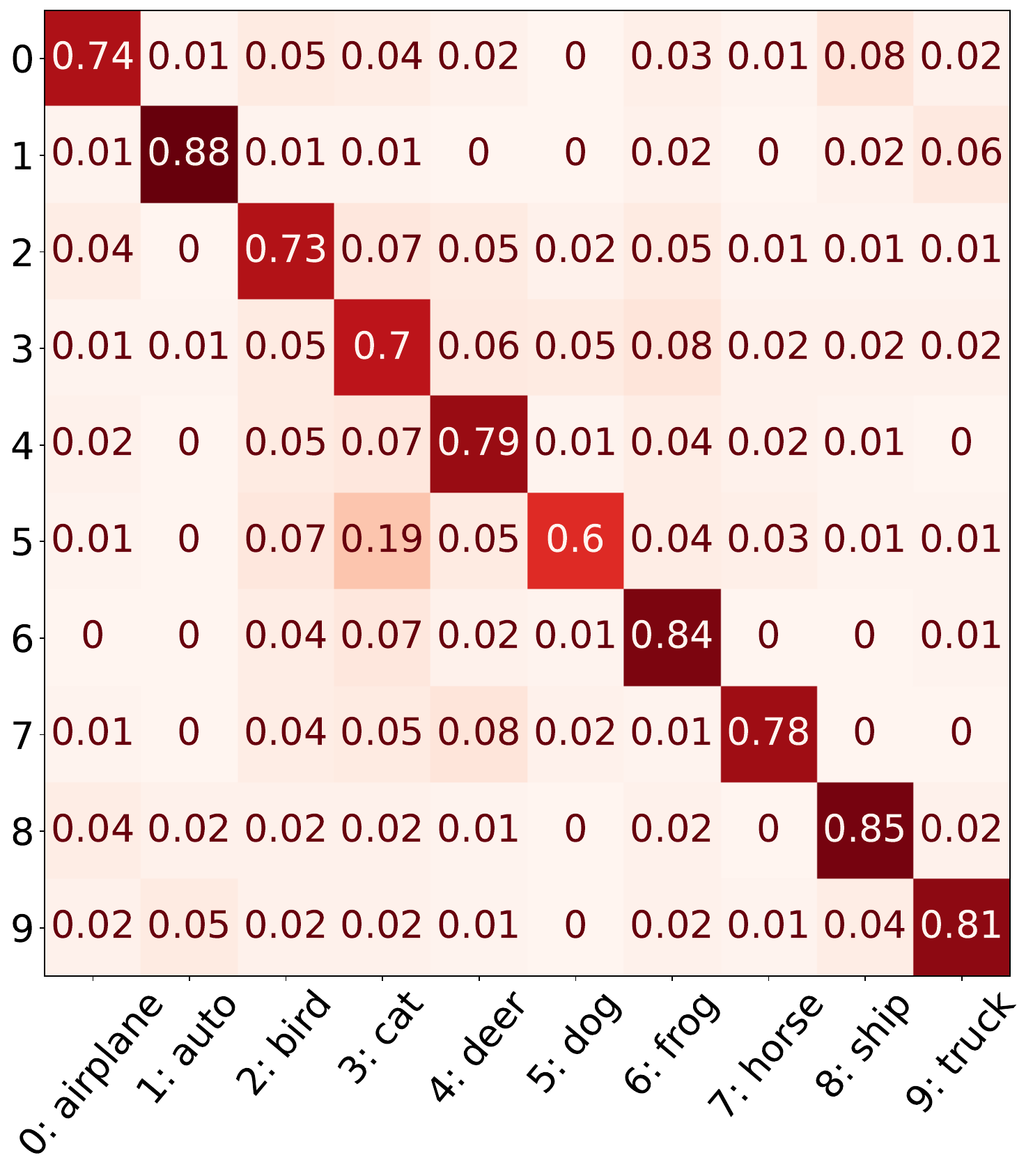}};
        \draw (\ox + \w * 2, -\h * \r) node[inner sep=0] {\includegraphics[width=\imgwidth cm]{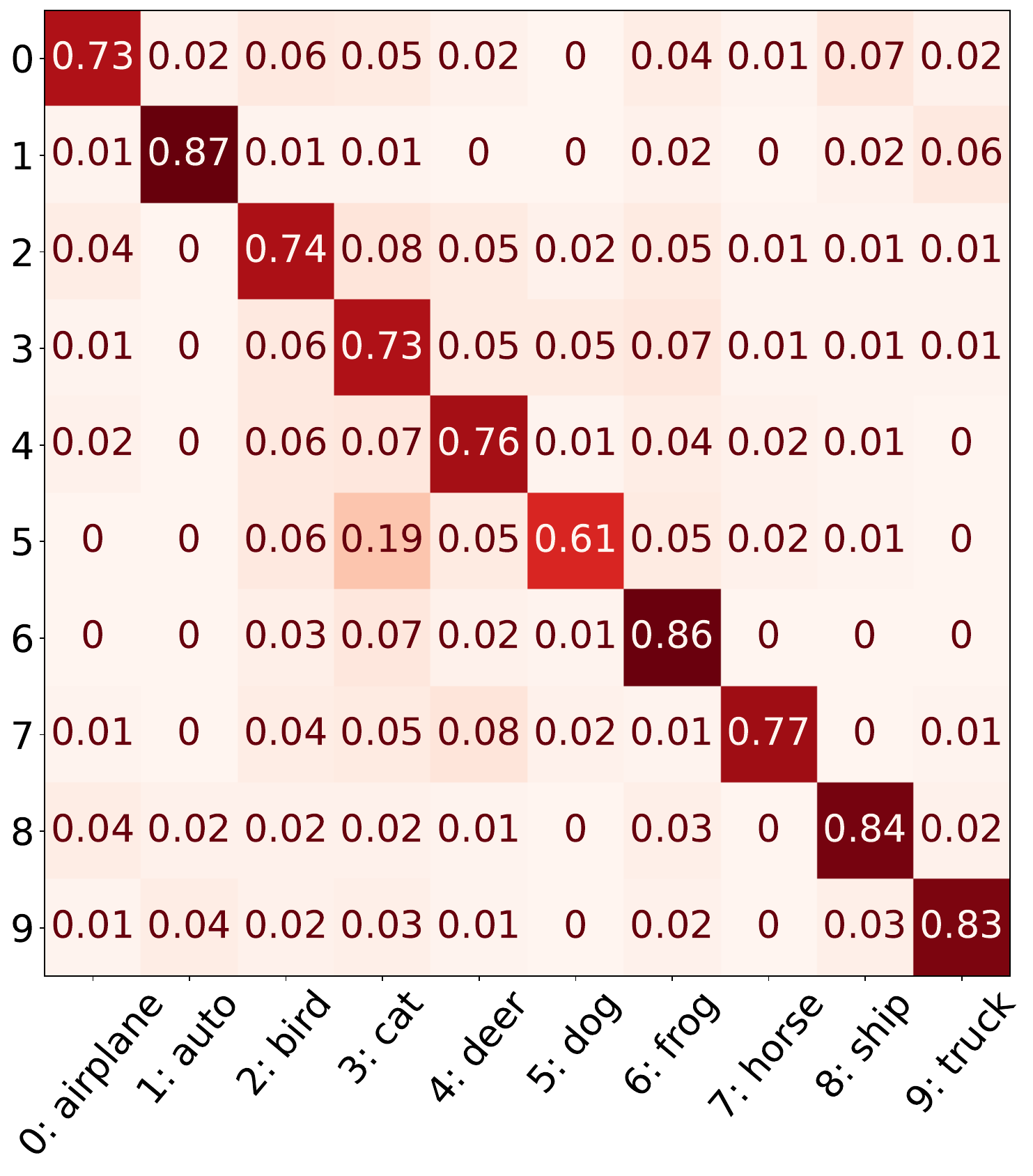}};
        \draw (\ox + \w * 3 ,-\h * \r) node[inner sep=0] {\includegraphics[width=\imgwidth cm]{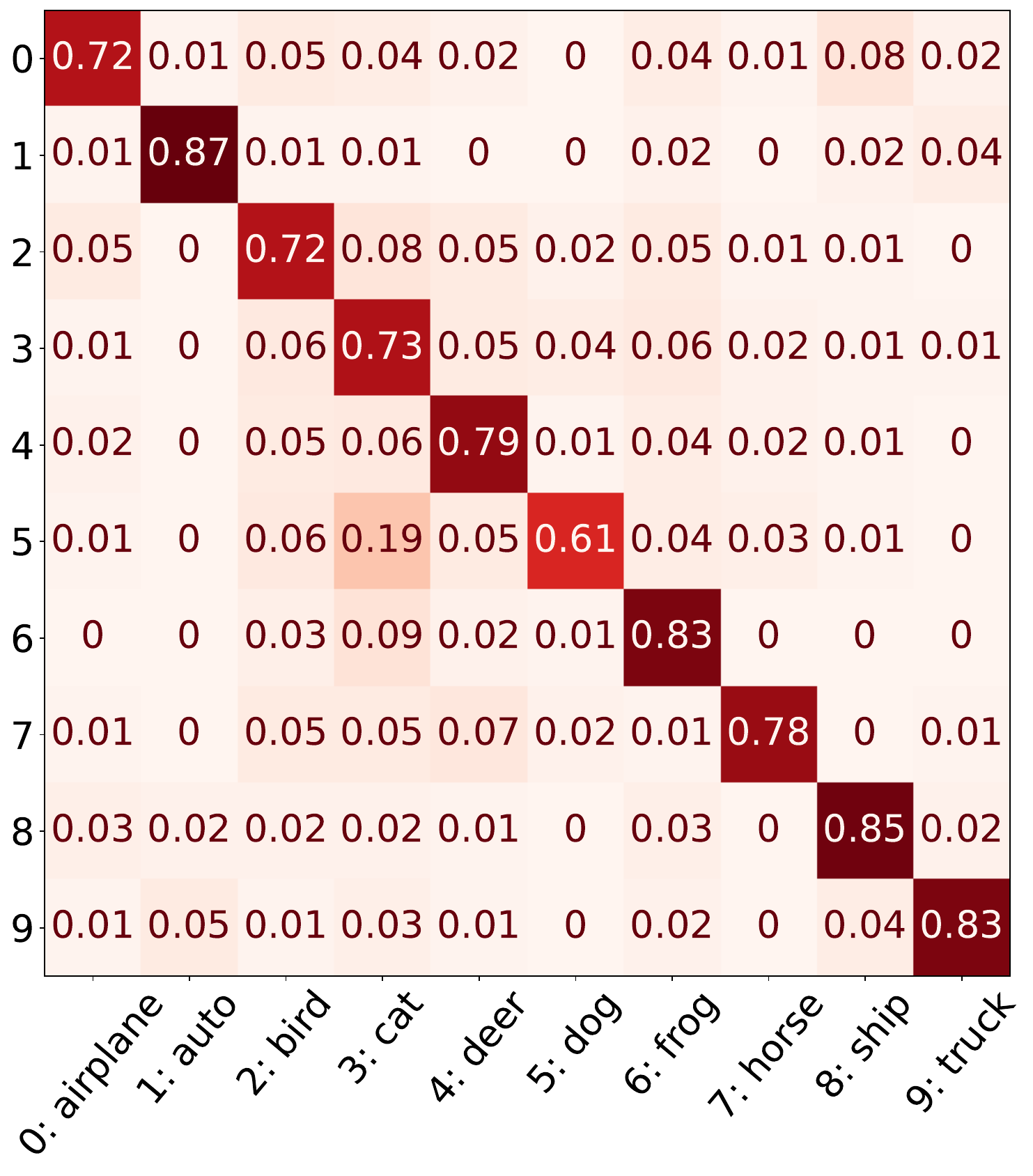}};

        \node at (-\imgwidth * 0.25, -\h * \r + 1, 2) [rotate=90] {\scriptsize{True label}};

        \node at (\ox + \w * 0, -\h * \r + \imgwidth * 0.55) [above] {\scriptsize{9\textsuperscript{th} visit}};
        \node at (\ox + \w * 1, -\h * \r + \imgwidth * 0.55) [above] {\scriptsize{10\textsuperscript{th} visit}};
        \node at (\ox + \w * 2, -\h * \r + \imgwidth * 0.55) [above] {\scriptsize{11\textsuperscript{th} visit}};
        \node at (\ox + \w * 3, -\h * \r + \imgwidth * 0.55) [above] {\scriptsize{12\textsuperscript{th} visit}};

        \def \r{3} 
        \draw (\ox + \w * 0, -\h * \r) node[inner sep=0] {\includegraphics[width=\imgwidth cm]{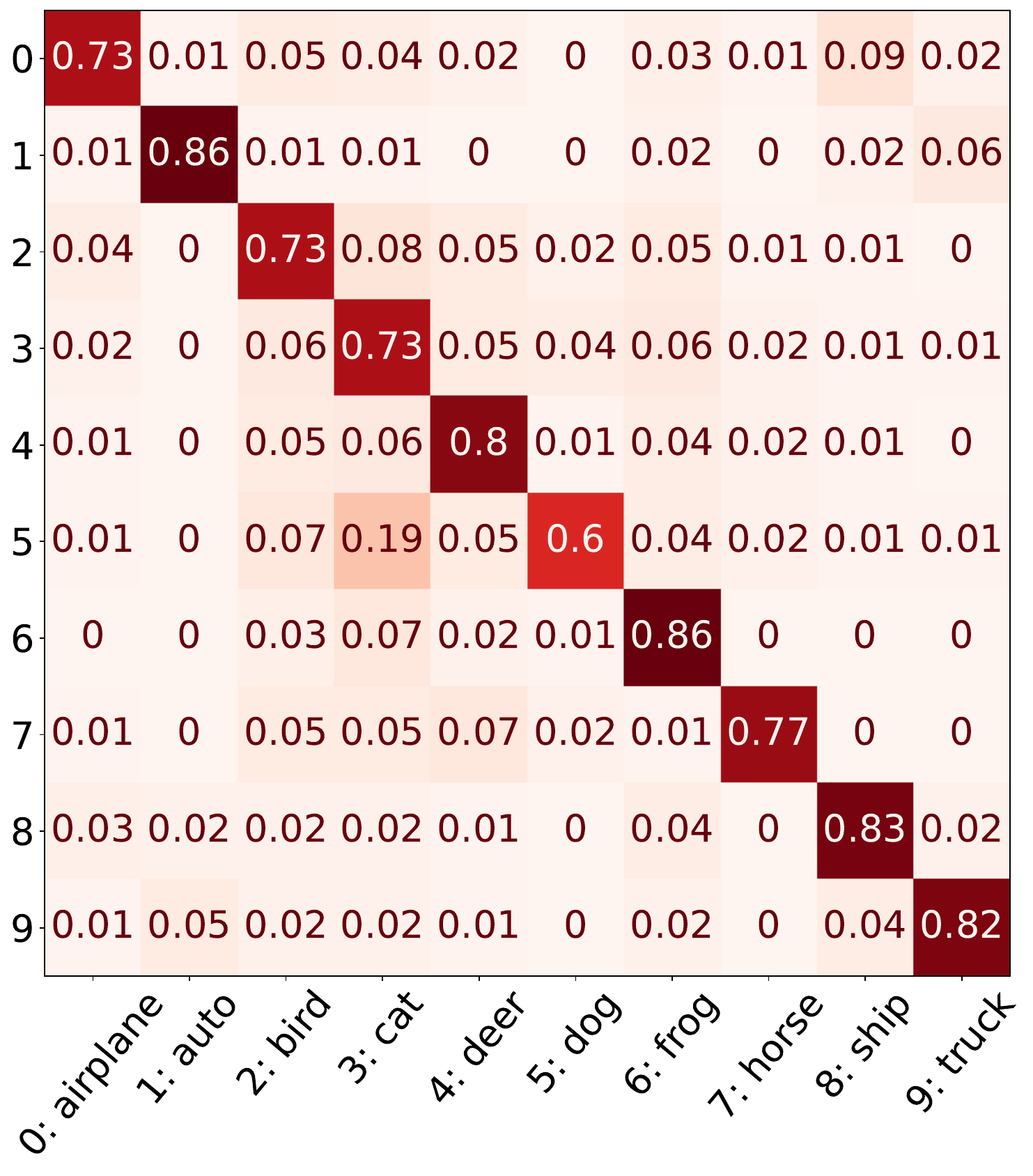}};
        \draw (\ox + \w * 1, -\h * \r) node[inner sep=0] {\includegraphics[width=\imgwidth cm]{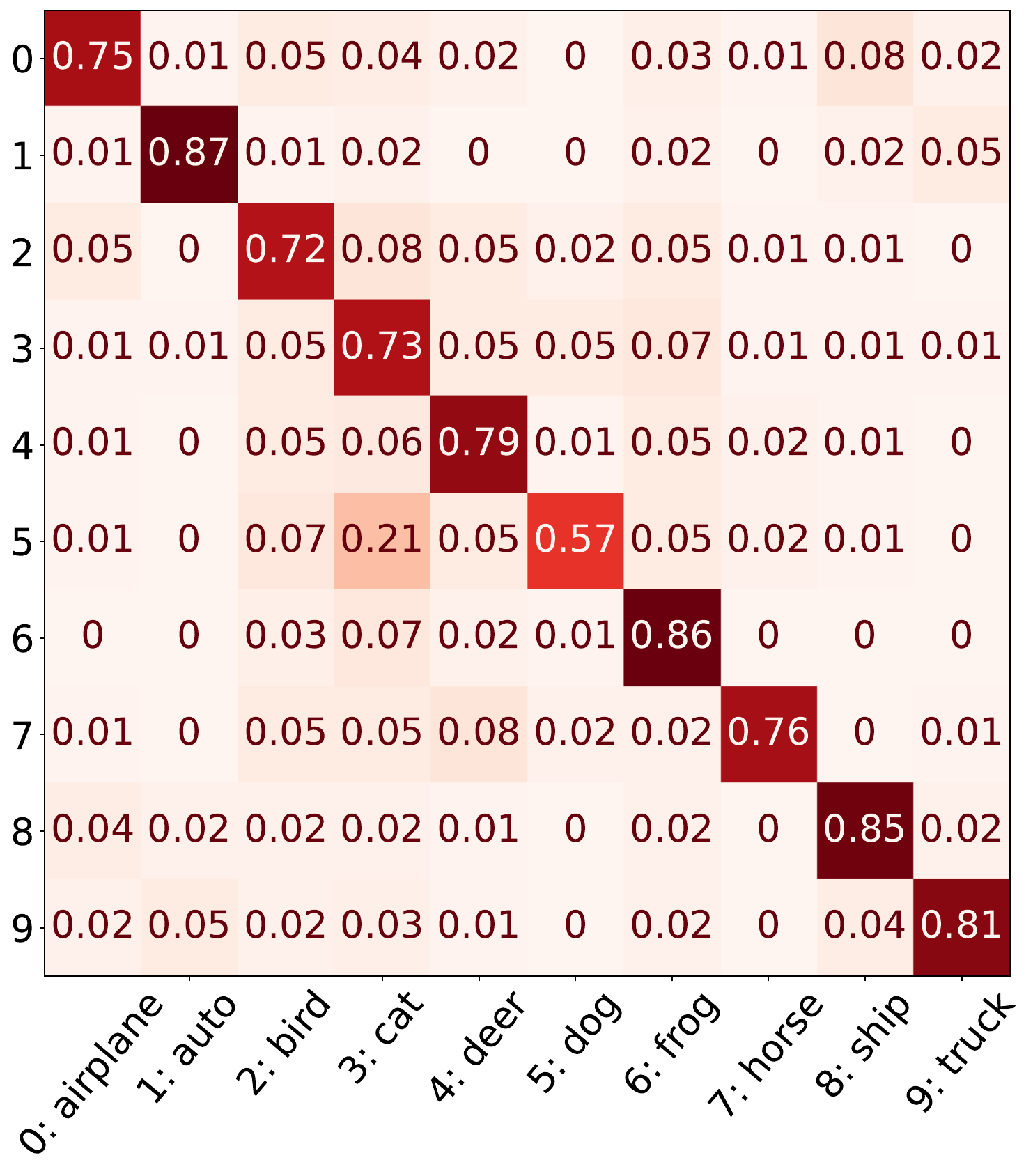}};
        \draw (\ox + \w * 2, -\h * \r) node[inner sep=0] {\includegraphics[width=\imgwidth cm]{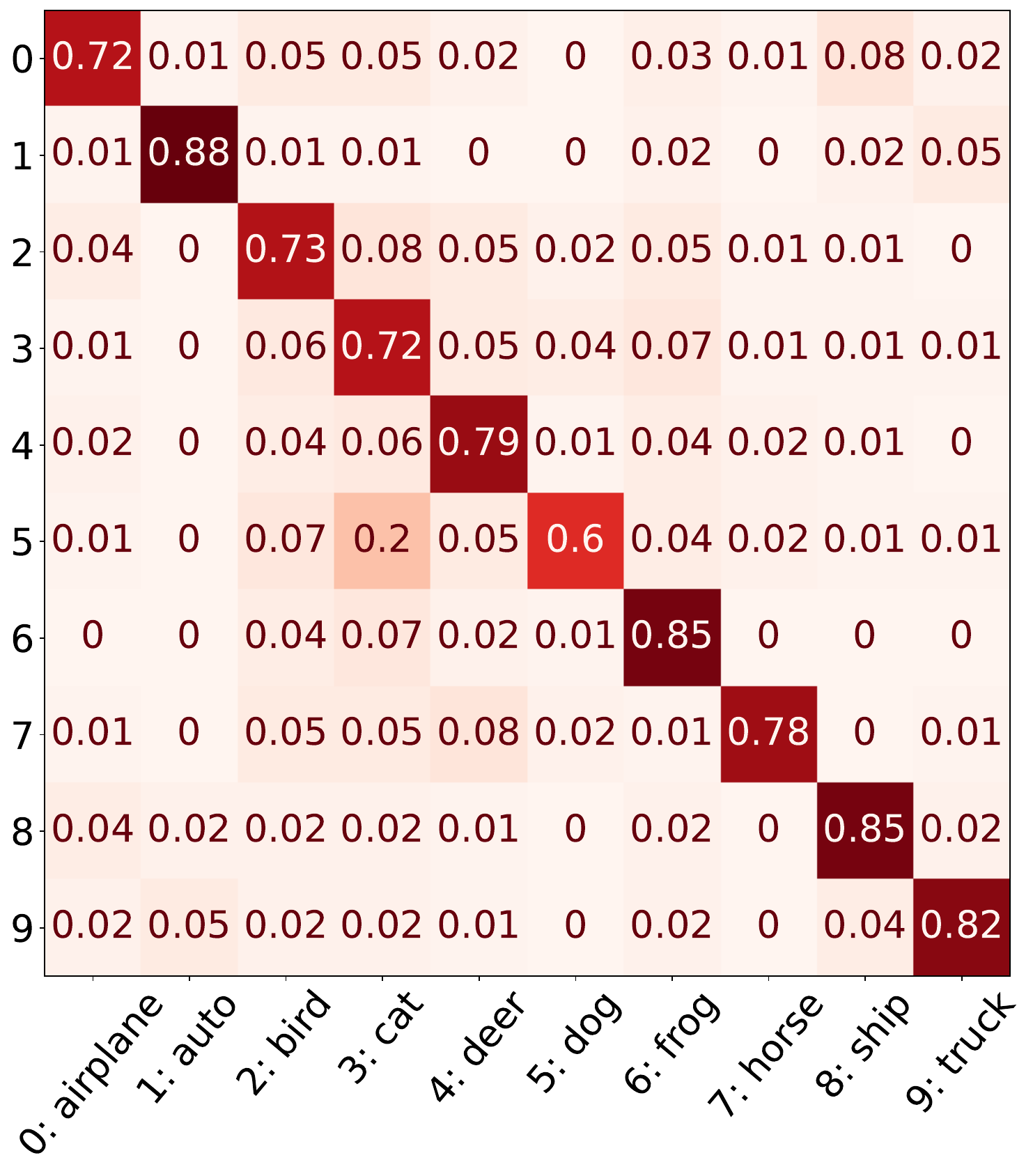}};
        \draw (\ox + \w * 3 ,-\h * \r) node[inner sep=0] {\includegraphics[width=\imgwidth cm]{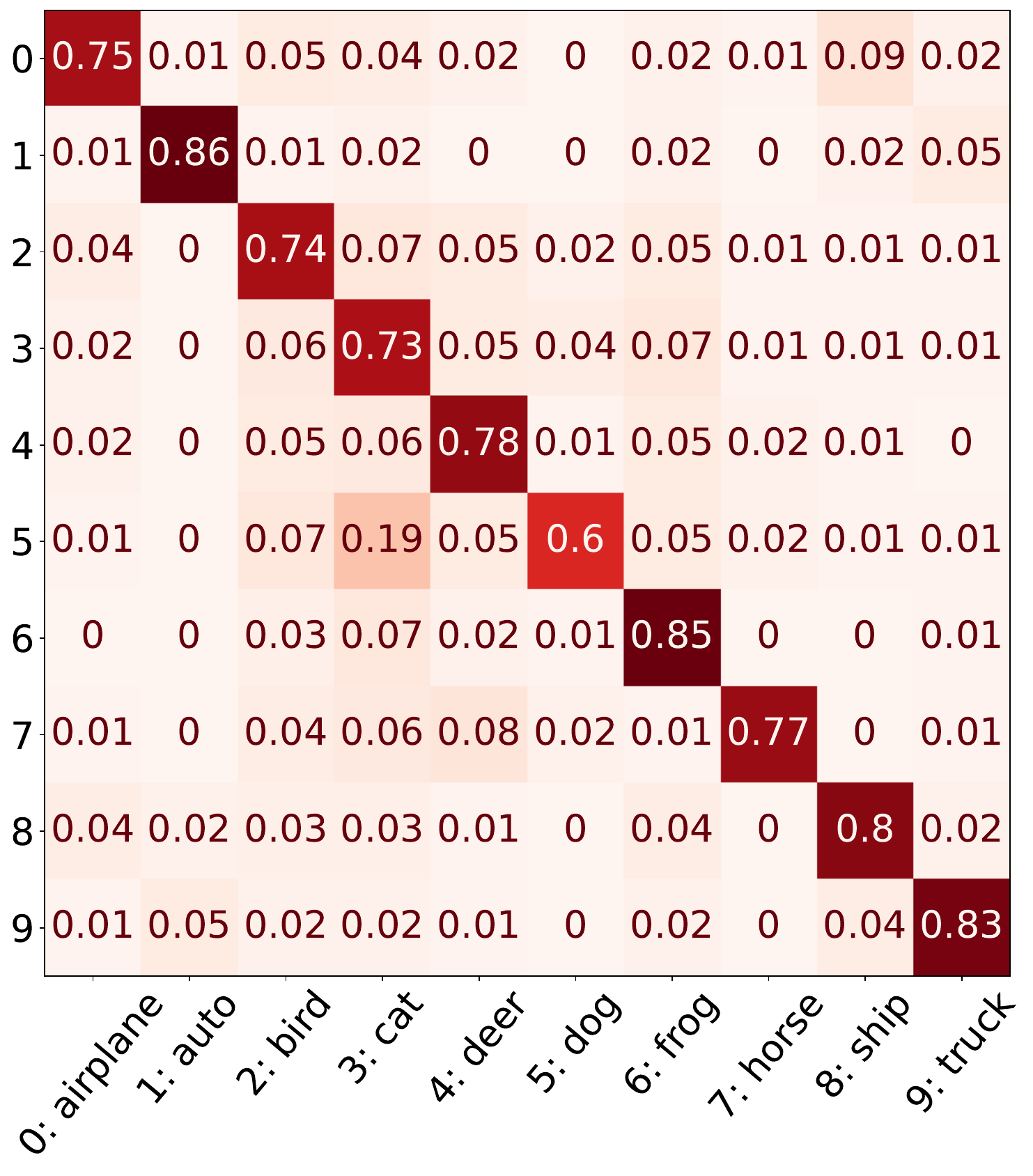}};

        \node at (-\imgwidth * 0.25, -\h * \r + 1, 2) [rotate=90] {\scriptsize{True label}};

        \node at (\ox + \w * 0, -\h * \r + \imgwidth * 0.55) [above] {\scriptsize{13\textsuperscript{th} visit}};
        \node at (\ox + \w * 1, -\h * \r + \imgwidth * 0.55) [above] {\scriptsize{14\textsuperscript{th} visit}};
        \node at (\ox + \w * 2, -\h * \r + \imgwidth * 0.55) [above] {\scriptsize{15\textsuperscript{th} visit}};
        \node at (\ox + \w * 3, -\h * \r + \imgwidth * 0.55) [above] {\scriptsize{16\textsuperscript{th} visit}};
        
        \def \r{4} 
        \draw (\ox + \w * 0, -\h * \r) node[inner sep=0] {\includegraphics[width=\imgwidth cm]{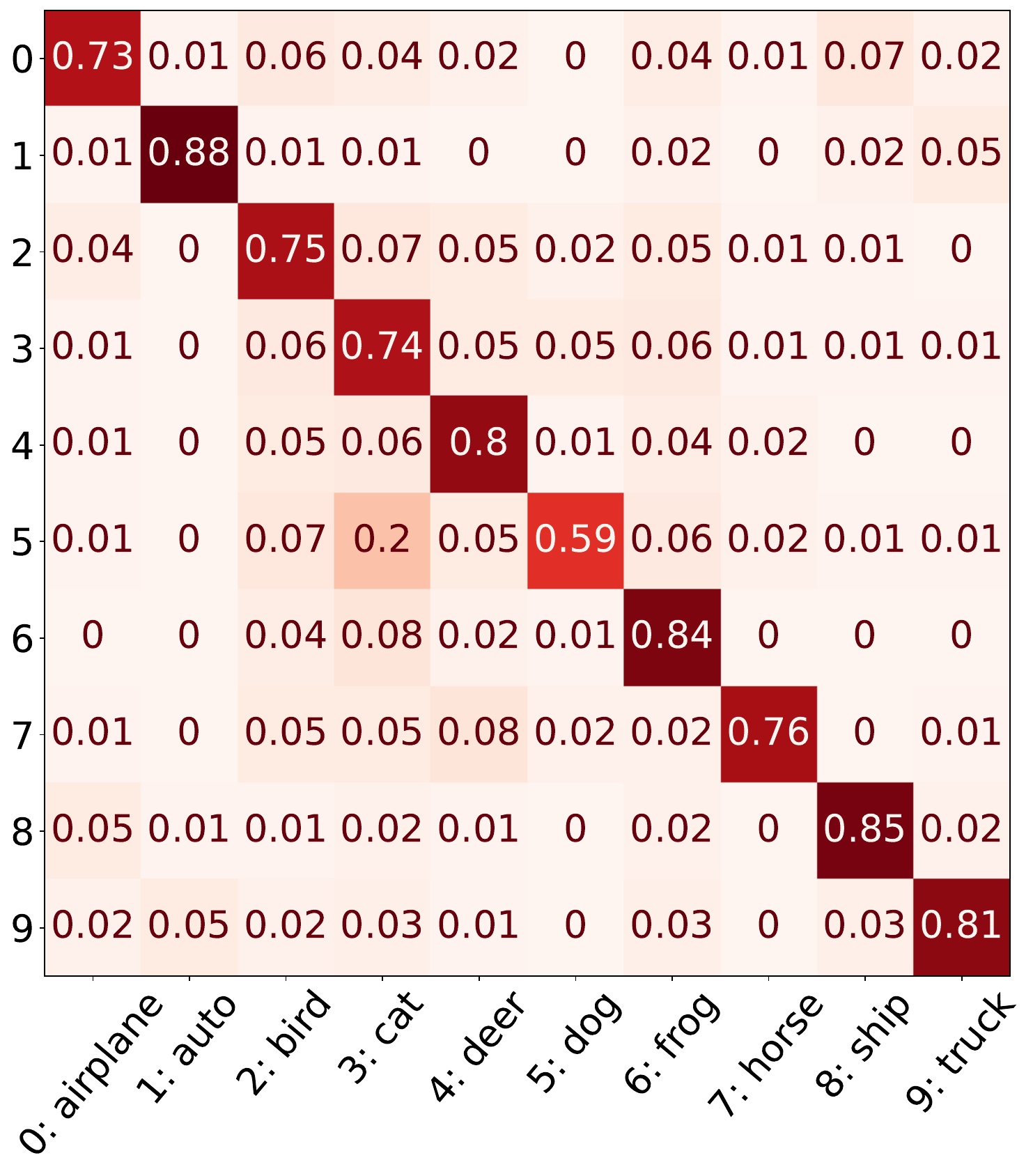}};
        \draw (\ox + \w * 1, -\h * \r) node[inner sep=0] {\includegraphics[width=\imgwidth cm]{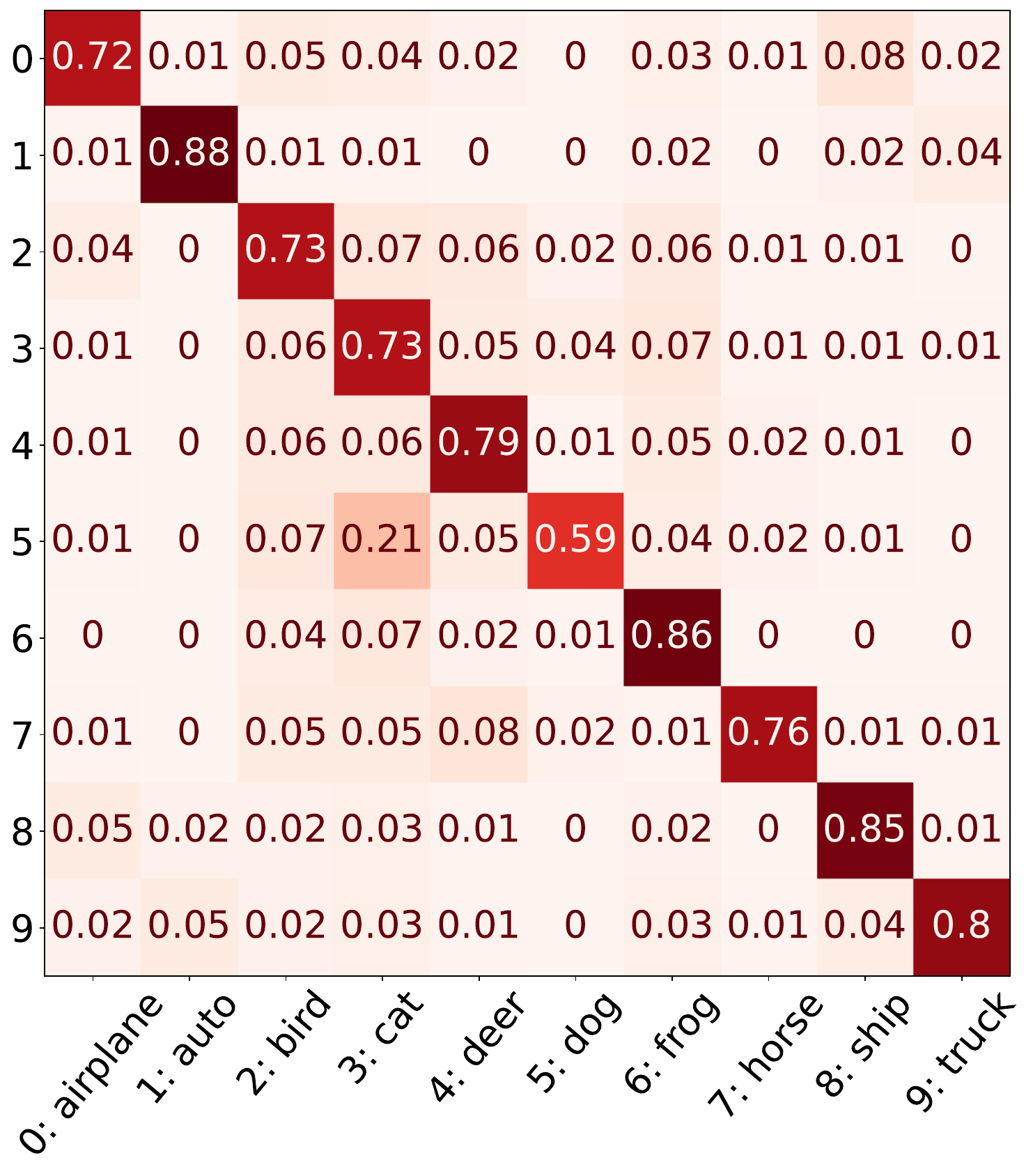}};
        \draw (\ox + \w * 2, -\h * \r) node[inner sep=0] {\includegraphics[width=\imgwidth cm]{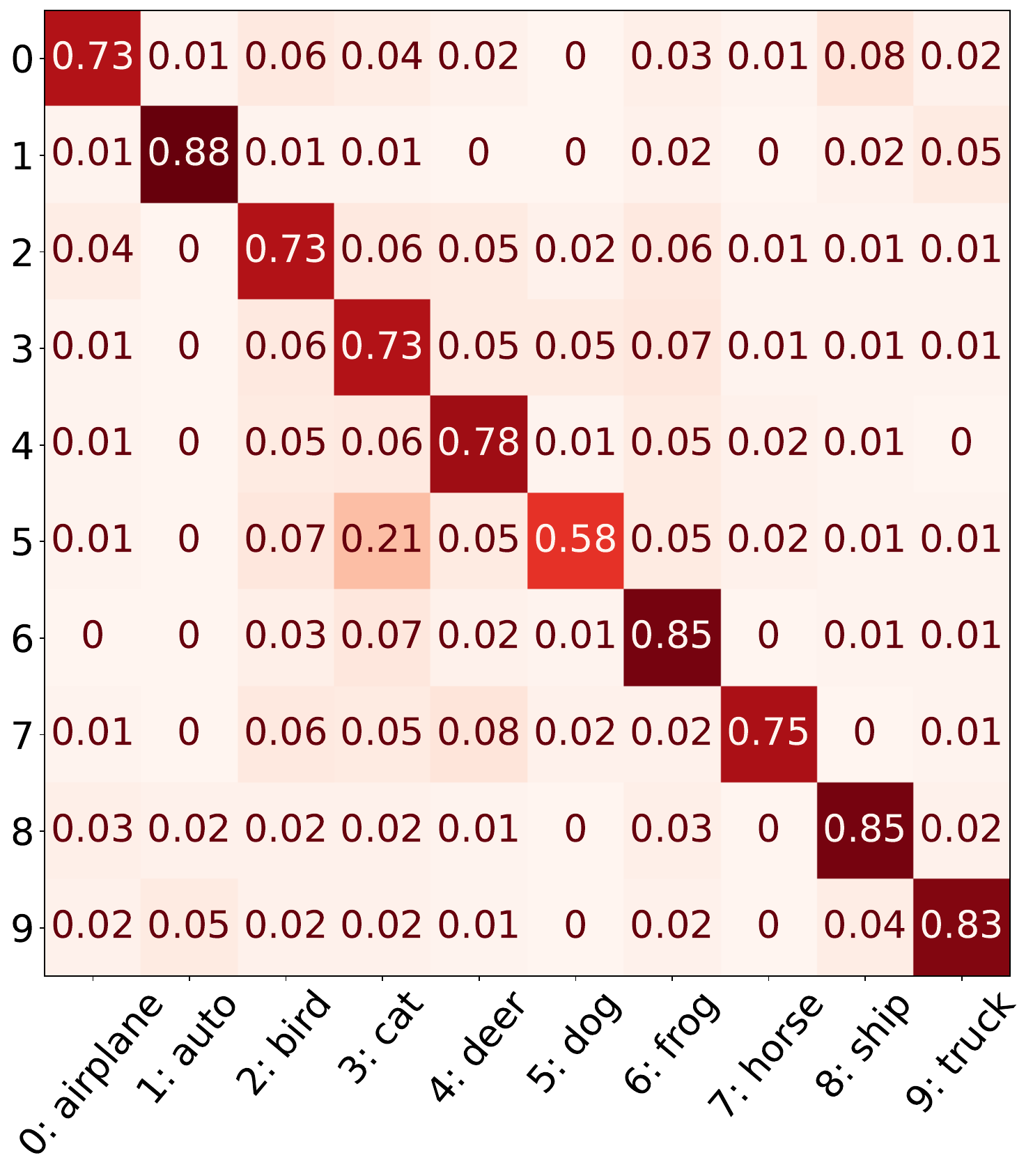}};
        \draw (\ox + \w * 3 ,-\h * \r) node[inner sep=0] {\includegraphics[width=\imgwidth cm]{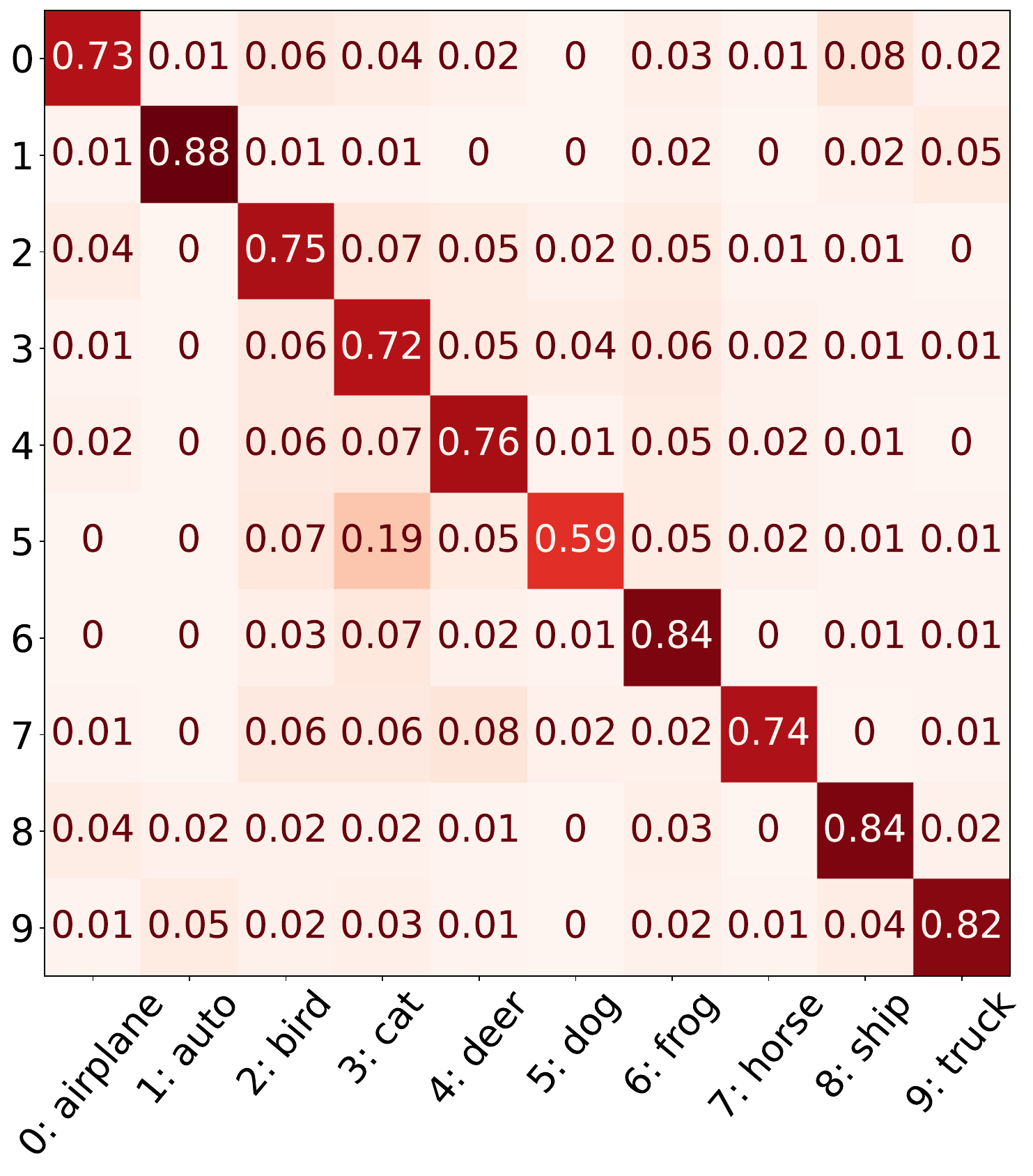}};

        \node at (\ox + \w * 0, -\h * \r - \imgwidth * 0.6) [above] {\scriptsize{Predicted label}};
        \node at (\ox + \w * 1, -\h * \r - \imgwidth * 0.6) [above] {\scriptsize{Predicted label}};
        \node at (\ox + \w * 2, -\h * \r - \imgwidth * 0.6) [above] {\scriptsize{Predicted label}};
        \node at (\ox + \w * 3, -\h * \r - \imgwidth * 0.6) [above] {\scriptsize{Predicted label}};
        
        \node at (-\imgwidth * 0.25, -\h * \r + 1, 2) [rotate=90] {\scriptsize{True label}};

        \node at (\ox + \w * 0, -\h * \r + \imgwidth * 0.55) [above] {\scriptsize{17\textsuperscript{th} visit}};
        \node at (\ox + \w * 1, -\h * \r + \imgwidth * 0.55) [above] {\scriptsize{18\textsuperscript{th} visit}};
        \node at (\ox + \w * 2, -\h * \r + \imgwidth * 0.55) [above] {\scriptsize{19\textsuperscript{th} visit}};
        \node at (\ox + \w * 3, -\h * \r + \imgwidth * 0.55) [above] {\scriptsize{20\textsuperscript{th} visit}};
        
        \end{tikzpicture}
    \vspace*{-0.5\baselineskip}
    \caption{The dynamic of the confusion matrix of \method (\textit{ours}) in episodic TTA with 20 visits.}
    \vspace*{-\baselineskip}
    \label{fig:full_cfs_mat_petta}
\end{figure*}

For the task CIFAR-10 $\rightarrow$ CIFAR-10-C~\cite{hendrycks2019robustness} in \textit{\setting TTA} setting (with 20 visits), we additionally showcase the confusion matrix of RoTTA~\cite{yuan2023robust} (\Fig{\ref{fig:full_cfs_mat_rotta}}) and our proposed \method (Fig.~\ref{fig:full_cfs_mat_petta}) at each visit. Our \method persistently achieves competitive performance across 20 visits while RoTTA~\cite{yuan2023robust} gradually degrades. 

\section{Experimental Details}
\label{sec:experimental_details}

\subsection{Computing Resources}
\label{ssec:computing_resources}
A computer cluster equipped with an Intel(R) Core(TM) $3.80$GHz i7-10700K CPU, 64 GB RAM, and one NVIDIA GeForce RTX 3090 GPU (24 GB VRAM) is used for our experiments. 

\subsection{Experiments on CCC Testing Stream}
In this section, we further evaluate the performance of our \method on the testing data stream of Continuous Changing Corruption (CCC)~\cite{press2023rdumb} setting. Here we use the baseline accuracy $20\%$, transition speed $1000$, and random seed $44$.\footnote{\url{https://github.com/oripress/CCC}}
The compared methods are source model (ResNet 50), \method, RoTTA~\cite{yuan2023robust}, and RDumb~\cite{press2023rdumb}. 
Noteworthy, different from \setting TTA, the class labels here are i.i.d. distributed.
The adaptation configuration of \method follows the same settings as used on ImageNet-C, while the same setting introduced in \Sec{\ref{appdx:model_reset}}, with $T=1000$ is used for RDumb~\cite{press2023rdumb}.  

\subsection{Test-time Adaptation Methods}
\noindent \textbf{Pre-trained Model on Source Distribution. } Following previous studies~\cite{wang2021tent, yuan2023robust, döbler2023robust, Wang_2022_CVPR}, only the batch norm layers are updated. As stated in \Sec{\ref{ssec:exp_setup}}, RobustBench~\cite{croce2021robustbench} and \texttt{torchvision}~\cite{torchvision2016} provide pre-trained models trained on source distributions. 
Specifically, for ImageNet-C and DomainNet experiments, a ResNet50 model~\cite{He2015} pre-trained on ImageNet V2 (specifically, checkpoint \texttt{ResNet50\_Weights.IMAGENET1K\_V2} of \texttt{torchvision}) is used. 
From \texttt{RobustBench}, the model with checkpoint \texttt{Standard} and \texttt{Hendrycks2020AugMix\_ResNeXt}~\cite{hendrycks2020augmix} are adopted for CIFAR10-C and CIFAR-100-C experiments, respectively. Lastly, experiments on DomainNet dataset utilize the checkpoint (\texttt{best\_real\_2020}) provided in AdaContrast~\cite{chen2022contrastive} study.\footnote{\url{https://github.com/DianCh/AdaContrast}}

\noindent \textbf{Optimizer. } Without specifically stated, Adam~\cite{KingmaB14} optimizer with learning rate equal $1e^{-3}$, and $\beta = (0.9, 0.999)$ is selected as a universal choice for all experiments. 

\noindent \textbf{More Details on \method. } Since designing the batch normalization layers, and the memory bank is not the key focus of \method, we conveniently adopt the implementation of the Robust Batch Norm layer and the Category-balanced Sampling strategy using a memory bank introduced in RoTTA~\cite{yuan2023robust}.

\subsection{The Use of Existing Assets}
\label{ssec:assets_license}
Many components of \method is utilized from the official repository of RoTTA~\cite{yuan2023robust}~\footnote{\url{https://github.com/BIT-DA/RoTTA}} and RMT~\cite{döbler2023robust}.~\footnote{\url{https://github.com/mariodoebler/test-time-adaptation}} These two assets are released under MIT license. All the datasets, including CIFAR-10-C, CIFAR-100-C and ImageNet-C~\cite{hendrycks2019robustness} are publicly available online, released under Apache-2.0 license.\footnote{\url{https://github.com/hendrycks/robustness}} DomainNet dataset~\cite{peng2019moment} (cleaned version) is also released for research purposes.\footnote{\url{https://ai.bu.edu/M3SDA/}}

\clearpage
\section*{NeurIPS Paper Checklist}

\begin{enumerate}

\item {\bf Claims}
    \item[] Question: Do the main claims made in the abstract and introduction accurately reflect the paper's contributions and scope?
    \item[] Answer: \answerYes{} %
    \item[] Justification: \justification{We have highlighted the three main claims and contributions of our work in both the abstract (highlighted in bold font) and the introduction section (listed as bullet points).}

\item {\bf Limitations}
    \item[] Question: Does the paper discuss the limitations of the work performed by the authors?
    \item[] Answer: \answerYes{} %
    \item[] Justification: \justification{We have discussed the limitations and potential future work of our study in \Sec{\ref{sec:discussions}}. Specifically, three main limitations are included: (1) Collapse prevention can not be guaranteed through regularization, \method requires (2) the use of a relatively small memory bank is available and (3) the empirical mean and covariant matrix of feature vectors on the source dataset is computable. We also include discussions in \Appdx{\ref{appdx:using_mem_bank}} and \Appdx{\ref{appdx:mean_cov_source_dataset}} to further elaborate (2), and (3) respectively.}

\item {\bf Theory Assumptions and Proofs}
    \item[] Question: For each theoretical result, does the paper provide the full set of assumptions and a complete (and correct) proof?
    \item[] Answer: \answerYes{} %
    \item[] Justification: \justification{We have provided the full proof of all lemmas and theorem in \Appdx{\ref{sec:convergence_proof}}.}

    \item {\bf Experimental Result Reproducibility}
    \item[] Question: Does the paper fully disclose all the information needed to reproduce the main experimental results of the paper to the extent that it affects the main claims and/or conclusions of the paper (regardless of whether the code and data are provided or not)?
    \item[] Answer: \answerYes{} %
    \item[] Justification: \justification{This study propose a new TTA approach - \method. A full description of this approach is given in \Sec{\ref{ssec:pers_tta}} with its pseudo-code provided in \Appdx{\ref{appdx:petta_pseudo_code}. The implementation of \method in Python is also attached as supplemental material. Additionally, \Sec{\ref{ssec:exp_setup}} and \Appdx{\ref{sec:experimental_details}}} are dedicated to providing further implementation details for reproducing the main experimental results. Lastly, the construction of \setting TTA is notably simple, and can be easily extended to other TTA streams. Its configuration on each tasks is described in the \Setting TTA paragraph of \Sec{\ref{ssec:exp_setup}}.}

\item {\bf Open access to data and code}
    \item[] Question: Does the paper provide open access to the data and code, with sufficient instructions to faithfully reproduce the main experimental results, as described in supplemental material?
    \item[] Answer: \answerYes{} %
    \item[] Justification: \justification{This study does not involve any private datasets. All datasets used in our experiments are publicly available online from previous works (more information in \Appdx{\ref{ssec:assets_license}}). The source code of \method is also attached as supplemental material.}

\item {\bf Experimental Setting/Details}
    \item[] Question: Does the paper specify all the training and test details (e.g., data splits, hyperparameters, how they were chosen, type of optimizer, etc.) necessary to understand the results?
    \item[] Answer: \answerYes{} %
    \item[] Justification: \justification{The experimental settings of the key results in the paper have been provided in \Sec{\ref{sec:eps_gmmc_result}} (Simulation Setup) and \Sec{\ref{ssec:exp_setup}} (Setup - Benchmark Datasets). In the supplementary material, any \textit{additional} experimental results beyond the main paper, such as those in \Appdx{\ref{appdx:repeating_random_orders}}, and \Appdx{\ref{appdx:model_reset}}, are consistently preceded by a subsection titled \textit{Experiment Setup} summarizing the experimental details before presenting the results.}

\item {\bf Experiment Statistical Significance}
    \item[] Question: Does the paper report error bars suitably and correctly defined or other appropriate information about the statistical significance of the experiments?
    \item[] Answer:  \answerYes{} %
    \item[] Justification: \justification{
    Due to the limited computing resources, we only extensively evaluate the performance of our proposed method \textit{(\method)} across 5 independent runs, with different random seeds. Specifically, the mean values in 5 runs are reported in \Tab{\ref{tab:cifar-10-performance}}, \Tab{\ref{tab:imagenet-c-performance}}, \Tab{\ref{tab:cifar-100-performance}}, and \Tab{\ref{tab:domainnet-performance}}. The corresponding standard deviation values are provided in \Appdx{\ref{appdx:additional_main_results}}.
    }

\item {\bf Experiments Compute Resources}
    \item[] Question: For each experiment, does the paper provide sufficient information on the computer resources (type of compute workers, memory, time of execution) needed to reproduce the experiments?
    \item[] Answer: \answerYes{} %
    \item[] Justification: \justification{We have provided the information on the computing resources used in our experiments in \Appdx{\ref{ssec:computing_resources}}.}
    
\item {\bf Code Of Ethics}
    \item[] Question: Does the research conducted in the paper conform, in every respect, with the NeurIPS Code of Ethics \url{https://neurips.cc/public/EthicsGuidelines}?
    \item[] Answer: \answerYes{} %
    \item[] Justification: \justification{The authors have reviewed and to the best of our judgment, this study has conformed to the NeurIPS Code of Ethics.}

\item {\bf Broader Impacts}
    \item[] Question: Does the paper discuss both potential positive societal impacts and negative societal impacts of the work performed?
    \item[] Answer: \answerNo{}  %
    \item[] Justification: \justification{This study advances the research in test-time adaptation area in general, and not tied to particular applications. Hence, there are no significant potential societal consequences of our work which we feel must be specifically highlighted here.}
    
\item {\bf Safeguards}
    \item[] Question: Does the paper describe safeguards that have been put in place for responsible release of data or models that have a high risk for misuse (e.g., pretrained language models, image generators, or scraped datasets)?
    \item[] Answer: \answerNA{} %
    \item[] Justification: \justification{To the best of our judgment, this study poses no risks for misuse.}

\item {\bf Licenses for existing assets}
    \item[] Question: Are the creators or original owners of assets (e.g., code, data, models), used in the paper, properly credited and are the license and terms of use explicitly mentioned and properly respected?
    \item[] Answer: \answerYes{} %
    \item[] Justification: \justification{The original papers that produced the code package or dataset have been properly cited throughout the paper. Further information on the licenses of used assets are provided in \Appdx{\ref{ssec:assets_license}}.}   

\item {\bf New Assets}
    \item[] Question: Are new assets introduced in the paper well documented and is the documentation provided alongside the assets?
    \item[] Answer: \answerNA{} %
    \item[] Justification: \justification{This study does not release new assets.}

\item {\bf Crowdsourcing and Research with Human Subjects}
    \item[] Question: For crowdsourcing experiments and research with human subjects, does the paper include the full text of instructions given to participants and screenshots, if applicable, as well as details about compensation (if any)? 
    \item[] Answer: \answerNA{} %
    \item[] Justification: \justification{This study does not involve crowdsourcing nor research with human subjects.}

\item {\bf Institutional Review Board (IRB) Approvals or Equivalent for Research with Human Subjects}
    \item[] Question: Does the paper describe potential risks incurred by study participants, whether such risks were disclosed to the subjects, and whether Institutional Review Board (IRB) approvals (or an equivalent approval/review based on the requirements of your country or institution) were obtained?
    \item[] Answer: \answerNA{} %
    \item[] Justification: \justification{This study does not involve crowdsourcing nor research with human subjects.}

\end{enumerate}

\end{document}